# OPTIMAL COMPUTATIONAL AND STATISTICAL RATES OF CONVERGENCE FOR SPARSE NONCONVEX LEARNING PROBLEMS

By Zhaoran Wang*, Han Liu* and Tong Zhang†

*Princeton University* and Rutgers University†*

We provide theoretical analysis of the statistical and computational properties of penalized $M$-estimators that can be formulated as the solution to a possibly nonconvex optimization problem. Many important estimators fall in this category, including least squares regression with nonconvex regularization, generalized linear models with nonconvex regularization, and sparse elliptical random design regression. For these problems, it is intractable to calculate the global solution due to the nonconvex formulation. In this paper, we propose an approximate regularization path following method for solving a variety of learning problems with nonconvex objective functions. Under a unified analytic framework, we simultaneously provide explicit statistical and computational rates of convergence for any local solution attained by the algorithm. Computationally, our algorithm attains a global geometric rate of convergence for calculating the full regularization path, which is optimal among all first-order algorithms. Unlike most existing methods that only attain geometric rates of convergence for one single regularization parameter, our algorithm calculates the full regularization path with the same iteration complexity. In particular, we provide a refined iteration complexity bound to sharply characterize the performance of each stage along the regularization path. Statistically, we provide sharp sample complexity analysis for all the approximate local solutions along the regularization path. In particular, our analysis improves upon existing results by providing a more refined sample complexity bound as well as an exact support recovery result for the final estimator. These results show that the final estimator attains an oracle statistical property due to the usage of nonconvex penalty.

**1. Introduction.** This paper considers the statistical and computational properties of a family of penalized $M$-estimators that can be formulated as

$$\widehat{\boldsymbol{\beta}}_\lambda \in \underset{\boldsymbol{\beta}\in\mathbb{R}^d}{\operatorname{argmin}}\Big\{\mathcal{L}(\boldsymbol{\beta}) + \mathcal{P}_\lambda(\boldsymbol{\beta})\Big\}, \tag{1.1}$$

where $\mathcal{L}(\boldsymbol{\beta})$ is a loss function, while $\mathcal{P}_\lambda(\boldsymbol{\beta})$ is a penalty function with regularization parameter $\lambda$. A familiar example is the Lasso estimator (Tibshirani, 1996), in which $\mathcal{L}(\boldsymbol{\beta}) = \|\mathbf{X}\boldsymbol{\beta} - \mathbf{y}\|_2^2/(2n)$ and $\mathcal{P}_\lambda(\boldsymbol{\beta}) = \lambda\|\boldsymbol{\beta}\|_1$. Here $\mathbf{X} = (\mathbf{x}_1, \dots, \mathbf{x}_n)^T \in \mathbb{R}^{n\times d}$ is the design matrix, $\mathbf{y} = (y_1, \dots, y_n)^T \in \mathbb{R}^n$ is the response vector, $\|\cdot\|_2$ is the Euclidean norm, and $\|\boldsymbol{\beta}\|_1 = \sum_{j=1}^d |\beta_j|$ is the $\ell_1$ norm of $\boldsymbol{\beta}$. In general, we prefer the settings where both the loss function $\mathcal{L}(\boldsymbol{\beta})$ and the penalty term $\mathcal{P}_\lambda(\boldsymbol{\beta})$ in (1.1) are convex, since convexity makes both statistical and computational analysis convenient.

Significant progress has been made on understanding convex penalized $M$-estimators (van de Geer, 2000, 2008; Rothman et al., 2008; Wainwright, 2009; Bickel et al., 2009; Zhang, 2009; Koltchinskii, 2009b; Raskutti et al., 2011; Negahban et al., 2012). Meanwhile, penalized $M$-estimators with nonconvex loss or penalty functions have recently attracted much interest because of their more attractive statistical properties. For example, unlike the $\ell_1$ penalty, which induces significant estimation bias for parameters with large absolute values (Zhang and Huang, 2008), nonconvex penalties such as the smoothly clipped absolute deviation (SCAD) penalty (Fan and Li, 2001) and minimax concave penalty (MCP) (Zhang, 2010a) can eliminate this estimation bias and attain more refined statistical rates of convergence. As another example of penalized $M$-estimators with nonconvex loss functions, we consider a semiparametric variant of the penalized least squares regression. Recall that a penalized least squares regression estimator can be formulated as

$$\begin{aligned}\widehat{\boldsymbol{\beta}}_\lambda &\in \underset{\boldsymbol{\beta}\in\mathbb{R}^d}{\operatorname{argmin}} \left\{ \frac{1}{2n}\|\mathbf{X}\boldsymbol{\beta} - \mathbf{y}\|_2^2 + \mathcal{P}_\lambda(\boldsymbol{\beta}) \right\} \\ &= \underset{\boldsymbol{\beta}\in\mathbb{R}^d}{\operatorname{argmin}} \left\{ \frac{1}{2}\left(1, -\boldsymbol{\beta}^T\right) \widehat{\mathbf{S}} \left(1, -\boldsymbol{\beta}^T\right)^T + \mathcal{P}_\lambda(\boldsymbol{\beta}) \right\},\end{aligned}$$

where $\widehat{\mathbf{S}} = (\mathbf{y}, \mathbf{X})^T(\mathbf{y}, \mathbf{X})/n$ is the sample covariance matrix of a random vector $(Y, \boldsymbol{X}^T)^T \in \mathbb{R}^{d+1}$. When the design matrix $\mathbf{X}$ contains heavy-tail data, we may resort to elliptical random design regression, which is a semiparametric extension of Gaussian random design regression. In detail, we replace the sample covariance matrix $\widehat{\mathbf{S}}$ with a possibly indefinite covariance matrix estimator $\widehat{\mathbf{K}}$ (to be defined in §2.2), which is more robust within the elliptical family. Since $\widehat{\mathbf{K}}$ does not guarantee to be positive semidefinite, the loss function $\mathcal{L}(\boldsymbol{\beta}) = (1, -\boldsymbol{\beta}^T)\widehat{\mathbf{K}}(1, -\boldsymbol{\beta}^T)^T/2$ could be nonconvex.

Though the global solutions of these nonconvex $M$-estimators enjoy nice statistical properties, it is in general computationally intractable to obtain the global solutions. Instead, a more realistic approach is to directly leverage standard optimization procedures to obtain a local solution $\widehat{\boldsymbol{\beta}}_\lambda$ that satisfies the first-order Karush-Kuhn-Tucker (KKT) condition

$$\mathbf{0} \in \partial \Big\{ \mathcal{L}(\widehat{\boldsymbol{\beta}}_\lambda) + \mathcal{P}_\lambda(\widehat{\boldsymbol{\beta}}_\lambda) \Big\}, \tag{1.2}$$

where $\partial(\cdot)$ denotes the subgradient operator.

In the context of least squares regression with nonconvex penalties, several numerical procedures have been proposed to find the local solutions, including local quadratic approximation (LQA) (Fan and Li, 2001), minorize-maximize (MM) algorithm (Hunter and Li, 2005), local linear approximation (LLA) (Zou and Li, 2008), concave convex procedure (CCCP) (Kim et al., 2008), and coordinate descent (Breheny and Huang, 2011; Mazumder et al., 2011). The theoretical properties of the local solutions obtained by these numerical procedures are in general unestablished. Only recently Zhang and Zhang (2012) showed that the gradient descent method initialized at a Lasso solution attains a unique local solution that has the same statistical properties as the global solution; Fan et al. (2014) proved that the LLA algorithm initialized with a Lasso solution attains a local solution with oracle statistical properties. The same conclusion was also obtained by Zhang (2010b); Zhang et al. (2013), where the LLA algorithm was referred to as multi-stage convex relaxation. In recent work, Wang et al. (2013) proposed a calibrated concave-convex procedure (CCCP) along with a high-dimensional BIC criterion that can achieve the oracle estimator. However, these works mainly focused on statistical recovery results, while the corresponding computational complexity results remain unclear. Also, they didn't consider nonconvex loss functions. In addition, their analysis relies on the assumption that all the computation (e.g., solving an optimization problem) can be carried out exactly, which is unrealistic in practice, since practical computational procedures can only attain finite numerical precision in finite time. Moreover, our method only requires the weakest possible minimum signal strength to attain the oracle estimator (Zhang and Zhang, 2012), while the procedures in Wang et al. (2013); Fan et al. (2014) rely on a stronger signal strength which is suboptimal. See §6 for a more detailed discussion.

In this paper, we propose an approximate regularization path following method for solving a general family of penalized $M$-estimators with possibly nonconvex loss or penalty functions. Our algorithm leverages the fast local convergence in the proximity of sparse solutions, which is also observed by Nesterov (2013); Wright et al. (2009); Agarwal et al. (2012); Xiao and Zhang (2013). More specifically, we consider a decreasing sequence of regularization parameters $\{\lambda_t\}_{t=0}^N$, where $\lambda_0$ corresponds to an all-zero solution, and $\lambda_N = \lambda_{\rm tgt}$ is the target regularization parameter that ensures the obtained estimator to achieve the optimal statistical rate of convergence. For each $\lambda_t$, we construct a sequence of local quadratic approximations of the loss function $\mathcal{L}(\boldsymbol{\beta})$, and utilize a variant of Nesterov's proximal-gradient method (Nesterov, 2013), which iterates over the updating step

$$\boldsymbol{\beta}_t^{k+1} \leftarrow \operatorname*{argmin}_{\boldsymbol{\beta}\in\mathbb{R}^d} \Big\{ \mathcal{L}(\boldsymbol{\beta}_t^k) + \nabla\mathcal{L}(\boldsymbol{\beta}_t^k)^T(\boldsymbol{\beta}-\boldsymbol{\beta}_t^k) + \frac{L_t^k}{2}\|\boldsymbol{\beta}-\boldsymbol{\beta}_t^k\|_2^2 + \mathcal{P}_{\lambda_t}(\boldsymbol{\beta}) \Big\}, \tag{1.3}$$

where $k = 1, 2, \ldots$. Here $\boldsymbol{\beta}_t^k$ and $L_t^k$ correspond to the $k$-th iteration of the proximal-gradient method for regularization parameter $\lambda_t$. Here $L_t^k$ is chosen by an adaptive line-search method, which will be specified in §3.2. Let $\widehat{\boldsymbol{\beta}}_{\lambda_t}$ be an exact local solution satisfying (1.2) with $\lambda = \lambda_t$. As illustrated in Figure 1, for each $\lambda_t$, our algorithm calculates an approximation $\widetilde{\boldsymbol{\beta}}_t$ of the exact local solution $\widehat{\boldsymbol{\beta}}_{\lambda_t}$ up to certain optimization precision. Such approximate local solution $\widetilde{\boldsymbol{\beta}}_t$ guarantees to be sparse, and therefore falls into the fast convergence region corresponding to $\lambda_{t+1}$. Consequently, the resulting procedure achieves a geometric rate of convergence within each path following stage, and therefore attains a global geometric rate of convergence for calculating the entire regularization path. Moreover, we establish the nonasymptotic statistical rates of convergence and oracle properties for all the approximate and exact local solutions along the full regularization path.

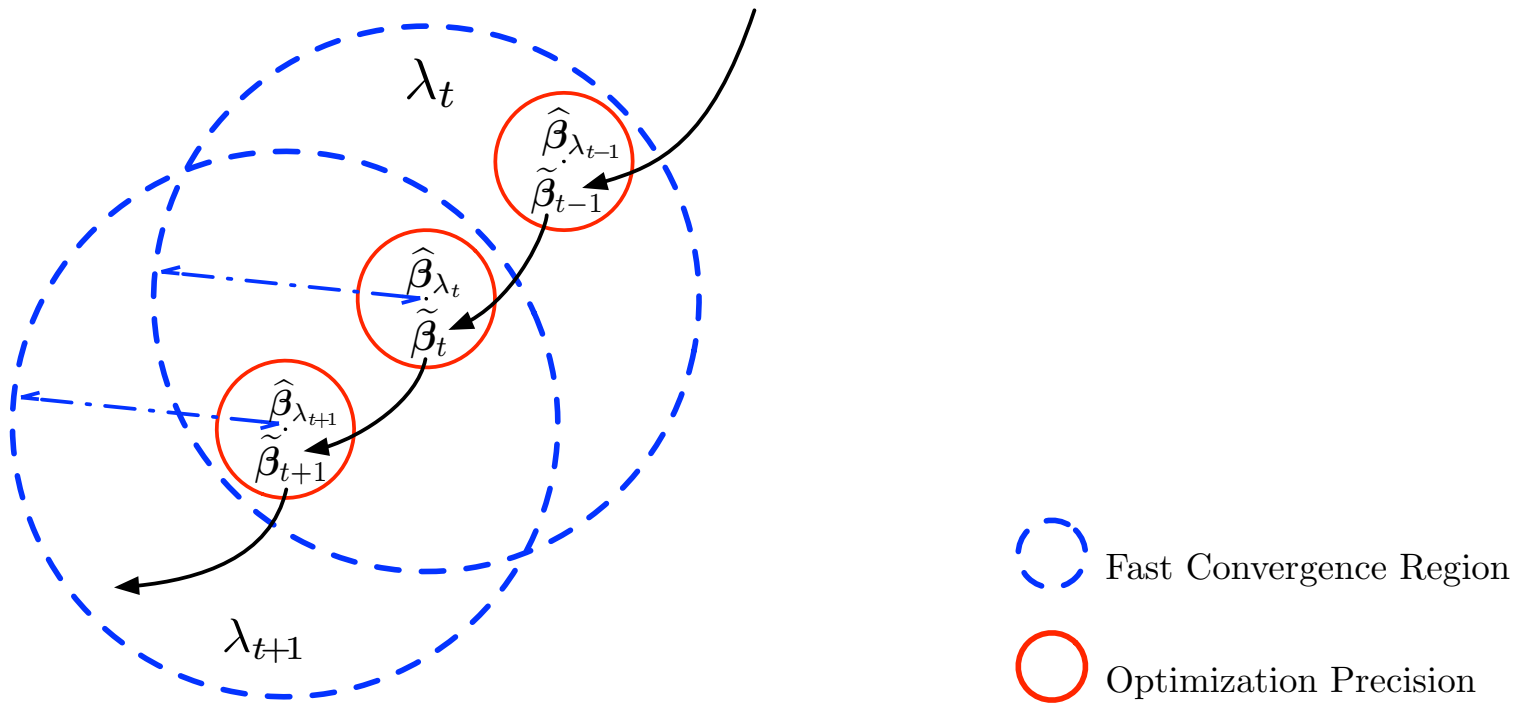

FIG 1. *For regularization parameter $\lambda_t$, $\widehat{\boldsymbol{\beta}}_{\lambda_t}$ is an exact local solution satisfying (1.2) with $\lambda = \lambda_t$. Within the $t$-th path following stage, our algorithm achieves an approximate local solution $\widetilde{\boldsymbol{\beta}}_t$, which approximates the exact local solution $\widehat{\boldsymbol{\beta}}_{\lambda_t}$ up to certain optimization precision. Our approximate path following algorithm ensures that $\widetilde{\boldsymbol{\beta}}_t$ is sparse, and therefore falls into the fast convergence region corresponding to regularization parameter $\lambda_{t+1}$.*

The idea of path following has been well-studied for sparse recovery problems (Efron et al., 2004; Hastie et al., 2005; Park and Hastie, 2007; Zhao and Yu, 2007; Rosset and Zhu, 2007; Friedman et al., 2010; Mazumder et al., 2011; Breheny and Huang, 2011; Xiao and Zhang, 2013; Mairal and Yu, 2012). Compared with these previous works, we consider a broader family of nonconvex $M$-estimators, including nonconvex penalty functions, such as SCAD and MCP, as well as nonconvex loss functions, such as semiparametric elliptical design loss. Moreover, we provide sharp computational and statistical analysis for all the approximate and exact local solutions attained by the proposed approximate path following method along the regularization path.

The contributions of this paper are two folds:

- Computationally, we propose an optimization algorithm that ensures a global geometric rate of convergence for nonconvex sparse learning problems. In detail, recall that $N$ is the total number of path following stages. Within the $N$-th path following stage, we denote by $\epsilon_{\mathrm{opt}}$ the desired optimization precision of the approximate local solution $\widetilde{\boldsymbol{\beta}}_N$. We need no more than a logarithmic number of the proximal-gradient update iterations defined in (1.3) to calculate the entire path:

$$\text{Total \# of proximal-gradient iterations} \le C \log\left(\frac{1}{\epsilon_{\mathrm{opt}}}\right),$$

  where $C > 0$ is a constant. This global geometric rate of convergence is optimal among all first-order methods, because it attains the lower bound for first-order methods on strongly convex and smooth objective function (Nesterov, 2004, Theorem 2.1.12), which is a subclass of the possibly nonconvex objective functions considered in this paper.
- Statistically, we prove that along the full regularization path, all the approximate local solutions obtained by our algorithm enjoy desirable statistical rates of convergence for estimating the true parameter vector $\boldsymbol{\beta}^*$. In detail, let $s^*$ be the number of nonzero entries of $\boldsymbol{\beta}^*$, the approximate local solution $\widetilde{\boldsymbol{\beta}}_t$'s satisfy

$$\|\widetilde{\boldsymbol{\beta}}_t - \boldsymbol{\beta}^*\|_2 \le C\lambda_t\sqrt{s^*}, \quad \text{for} \;\; t = 1, \dots, N \tag{1.4}$$

  with high probability. In particular, within the $N$-th path following stage, we have $\lambda_N = \lambda_{\mathrm{tgt}} = C'\sqrt{\log d/n}$. Here $C$ and $C'$ are positive constants that do not dependent on $d$ and $n$. In the $d \gg n$ regime, the final approximate local solution $\widetilde{\boldsymbol{\beta}}_N$ achieves the optimal statistical rate of convergence. Furthermore, we prove that, within the $t$-th path following stage, the iterative solution sequence $\{\boldsymbol{\beta}_t^k\}_{k=0}^{\infty}$ produced by (1.3) converges towards a unique exact local solution $\widehat{\boldsymbol{\beta}}_{\lambda_t}$, which enjoys a more refined oracle statistical property. More specifically, let $s_1^*$ be the number of "large" nonzero coefficients of $\boldsymbol{\beta}^*$ and $s_2^* = s^* - s_1^*$ be the number of "small" nonzero coefficients (detailed definitions of $s_1^*$ and $s_2^*$ are provided in Theorem 4.8), we have

$$\big\|\widehat{\boldsymbol{\beta}}_{\lambda_t} - \boldsymbol{\beta}^*\big\|_2 \le C\sqrt{\frac{s_1^*}{n}} + C'\sqrt{s_2^*}\lambda_t, \quad \text{for} \quad t = 1, \ldots, N \tag{1.5}$$

with high probability. In particular, for the final stage we have $\lambda_N = \lambda_{\text{tgt}} = C''\sqrt{\log d/n}$. Here $C$, $C'$ and $C''$ are positive constants. Note that the oracle statistical property in (1.5) is significantly sharper than the rate of convergence in (1.4), e.g., when $s^* = s_1^*$ and $t = N$, the right-hand side of (1.4) is of the order of $\sqrt{s^* \log d/n}$, while the right-hand side of (1.5) is of the order of $\sqrt{s^*/n}$. Moreover, we prove that when the absolute values of the nonzero coefficients of $\boldsymbol{\beta}^*$ are larger than $C'''\sqrt{\log d/n}$, $\widehat{\boldsymbol{\beta}}_{\lambda_t}$ exactly recovers the support of $\boldsymbol{\beta}^*$, i.e.,

$$\text{supp}\big(\widehat{\boldsymbol{\beta}}_{\lambda_t}\big) = \text{supp}(\boldsymbol{\beta}^*).$$

In summary, our joint analysis of the statistical and computational properties provides a theoretical characterization of the entire regularization path.

In independent work, Loh and Wainwright (2013) discussed similar problems. In detail, they provided sufficient conditions under which local optima have desired theoretical properties, and verified that the approximate local solution attained by the composite gradient descent method satisfies these conditions. Our work differs from theirs in three aspects:

(i) Our statistical recovery result in (1.4) covers all the approximate local solutions along the entire regularization path. They provided a similar statistical result, but only for the target regularization parameter, i.e., $\lambda_N = \lambda_{\text{tgt}}$ in (1.4).

(ii) As results of independent interest, we prove the oracle statistical properties of the exact local solutions along the regularization path, including the refined statistical rates of convergence in (1.5) and the guarantee of exact support recovery, while they didn't provide such results. Since the statistical result in (1.4) is also achievable using convex regularization, e.g., the $\ell_1$ penalty, these oracle properties are essential for justifying the benefits of using nonconvex penalty functions.

(iii) Our analysis technique is different from theirs. In detail, our statistical analysis is embedded in the analysis of the optimization procedure. In particular, we provide fine-grained analysis of the sparsity pattern of all the intermediate solutions obtained from the proximal-gradient iterations. In contrast, they provided characterizations of local solutions under a global restricted strongly convex/smoothness condition.

The rest of this paper is organized as follows. First we briefly introduce some useful notation. In §2 we introduce $M$-estimators with possibly noncon-vex loss and penalty functions. In §3 we present the proposed approximate regularization path following method. In §4 we present the main theoretical results on the computational efficiency and statistical accuracy of the proposed procedure. In §5 we prove the theoretical results in §4. In §6 we provide a detailed comparison between our method and the existing nonconvex procedures. Numerical results are presented in §7.

**Notation:** For $q \in [1, +\infty)$, the $\ell_q$ norm of $\boldsymbol{\beta} = (\beta_1, \ldots, \beta_d)^T \in \mathbb{R}^d$ is denoted by $\|\boldsymbol{\beta}\|_q = \left(\sum_{j=1}^d |\beta_j|^q\right)^{1/q}$. Specifically, we define $\|\boldsymbol{\beta}\|_\infty = \max_{1\le j\le d}\{|\beta_j|\}$ and $\|\boldsymbol{\beta}\|_0 = \operatorname{card}\{\operatorname{supp}(\boldsymbol{\beta})\}$, where $\operatorname{supp}(\boldsymbol{\beta}) = \{j : \beta_j \neq 0\}$ and $\operatorname{card}\{\cdot\}$ is the cardinality of a set. Correspondingly, we denote the $\ell_q$ ball $\{\boldsymbol{\beta} : \|\boldsymbol{\beta}\|_q \le R\}$ by $B_q(R)$. For a set $S$, we denote its cardinality by $|S|$ and its complement by $\bar{S}$. For $S, \bar{S} \subseteq \{1, \ldots, d\}$, we define $\boldsymbol{\beta}_S \in \mathbb{R}^d$ and $\boldsymbol{\beta}_{\bar{S}} \in \mathbb{R}^d$ as $(\boldsymbol{\beta}_S)_j = \mathbb{1}(j \in S)\cdot\beta_j$ and $(\boldsymbol{\beta}_{\bar{S}})_j = \mathbb{1}(j \notin S)\cdot\beta_j$ for $j = 1, \ldots, d$, where $\mathbb{1}(\cdot)$ is the indicator function. We denote all-zero matrices by $\mathbf{0}$. For notational simplicity, we use generic absolute constants $C, C', \ldots$, whose values may change from line to line.

Throughout, we denote the exact and approximate local solutions by $\widehat{\boldsymbol{\beta}}$ and $\widetilde{\boldsymbol{\beta}}$ respectively. We index $\widehat{\boldsymbol{\beta}}$ with the corresponding regulation parameter $\lambda$, e.g., $\widehat{\boldsymbol{\beta}}_\lambda$. For the proposed path following method, we use subscript $t$ to index the path following stages, e.g, the approximate local solution obtained within the $t$-th stage is denoted by $\widetilde{\boldsymbol{\beta}}_t$. Within the $t$-th stage, we index the proximal-gradient iterations with superscript $k$, e.g., $\boldsymbol{\beta}_t^k$.

## 2. Some Nonconvex Sparse Learning Problems.

Many theoretical results on penalized $M$-estimators rely on the condition that the loss and penalty functions are convex, since convexity makes both computational and statistical analysis convenient. However, the statistical performance of the estimator obtained from these convex formulations could be suboptimal in some settings. In the following, we introduce several nonconvex sparse learning problems as motivating examples.

### 2.1. *Nonconvex Penalty.*

Throughout this paper, we consider decomposable penalty functions

$$\mathcal{P}_\lambda(\boldsymbol{\beta}) = \sum_{j=1}^d p_\lambda(\beta_j),$$

e.g., the $\ell_1$ penalty $\lambda\|\boldsymbol{\beta}\|_1 = \sum_{i=1}^d \lambda|\beta_j|$. When the minimum of $|\beta_j^*| > 0$ is not close to zero, the $\ell_1$ penalty introduces large bias in parameter estimation. To remedy this effect, Fan and Li (2001) proposed the SCAD penalty

$$p_\lambda(\beta_j) = \lambda \int_0^{|\beta_j|} \left\{ \mathbb{1}(z \le \lambda) + \frac{(a\lambda - z)_+}{(a-1)\lambda} \mathbb{1}(z > \lambda) \right\} \mathrm{d}z, \quad a > 2, \tag{2.1}$$

and Zhang (2010a) proposed the MCP penalty

$$p_\lambda(\beta_j) = \lambda \int_0^{|\beta_j|} \left(1 - \frac{z}{\lambda b}\right)_+ \mathrm{d}z, \quad b > 0. \tag{2.2}$$

See Zhang and Zhang (2012) for a detailed survey. These nonconvex penalty functions are illustrated in Figure 2(a). In fact, these nonconvex penalties can be formulated as the sum of the $\ell_1$ penalty and a concave part

$$p_\lambda(\beta_j) = \lambda|\beta_j| + q_\lambda(\beta_j). \tag{2.3}$$

The concave components $q_\lambda(\beta_j)$ of SCAD and MCP are illustrated in Figure 2(b), while the corresponding derivatives $q'_\lambda(\beta_j)$ are illustrated in Figure 2(c). See §A.1 of the supplementary material (Wang et al., 2014b) for the detailed analytical forms of $p_\lambda(\beta_j)$ and $q_\lambda(\beta_j)$ for SCAD and MCP.

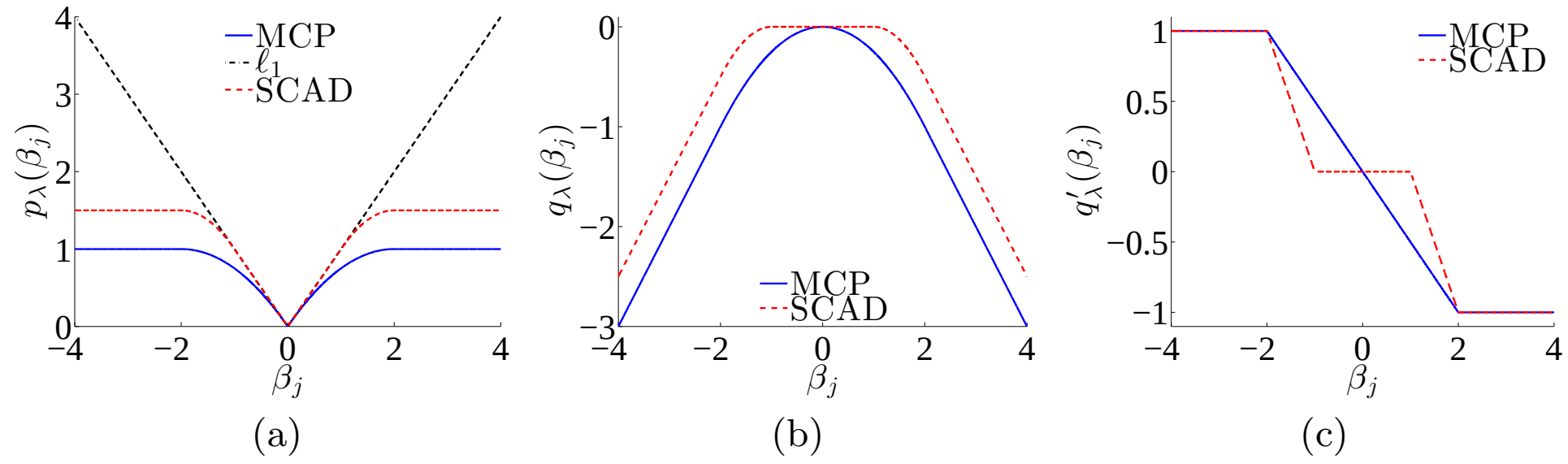

FIG 2. *An illustration of nonconvex penalties: (a) Plots of $p_\lambda(\beta_j)$ for MCP, $\ell_1$, and SCAD; (b) Plots of $q_\lambda(\beta_j)$ for MCP and SCAD; (c) Plots of $q'_\lambda(\beta_j)$ for MCP and SCAD. Here $p_\lambda(\beta_j)$ is the penalty function evaluated at the $j$-th dimension of $\boldsymbol{\beta}$, $q_\lambda(\beta_j)$ is the concave component of $p_\lambda(\beta_j)$, and $q'_\lambda(\beta_j)$ is the derivative of $q_\lambda(\beta_j)$. Here we set $a = 2.1$ for SCAD, $b = 2$ for MCP, and $\lambda = 1$.*

In fact, our method and theory are not limited to SCAD and MCP. More generally, we only rely on the following regularity conditions on the concave component $q_\lambda(\beta_j)$:

### Regularity Conditions on Nonconvex Penalty

(a) $q'_\lambda(\beta_j)$ is monotone and Lipschitz continuous, i.e., for $\beta'_j > \beta_j$, there exist two constants $\zeta_- \geq 0$ and $\zeta_+ \geq 0$ such that

$$-\zeta_- \leq \frac{q'_\lambda(\beta'_j) - q'_\lambda(\beta_j)}{\beta'_j - \beta_j} \leq -\zeta_+ \leq 0;$$

(b) $q_\lambda(\beta_j)$ is symmetric, i.e., $q_\lambda(-\beta_j) = q_\lambda(\beta_j)$ for any $\beta_j$;
(c) $q_\lambda(\beta_j)$ and $q'_\lambda(\beta_j)$ pass through the origin, i.e., $q_\lambda(0) = q'_\lambda(0) = 0$;
(d) $q'_\lambda(\beta_j)$ is bounded, i.e., $|q'_\lambda(\beta_j)| \leq \lambda$ for any $\beta_j$;
(e) $q'_\lambda(\beta_j)$ has bounded difference with respect to $\lambda$: $\left|q'_{\lambda_1}(\beta_j) - q'_{\lambda_2}(\beta_j)\right| \leq$

$|\lambda_1 - \lambda_2|$ for any $\beta_j$.

In regularity condition (a), $\zeta_-$ and $\zeta_+$ are two parameters that control the concavity of $q_\lambda(\beta_j)$. Note that the second order derivative of a function characterizes its convexity/concavity. Taking $\beta'_j \to \beta_j$ in regularity condition (a), we have $q''_\lambda(\beta_j) \in [-\zeta_-, -\zeta_+]$ (ignoring those $\beta_j$'s where $q''_\lambda(\beta_j)$ doesn't exist), which suggests larger $\zeta_-$ and $\zeta_+$ allow $q_\lambda(\beta_j)$ to be more concave. For SCAD we have $\zeta_- = 1/(a-1)$ and $\zeta_+ = 0$, while for MCP we have $\zeta_- = 1/b$ and $\zeta_+ = 0$. In Figure 2(b) and Figure 2(c), we can verify that regularity conditions (a)-(d) hold for MCP and SCAD. In addition, we illustrate regularity condition (e) for MCP and SCAD in §A.2 of the supplementary material (Wang et al., 2014b).

From (2.3) we have $\mathcal{P}_\lambda(\boldsymbol{\beta}) = \sum_{j=1}^d p_\lambda(\beta_j) = \lambda\|\boldsymbol{\beta}\|_1 + \sum_{j=1}^d q_\lambda(\beta_j)$. For notational simplicity, we define

$$\mathcal{Q}_\lambda(\boldsymbol{\beta}) = \sum_{j=1}^d q_\lambda(\beta_j) = \mathcal{P}_\lambda(\boldsymbol{\beta}) - \lambda\|\boldsymbol{\beta}\|_1. \tag{2.4}$$

Hence $\mathcal{Q}_\lambda(\boldsymbol{\beta})$ denotes the decomposable concave component of the nonconvex penalty $\mathcal{P}_\lambda(\boldsymbol{\beta})$.

2.2. *Nonconvex Loss Function.* In this paper, we focus on an example of nonconvex loss function named semiparametric elliptical design regression. More specifically, we have $n$ pairs of observations $\mathbf{z}_1 = (y_1, \mathbf{x}_1^T)^T, \ldots, \mathbf{z}_n = (y_n, \mathbf{x}_n^T)^T$ of a random vector $\boldsymbol{Z} = (Y, \boldsymbol{X}^T)^T \in \mathbb{R}^{d+1}$ that follows a $(d+1)$-dimensional elliptical distribution. (See §A.3 of the supplementary material (Wang et al., 2014b) for a detailed introduction to elliptical distribution.) Then we can verify that $(Y|\boldsymbol{X} = \mathbf{x})$ follows a univariate elliptical distribution. If we assume that $\mathbb{E}(Y|\boldsymbol{X} = \mathbf{x}) = \mathbf{x}^T\boldsymbol{\beta}^*$, the population version of the semiparametric elliptical design regression estimator can be defined as

$$\begin{aligned}\breve{\boldsymbol{\beta}} &= \operatorname*{argmin}_{\boldsymbol{\beta}\in\mathbb{R}^d} \left\{ \frac{1}{2}\mathbb{E}_{\boldsymbol{X},Y}\left( \left(Y - \boldsymbol{X}^T\boldsymbol{\beta}\right)^2 \right) + \mathcal{P}_\lambda(\boldsymbol{\beta}) \right\} \\ &= \operatorname*{argmin}_{\boldsymbol{\beta}\in\mathbb{R}^d} \left\{ \frac{1}{2}\left(1, -\boldsymbol{\beta}^T\right)\boldsymbol{\Sigma}_{\boldsymbol{Z}}\left(1, -\boldsymbol{\beta}^T\right)^T + \mathcal{P}_\lambda(\boldsymbol{\beta}) \right\}.\end{aligned} \tag{2.5}$$

The above procedure is not practically implementable, since the population covariance matrix $\boldsymbol{\Sigma}_{\boldsymbol{Z}}$ in (2.5) is unknown. In practice, we need to estimate the population covariance matrix $\boldsymbol{\Sigma}_{\boldsymbol{Z}}$. For this purpose, we propose a rank-based covariance matrix estimator $\widehat{\mathbf{K}}_{\boldsymbol{Z}}$, which is calculated by a two-step procedure described in §A.4 of the supplementary material (Wang et al., 2014b). Since $\widehat{\mathbf{K}}_{\boldsymbol{Z}}$ is not necessarily positive semidefinite, the loss function in semiparametric elliptical design regression, i.e.,

$$\mathcal{L}(\boldsymbol{\beta}) = \frac{1}{2}\left(1, -\boldsymbol{\beta}^T\right) \widehat{\mathbf{K}}_{\boldsymbol{Z}} \left(1, -\boldsymbol{\beta}^T\right)^T, \tag{2.6}$$

is possibly nonconvex.

**3. Approximate Regularization Path Following Method.** Before we go into details, we first present the high level idea of approximate regularization path following. We then introduce the basic building block of our path following method — a proximal-gradient method tailored to nonconvex problems.

3.1. *Approximate Regularization Path Following.* Fast local geometric convergence in the proximity of sparse solutions has been observed by many authors (Wright et al., 2009; Blumensath and Davies, 2009; Agarwal et al., 2012; Xiao and Zhang, 2013). We exploit such fast local convergence under an approximate path framework to achieve fast global convergence.

**Initialization:** In (1.1), when the regularization parameter $\lambda$ is sufficiently large, the solution to sparse learning problems is an all-zero vector. Recall that any exact local solution $\widehat{\boldsymbol{\beta}}_\lambda$ satisfies the first-order optimality condition, $\mathbf{0} \in \partial\left\{\mathcal{L}\left(\widehat{\boldsymbol{\beta}}_\lambda\right) + \mathcal{P}_\lambda\left(\widehat{\boldsymbol{\beta}}_\lambda\right)\right\}$. Since the nonconvex penalty $\mathcal{P}_\lambda(\boldsymbol{\beta})$ can be formulated as $\mathcal{P}_\lambda(\boldsymbol{\beta}) = \mathcal{Q}_\lambda(\boldsymbol{\beta}) + \lambda\|\boldsymbol{\beta}\|_1$, where $\mathcal{Q}_\lambda(\boldsymbol{\beta})$ is defined in (2.4), the first-order optimality condition implies there should exist some subgradient $\boldsymbol{\xi} \in \partial\|\widehat{\boldsymbol{\beta}}_\lambda\|_1$ such that

$$\mathbf{0} = \nabla\mathcal{L}\left(\widehat{\boldsymbol{\beta}}_\lambda\right) + \nabla\mathcal{Q}_\lambda\left(\widehat{\boldsymbol{\beta}}_\lambda\right) + \lambda\boldsymbol{\xi}. \tag{3.1}$$

Let $\lambda$ be chosen such that $\widehat{\boldsymbol{\beta}}_\lambda = \mathbf{0}$. Then regularity condition (c) implies $\nabla\mathcal{Q}_\lambda(\mathbf{0}) = \mathbf{0}$. Meanwhile, since $\boldsymbol{\xi} \in \partial\|\mathbf{0}\|_1$, we have $\|\boldsymbol{\xi}\|_\infty \leq 1$, which implies $\|\nabla\mathcal{L}(\mathbf{0})\|_\infty \leq \lambda$ in (3.1). Hence, $\lambda_0 = \|\nabla\mathcal{L}(\mathbf{0})\|_\infty$ is the smallest regularization parameter such that any exact local solution $\widehat{\boldsymbol{\beta}}_\lambda$ to the minimization problem (1.1) is all-zero. We choose this $\lambda_0$ to be the initial parameter of our regularization path.

**Approximate Path Following:** Let $\lambda_{\mathrm{tgt}} \in (0, \lambda_0)$ be the target regularization parameter in (1.1). In practice, we may choose $\lambda_{\mathrm{tgt}}$ by cross-validation or the high-dimensional BIC criterion proposed by Wang et al. (2013). We consider a decreasing sequence of regularization parameters $\{\lambda_t\}_{t=0}^N$, where

$$\lambda_t = \eta^t\lambda_0 \quad (t = 0, \ldots, N), \qquad \lambda_N = \lambda_{\mathrm{tgt}}, \qquad \text{and} \quad \eta \in [0.9, 1). \tag{3.2}$$

Here $\eta$ is an absolute constant that doesn't scale with sample size $n$ and dimension $d$. In §4 and §5 we will prove that, $\eta \in [0.9, 1)$ ensures the global geometric rate of convergence. Consequently, since we have $\lambda_{\mathrm{tgt}} = \lambda_0\eta^N$ by

(3.2), the number of path following stages is

$$N = \frac{\log(\lambda_0/\lambda_{\text{tgt}})}{\log(\eta^{-1})}. \tag{3.3}$$

Without loss of generality, we assume that $\eta$ is properly chosen such that $N$ is an integer. We will show in §4 that, $\lambda_{\text{tgt}}$ scales with sample size $n$ and dimension $d$. Since $\eta$ is a constant, the number of stages $N$ also scales with $n$ and $d$. Within the $t$-th ($t = 1, \ldots, N$) path following stage, we aim to obtain a local solution to the minimization problem $\min_{\boldsymbol{\beta}}\big\{\mathcal{L}(\boldsymbol{\beta}) + \mathcal{P}_{\lambda_t}(\boldsymbol{\beta})\big\}$.

As shown in Lines 5-9 of Algorithm 1, within the $t$-th ($t = 1, \ldots, N-1$) path following stage, we employ a variant of proximal-gradient method (Algorithm 3) to obtain an approximate local solution $\widetilde{\boldsymbol{\beta}}_t$ for regularization parameter $\lambda_t = \eta^t\lambda_0$. To ensure that each path following stage enjoys a fast geometric rate of convergence, we propose an approximation path following strategy. More specifically, we use the approximate local solution $\widetilde{\boldsymbol{\beta}}_{t-1}$ obtained within the $(t-1)$-th path following stage to initialize the $t$-th stage (Line 8 and Line 12 of Algorithm 1). Recall that we need to adaptively search for the best $L_t^k$ ($k = 0, 1, \ldots$) in (1.3). To achieve computational efficiency, within the $(t-1)$-th path following stage, we store the chosen $L_{t-1}^k$ at the last proximal-gradient iteration of the $(t-1)$-th stage as $L_{t-1}$. Within the $t$-th stage we initialize the search for $L_t^0$ with $L_{t-1}$ (Line 8 and Line 12 of Algorithm 1), which will be explained in §3.2.

**Configuration of Optimization Precision:** We set the optimization precision $\epsilon_t$ for the $t$-th ($t = 1, \ldots, N-1$) stage to be $\lambda_t/4$ (Line 7 of Algorithm 1). Within the $N$-th path following stage where $\lambda_N = \lambda_{\text{tgt}}$ (Line 10), we solve up to high optimization precision $\epsilon_{\text{opt}} \ll \lambda_{\text{tgt}}/4$ (Line 11). The intuition behind this configuration of optimization precision is explained as follows:

- For $t = 1, \ldots, N-1$, recall the exact local solution $\widehat{\boldsymbol{\beta}}_{\lambda_t}$ is an estimator of the true parameter vector $\boldsymbol{\beta}^*$ corresponding to the regularization parameter $\lambda_t$. According to high-dimensional statistical theory, the statistical error of $\widehat{\boldsymbol{\beta}}_{\lambda_t}$ should be upper bounded by $C\lambda_t\sqrt{s^*}$ with high probability, where $s^* = \|\boldsymbol{\beta}^*\|_0$. In Lemma 5.1 we will prove that, if the optimization error of the approximate local solution $\widetilde{\boldsymbol{\beta}}_t$ is at most $\lambda_t/4$, then $\widetilde{\boldsymbol{\beta}}_t$ lies within a ball of radius $C'\lambda_t\sqrt{s^*}$ centered at $\boldsymbol{\beta}^*$ with high probability. That is to say, the approximate local solution $\widetilde{\boldsymbol{\beta}}_t$ has the same order of statistical error as the exact solution $\widehat{\boldsymbol{\beta}}_{\lambda_t}$, and therefore enjoys desired statistical recovery properties. In particular, in Theorem 5.5 we will prove that, $\widetilde{\boldsymbol{\beta}}_t$ is guaranteed to be sparse, and thus falls into the fast convergence region of the next path following stage.
- However, for $t = N$, we need to solve up to high optimization precision

**Algorithm 1** The approximate path following method, which solves for a decreasing sequence of regularization parameters $\{\lambda_t\}_{t=0}^N$. Within the $t$-th path following stage, we employ the proximal-gradient method illustrated in Algorithm 3 to achieve an approximate local solution $\widetilde{\boldsymbol{\beta}}_t$ for $\lambda_t$. This approximate local solution is then used to initialize the $(t+1)$-th stage.

1: $\{\widetilde{\boldsymbol{\beta}}_t\}_{t=1}^N \leftarrow$ Approximate-Path-Following$(\lambda_{\text{tgt}}, \epsilon_{\text{opt}})$
2: **input:** $\lambda_{\text{tgt}} > 0, \epsilon_{\text{opt}} > 0$ {Here we set $\epsilon_{\text{opt}} \ll \lambda_{\text{tgt}}/4$.}
3: **parameters:** $\eta \in [0.9, 1), R > 0,\ L_{\min} > 0,\ \lambda_0 = \|\nabla\mathcal{L}(\mathbf{0})\|_\infty$
{For logistic loss, we set $R \in (0, +\infty)$; For other loss functions, we set $R = +\infty$.}
{In practice, we set $L_{\min}$ to be a sufficiently small value, e.g., $10^{-6}$.}
4: **initialize:** $\widetilde{\boldsymbol{\beta}}_0 \leftarrow \mathbf{0}, L_0 \leftarrow L_{\min}, N \leftarrow \log(\lambda_0/\lambda_{\text{tgt}})/\log(\eta^{-1})$
5: **for** $t = 1, \ldots, N-1$ **do**
6: $\quad\lambda_t \leftarrow \eta^t\lambda_0$
7: $\quad\epsilon_t \leftarrow \lambda_t/4$
8: $\quad\{\widetilde{\boldsymbol{\beta}}_t, L_t\} \leftarrow$ Proximal-Gradient$\big(\lambda_t, \epsilon_t, \widetilde{\boldsymbol{\beta}}_{t-1}, L_{t-1}, R\big)$ as in Algorithm 3
9: **end for**
10: $\lambda_N \leftarrow \lambda_{\text{tgt}}$
11: $\epsilon_N \leftarrow \epsilon_{\text{opt}}$
12: $\{\widetilde{\boldsymbol{\beta}}_N, L_N\} \leftarrow$ Proximal-Gradient$\big(\lambda_N, \epsilon_N, \widetilde{\boldsymbol{\beta}}_{N-1}, L_{N-1}, R\big)$
13: **return** $\{\widetilde{\boldsymbol{\beta}}_t\}_{t=1}^N$

$\epsilon_{\text{opt}} \ll \lambda_{\text{tgt}}/4$. This is because, even though $\widetilde{\boldsymbol{\beta}}_t$ and $\widehat{\boldsymbol{\beta}}_{\lambda_t}$ both have statistical error of the order $\lambda_t\sqrt{s^*}$, in certain regimes (to be specified in Theorem 4.8), the exact local solution $\widehat{\boldsymbol{\beta}}_{\lambda_t}$ can achieve an improved recovery performance (as shown in (1.5)) due to the usage of nonconvex penalties. Therefore, within the final stage we need to obtain an approximate solution $\widetilde{\boldsymbol{\beta}}_N$ as close to the exact local solution $\widehat{\boldsymbol{\beta}}_{\lambda_{\text{tgt}}}$ as possible, so that $\widetilde{\boldsymbol{\beta}}_N$ has a sharper statistical rate of convergence.

In Algorithm 1, $R > 0$ (Line 3) is a parameter that determines the radius of the constraint used in the proximal-gradient method (Line 8 and Line 12). For least squares loss and semiparametric elliptical design loss, we don't need any constraint. Therefore, we set $R = +\infty$. However, for logistic loss we need to impose an $\ell_2$ constraint of radius $R \in (0, +\infty)$. Here $L_{\min}$ is a parameter used in the proximal-gradient method (Line 3 of Algorithm 3), which is often set to be a sufficiently small value in practice, e.g., $L_{\min} = 10^{-6}$. We will explain with details in §3.2.

3.2. *Proximal-Gradient Method for Nonconvex Problems.* Before we introduce our proximal-gradient method which is tailored to nonconvex problems, we first give a brief introduction to Nesterov's proximal-gradient method

(Nesterov, 2013), which solves the following convex optimization problem

$$\text{minimize } \phi_\lambda(\boldsymbol{\beta}), \quad \text{where } \phi_\lambda(\boldsymbol{\beta}) = \mathcal{L}(\boldsymbol{\beta}) + \mathcal{P}_\lambda(\boldsymbol{\beta}), \quad \boldsymbol{\beta} \in \Omega. \tag{3.4}$$

Here $\mathcal{L}(\boldsymbol{\beta})$ is convex and differentiable, $\mathcal{P}_\lambda(\boldsymbol{\beta})$ is convex but possibly nonsmooth, and $\Omega$ is a closed convex set.

Recall that $\boldsymbol{\beta}_t^k$ corresponds to the $k$-th iteration of the proximal-gradient method within the $t$-th path following stage. Nesterov's proximal-gradient method updates $\boldsymbol{\beta}_t^k$ to be the minimizer of the following local quadratic approximation of $\phi_{\lambda_t}(\boldsymbol{\beta})$ at $\boldsymbol{\beta}_t^{k-1}$

$$\begin{aligned}\psi_{L_t^k,\lambda_t}(\boldsymbol{\beta};\boldsymbol{\beta}_t^{k-1}) &= \mathcal{L}(\boldsymbol{\beta}_t^{k-1}) + \nabla\mathcal{L}(\boldsymbol{\beta}_t^{k-1})^T(\boldsymbol{\beta}-\boldsymbol{\beta}_t^{k-1}) \\ &\quad + \frac{L_t^k}{2}\|\boldsymbol{\beta}-\boldsymbol{\beta}_t^{k-1}\|_2^2 + \mathcal{P}_{\lambda_t}(\boldsymbol{\beta}),\end{aligned} \tag{3.5}$$

where $L_t^k > 0$ is chosen by line-search.

Nesterov's proximal-gradient method requires that both $\mathcal{L}(\boldsymbol{\beta})$ and $\mathcal{P}_\lambda(\boldsymbol{\beta})$ in (3.4) are convex. However, in the optimization problem (1.1) considered in this paper, $\mathcal{L}(\boldsymbol{\beta})$ and $\mathcal{P}_\lambda(\boldsymbol{\beta})$ may be no longer convex. In this case, directly plugging $\mathcal{L}(\boldsymbol{\beta})$ and $\mathcal{P}_\lambda(\boldsymbol{\beta})$ into Nesterov's proximal-gradient might lead to the phenomenon of bad local optima under a path following scheme, as observed by She (2009, 2012). To extend the proximal-gradient method to nonconvex settings, we adopt an alternative formulation of the objective function.

Recall that the nonconvex penalty can be decomposed as $\mathcal{P}_\lambda(\boldsymbol{\beta}) = \lambda\|\boldsymbol{\beta}\|_1 + \mathcal{Q}_\lambda(\boldsymbol{\beta})$, where $\mathcal{Q}_\lambda(\boldsymbol{\beta})$ is defined in (2.4). For notational simplicity, we denote $\mathcal{L}(\boldsymbol{\beta}) + \mathcal{Q}_\lambda(\boldsymbol{\beta})$ by $\widetilde{\mathcal{L}}_\lambda(\boldsymbol{\beta})$. Therefore, the objective function $\phi_\lambda(\boldsymbol{\beta}) = \mathcal{L}(\boldsymbol{\beta}) + \mathcal{P}_\lambda(\boldsymbol{\beta}) = \mathcal{L}(\boldsymbol{\beta}) + \mathcal{Q}_\lambda(\boldsymbol{\beta}) + \lambda\|\boldsymbol{\beta}\|_1$ can be reformulated as

$$\phi_\lambda(\boldsymbol{\beta}) = \widetilde{\mathcal{L}}_\lambda(\boldsymbol{\beta}) + \lambda\|\boldsymbol{\beta}\|_1, \tag{3.6}$$

where we can view $\widetilde{\mathcal{L}}_\lambda(\boldsymbol{\beta})$ as a surrogate loss function and $\lambda\|\boldsymbol{\beta}\|_1$ as a new penalty function. This reformulation ensures the convexity of the new penalty function. Moreover, in Lemma 5.1 we will prove that, the surrogate loss function $\widetilde{\mathcal{L}}_\lambda(\boldsymbol{\beta})$ is actually strongly convex on a sparse set. Correspondingly, we modify Nesterov's proximal-gradient method to minimize the local quadratic approximation defined as

$$\begin{aligned}\psi_{L_t^k,\lambda_t}(\boldsymbol{\beta};\boldsymbol{\beta}_t^{k-1}) &= \widetilde{\mathcal{L}}_{\lambda_t}(\boldsymbol{\beta}_t^{k-1}) + \nabla\widetilde{\mathcal{L}}_{\lambda_t}(\boldsymbol{\beta}_t^{k-1})^T(\boldsymbol{\beta}-\boldsymbol{\beta}_t^{k-1}) \\ &\quad + \frac{L_t^k}{2}\|\boldsymbol{\beta}-\boldsymbol{\beta}_t^{k-1}\|_2^2 + \lambda_t\|\boldsymbol{\beta}\|_1.\end{aligned} \tag{3.7}$$

Note that, unlike (3.5), we use a quadratic approximation to the surrogate loss function $\widetilde{\mathcal{L}}_{\lambda_t}(\boldsymbol{\beta})$ in (3.7), instead of the original loss function $\mathcal{L}(\boldsymbol{\beta})$. At the $k$-th iteration of the proximal-gradient method, we update $\boldsymbol{\beta}_t^k$ to be the minimizer of the quadratic approximation defined in (3.7), i.e.,

$$\boldsymbol{\beta}_t^k \leftarrow \underset{\boldsymbol{\beta}\in\Omega}{\operatorname{argmin}} \left\{ \psi_{L_t^k,\lambda_t}\big(\boldsymbol{\beta};\boldsymbol{\beta}_t^{k-1}\big) \right\}. \tag{3.8}$$

Now we specify the constraint set $\Omega$ in (3.8). For $\mathcal{L}(\boldsymbol{\beta})$ being least squares or semiparametric elliptical design loss, we set $\Omega = \mathbb{R}^d$. For logistic loss, we set $\Omega = B_2(R)$ with $R \in (0, +\infty)$, where $B_2(R)$ is a centered $\ell_2$ ball of radius $R$. In Lemma 5.1 we will show that, in the setting of logistic loss, the boundedness of $\|\boldsymbol{\beta}_t^k\|_2$'s is essential for establishing the strong convexity of the surrogate loss function $\widetilde{\mathcal{L}}_{\lambda_t}(\boldsymbol{\beta})$ along the full regularization path. To unify the notation, we consider $\Omega = B_2(R)$ throughout — when the constraint set $\Omega = \mathbb{R}^d$, we set $R = +\infty$. Correspondingly, we denote (3.8) by

$$\boldsymbol{\beta}_t^k \leftarrow \mathcal{T}_{L_t^k,\lambda_t}\big(\boldsymbol{\beta}_t^{k-1}; R\big). \tag{3.9}$$

In the sequel, we provide the closed-form expression of update scheme (3.9):

**Update Scheme of Proximal-Gradient Method for Nonconvex Problems**

- For $\Omega = \mathbb{R}^d$, i.e., $R = +\infty$, $\mathcal{T}_{L_t^k,\lambda_t}\big(\boldsymbol{\beta}_t^{k-1}; +\infty\big)$ is a soft-thresholding operator taking the form of

$$\Big(\mathcal{T}_{L_t^k,\lambda_t}\big(\boldsymbol{\beta}_t^{k-1}; +\infty\big)\Big)_j = \begin{cases} 0 & \text{if } |\bar{\beta}_j| \le \lambda_t/L_t^k, \\ \operatorname{sign}(\bar{\beta}_j)\big(|\bar{\beta}_j| - \lambda_t/L_t^k\big) & \text{if } |\bar{\beta}_j| > \lambda_t/L_t^k, \end{cases} \tag{3.10}$$

  for $j = 1, \ldots, d$, where

$$\begin{aligned} \bar{\boldsymbol{\beta}} &= \boldsymbol{\beta}_t^{k-1} - \frac{1}{L_t^k}\nabla\widetilde{\mathcal{L}}_{\lambda_t}\big(\boldsymbol{\beta}_t^{k-1}\big) \\ &= \boldsymbol{\beta}_t^{k-1} - \frac{1}{L_t^k}\Big(\nabla\mathcal{L}\big(\boldsymbol{\beta}_t^{k-1}\big) + \nabla\mathcal{Q}_{\lambda_t}\big(\boldsymbol{\beta}_t^{k-1}\big)\Big), \end{aligned} \tag{3.11}$$

  and $\bar{\beta}_j$ is the $j$-th dimension of $\bar{\boldsymbol{\beta}}$.
- For $\Omega = B_2(R)$ with $R \in (0, +\infty)$, $\mathcal{T}_{L_t^k,\lambda_t}\big(\boldsymbol{\beta}_t^{k-1}; R\big)$ can be obtained by projecting $\mathcal{T}_{L_t^k,\lambda_t}\big(\boldsymbol{\beta}_t^{k-1}; +\infty\big)$ defined in (3.10) onto $B_2(R)$, i.e.,

$$\mathcal{T}_{L_t^k,\lambda_t}\big(\boldsymbol{\beta}_t^{k-1}; R\big) = \begin{cases} \mathcal{T}_{L_t^k,\lambda_t}\big(\boldsymbol{\beta}_t^{k-1}; +\infty\big) & \text{if } \big\|\mathcal{T}_{L_t^k,\lambda_t}\big(\boldsymbol{\beta}_t^{k-1}; +\infty\big)\big\|_2 < R, \\ \dfrac{R \cdot \mathcal{T}_{L_t^k,\lambda_t}\big(\boldsymbol{\beta}_t^{k-1}; +\infty\big)}{\big\|\mathcal{T}_{L_t^k,\lambda_t}\big(\boldsymbol{\beta}_t^{k-1}; +\infty\big)\big\|_2} & \text{if } \big\|\mathcal{T}_{L_t^k,\lambda_t}\big(\boldsymbol{\beta}_t^{k-1}; +\infty\big)\big\|_2 \ge R. \end{cases} \tag{3.12}$$

See §B.2 of the supplementary material (Wang et al., 2014b) for a detailed derivation. In §B.1 of the supplementary material (Wang et al., 2014b), we provide the specific forms of $\nabla\mathcal{L}(\boldsymbol{\beta})$ and $\nabla\mathcal{Q}_{\lambda_t}(\boldsymbol{\beta})$ in (3.11) for the nonconvex problems discussed in §2.

**Line-Search Method:** Before we present the proposed proximal-gradient method in detail, we briefly introduce a line-search algorithm, which adaptively searches for the best quadratic coefficient $L_t^k$ of the local quadratic approximation (3.7). As shown in Lines 4-7 of Algorithm 2, the main idea of line-search is to iteratively increase $L_t^k$ by a factor of two and compute the corresponding $\boldsymbol{\beta}_t^k$, until the local approximation $\psi_{L_t^k,\lambda_t}\big(\boldsymbol{\beta}_t^k;\boldsymbol{\beta}_t^{k-1}\big)$ becomes a tight upper bound of the objective function $\phi_{\lambda_t}\big(\boldsymbol{\beta}_t^k\big)$. We will theoretically characterize the computational complexity of this line-search algorithm in Remark 4.6, and specify the range of $L_t^k$ in Theorem 5.5.

**Algorithm 2** The line-search method used to search for the best $L_t^k$ and compute the corresponding $\boldsymbol{\beta}_t^k$. Here $\phi_{\lambda_t}(\boldsymbol{\beta})$ is the objective function defined in (3.4), and $\psi_{L_t^k,\lambda_t}\big(\boldsymbol{\beta};\boldsymbol{\beta}_t^{k-1}\big)$ is the local quadratic approximation of $\phi_{\lambda_t}(\boldsymbol{\beta})$ defined in (3.7).

1: $\big\{\boldsymbol{\beta}_t^k, L_t^k\big\} \leftarrow \text{Line-Search}\big(\lambda_t, \boldsymbol{\beta}_t^{k-1}, L_{\text{init}}, R\big)$
2: **input:** $\lambda_t > 0, \boldsymbol{\beta}_t^{k-1} \in \mathbb{R}^d, L_{\text{init}} > 0, R > 0$
3: **initialize:** $L_t^k \leftarrow L_{\text{init}}$
4: **repeat**
5: $\quad\boldsymbol{\beta}_t^k \leftarrow \mathcal{T}_{L_t^k,\lambda_t}\big(\boldsymbol{\beta}_t^{k-1};R\big)$ as defined in (3.9)
6: $\quad$**if** $\phi_{\lambda_t}\big(\boldsymbol{\beta}_t^k\big) > \psi_{L_t^k,\lambda_t}\big(\boldsymbol{\beta}_t^k;\boldsymbol{\beta}_t^{k-1}\big)$ **then** $L_t^k \leftarrow 2L_t^k$
7: **until** $\phi_{\lambda_t}\big(\boldsymbol{\beta}_t^k\big) \le \psi_{L_t^k,\lambda_t}\big(\boldsymbol{\beta}_t^k;\boldsymbol{\beta}_t^{k-1}\big)$
8: **return** $\big\{\boldsymbol{\beta}_t^k, L_t^k\big\}$

**Stopping Criterion:** In the following, we introduce the stopping criterion of our proximal-gradient method. In other words, we specify the optimality conditions that should be satisfied by the approximate solution $\widetilde{\boldsymbol{\beta}}_t$ attained by our proximal-gradient method.

It is known that any exact local solution $\widehat{\boldsymbol{\beta}}_\lambda$ to the optimization problem

$$\text{minimize } \phi_\lambda(\boldsymbol{\beta}), \quad \text{where } \phi_\lambda(\boldsymbol{\beta}) = \widetilde{\mathcal{L}}_\lambda(\boldsymbol{\beta}) + \lambda\|\boldsymbol{\beta}\|_1, \;\; \boldsymbol{\beta}\in\Omega$$

satisfies the optimality condition, i.e, there exists some $\boldsymbol{\xi} \in \partial\big\|\widehat{\boldsymbol{\beta}}_\lambda\big\|_1$ such that

$$\big(\widehat{\boldsymbol{\beta}}_\lambda - \boldsymbol{\beta}\big)^T\Big(\nabla\widetilde{\mathcal{L}}_\lambda\big(\widehat{\boldsymbol{\beta}}_\lambda\big) + \lambda\boldsymbol{\xi}\Big) \le 0, \quad \text{for any } \boldsymbol{\beta}\in\Omega. \tag{3.13}$$

We can understand this optimality condition as follows: Locally at $\widehat{\boldsymbol{\beta}}_\lambda$, any feasible direction pointed at $\widehat{\boldsymbol{\beta}}_\lambda$, i.e., $\big(\widehat{\boldsymbol{\beta}}_\lambda - \boldsymbol{\beta}\big)$ where $\boldsymbol{\beta}\in\Omega$, leads to a decrease in the objective function value $\phi_\lambda(\boldsymbol{\beta})$, because as shown in (3.13), such direction forms an obtuse angle with the (sub)gradient vector of $\phi_\lambda(\boldsymbol{\beta})$ evaluated at $\widehat{\boldsymbol{\beta}}_\lambda$. If $\widehat{\boldsymbol{\beta}}_\lambda$ lies in the interior of $\Omega$, e.g., $\Omega = \mathbb{R}^d$, then (3.13) reduces to the well-known first-order KKT condition,[1]

$$\nabla \widetilde{\mathcal{L}}_\lambda\big(\widehat{\boldsymbol{\beta}}_\lambda\big) + \lambda \boldsymbol{\xi} = \mathbf{0}, \quad \text{where } \ \boldsymbol{\xi} \in \partial \big\|\widehat{\boldsymbol{\beta}}_\lambda\big\|_1. \tag{3.14}$$

Based on the optimality condition in (3.13), we measure the suboptimality of a $\boldsymbol{\beta} \in \Omega$ with

$$\omega_\lambda(\boldsymbol{\beta}) = \min_{\boldsymbol{\xi}' \in \partial \|\boldsymbol{\beta}\|_1} \max_{\boldsymbol{\beta}' \in \Omega} \left\{ \frac{(\boldsymbol{\beta} - \boldsymbol{\beta}')^T}{\|\boldsymbol{\beta} - \boldsymbol{\beta}'\|_1} \big(\nabla \widetilde{\mathcal{L}}_\lambda(\boldsymbol{\beta}) + \lambda \boldsymbol{\xi}'\big) \right\}. \tag{3.15}$$

To understand this measure of suboptimality, first note that, if $\boldsymbol{\beta}$ is an exact local solution, then we have $\omega_\lambda(\boldsymbol{\beta}) \leq 0$ by (3.13). Otherwise, if $\boldsymbol{\beta}$ is close to some exact local solution, then $\omega_\lambda(\boldsymbol{\beta})$ is some small positive value. When $\boldsymbol{\beta}$ lies in the interior of $\Omega$, then (3.15) reduces to a more straightforward

$$\omega_\lambda(\boldsymbol{\beta}) = \min_{\boldsymbol{\xi}' \in \partial \|\boldsymbol{\beta}\|_1} \Big\{ \big\|\nabla \widetilde{\mathcal{L}}_\lambda(\boldsymbol{\beta}) + \lambda \boldsymbol{\xi}'\big\|_\infty \Big\}. \tag{3.16}$$

Because for any fixed $\boldsymbol{v} \in \mathbb{R}^d$, we have $(\boldsymbol{\beta} + C\boldsymbol{v}) \in \Omega$ for $C > 0$ sufficiently small. Setting $\boldsymbol{\beta}$ to be this value in (3.15), we have

$$\omega_\lambda(\boldsymbol{\beta}) = \min_{\boldsymbol{\xi}' \in \partial \|\boldsymbol{\beta}\|_1} \max_{\boldsymbol{v} \in \mathbb{R}^d} \left\{ \frac{\boldsymbol{v}^T}{\|\boldsymbol{v}\|_1} \big(\nabla \widetilde{\mathcal{L}}_\lambda(\boldsymbol{\beta}) + \lambda \boldsymbol{\xi}'\big) \right\} = \min_{\boldsymbol{\xi}' \in \partial \|\boldsymbol{\beta}\|_1} \Big\{ \big\|\nabla \widetilde{\mathcal{L}}_\lambda(\boldsymbol{\beta}) + \lambda \boldsymbol{\xi}'\big\|_\infty \Big\},$$

where the second equality follows from the duality between $\ell_1$ and $\ell_\infty$ norm.

Equipped with the suboptimality measure $\omega_\lambda(\boldsymbol{\beta})$ defined in (3.15), now we can define the stopping criterion of our proximal-gradient method to be $\omega_{\lambda_t}\big(\boldsymbol{\beta}_t^k\big) \leq \epsilon_t$, where $\epsilon_t > 0$ is the desired optimization precision within the $t$-th path following stage (Line 9 of Algorithm 3). Therefore, the proximal-gradient method achieves an approximate local solution $\widetilde{\boldsymbol{\beta}}_t$ with suboptimality $\epsilon_t$. Recall that within the $t$-th path following stage ($t = 1, \ldots, N-1$), we set $\epsilon_t$ to be $\lambda_t/4$ (Line 7 of Algorithm 1), while within the $N$-th path following stage, we set $\epsilon_t = \epsilon_{\text{opt}} \ll \lambda_{\text{tgt}}/4$ (Line 11 of Algorithm 1).

**Proposed Proximal-Gradient Method:** We are now ready to present the proposed proximal-gradient method in detail. Recall that, within the $t$-th stage of our path following algorithm, we employ the proximal-gradient method to obtain the approximate local solution $\widetilde{\boldsymbol{\beta}}_t$ (Line 8 and Line 12 of Algorithm 1). As shown in Line 8 of Algorithm 3, at the $k$-th iteration of our proximal-gradient method, we employ the line-search method (Algorithm 2) to search for the best $L_t^k$ and calculate the corresponding $\boldsymbol{\beta}_t^k$.

At the $k$-th iteration of the proximal-gradient method, we set the initial value $L_{\text{init}}$ of line-search to be $\max\big\{L_{\min}, L_t^{k-1}/2\big\}$ (Line 7 of Algorithm 3),

[1] Because given that $\widehat{\boldsymbol{\beta}}_\lambda$ lies in the interior of $\Omega$, we have $\big(\widehat{\boldsymbol{\beta}}_\lambda + C\boldsymbol{v}\big) \in \Omega$ and $\big(\widehat{\boldsymbol{\beta}}_\lambda - C\boldsymbol{v}\big) \in \Omega$ for any fixed $\boldsymbol{v} \in \mathbb{R}^d$ and $C > 0$ sufficiently small. Setting $\boldsymbol{\beta}$ in (3.13) to be these two values, we obtain $\boldsymbol{v}^T\big(\nabla \widetilde{\mathcal{L}}_\lambda\big(\widehat{\boldsymbol{\beta}}_\lambda\big) + \boldsymbol{\xi}\big) = 0$, which implies (3.14) since $\boldsymbol{v}$ is arbitrarily chosen.

**Algorithm 3** The proximal-gradient method for nonconvex problems, which iteratively leverages the line-search method illustrated in Algorithm 2 at each iteration.

1: $\{\widetilde{\boldsymbol{\beta}}_t, L_t\} \leftarrow$ Proximal-Gradient$\big(\lambda_t, \epsilon_t, \boldsymbol{\beta}_t^0, L_t^0, R\big)$
2: **input:** $\lambda_t > 0, \epsilon_t > 0, \boldsymbol{\beta}_t^0 \in \mathbb{R}^d, L_t^0 > 0, R > 0$
3: **parameter:** $L_{\min} > 0$
4: **initialize:** $k \leftarrow 0$
5: **repeat**
6: $\quad k \leftarrow k + 1$
7: $\quad L_{\text{init}} \leftarrow \max\big\{L_{\min}, L_t^{k-1}/2\big\}$
8: $\quad \boldsymbol{\beta}_t^k, L_t^k \leftarrow$ Line-Search$\big(\lambda_t, \boldsymbol{\beta}_t^{k-1}, L_{\text{init}}, R\big)$ as in Algorithm 2
9: **until** $\omega_{\lambda_t}\big(\boldsymbol{\beta}_t^k\big) \le \epsilon_t$ as defined in (3.15)
10: $\widetilde{\boldsymbol{\beta}}_t \leftarrow \boldsymbol{\beta}_t^k$
11: $L_t \leftarrow L_t^k$
12: **return** $\{\widetilde{\boldsymbol{\beta}}_t, L_t\}$

where $L_{\min} > 0$ is used to prevent $L_{\text{init}}$ from being too small. In practice, $L_{\min}$ is often set to be a sufficiently small value, e.g., $L_{\min} = 10^{-6}$. The intuition behind such initialization can be understood as follows: As shown in (3.7), $L_t^{k-1}$ and $L_t^k$ are the quadratic coefficients of the local quadratic approximations of the objective function at $\boldsymbol{\beta}_t^{k-2}$ and $\boldsymbol{\beta}_t^{k-1}$ respectively. Intuitively speaking, $\boldsymbol{\beta}_t^{k-2}$ and $\boldsymbol{\beta}_t^{k-1}$ are close to each other, which implies that $L_t^{k-1}$ is a good guess for $L_t^k$. Hence, we can initialize the line-search method for $L_t^k$ with a value slightly smaller than $L_t^{k-1}$, e.g., $L_t^{k-1}/2$.

When the stopping criterion $\omega_{\lambda_t}\big(\boldsymbol{\beta}_t^k\big) \le \epsilon_t$ is satisfied (Line 9 of Algorithm 3), the proximal-gradient method stops and outputs the approximate local solution $\widetilde{\boldsymbol{\beta}}_t = \boldsymbol{\beta}_t^k$ (Line 10 of Algorithm 3). We also keep track of $L_t = L_t^k$ to accelerate the line-search procedure within the next path following stage.

## 4. Theoretical Results.

We establish theoretical results on the iteration complexity and statistical performance of our approximate regularization path following method for nonconvex learning problems.

### 4.1. *Assumptions.*

We first list the required assumptions. The first assumption is about the relationship between $\lambda_{\text{tgt}}$ and $\|\nabla\mathcal{L}(\boldsymbol{\beta}^*)\|_\infty$.

**Assumption 4.1.** For least squares loss and logistic loss, we set $\lambda_{\text{tgt}} = C\sqrt{\log d/n}$. Meanwhile, for semiparametric elliptical design loss, we set $\lambda_{\text{tgt}} = C'\|\boldsymbol{\beta}^*\|_1\sqrt{\log d/n}$. We assume

$$\|\nabla\mathcal{L}(\boldsymbol{\beta}^*)\|_\infty \le \lambda_{\text{tgt}}/8. \tag{4.1}$$

Assumption 4.1 is a common condition that $\lambda_{\text{tgt}}$ should be large enough to dominate the noise. For instance, for least squares loss we have

$$\nabla\mathcal{L}(\boldsymbol{\beta}^*) = \frac{1}{n}\mathbf{X}^T\big(\mathbf{X}\boldsymbol{\beta}^* - \mathbf{y}\big),$$

where $\mathbf{X}\boldsymbol{\beta}^* - \mathbf{y}$ is in fact the noise vector. In Lemma C.1 in §C.1 of the supplementary material (Wang et al., 2014b) we will show that, for least squares loss and logistic loss, we have that $\|\nabla\mathcal{L}(\boldsymbol{\beta}^*)\|_\infty \le C\sqrt{\log d/n}$ holds with high probability. Similarly, in Lemma C.2 in §C.1 of the supplementary material (Wang et al., 2014b) we will prove that, for semiparametric elliptical design loss, $\|\nabla\mathcal{L}(\boldsymbol{\beta}^*)\|_\infty \le C'\|\boldsymbol{\beta}^*\|_1\sqrt{\log d/n}$ holds with high probability. Thus, our assumption on $\lambda_{\text{tgt}}$ and $\|\nabla\mathcal{L}(\boldsymbol{\beta}^*)\|_\infty$ holds with high probability.

In the sequel, we lay out another assumption on the sparse eigenvalues of $\nabla^2\mathcal{L}(\boldsymbol{\beta})$, which are defined as follows.

**Definition 4.2** (Sparse Eigenvalues)**.** Let $s$ be a positive integer. The largest and smallest $s$-sparse eigenvalues of the Hessian matrix $\nabla^2\mathcal{L}(\boldsymbol{\beta})$ are

$$\rho_+\big(\nabla^2\mathcal{L}, s\big) = \sup\Big\{\boldsymbol{v}^T\nabla^2\mathcal{L}(\boldsymbol{\beta})\boldsymbol{v} : \|\boldsymbol{v}\|_0 \le s,\ \|\boldsymbol{v}\|_2 = 1,\ \boldsymbol{\beta}\in\mathbb{R}^d\Big\},$$
$$\rho_-\big(\nabla^2\mathcal{L}, s\big) = \inf\Big\{\boldsymbol{v}^T\nabla^2\mathcal{L}(\boldsymbol{\beta})\boldsymbol{v} : \|\boldsymbol{v}\|_0 \le s,\ \|\boldsymbol{v}\|_2 = 1,\ \boldsymbol{\beta}\in\mathbb{R}^d\Big\}.$$

For least squares loss and semiparametric elliptical design loss, $\nabla^2\mathcal{L}(\boldsymbol{\beta})$ doesn't depend on $\boldsymbol{\beta}$. However, for logistic loss we have

$$\nabla^2\mathcal{L}(\boldsymbol{\beta}) = \frac{1}{n}\sum_{i=1}^n \mathbf{x}_i\mathbf{x}_i^T \cdot \frac{1}{1+\exp(-\mathbf{x}_i^T\boldsymbol{\beta})}\cdot\frac{1}{1+\exp(\mathbf{x}_i^T\boldsymbol{\beta})}, \tag{4.2}$$

which depends on $\boldsymbol{\beta}$. Note in Definition 4.2, the smallest $s$-sparse eigenvalue $\rho_-\big(\nabla^2\mathcal{L}, s\big)$ is obtained by taking infimum over all $\boldsymbol{\beta}\in\mathbb{R}^d$. Consequently, for logistic loss, $\rho_-\big(\nabla^2\mathcal{L}, s\big)$ is always zero, because in (4.2) we can take $\boldsymbol{\beta}$ such that $|\mathbf{x}_i^T\boldsymbol{\beta}| \to +\infty$ for all nonzero $\mathbf{x}_i$'s, which implies that $\nabla^2\mathcal{L}(\boldsymbol{\beta})$ goes to an all-zero matrix. To avoid this degenerate case, for logistic loss we define the sparse eigenvalues by taking infimum/supremum over all $\boldsymbol{\beta}$ with $\|\boldsymbol{\beta}\|_2$ bounded instead of over all $\boldsymbol{\beta}\in\mathbb{R}^d$.

**Definition 4.3** (Sparse Eigenvalues for Logistic Loss)**.** Let $s$ be a positive integer. For logistic loss, we define the largest and smallest $s$-sparse eigenvalues of $\nabla^2\mathcal{L}(\boldsymbol{\beta})$ to be

$$\rho_+\big(\nabla^2\mathcal{L}, s, R\big) = \sup\big\{\boldsymbol{v}^T\nabla^2\mathcal{L}(\boldsymbol{\beta})\boldsymbol{v} : \|\boldsymbol{v}\|_0 \le s,\ \|\boldsymbol{v}\|_2 = 1,\ \|\boldsymbol{\beta}\|_2 \le R\big\},$$
$$\rho_-\big(\nabla^2\mathcal{L}, s, R\big) = \inf\big\{\boldsymbol{v}^T\nabla^2\mathcal{L}(\boldsymbol{\beta})\boldsymbol{v} : \|\boldsymbol{v}\|_0 \le s,\ \|\boldsymbol{v}\|_2 = 1,\ \|\boldsymbol{\beta}\|_2 \le R\big\},$$

where $R\in(0,+\infty)$ is an absolute constant such that $\|\boldsymbol{\beta}^*\|_2 \le R$.

In Definition 4.3, we implicitly assume that $\|\boldsymbol{\beta}^*\|_2$ is upper bounded by some known absolute constant. Although it seems rather restrictive, this assumption is essential for logistic loss. Otherwise, $\nabla^2\mathcal{L}(\boldsymbol{\beta}^*)$ may go to an all-zero matrix when $\|\boldsymbol{\beta}^*\|_2 \to +\infty$. In this case, the curvature of the objective function at $\boldsymbol{\beta}^*$ is zero, a consistent estimation of $\boldsymbol{\beta}^*$ is impossible. Although such assumption is necessary for theoretical purposes, we require no prior knowledge about the exact value of $\|\boldsymbol{\beta}^*\|_2$ in practice, since we can always set $R$ to be a sufficiently large constant in our algorithm (Line 3 of Algorithm 1). To unify the later analysis for different loss functions, we omit the extra term $R$ in Definition 4.3 unless its necessary.

Recall that we impose an $\ell_2$ constraint of radius $R$ for all the proximal-gradient iterations within each path following stage (Line 8 and Line 12 of Algorithm 1). Therefore, we have $\|\boldsymbol{\beta}_t^k\|_2 \le R$ during the whole iterative procedure (for least squares loss and semiparametric elliptical design loss, $R = +\infty$; for logistic loss, $R \in (0, +\infty)$). Now we are ready to present the assumption on the sparse eigenvalues of the Hessian matrix.

**Assumption 4.4.** Let $s^* = \|\boldsymbol{\beta}^*\|_0$. We assume:

- There exists an integer $\widetilde{s} > Cs^*$ such that
$$\rho_+\bigl(\nabla^2\mathcal{L}, s^* + 2\widetilde{s}\bigr) < +\infty, \quad \rho_-\bigl(\nabla^2\mathcal{L}, s^* + 2\widetilde{s}\bigr) > 0$$
are two absolute constants. The constant $C > 0$ is specified in (4.4).
- The concavity parameter $\zeta_-$ defined in regularity condition (a) satisfies
$$\zeta_- \le C'\rho_-\bigl(\nabla^2\mathcal{L}, s^* + 2\widetilde{s}\bigr) \tag{4.3}$$
with constant $C' < 1$.

In Assumption 4.4, the constant
$$C = 144\kappa^2 + 250\kappa, \tag{4.4}$$
where $\kappa$ is a condition number defined as
$$\kappa = \frac{\rho_+\bigl(\nabla^2\mathcal{L}, s^* + 2\widetilde{s}\bigr) - \zeta_+}{\rho_-\bigl(\nabla^2\mathcal{L}, s^* + 2\widetilde{s}\bigr) - \zeta_-}. \tag{4.5}$$
The constant in (4.4) is rather large for practical purposes. We could expect it to be much smaller if we manage to get smaller constants in the technical proof. However, we mainly focus on providing novel theoretical insights in this paper, without paying too much effort on optimizing constants.

Recall that regularity condition (a) implies $\zeta_+ \le \zeta_-$. Meanwhile, we have $\rho_-\bigl(\nabla^2\mathcal{L}, s^* + 2\widetilde{s}\bigr) \le \rho_+\bigl(\nabla^2\mathcal{L}, s^* + 2\widetilde{s}\bigr)$ by definition. Thus, (4.3) implies
$$\zeta_+ \le C'\rho_+\bigl(\nabla^2\mathcal{L}, s^* + 2\widetilde{s}\bigr), \tag{4.6}$$
where $C' < 1$ is the same constant as in (4.3). Therefore, we have $\kappa \in [1, +\infty)$. Restrictions (4.3) and (4.6) on the concavity parameters suggest that, the concavity of the concave component $\mathcal{Q}_\lambda(\boldsymbol{\beta}) = \sum_{j=1}^d q_\lambda(\beta_j)$ of the nonconvex penalty should not outweigh the convexity of the loss function on a sparse set. It is also worth noting the concavity parameters are independent from the regularization parameter, e.g., for MCP in (2.2), $b = 1/\zeta_-$ and $\lambda$ are two independent parameters. Thus, Assumption 4.4 doesn't depend on $\lambda$ at all.

Assumption 4.4 is closely related to the restricted isometry property (RIP) condition proposed by Candés and Tao (2005). Similar conditions have been studied by Bickel et al. (2009); Raskutti et al. (2010); Negahban et al. (2012); Zhang (2010b); Zhang et al. (2013); Xiao and Zhang (2013). In detail, for least squares loss, the RIP condition assumes there exists an integer $s$ and some constant $\delta \in (0, 1)$ such that

$$1 - \delta \le \rho_-\big(\nabla^2 \mathcal{L}, s\big) \le \rho_+\big(\nabla^2 \mathcal{L}, s\big) \le 1 + \delta. \tag{4.7}$$

Now we justify Assumption 4.4 for least squares loss with an example.

To show Assumption 4.4 is well defined, we assume the RIP condition in (4.7) holds with $s = 877s^*$ and $\delta = 0.01$. We set the concavity parameters of the nonconvex penalty in (a) to be $\zeta_+ = 0$ and $\zeta_- = \rho_-\big(\nabla^2 \mathcal{L}, s\big)/20$, e.g., for MCP defined in (2.2), we take $b = 1/\zeta_- = 20/\rho_-\big(\nabla^2 \mathcal{L}, s\big)$. In the following, we verify there exists an integer $\widetilde{s} = 438s^*$ that satisfies Assumption 4.4.

First, according to the RIP condition, we have

$$\begin{aligned} \rho_+\big(\nabla^2 \mathcal{L}, s^* + 2\widetilde{s}\big) = \rho_+\big(\nabla^2 \mathcal{L}, 877s^*\big) &= \rho_+\big(\nabla^2 \mathcal{L}, s\big) \\ &\le (1 + \delta) = 1.01 < +\infty, \end{aligned} \tag{4.8}$$

$$\begin{aligned} \rho_-\big(\nabla^2 \mathcal{L}, s^* + 2\widetilde{s}\big) = \rho_-\big(\nabla^2 \mathcal{L}, 877s^*\big) &= \rho_-\big(\nabla^2 \mathcal{L}, s\big) \\ &\ge (1 - \delta) = 0.99 > 0. \end{aligned} \tag{4.9}$$

Second, we calculate the value of $\widetilde{s}$ in detail. Since the condition number $\kappa$ defined in (4.5) satisfies

$$\begin{aligned} 1 \le \kappa = \frac{\rho_+\big(\nabla^2 \mathcal{L}, s^* + 2\widetilde{s}\big) - \zeta_+}{\rho_-\big(\nabla^2 \mathcal{L}, s^* + 2\widetilde{s}\big) - \zeta_-} &= \frac{\rho_+\big(\nabla^2 \mathcal{L}, s\big) - \zeta_+}{\rho_-\big(\nabla^2 \mathcal{L}, s\big) - \zeta_-} \\ &= \frac{20}{19} \cdot \frac{\rho_+\big(\nabla^2 \mathcal{L}, s\big)}{\rho_-\big(\nabla^2 \mathcal{L}, s\big)} \le \frac{20}{19} \cdot \frac{1 + \delta}{1 - \delta} < 1.08. \end{aligned}$$

We now verify that $\widetilde{s}$ satisfies $\widetilde{s} > Cs^*$ in Assumption 4.4, where $C$ is defined in (4.4). Plugging the range $1 \le \kappa < 1.08$ into the definition of $C$, we obtain $C = 144\kappa^2 + 250\kappa < 438$. Therefore, as long as the RIP condition holds with $s = 877s^*$ and $\delta = 0.01$, we can find an integer $\widetilde{s} = 438s^*$ that satisfies Assumption 4.4, which also implies Assumption 4.4 is a weaker assumption than the RIP condition. For least squares loss, the RIP condition is known to hold for a variety of design matrices with high probability, which implies that Assumption 4.4 also holds with high probability for these designs.

Furthermore, we will justify Assumption 4.4 for $\mathcal{L}(\boldsymbol{\beta})$ being semiparametric elliptical design loss and logistic loss in §C.2 of the supplementary material (Wang et al., 2014b). Also, in the discussion for logistic loss in §C.2, we prove that the assumption of restricted strong convexity/smoothness in Loh and Wainwright (2013) is stronger than our Assumption 4.4.

Hereafter, we use the shorthands

$$\rho_+ = \rho_+\big(\nabla^2\mathcal{L}, s^* + 2\widetilde{s}\big), \quad \rho_- = \rho_-\big(\nabla^2\mathcal{L}, s^* + 2\widetilde{s}\big) \tag{4.10}$$

for notational simplicity.

4.2. *Main Theorems.* We first provide the main results about the computational rate of convergence. We then establish the statistical properties of the local solutions obtained by our approximate path following method.

4.2.1. *Computational Theory.* The next theorem shows that the proposed approximate regularization path following method achieves a global geometric rate of convergence for calculating the entire regularization path, which is the optimal rate among all first-order optimization methods.

Recall that $\epsilon_{\rm opt} \ll \lambda_{\rm tgt}/4$ is the desired optimization precision of the final path following stage (Line 12 of Algorithm 1), and $N = \log(\lambda_0/\lambda_{\rm tgt})/\log(\eta^{-1})$ is the total number of approximate path following stages, where $\eta \in [0.9, 1)$ is an absolute constant. Meanwhile, remind that $\rho_- = \rho_-\big(\nabla^2\mathcal{L}, s^* + 2\widetilde{s}\big) > 0$ is the smallest sparse eigenvalue specified in Assumption 4.4; As defined in regularity condition (a), $\zeta_- > 0$ is the concavity parameter of the nonconvex penalty, which satisfies (4.3) in Assumption 4.4.

**Theorem 4.5** (Geometric Rate of Convergence)**.** Under Assumption 4.1 and Assumption 4.4, we have the following results:

1. **Geometric Rate of Convergence within the $t$-th Stage:** Within the $t$-th ($t = 1, \ldots, N$) path following stage (Lines 8 and 12 of Algorithm 1), the iterative sequence $\big\{\boldsymbol{\beta}_t^k\big\}_{k=0}^{\infty}$ produced by the proximal-gradient method (Algorithm 3) converges to a unique local solution $\widehat{\boldsymbol{\beta}}_{\lambda_t}$.
   - Within the $t$-th path following stage ($t = 1, \ldots, N-1$), the total number of proximal-gradient iterations (Lines 5-9 of Algorithm 3) is no more than $C' \log\big(4C\sqrt{s^*}\big)$.
   - Within the $N$-th stage ($\lambda_N = \lambda_{\rm tgt}$), the total number of proximal-gradient iterations is no more than $\max\big\{0, C' \log\big(C\lambda_{\rm tgt}\sqrt{s^*}/\epsilon_{\rm opt}\big)\big\}$.

   Here $s^* = \|\boldsymbol{\beta}^*\|_0$ and

$$C = 2\sqrt{21} \cdot \sqrt{\kappa}(1+\kappa), \quad C' = 2\Big/\log\left(\frac{1}{1 - 1/(8\kappa)}\right), \tag{4.11}$$

   where $\kappa \in [1, +\infty)$ is the condition number defined in (4.5).

2. **Geometric Rate of Convergence over the Full Path:** To compute the entire path, we need no more than

$$(4.12)\qquad \underbrace{(N-1)C'\log\left(4C\sqrt{s^*}\right)}_{1,\ldots,(N-1)-\text{th Stages}} + \underbrace{C'\log\left(\frac{C\lambda_{\text{tgt}}\sqrt{s^*}}{\epsilon_{\text{opt}}}\right)}_{N-\text{th Stage}}$$

proximal-gradient iterations, where $C$, $C'$ are specified in (4.11).

3. **Geometric Rate of Convergence of Objective Function Value:** Let $\widetilde{\boldsymbol{\beta}}_t$ be the approximate local solution obtained within the $t$-th stage.
   - For $t = 0, \ldots, N-1$, the value of the objective function decays exponentially towards the value at the final exact local solution $\widehat{\boldsymbol{\beta}}_{\lambda_{\text{tgt}}}$, i.e.,

     $$(4.13)\qquad \phi_{\lambda_{\text{tgt}}}\left(\widetilde{\boldsymbol{\beta}}_t\right) - \phi_{\lambda_{\text{tgt}}}\left(\widehat{\boldsymbol{\beta}}_{\lambda_{\text{tgt}}}\right) \le C\lambda_0^2 s^* \cdot \eta^{2(t+1)},$$

     where $C = 105/(\rho_- - \zeta_-)$.
   - For $t = N$, we have

     $$(4.14)\qquad \phi_{\lambda_{\text{tgt}}}\left(\widetilde{\boldsymbol{\beta}}_N\right) - \phi_{\lambda_{\text{tgt}}}\left(\widehat{\boldsymbol{\beta}}_{\lambda_{\text{tgt}}}\right) \le \left(C'\lambda_{\text{tgt}} s^*\right) \cdot \epsilon_{\text{opt}},$$

     where $C' = 21/(\rho_- - \zeta_-)$.

PROOF. See the next section for a detailed proof. □

Result 1 suggests that, within each path following stage, the proximal-gradient algorithm attains a geometric rate of convergence. More specifically, within the $t$-th ($t = 1, \ldots, N$) stage (Line 8 and Line 12 of Algorithm 1), we only need a logarithmic number of proximal-gradient update iterations (Lines 5-9 of Algorithm 3) to compute an approximate local solution $\widetilde{\boldsymbol{\beta}}_t$. Furthermore, within the $t$-th path following stage, the iterative sequence $\{\boldsymbol{\beta}_t^k\}_{k=0}^{\infty}$ produced by Algorithm 3 converges towards a unique local solution $\widehat{\boldsymbol{\beta}}_{\lambda_t}$. In Theorem 4.8, we will show that $\widehat{\boldsymbol{\beta}}_{\lambda_t}$ enjoys a more refined statistical rate of convergence due to the usage of nonconvex penalty.

Result 2 suggests that our approximate path following method attains a global geometric rate of convergence. From the perspective of high-dimensional statistics, the total number of stages $N$ scales with dimension $d$ and sample size $n$, because $N = \log(\lambda_0/\lambda_{\text{tgt}})/\log(\eta^{-1})$, where $\eta$ is an absolute constant. From the perspective of optimization, given dimension $d$ and sample size $n$, when the optimization precision $\epsilon_{\text{opt}}$ is sufficiently small such that in (4.12) the second term dominates its first term, then the total iteration complexity is $C\log(1/\epsilon_{\text{opt}})$. In other words, we only need to conduct a logarithmic number of proximal-gradient iterations to compute the full regularization path.

Recall that we measure the suboptimality of an approximate solution with $\omega_\lambda(\boldsymbol{\beta})$ defined in (3.15), which doesn't directly reflect the suboptimality of the objective function value. Hence we provide result 3 to characterize the decay of the objective gap $\phi_{\lambda_{\text{tgt}}}(\widetilde{\boldsymbol{\beta}}_t) - \phi_{\lambda_{\text{tgt}}}(\widehat{\boldsymbol{\beta}}_{\lambda_{\text{tgt}}})$. In detail, (4.13) illustrates the exponential decay of the objective gap along the regularization path, i.e., $t = 1, \dots, N-1$, while (4.14) suggests that, the final objective function value evaluated at $\widetilde{\boldsymbol{\beta}}_N$ is sufficiently close to the value at the exact local solution $\widehat{\boldsymbol{\beta}}_{\lambda_{\text{tgt}}}$, as long as the optimization precision $\epsilon_{\text{opt}}$ is sufficiently small.

Recall the largest sparse eigenvalue $\rho_+ = \rho_+(\nabla^2\mathcal{L}, s^* + 2\widetilde{s}) > 0$ is specified in Assumption 4.4; As defined in regularity condition (a), $\zeta_+ > 0$ is the concavity parameter of the nonconvex penalty, which satisfies (4.6) in Assumption 4.4; $L_{\min}$ is a parameter of Algorithm 3 (Line 3).

**Remark 4.6.** Nesterov (2013) proved that the total number of line-search steps (Lines 4-7 of Algorithm 2) within the $k$-th proximal-gradient iteration (Line 8 of Algorithm 3) is no more than

$$2(k+1) + \max\left\{0, \frac{\log(\rho_+ - \zeta_+) - \log L_{\min}}{\log 2}\right\}.$$

Piecing the above results together, we conclude that, the total number of line-search iterations (Lines 4-7 of Algorithm 2) required to compute the full regularization path is of the same order as (4.12).

4.2.2. *Statistical Theory.* We present two types of statistical results. Recall that $\widetilde{\boldsymbol{\beta}}_t$ is the approximate local solution obtained within the $t$-th path following stage, while $\widehat{\boldsymbol{\beta}}_{\lambda_t}$ is the corresponding exact local solution that satisfies the exact optimality condition in (3.13). In Theorem 4.7, we will provide a statistical characterization of all the approximate local solutions $\{\widetilde{\boldsymbol{\beta}}_t\}_{t=1}^N$ attained along the full regularization path. Remind in Theorem 4.5 we prove that within the $t$-th stage, the iterative sequence $\{\boldsymbol{\beta}_t^k\}_{k=0}^\infty$ produced by the proximal-gradient method converges towards a unique exact local solution $\widehat{\boldsymbol{\beta}}_{\lambda_t}$. In Theorem 4.8, we will provide more refined statistical properties of these exact local solutions $\{\widehat{\boldsymbol{\beta}}_{\lambda_t}\}_{t=1}^N$ along the full regularization path. Since $\widehat{\boldsymbol{\beta}}_{\lambda_N} = \widehat{\boldsymbol{\beta}}_{\lambda_{\text{tgt}}}$, this result justifies the statistical property of the final estimator.

**Theorem 4.7** (Statistical Rates of Convergence of Approximate Local Solutions)**.** Recall that $\widetilde{\boldsymbol{\beta}}_t$ is the approximate local solution obtained within the $t$-th path following stage (Line 8 and Line 12 of Algorithm 1). Under Assumption 4.1 and Assumption 4.4, we have

$$\|\widetilde{\boldsymbol{\beta}}_t - \boldsymbol{\beta}^*\|_2 \le C\lambda_t\sqrt{s^*}, \quad \text{for } t = 1, \dots, N, \tag{4.15}$$

where $s^* = \|\boldsymbol{\beta}^*\|_0$ and $C = (21/8)/(\rho_- - \zeta_-)$.

PROOF. See the next section for a detailed proof. □

Theorem 4.7 provides statistical rates of convergence of all the approximate local solutions attained by our algorithm along the regularization path. Recall that in Assumption 4.1, we set $\lambda_{\text{tgt}} = C\sqrt{\log d/n}$ for least squares and logistic loss, and $\lambda_{\text{tgt}} = C'\|\boldsymbol{\beta}^*\|_1\sqrt{\log d/n}$ for semiparametric elliptical design loss. For least squares and logistic loss, taking $t = N$ in Theorem 4.7, we have

$$\big\|\widetilde{\boldsymbol{\beta}}_N - \boldsymbol{\beta}^*\big\|_2 \le \frac{21/8}{\rho_- - \zeta_-}\lambda_{\text{tgt}}\sqrt{s^*} = \frac{21/8 \cdot C}{\rho_- - \zeta_-}\sqrt{\frac{s^*\log d}{n}}.$$

Hence, the final approximate local solution $\widetilde{\boldsymbol{\beta}}_N$ attains the minimax rate of convergence for parameter estimation. Similarly, for semiparametric elliptical design loss, we have

$$\big\|\widetilde{\boldsymbol{\beta}}_N - \boldsymbol{\beta}^*\big\|_2 \le \frac{21/8 \cdot C'}{\rho_- - \zeta_-}\|\boldsymbol{\beta}^*\|_1\sqrt{\frac{s^*\log d}{n}},$$

which suggests that the rate of convergence of the final approximate local solution is also optimal in the regime where $\|\boldsymbol{\beta}^*\|_1$ is upper bounded by a constant. Moreover, since $\eta$ is an absolute constant, for $\widetilde{\boldsymbol{\beta}}_{N-K}$ with $K$ being a positive integer constant, Theorem 4.7 gives

$$\big\|\widetilde{\boldsymbol{\beta}}_{N-K} - \boldsymbol{\beta}^*\big\|_2 \le \frac{21/8}{\rho_- - \zeta_-}\lambda_{N-K}\sqrt{s^*} \le \frac{21/8 \cdot \eta^{-K}}{\rho_- - \zeta_-}\lambda_{\text{tgt}}\sqrt{s^*},$$

which suggests that, the approximate local solution $\widetilde{\boldsymbol{\beta}}_{N-K}$ enjoys the same rate of convergence as the final approximate local solution $\widetilde{\boldsymbol{\beta}}_N$, but with a larger constant $C = (21/8)\cdot\eta^{-K}/(\rho_- - \zeta_-) > (21/8)/(\rho_- - \zeta_-)$.

In independent work, Theorem 1 and Corollaries 1-3 of Loh and Wainwright (2013) showed the approximate local solution $\widetilde{\boldsymbol{\beta}}$ attained by their optimization procedure satisfies $\big\|\widetilde{\boldsymbol{\beta}} - \boldsymbol{\beta}^*\big\|_2 \le C\lambda_{\text{tgt}}\sqrt{s^*}$. A comparison between Theorem 4.7 and their result suggests that, our approximate local solution $\widetilde{\boldsymbol{\beta}}_N$ obtained within the final path following stage has the same statistical rate of convergence as the approximate local solution attained by their procedure. Meanwhile, Theorem 4.7 provides additional statistical characterizations for the other regularization parameters along the regularization path, i.e., $\lambda_1, \ldots, \lambda_{N-1}$.

In the next theorem, we provide a refined statistical rate of convergence. Recall within the $t$-th path following stage, the iterative sequence $\{\boldsymbol{\beta}_t^k\}_{k=0}^{\infty}$ produced by the proximal-gradient method converges towards a unique exact local solution $\widehat{\boldsymbol{\beta}}_{\lambda_t}$. The next theorem states that $\widehat{\boldsymbol{\beta}}_{\lambda_t}$ benefits from nonconvex regularization and possesses an improved statistical rate of convergence.

**Theorem 4.8** (Refined Statistical Rates of Convergence of Exact Local So-

lutions). For the regularization parameter $\lambda_t$, we assume that the nonconvex penalty $\mathcal{P}_{\lambda_t}(\boldsymbol{\beta}) = \sum_{j=1}^{d} p_{\lambda_t}(\beta_j)$ satisfies

$$p'_{\lambda_t}(\beta_j) = 0, \quad \text{for} \quad |\beta_j| \geq \nu_t, \tag{4.16}$$

for some $\nu_t > 0$. Let $S_1^* \cup S_2^* = S^* = \operatorname{supp}(\boldsymbol{\beta}^*)$ with $|S_1^*| = s_1^*$, $|S_2^*| = s_2^*$ and $|S^*| = s^* = s_1^* + s_2^*$. For $j \in S_1^* \subseteq S^*$, we assume $|\beta_j^*| \geq \nu_t$, while for $j \in S_2^* \subseteq S^*$, we assume $|\beta_j^*| < \nu_t$. Under Assumption 4.1 and Assumption 4.4, we have

$$\|\widehat{\boldsymbol{\beta}}_{\lambda_t} - \boldsymbol{\beta}^*\|_2 \leq \underbrace{C\|(\nabla\mathcal{L}(\boldsymbol{\beta}^*))_{S_1^*}\|_2}_{S_1^*:\ \text{Large } |\beta_j|'\text{s}} + \underbrace{C'\lambda_t\sqrt{s_2^*}}_{S_2^*:\ \text{Small } |\beta_j|'\text{s}}, \quad \text{for } t = 1, \dots, N, \tag{4.17}$$

where $C = 1/(\rho_- - \zeta_-)$ and $C' = 3/(\rho_- - \zeta_-)$.

PROOF. See the next section for a detailed proof. □

In Theorem 4.8, the assumption in (4.16) applies to a variety of nonconvex penalty functions. For SCAD in (2.1), we have $\nu_t = a\lambda_t$; While for MCP in (2.2), we have $\nu_t = b\lambda_t$. Theorem 4.8 suggests that, for "small" coefficients such that $|\beta_j| < \nu_t$, the second part on the right-hand side of (4.17) has the same recovery performance as in Theorem 4.7, while for "large" coefficients such that $|\beta_j| \geq \nu_t$, the first part in (4.17) possesses a more refined rate of convergence. To understand this, we consider an example with $\mathcal{L}(\boldsymbol{\beta})$ being least squares loss. We assume that $(Y|\boldsymbol{X} = \mathbf{x}_i)$ follows a sub-Gaussian distribution with mean $\mathbf{x}_i^T\boldsymbol{\beta}^*$ and variance proxy $\sigma^2$. Moreover, we assume that the columns of $\mathbf{X}$ are normalized in such a way that $\max_{j\in\{1,\dots,d\}}\{\|\mathbf{X}_j\|_2\} \leq \sqrt{n}$. Then we have

$$\|(\nabla\mathcal{L}(\boldsymbol{\beta}^*))_{S_1^*}\|_2 \leq C\sigma\sqrt{\frac{s_1^*}{n}} \tag{4.18}$$

with high probability. Clearly, this $\sqrt{s_1^*/n}$ rate of convergence on the right-hand side of (4.18) is significantly faster than the usual $\sqrt{s^*\log d/n}$ rate, since it gets rid of the $\log d$ term, and $s_1^* \leq s^*$. In fact, $\nu_t$ is the minimum signal strength above which we are able to obtain this refined rate of convergence. In the examples of SCAD and MCP, we have $\nu_t = C\lambda_t$. Recall that $\{\lambda_t\}_{t=0}^N$ is a decreasing sequence. Hence, we are able to achieve this more refined rate of convergence for smaller and smaller signal strength along the regularization path. Moreover, for $t = N$, the minimum signal strength $\nu_N = \lambda_N = \lambda_{\text{tgt}} = C\sqrt{\log d/n}$. Hence, the required minimum signal strength goes to zero as the sample size increases. Following a similar proof of Lemma C.1 and Lemma C.2 in the supplementary material (Wang et al., 2014b), we can also obtain similar results for logistic loss and semiparametric elliptical design loss. This refined rate of convergence is sharper than the result in Theorem 4.7, which is also achievable via convex regularization, e.g., the $\ell_1$ penalty. Therefore, Theorem 4.8 clearly justifies the benefits of using nonconvex regularization. Moreover, in §6 we will show that our requirement on the minimum signal strength to achieve this refined rate of convergence is optimal, and is a weaker requirement than the suboptimal requirements in Wang et al. (2013); Fan et al. (2014).

In addition to the refined rate of convergence for parameter estimation in Theorem 4.8, in the next theorem we prove that the exact local solution $\widehat{\boldsymbol{\beta}}_{\lambda_t}$ also recovers the support of $\boldsymbol{\beta}^*$. Before we present the next theorem, we introduce the definition of an oracle estimator, denoted by $\widehat{\boldsymbol{\beta}}_{\mathrm{O}}$. Recall that $S^* = \mathrm{supp}(\boldsymbol{\beta}^*)$. The oracle estimator $\widehat{\boldsymbol{\beta}}_{\mathrm{O}}$ is defined as

$$\widehat{\boldsymbol{\beta}}_{\mathrm{O}} = \mathop{\mathrm{argmin}}_{\substack{\mathrm{supp}\,(\boldsymbol{\beta})\subseteq S^* \\ \boldsymbol{\beta}\in\Omega}} \mathcal{L}(\boldsymbol{\beta}), \tag{4.19}$$

where $\Omega = \mathbb{R}^d$ for least squares loss and semiparametric elliptical design loss, while $\Omega = B_2(R)$ for logistic loss with $R \geq \|\boldsymbol{\beta}^*\|_2$. In the next lemma, we show that $\widehat{\boldsymbol{\beta}}_{\mathrm{O}}$ is the unique global solution to the minimization problem in (4.19) even for nonconvex loss functions, and has nice statistical properties.

**Lemma 4.9.** Under Assumption 4.4, the oracle estimator $\widehat{\boldsymbol{\beta}}_{\mathrm{O}}$ is the unique global minimizer of (4.19). For $\mathcal{L}(\boldsymbol{\beta})$ being least squares loss, we assume that $(Y|\boldsymbol{X} = \mathbf{x}_i)$ follows a sub-Gaussian distribution with mean $\mathbf{x}_i^T\boldsymbol{\beta}^*$ and variance proxy $\sigma^2$, then the oracle estimator satisfies

$$\big\|\widehat{\boldsymbol{\beta}}_{\mathrm{O}} - \boldsymbol{\beta}^*\big\|_\infty \leq C\sigma\sqrt{2/\rho_-}\cdot\sqrt{\frac{\log s^*}{n}} \tag{4.20}$$

with high probability for some constant $C$.

Proof. See the supplementary material (Wang et al., 2014b) for a detailed proof. □

Statistical recovery results similar to (4.20) also hold for logistic loss and semiparametric elliptical design loss under different conditions. These results are omitted here for simplicity. Lemma 4.9 suggests that, for a sufficiently large $n$ and sufficient minimum signal strength, the oracle estimator $\widehat{\boldsymbol{\beta}}_{\mathrm{O}}$ exactly recovers the support of $\boldsymbol{\beta}^*$. More specifically, if the minimum signal strength satisfies $\min_{j\in S^*}|\beta_j^*| \geq 2\nu$ for $\nu > 0$, then with high probability

$$\min_{j\in S^*}\big|\big(\widehat{\boldsymbol{\beta}}_{\mathrm{O}}\big)_j\big| \geq \min_{j\in S^*}|\beta_j^*| - \big\|\widehat{\boldsymbol{\beta}}_{\mathrm{O}} - \boldsymbol{\beta}^*\big\|_\infty \geq 2\nu - \sigma\sqrt{2/\rho_-}\cdot\sqrt{\frac{\log s^*}{n}},$$

which implies $\min_{j\in S^*}\big|\big(\widehat{\boldsymbol{\beta}}_{\rm O}\big)_j\big| \geq \nu > 0$ for $n$ sufficiently large. Meanwhile, recall that $\operatorname{supp}\big(\widehat{\boldsymbol{\beta}}_{\rm O}\big) \subseteq S^*$ by definition. Hence we have $\operatorname{supp}\big(\widehat{\boldsymbol{\beta}}_{\rm O}\big) = S^*$.

The next theorem states, under the condition of sufficient minimum signal strength, $\widehat{\boldsymbol{\beta}}_{\lambda_t}$ is the oracle estimator, and exactly recovers the support of $\boldsymbol{\beta}^*$.

**Theorem 4.10** (Support Recovery)**.** For the regularization parameter $\lambda_t$, suppose that the nonconvex penalty $\mathcal{P}_{\lambda_t}(\boldsymbol{\beta}) = \sum_{j=1}^d p_{\lambda_t}(\beta_j)$ satisfies (4.16) for some $\nu_t > 0$. For least squares loss, we assume that $(Y|\boldsymbol{X} = \mathbf{x}_i)$ follows a sub-Gaussian distribution with mean $\mathbf{x}_i^T\boldsymbol{\beta}^*$ and variance proxy $\sigma^2$. Under Assumption 4.1 and Assumption 4.4, if the minimum signal strength satisfies $\min_{j\in S^*}|\beta_j^*| \geq 2\nu_t$, then for $n$ sufficiently large, $\widehat{\boldsymbol{\beta}}_{\lambda_t} = \widehat{\boldsymbol{\beta}}_{\rm O}$, and $\operatorname{supp}\big(\widehat{\boldsymbol{\beta}}_{\lambda_t}\big) = \operatorname{supp}\big(\widehat{\boldsymbol{\beta}}_{\rm O}\big) = \operatorname{supp}(\boldsymbol{\beta}^*)$ with high probability.

PROOF. See the next section for a detailed proof. □

Recall the assumption in (4.16) applies to a variety of nonconvex penalties, including SCAD and MCP, for which we have $\nu_t = C\lambda_t$ with $C > 0$. Hence, the minimum signal strength that is required to achieve the oracle estimator and exact support recovery actually shrinks with the decreasing sequence $\{\lambda_t\}_{t=0}^N$ along the regularization path. For least squares loss, we have $\nu_N = C\lambda_{\rm tgt} = C'\sqrt{\log d/n}$ for $t = N$. Hence, within the final path following stage, the required minimum signal strength goes to zero as sample size $n \to \infty$. Furthermore, such requirement on the minimum signal strength for achieving the oracle estimator is optimal, i.e., no weaker requirement exists (Zhang and Zhang, 2012). In §6 we will show that, for least squares loss, some of recent works (Fan et al., 2014; Wang et al., 2013) require a stronger minimum signal strength to achieve the oracle estimator in the same setting of least squares regression. Similar results to Theorem 4.10 also hold for other loss functions, but under different conditions. They are omitted here for simplicity.

## 5. Proof of Main Results.

In this section we present the proof sketch of the main results. The desired computational and statistical results rely on the strong convexity of the surrogate loss function $\widetilde{\mathcal{L}}_\lambda(\boldsymbol{\beta})$, e.g., we need $\widetilde{\mathcal{L}}_\lambda(\boldsymbol{\beta})$ to be strongly convex to establish the geometric rate of convergence of the proximal-gradient method within each path following stage. However, $\widetilde{\mathcal{L}}_\lambda(\boldsymbol{\beta})$ is nonconvex in general, since $\widetilde{\mathcal{L}}_\lambda(\boldsymbol{\beta}) = \mathcal{L}(\boldsymbol{\beta}) + \mathcal{Q}_\lambda(\boldsymbol{\beta})$, where $\mathcal{L}(\boldsymbol{\beta})$ is possibly nonconvex and $\mathcal{Q}_\lambda(\boldsymbol{\beta})$ is concave. In the following lemma, we prove that $\widetilde{\mathcal{L}}_\lambda(\boldsymbol{\beta}) = \mathcal{L}(\boldsymbol{\beta}) + \mathcal{Q}_\lambda(\boldsymbol{\beta})$ is strongly convex for $\boldsymbol{\beta}$ on a sparse set. In a similar way, we establish the strong smoothness of $\widetilde{\mathcal{L}}_\lambda(\boldsymbol{\beta})$ on a sparse set.

Recall $\rho_- = \rho_-\big(\nabla^2\mathcal{L}, s^* + 2\widetilde{s}\big)$ and $\rho_+ = \rho_+\big(\nabla^2\mathcal{L}, s^* + 2\widetilde{s}\big)$ are the sparse eigenvalues specified in Assumption 4.4. As defined in regularity condition (a), $\zeta_-, \zeta_+ > 0$ are the concavity parameters of the nonconvex penalty, which satisfy (4.3) and (4.6).

**Lemma 5.1.** Let $\boldsymbol{\beta}, \boldsymbol{\beta}' \in \mathbb{R}^d$ be two sparse vectors, which satisfy $\|(\boldsymbol{\beta} - \boldsymbol{\beta}')_{\overline{S^*}}\|_0 \leq 2\widetilde{s}$, where $\widetilde{s}$ is specified in Assumption 4.4 and $S^* = \mathrm{supp}(\boldsymbol{\beta}^*)$. For $\mathcal{L}(\boldsymbol{\beta})$ being logistic loss, we further assume $\|\boldsymbol{\beta}\|_2 \leq R$ and $\|\boldsymbol{\beta}'\|_2 \leq R$, where $R$ is a constant specified in Definition 4.3. Then the surrogate loss function $\widetilde{\mathcal{L}}_\lambda(\boldsymbol{\beta}) = \mathcal{L}(\boldsymbol{\beta}) + \mathcal{Q}_\lambda(\boldsymbol{\beta})$ satisfies the restricted strong convexity

$$\widetilde{\mathcal{L}}_\lambda(\boldsymbol{\beta}') \geq \widetilde{\mathcal{L}}_\lambda(\boldsymbol{\beta}) + \nabla\widetilde{\mathcal{L}}_\lambda(\boldsymbol{\beta})^T(\boldsymbol{\beta}' - \boldsymbol{\beta}) + \frac{\rho_- - \zeta_-}{2}\|\boldsymbol{\beta}' - \boldsymbol{\beta}\|_2^2,$$

and the restricted strong smoothness

$$\widetilde{\mathcal{L}}_\lambda(\boldsymbol{\beta}') \leq \widetilde{\mathcal{L}}_\lambda(\boldsymbol{\beta}) + \nabla\widetilde{\mathcal{L}}_\lambda(\boldsymbol{\beta})^T(\boldsymbol{\beta}' - \boldsymbol{\beta}) + \frac{\rho_+ - \zeta_+}{2}\|\boldsymbol{\beta}' - \boldsymbol{\beta}\|_2^2.$$

Proof. See §D.2 in the supplementary material (Wang et al., 2014b) for a detailed proof. □

A similar result has been discussed by Negahban et al. (2012). The main difference is that, our constraint set where $\widetilde{\mathcal{L}}_\lambda(\boldsymbol{\beta})$ is strongly convex/smooth is a sparse subspace, while that of Negahban et al. (2012) is a cone.

Note that in Lemma 5.1, the strong convexity and smoothness of $\widetilde{\mathcal{L}}_\lambda(\boldsymbol{\beta})$ rely on the sparsity of $\boldsymbol{\beta}$ and $\boldsymbol{\beta}'$. Hence, we need to establish results regarding the sparsity of $\boldsymbol{\beta}_t^k$ throughout the whole iterative procedure. In the sequel, we provide several important lemmas: Lemma 5.2 and Lemma 5.3 characterize the statistical properties of any sparse $\boldsymbol{\beta}$; Based on such statistical properties, Lemma 5.4 proves that, any proximal-gradient update iteration with a sparse input produces a sparse output. Equipped with these lemmas, we can establish the sparsity of the solution path by mathematical induction in Theorem 5.5.

The next lemma provides a characterization of any sparse $\boldsymbol{\beta}$ with certain suboptimality.

**Lemma 5.2.** We assume that $\boldsymbol{\beta}$ satisfies

$$\|\boldsymbol{\beta}_{\overline{S^*}}\|_0 \leq \widetilde{s}, \quad \omega_\lambda(\boldsymbol{\beta}) \leq \lambda/2 \tag{5.1}$$

with $\lambda \geq \lambda_{\mathrm{tgt}}$, where $\omega_\lambda(\boldsymbol{\beta})$ is the measure of suboptimality defined in (3.15). For logistic loss, we assume $\|\boldsymbol{\beta}\|_2 \leq R$, where $R > 0$ is a constant specified in Definition 4.3. Under Assumption 4.1 and Assumption 4.4, $\boldsymbol{\beta}$ satisfies

$$\|\boldsymbol{\beta} - \boldsymbol{\beta}^*\|_2 \leq C\lambda\sqrt{s^*}, \quad \text{where} \quad C = \frac{21/8}{\rho_- - \zeta_-}.$$

Meanwhile, the objective function value evaluated at $\boldsymbol{\beta}$ satisfies

$$\phi_\lambda(\boldsymbol{\beta}) - \phi_\lambda(\boldsymbol{\beta}^*) \le C'\lambda^2 s^*, \quad \text{where} \quad C' = \frac{21/2}{\rho_- - \zeta_-}.$$

PROOF. See §D.3 of the supplementary material (Wang et al., 2014b) for a detailed proof. □

Recall that we use the approximate local solution $\widetilde{\boldsymbol{\beta}}_{t-1}$ obtained within the $(t-1)$-th path following stage to be the initialization of the $t$-th stage (Line 8 of Algorithm 1), i.e., $\boldsymbol{\beta}_t^0 = \widetilde{\boldsymbol{\beta}}_{t-1}$. By setting $\boldsymbol{\beta} = \widetilde{\boldsymbol{\beta}}_{t-1} = \boldsymbol{\beta}_t^0$ and $\lambda = \lambda_t$ in Lemma 5.2, we can see that, if $\widetilde{\boldsymbol{\beta}}_{t-1}$ is sparse and $(\lambda_t/2)$-suboptimal, then the initial point $\boldsymbol{\beta}_t^0$ of the $t$-th stage has nice statistical recovery performance. However, it is unclear whether the rest of $\boldsymbol{\beta}_t^k$'s $(k = 1, 2, \ldots)$ within the $t$-th stage also have similar recovery performance. To prove this, we first present Lemma 5.3, which shows that under the condition that $\boldsymbol{\beta}$ is sparse and $\phi_\lambda(\boldsymbol{\beta})$ is close to $\phi_\lambda(\boldsymbol{\beta}^*)$, $\boldsymbol{\beta}$ has desired statistical properties. After Lemma 5.3, we will explain that if $\boldsymbol{\beta}_t^0$ satisfies this condition, then all the $\boldsymbol{\beta}_t^k$'s $(k = 1, 2, \ldots)$ within the same path following stage also satisfy this condition, and therefore enjoys nice statistical properties.

**Lemma 5.3.** Suppose that, for $\lambda \ge \lambda_{\text{tgt}}$, $\boldsymbol{\beta}$ satisfies

$$\|\boldsymbol{\beta}_{\overline{S^*}}\|_0 \le \widetilde{s}, \quad \phi_\lambda(\boldsymbol{\beta}) - \phi_\lambda(\boldsymbol{\beta}^*) \le C\lambda^2 s^*, \quad \text{where} \quad C = \frac{21/2}{\rho_- - \zeta_-}.$$

For logistic loss, we further assume $\|\boldsymbol{\beta}\|_2 \le R$, where $R$ is a constant specified in Definition 4.3. Under Assumption 4.1 and Assumption 4.4, we have

$$\|\boldsymbol{\beta} - \boldsymbol{\beta}^*\|_2 \le C'\lambda\sqrt{s^*}, \quad \text{where} \quad C' = \frac{15/2}{\rho_- - \zeta_-}.$$

PROOF. See §D.4 of the supplementary material (Wang et al., 2014b) for a detailed proof. □

Let $\lambda = \lambda_t$ and $\boldsymbol{\beta} = \boldsymbol{\beta}_t^k$ in Lemma 5.3. It suggests that within the $t$-th path following stage, all $\boldsymbol{\beta}_t^k$'s $(k = 1, 2, \ldots)$ have nice statistical recovery performance under three sufficient conditions: (i) Each $\boldsymbol{\beta}_t^k$ is sparse; (ii) The objective function value $\phi_{\lambda_t}(\boldsymbol{\beta}_t^k)$ is close to $\phi_{\lambda_t}(\boldsymbol{\beta}^*)$; (iii) For logistic loss, we further need $\|\boldsymbol{\beta}_t^k\|_2 \le R$. For condition (ii), recall that if we set $\boldsymbol{\beta} = \boldsymbol{\beta}_t^0$ and $\lambda = \lambda_t$ in Lemma 5.2, then $\boldsymbol{\beta}_t^0$ being sparse and $(\lambda_t/2)$-suboptimal implies that $\phi_{\lambda_t}(\boldsymbol{\beta}_t^0)$ is close to $\phi_{\lambda_t}(\boldsymbol{\beta}^*)$. Since the proximal-gradient method ensures the monotone decrease of $\{\phi_{\lambda_t}(\boldsymbol{\beta}_t^k)\}_{k=0}^\infty$ within the $t$-th stage (see Lemma D.1 of the supplementary material (Wang et al., 2014b)), condition (ii) also holds. Meanwhile, condition (iii) obviously holds because of the $\ell_2$ constraint. To establish the statistical recovery performance of all the $\boldsymbol{\beta}_t^k$'s within the $t$-th stage, we still need to establish the sparsity of $\boldsymbol{\beta}_t^k$'s to guarantee condition (i) holds. To prove this, we present Lemma 5.4, which states that if $\boldsymbol{\beta}$ is sparse, then a proximal-gradient update operation (3.9) on $\boldsymbol{\beta}$ produces a sparse solution.

**Lemma 5.4.** Suppose that, for $\lambda \geq \lambda_{\text{tgt}}$, $\boldsymbol{\beta}$ satisfies

$$\|\boldsymbol{\beta}_{\overline{S^*}}\|_0 \leq \widetilde{s}, \quad \phi_\lambda(\boldsymbol{\beta}) - \phi_\lambda(\boldsymbol{\beta}^*) \leq C\lambda^2 s^*, \quad \text{and} \quad L < 2(\rho_+ - \zeta_+),$$

where $C = (21/2)/(\rho_- - \zeta_-)$. For logistic loss, we assume $\|\boldsymbol{\beta}\|_2 \leq R$, where $R$ is specified in Definition 4.3. Under Assumption 4.1 and Assumption 4.4, the proximal-gradient update operation defined in (3.9) produces a sparse solution, i.e.,

$$\big\|\big(\mathcal{T}_{L,\lambda}(\boldsymbol{\beta}; R)\big)_{\overline{S^*}}\big\|_0 \leq \widetilde{s}.$$

Here we set $R = +\infty$ if the domain $\Omega$ in (3.8) is $\mathbb{R}^d$.

Proof. See §D.5 of the supplementary material (Wang et al., 2014b) for a detailed proof. □

For $\boldsymbol{\beta} = \boldsymbol{\beta}_t^{k-1}$, $\lambda = \lambda_t$ and $L = L_t^k$, Lemma 5.4 states that, if $\boldsymbol{\beta}_t^{k-1}$ is sparse and the objective function value $\phi_{\lambda_t}\big(\boldsymbol{\beta}_t^{k-1}\big)$ is close to $\phi_{\lambda_t}(\boldsymbol{\beta}^*)$, then $\boldsymbol{\beta}_t^k = \mathcal{T}_{L_t^k,\lambda_t}\big(\boldsymbol{\beta}_t^{k-1}; R\big)$ produced by the proximal-gradient update step (3.8) is also sparse. Within the $t$-th path following stage, if $\boldsymbol{\beta}_t^0$ is sparse, $\omega_{\lambda_t}\big(\boldsymbol{\beta}_t^0\big) \leq \lambda_t/2$, and for logistic loss $\big\|\boldsymbol{\beta}_t^0\big\|_2 \leq R$, then by Lemma 5.2 we have

$$\phi_{\lambda_t}\big(\boldsymbol{\beta}_t^0\big) - \phi_{\lambda_t}(\boldsymbol{\beta}^*) \leq \frac{21/2}{\rho_- - \zeta_-}\lambda_t^2 s^*.$$

Since $\big\{\phi_{\lambda_t}\big(\boldsymbol{\beta}_t^k\big)\big\}_{k=0}^{\infty}$ decreases monotonically, we have

$$\phi_{\lambda_t}\big(\boldsymbol{\beta}_t^k\big) - \phi_{\lambda_t}(\boldsymbol{\beta}^*) \leq \phi_{\lambda_t}\big(\boldsymbol{\beta}_t^0\big) - \phi_{\lambda_t}(\boldsymbol{\beta}^*) \leq \frac{21/2}{\rho_- - \zeta_-}\lambda_t^2 s^*, \quad \text{for} \quad k = 1, 2, \ldots.$$

Assume that we have $L_t^k \leq 2(\rho_+ - \zeta_+)$ (which will be proved in Theorem 5.5). Applying Lemma 5.4 recursively, we obtain $\big\|\big(\boldsymbol{\beta}_t^k\big)_{\overline{S^*}}\big\|_0 \leq \widetilde{s}$ $(k = 1, 2, \ldots)$. Meanwhile, we have $\big\|\boldsymbol{\beta}_t^k\big\|_2 \leq R$ due to the $\ell_2$ constraint. Then according to Lemma 5.3, all $\boldsymbol{\beta}_t^k$'s within the $t$-th path following stage have nice recovery performance, i.e.,

$$\big\|\boldsymbol{\beta}_t^k - \boldsymbol{\beta}^*\big\|_2 \leq \frac{15/2}{\rho_- - \zeta_-}\lambda_t\sqrt{s^*}, \quad \text{for} \quad k = 1, 2, \ldots.$$

Furthermore, by Lemma 5.1 the sparsity of $\boldsymbol{\beta}_t^k$'s implies the restricted strong convexity and smoothness of $\widetilde{\mathcal{L}}_{\lambda_t}(\boldsymbol{\beta})$, which enable us to establish the geo-metric rate of convergence within the $t$-th path following stage. These results are formally presented in Theorem 5.5.

**Theorem 5.5.** Suppose within the $t$-th path following stage, the proximal-gradient method in Algorithm 3 is initialized by $\beta_t^0$ and $L_t^0$, which satisfy

$$\|(\beta_t^0)_{\overline{S^*}}\|_0 \le \widetilde{s}, \quad \omega_{\lambda_t}(\beta_t^0) \le \lambda_t/2, \quad \text{and} \quad L_t^0 \le 2(\rho_+ - \zeta_+).$$

For logistic loss we further assume $\|\beta_t^0\|_2 \le R$ with $R$ specified in Definition 4.3. Then we have the following results:

- For $k = 1, 2, \ldots$, we have

$$\|(\beta_t^k)_{\overline{S^*}}\|_0 \le \widetilde{s}, \ \|\beta_t^k - \beta^*\|_2 \le \frac{15/2}{\rho_- - \zeta_-}\lambda_t\sqrt{s^*}, \ L_t^k \le 2(\rho_+ - \zeta_+). \tag{5.2}$$

- The iterative sequence $\{\beta_t^k\}_{k=0}^{\infty}$ converges towards a unique exact local solution $\widehat{\beta}_{\lambda_t}$, which satisfies $\|(\widehat{\beta}_{\lambda_t})_{\overline{S^*}}\|_0 \le \widetilde{s}$ and the exact optimality condition that $\omega_{\lambda_t}(\beta_t^k) \le 0$.
- To achieve an approximate local solution $\widetilde{\beta}_t$ that satisfies $\omega_{\lambda_t}(\widetilde{\beta}_t) \le \lambda_t/4$, we need no more than $C' \log(4C\sqrt{s^*})$ proximal-gradient iterations defined in Lines 5-9 of Algorithm 3. Here

$$C = 2\sqrt{21} \cdot \sqrt{\kappa}(1+\kappa), \quad C' = 2 \Big/ \log\left(\frac{1}{1 - 1/(8\kappa)}\right). \tag{5.3}$$

- To obtain an approximate local solution $\widetilde{\beta}_t$ such that $\omega_{\lambda_t}(\widetilde{\beta}_t) \le \epsilon_{\text{opt}}$, we need no more than $C' \log(C\lambda_t\sqrt{s^*}/\epsilon_{\text{opt}})$ proximal-gradient iterations. Here $C$ and $C'$ are defined in (5.3).

PROOF. See §D.6 of the supplementary material (Wang et al., 2014b) for a detailed proof. □

To prove the geometric rate of convergence and desired statistical recovery results hold within all path following stages, i.e., $t = 0, \ldots, N$, we need to verify that the conditions of Theorem 5.5 hold within each stage. We prove by induction. We assume the initialization of $(t-1)$-th path following stage satisfies

$$\|(\beta_{t-1}^0)_{\overline{S^*}}\|_0 \le \widetilde{s}, \ \omega_{\lambda}(\beta_{t-1}^0) \le \lambda_t/2, \quad \text{and} \quad L_{t-1}^0 \le 2(\rho_+ - \zeta_+). \tag{5.4}$$

Applying Theorem 5.5, we obtain

$$\|(\beta_{t-1}^k)_{\overline{S^*}}\|_0 \le \widetilde{s}, \ \ L_{t-1}^k \le 2(\rho_+ - \zeta_+), \quad \text{for} \ \ k = 1, 2, \ldots.$$

Consequently, the approximate solution $\widetilde{\beta}_{t-1}$ produced by the $(t-1)$-th stage satisfies $\|(\widetilde{\beta}_{t-1})_{\overline{S^*}}\|_0 \le \widetilde{s}$, while $L_{t-1}$ satisfies $L_{t-1} \le 2(\rho_+ - \zeta_+)$. Since

we warm start the $t$-th path following stage with $\boldsymbol{\beta}_t^0 = \widetilde{\boldsymbol{\beta}}_{t-1}$ and $L_t^0 = L_{t-1}$ (Line 8 of Algorithm 1), we have

$$\big\|\big(\boldsymbol{\beta}_t^0\big)_{\overline{S^*}}\big\|_0 \le \widetilde{s}, \quad L_t^0 \le 2(\rho_+ - \zeta_+). \tag{5.5}$$

Moreover, note that the stopping criterion of the proximal-gradient method ensures $\omega_{\lambda_{t-1}}\big(\widetilde{\boldsymbol{\beta}}_{t-1}\big) \le \lambda_{t-1}/4$ (Line 9 of Algorithm 3), which implies $\omega_{\lambda_t}\big(\widetilde{\boldsymbol{\beta}}_{t-1}\big) \le \lambda_t/2$ according to Lemma D.4 of the supplementary material (Wang et al., 2014b). Thus we have

$$\omega_{\lambda_t}\big(\boldsymbol{\beta}_t^0\big) \le \lambda_t/2. \tag{5.6}$$

Therefore, we know that (5.4) implies (5.5) and (5.6). We will verify (5.5) and (5.6) hold for $t = 0$ in the proof of Theorem 4.5 in the supplementary material (Wang et al., 2014b). By induction, we have that (5.5) and (5.6) hold for $t = 0, \ldots, N$. As a consequence of Theorem 5.5, all path following stages have geometric rates of convergence along the solution path, which implies the global geometric rate of convergence in Theorem 4.5. See the supplementary material (Wang et al., 2014b) for a detail proof. Meanwhile, all $\boldsymbol{\beta}_t^k$'s have desired statistical properties, i.e.,

$$\big\|\boldsymbol{\beta}_t^k - \boldsymbol{\beta}^*\big\|_2 \le \frac{15/2}{\rho_- - \zeta_-}\lambda_t\sqrt{s^*}, \quad \text{for} \quad t = 1, \ldots, N \quad \text{and} \quad k = 0, 1, \ldots,$$

which leads to the statistical rates of convergence of the approximate local solutions $\big\{\widetilde{\boldsymbol{\beta}}_t\big\}_{t=1}^N$ in Theorem 4.7, the more refined rates of convergence of the exact local solutions $\big\{\widehat{\boldsymbol{\beta}}_{\lambda_t}\big\}_{t=1}^N$ in Theorem 4.8, and the support recovery results in Theorem 4.10. See §D.8-§D.10 of the supplementary material (Wang et al., 2014b) for detailed proofs respectively.

## 6. Discussion.

Our work is related to recent works on understanding nonconvex regularization in the context of least squares regression. Zhang (2010a) proposed an MC+ procedure for MCP penalized least squares regression. However, the computation of MC+ might be inefficient because there can be exponentially many switching points on its solution path. To remedy this issue, Zhang (2010b); Zhang et al. (2013) proposed the multi-stage convex relaxation method, which iteratively solves

$$\widehat{\boldsymbol{\beta}}^k \leftarrow \mathop{\mathrm{argmin}}_{\boldsymbol{\beta}\in\mathbb{R}^d}\bigg\{\mathcal{L}(\boldsymbol{\beta}) + \sum_{j=1}^d p'_\lambda\big(\big|\widehat{\beta}_j^{k-1}\big|\big)|\beta_j|\bigg\}, \quad k = 1, 2, \ldots, \tag{6.1}$$

where $p_\lambda(\beta_j)$ is defined in §2, and the initialization $\widehat{\boldsymbol{\beta}}^0$ is set to be the Lasso estimator corresponding to $\lambda$. For $k$ sufficiently large, $\widehat{\boldsymbol{\beta}}^k$ has the same oracle properties as in Theorem 4.8 and Theorem 4.10. However, for each $k$ we need to solve the minimization problem in (6.1) exactly, which is not realistic in practice, since practical optimization methods only attain finite numerical precision in finite iterations. In contrast, we provide simultaneous statistical and computational analysis by explicitly taking the numerical precision into account, and establish the global geometric rate of convergence in terms of iteration complexity for calculating the full regularization path.

This multi-stage convex relaxation method was previously referred to as local linear approximation (LLA), and was analyzed on fixed dimensional models by Zou and Li (2008). Fan et al. (2014) recently provided nonasymptotic analysis of LLA, and proved that LLA finds the oracle estimator in two iterations. However, their results rely on that the Lasso initialization satisfies $\|\widehat{\boldsymbol{\beta}}^0 - \boldsymbol{\beta}^*\|_\infty \le C\lambda$ with high probability, which requires $\lambda$ to take the value of $C'\sqrt{s^* \log d/n}$. Consequently, their requirement on the minimum signal strength is of the order of $\sqrt{s^* \log d/n}$, which is suboptimal. In contrast, we only require a minimum signal strength of the order of $\sqrt{\log d/n}$, which is optimal (Zhang and Zhang, 2012). Also, they didn't analyze the iteration complexity for computing each step of LLA, i.e., solving (6.1).

Very recently, Wang et al. (2013) considered a two-step approach similar to the two-step LLA procedure, named the calibrated CCCP. It differs from the two-step LLA in that, its Lasso initialization $\widehat{\boldsymbol{\beta}}^0$ is obtained using the regularization parameter $\tau\lambda$, where $\tau = o(1)$ and $\lambda = \sqrt{\log d/n}$. It attains the oracle estimator under the restricted eigenvalue (RE) condition (Bickel et al., 2009), but requires the minimum signal strength to be larger than $Cs^*\sqrt{\log d/n}$. Under a stronger assumption than the RE condition, namely the relaxed sparse Riesz condition, a minimum signal strength of the order of $\sqrt{\log d/n}/\tau$ is required. Such requirement is still suboptimal, but is close to the optimal scaling of $\sqrt{\log d/n}$ in our results, since $\tau$ can take $1/\log n$. They proposed a novel high-dimensional BIC criterion, which can be used to choose the best $\lambda_{\mathrm{tgt}}$ in our procedure. Also, they provided extensions to logistic regression.

The iterative hard thresholding (IHT) algorithm (Blumensath and Davies, 2009) can also achieve a local solution with desired statistical recovery performance at a global geometric rate of convergence. However, the theoretical results of IHT are not directly comparable with ours because of the usage of different noise models. If we have to cast the theoretical results of IHT into our model, their results are much weaker than ours. In detail, IHT attains an approximate local solution $\widetilde{\boldsymbol{\beta}}$, which satisfies

$$\|\widetilde{\boldsymbol{\beta}} - \boldsymbol{\beta}^*\|_2 \le 6\|\mathbf{e}\|_2 \tag{6.2}$$

with high probability. Here $\mathbf{e} \in \mathbb{R}^n$ is the noise vector in their setting, which is often considered to be perturbation noise. Note that a proper normalization gives $\mathbf{e} = (\mathbf{y} - \mathbf{X}\boldsymbol{\beta}^*)/\sqrt{n}$, where $\mathbf{y} - \mathbf{X}\boldsymbol{\beta}^*$ is considered to be the sub-Gaussian noise with zero mean and variance proxy $\sigma^2$ in our setting. Then (6.2) gives

$$\big\|\widetilde{\boldsymbol{\beta}} - \boldsymbol{\beta}^*\big\|_2 \leq 6\|\mathbf{e}\|_2 = 6\|\mathbf{y} - \mathbf{X}\boldsymbol{\beta}^*\|_2/\sqrt{n} \leq 6 \cdot 4\sigma\sqrt{n}/\sqrt{n} = 24\sigma \tag{6.3}$$

with high probability. Note that the upper bound on the right-hand side of (6.3) doesn't depend on $s^*$ and $d$, and fails to converge to zero as $n \to \infty$. In summary, casting the results of IHT into our setting of sub-Gaussian noise yields a rather weak result. Also, IHT requires prior knowledge on the true sparsity level $s^*$ to achieve fast global convergence, while our method doesn't.

In addition, the difference between our work and the independent work by Loh and Wainwright (2013) has been discussed in §1 and §4 with details.

**7. Numerical Results.** We provide numerical results illustrating the computational efficiency and statistical accuracy of the proposed method. In detail, first we illustrate the effectiveness of our method on a problem with both nonconvex loss and penalty functions. Then we conduct comparison between our method and existing nonconvex procedures.

In the first experiment, we consider $\mathcal{L}(\boldsymbol{\beta})$ being semiparametric elliptical random design loss defined in (2.6) and $\mathcal{P}_\lambda(\boldsymbol{\beta})$ being the MCP penalty defined in (2.2). We test on a synthetic dataset with $n = 500$ samples and $d = 2500$ dimensions. See the supplementary material (Wang et al., 2014b) for the detailed settings.

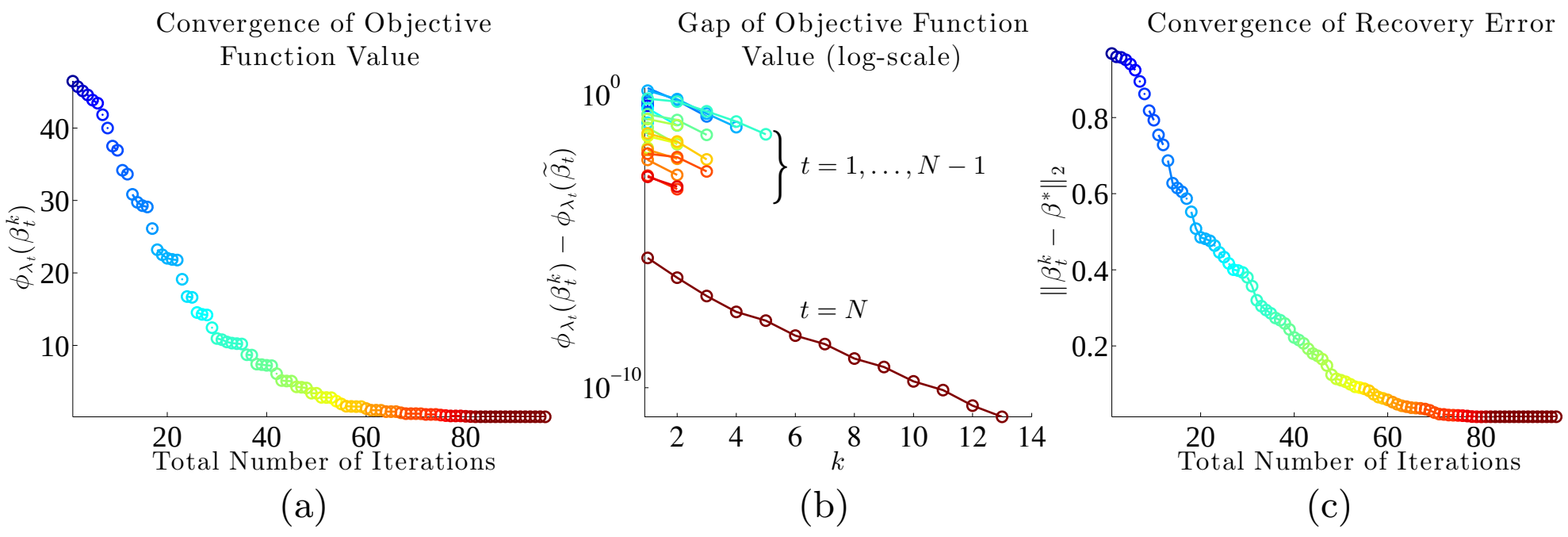

Fig 3. *Semiparametric elliptical design regression with MCP: (a) Plot of the objective function value $\phi_{\lambda_t}(\boldsymbol{\beta}_t^k)$ along the regularization path; (b) Plot of $\phi_{\lambda_t}(\boldsymbol{\beta}_t^k) - \phi_{\lambda_t}(\widetilde{\boldsymbol{\beta}}_t)$ (log-scale) within each path following stage; (c) Plot of the recovery error $\big\|\boldsymbol{\beta}_t^k - \boldsymbol{\beta}^*\big\|_2$. Here we illustrate each path following stage ($t = 1, \ldots, N$) with a different color. Note that each point in the figure denotes $\boldsymbol{\beta}_t^k$, which corresponds to the $k$-th iteration of the proximal-gradient method (Algorithm 3) within the $t$-th path following stage.*

As shown in Figure 3(a), the objective function value $\phi_\lambda(\boldsymbol{\beta}_t^k)$ is monotone decreasing along the regularization path, as characterized by our theory (see Lemma D.1 of the supplementary material (Wang et al., 2014b)), and converges eventually.

Figure 3(b) illustrates the geometric rate of convergence within each path following stage. In detail, each line denotes a path following stage. It shows the objective function value gap, i.e., $\phi_{\lambda_t}(\boldsymbol{\beta}_t^k) - \phi_{\lambda_t}(\widetilde{\boldsymbol{\beta}}_t)$, decays exponentially with $k$ within each stage. Note that

$$\begin{aligned}
&\phi_{\lambda_t}(\boldsymbol{\beta}_t^{k-1}) - \phi_{\lambda_t}(\widetilde{\boldsymbol{\beta}}_t) \geq \phi_{\lambda_t}(\boldsymbol{\beta}_t^{k-1}) - \phi_{\lambda_t}(\boldsymbol{\beta}_t^k) \\
&\qquad \geq \frac{L_t^k}{2}\|\boldsymbol{\beta}_t^k - \boldsymbol{\beta}_t^{k-1}\|_2^2 \geq \frac{L_t^k}{2}\frac{\omega_{\lambda_t}^2(\boldsymbol{\beta}_t^k)}{(L_t^k + \rho_+ - \zeta_-)^2}. \qquad (7.1)
\end{aligned}$$

Here the first inequality is because the objective function $\phi_{\lambda_t}(\boldsymbol{\beta}_t^k)$ is monotone decreasing, while the second and third inequalities follow from Lemma D.1 and Lemma D.2 of the supplementary material (Wang et al., 2014b) respectively. Therefore, $\omega_{\lambda_t}(\boldsymbol{\beta}_t^k)$ also decays exponentially within each stage, which implies that we only need a logarithmic number of iterations to attain the desired approximate local solution within each path following stage, as characterized by Theorem 4.5.

Figure 3(b) illustrates the success of the path following scheme in Figure 1: The $k = 1$ point on each line denotes the initialization of the corresponding path following stage, e.g., the $t$-th stage. Recall that such initialization is set to be the approximate local solution $\widetilde{\boldsymbol{\beta}}_{t-1}$ obtained within the $(t-1)$-th stage, which falls into the region of optimization precision in Figure 1. Meanwhile, the fast convergence within the $t$-th stage suggests that $\widetilde{\boldsymbol{\beta}}_{t-1}$ also falls into the region of fast convergence in Figure 1. Thus, the path following scheme works exactly as we have described in Figure 1 empirically.

Figure 3(c) shows that the $\ell_2$ recovery error decays towards a small value as the optimization method proceeds, which implies the attained approximate local solution has desired statistical properties, as predicted by Theorem 4.7.

In the second experiment, we compare our method with several existing nonconvex procedures on statistical performance, including LLA (Zou and Li, 2008), the calibrated CCCP (Wang et al., 2013), SparseNet (Mazumder et al., 2011), and the multi-stage convex relaxation method (Zhang, 2010b; Zhang et al., 2013). We consider an example of least squares regression with MCP, where $n = 200$, $d = 2000$ and $\|\boldsymbol{\beta}^*\|_0 = 10$. See the supplementary material (Wang et al., 2014b) for the detailed settings.

We compare the support recovery performance and $\ell_2$ recovery error of the estimators obtained from these procedures in Table 1, where we use the Lasso estimator and the oracle estimator defined in (4.19) as references. For support recovery, we are interested in the cardinality of true positive sets (TPS) and false positive set (FPS), both of which are defined in Table 1.

Ideally, the cardinality of TPS should be as large as $\|\boldsymbol{\beta}^*\|_0$ (which is 10 in this example), since a good procedure should exactly identify $S^* = \text{supp}(\boldsymbol{\beta}^*)$. Meanwhile, the cardinality of FPS should be close to zero, i.e., few of the coordinates in $\overline{S^*}$ is wrongly identified as nonzero. Table 1 shows that all nonconvex procedures significantly outperform the Lasso, which produces a less sparse estimator with larger $\ell_2$ recovery error. In this specific example, our method outperforms the existing nonconvex procedures. Moreover, our method almost recovers $S^*$ exactly, and achieves a small $\ell_2$ recovery error that is very close to the $\ell_2$ error of the oracle estimator, as characterized by Theorem 4.8.

TABLE 1
*Comparing statistical performance of nonconvex procedures: TPS/FPS denote the true/false positive sets, which are defined as $\{j \in S^* : \widehat{\beta}_j \neq 0\}$ and $\{j \in \overline{S^*} : \widehat{\beta}_j \neq 0\}$ respectively, and $|\cdot|$ denotes their cardinality. The $\ell_2$ recovery error is defined as $\|\widehat{\boldsymbol{\beta}} - \boldsymbol{\beta}^*\|_2$, where $\widehat{\boldsymbol{\beta}}$ is the estimator. Standard deviations are present in the parentheses.*

| Method | \|TPS\| | \|FPS\| | $\ell_2$ Error |
|---|---|---|---|
| **Approximate Path Following** | **10** (0) | **0.180** (0.0411) | **0.702** (0.0278) |
| SparseNet | 10 (0) | 0.950 (0.108) | 0.848 (0.0230) |
| Multi-stage Convex Relaxation | 10 (0) | 2.21 (0.146) | 1.28 (0.0753) |
| LLA | 10 (0) | 2.98 (0.304) | 1.28 (0.0996) |
| Calibrated CCCP | 9.99 (0.01) | 3.28 (0.308) | 1.40 (0.122) |
| Lasso | 9.98 (0.0141) | 31.15 (0.799) | 2.63 (0.0460) |
| Oracle Estimator | 10 (0) | 0 (0) | 0.484 (0.0221) |

## 8. Conclusion.

In this paper, we provided unified theory for penalized $M$-estimators with possibly nonconvex loss and penalty functions. These problems are motivated by generalized linear models with nonconvex penalties and semiparametric elliptical design regression, as well as a broad range of other applications. Because it is intractable to compute the global solutions of these problems due to the nonconvex formulation, we need to establish theory that characterizes both the computational and statistical properties of the local solutions obtained by specific algorithms. For this purpose, we proposed an approximate regularization path following method, which serves as a unified framework for solving a variety of high-dimensional sparse learning problems with nonconvexity. Computationally, our method enjoys a fast global geometric rate of convergence for calculating the entire regularization path; Statistically, all the approximate and exact local solutions attained by our method along the regularization path possess sharp statistical rate of convergence in both estimation and support recovery. In particular, we provide sharp theoretical analysis that demonstrates the advantage of using nonconvex penalties. This paper shows that, under suitable conditions, we can efficiently obtain the entire regularization path of a broad class of nonconvex sparse learning problems.

Our work can be extended in many directions: Our method and theory for least squares loss and logistic loss can be easily extended to other generalized linear models (see §C.2 of the supplementary material (Wang et al., 2014b) for details); For inverse covariance matrix estimation, our work is directly applicable to the Sparse Column Inverse Operator (SCIO) (Liu and Luo, 2012); Meanwhile, it might need more effort than verifying Assumption 4.1 and Assumption 4.4 to adapt the graphical Lasso into our framework, e.g., the optimization algorithm also has to be modified to enforce the positive semidefinite constraint; It is also interesting to consider other loss functions, e.g., quantile regression (Wang et al., 2012), for which Assumption 4.4 may no longer hold.

**Acknowledgement.** We sincerely thank Po-Ling Loh, Martin Wainwright and Yiyuan She for their helpful personal communications. We are grateful to the Editor, Associate Editor and referees for their insightful comments.

Han Liu is supported by NSF Grants III-1116730 and NSF III-1332109, NIH R01MH102339, NIH R01GM083084, and NIH R01HG06841, and FDA HHSF223201000072C. Tong Zhang is supported by the following grants: NSF IIS-1016061, NSF DMS-1007527, and NSF IIS-1250985.

## SUPPLEMENTARY MATERIAL

**Supplementary material for: Optimal Computational and Statistical Rates of Convergence for Sparse Nonconvex Learning Problems** (DOI: To Be Assigned; .pdf). We provide the detailed proof in the supplement (Wang et al., 2014b).

## References.

DEPARTMENT OF OPERATIONS RESEARCH
AND FINANCIAL ENGINEERING
PRINCETON UNIVERSITY
PRINCETON, NEW JERSEY 08544
USA
E-MAIL: zhaoran@princeton.edu
hanliu@princeton.edu

DEPARTMENT OF STATISTICS
RUTGERS UNIVERSITY
PISCATAWAY, NEW JERSEY 08854
USA
E-MAIL: tzhang@stat.rutgers.edu

# SUPPLEMENTARY MATERIAL FOR: OPTIMAL COMPUTATIONAL AND STATISTICAL RATES OF CONVERGENCE FOR SPARSE NONCONVEX LEARNING PROBLEMS

By Zhaoran Wang*, Han Liu* and Tong Zhang†

## APPENDIX A: NONCONVEX PENALTY AND LOSS FUNCTIONS

We provide detailed descriptions of the nonconvex penalty and loss functions discussed in §2 of Wang et al. (2014a). Specifically, for the nonconvex penalties, i.e., SCAD and MCP, we provide their analytical forms in §A.1, and illustrate regularity condition (e) of Wang et al. (2014a) in §A.2. For the nonconvex loss, i.e., semiparametric elliptical design loss, we provide further details on elliptical distribution in §A.3, and define the two-step elliptical covariance matrix estimation procedure for semiparametric elliptical design regression in §A.4.

**A.1. Analytical Forms of SCAD and MCP.** The SCAD penalty in (2.1) of Wang et al. (2014a) can be written as

$$p_\lambda(\beta_j) = \lambda|\beta_j| \cdot \mathbb{1}(|\beta_j| \le \lambda) - \frac{\beta_j^2 - 2a\lambda|\beta_j| + \lambda^2}{2(a-1)} \cdot \mathbb{1}(\lambda < |\beta_j| \le a\lambda) + \frac{(a+1)\lambda^2}{2} \cdot \mathbb{1}(|\beta_j| > a\lambda), \qquad a > 2,$$

and the MCP penalty in (2.2) of Wang et al. (2014a) can be written as

$$p_\lambda(\beta_j) = \left(\lambda|\beta_j| - \frac{\beta_j^2}{2b}\right) \cdot \mathbb{1}(|\beta_j| \le b\lambda) + \frac{b\lambda^2}{2} \cdot \mathbb{1}(|\beta_j| > b\lambda), \quad b > 0.$$

Correspondingly, the specific forms of the concave component $q_\lambda(\beta_j)$ are

$$q_\lambda(\beta_j) = \begin{cases} \dfrac{2\lambda|\beta_j| - \beta_j^2 - \lambda^2}{2(a-1)} \cdot \mathbb{1}(\lambda < |\beta_j| \le a\lambda) + \dfrac{(a+1)\lambda^2 - 2\lambda|\beta_j|}{2} \cdot \mathbb{1}(|\beta_j| > a\lambda), & \text{SCAD}, \\ -\dfrac{\beta_j^2}{2b} \cdot \mathbb{1}(|\beta_j| \le b\lambda) + \left(\dfrac{b\lambda^2}{2} - \lambda|\beta_j|\right) \cdot \mathbb{1}(|\beta_j| > b\lambda), & \text{MCP}. \end{cases}$$

**A.2. Illustration of Regularity Condition (e) for SCAD and MCP.** We verify regularity condition (e) in Wang et al. (2014a) holds for SCAD and MCP in Figure 4: For MCP, we illustrate with Figure 4(a); For SCAD, we illustrate with Figure 4(b) (for $\lambda_2 \geq a\lambda_1$) and Figure 4(c).

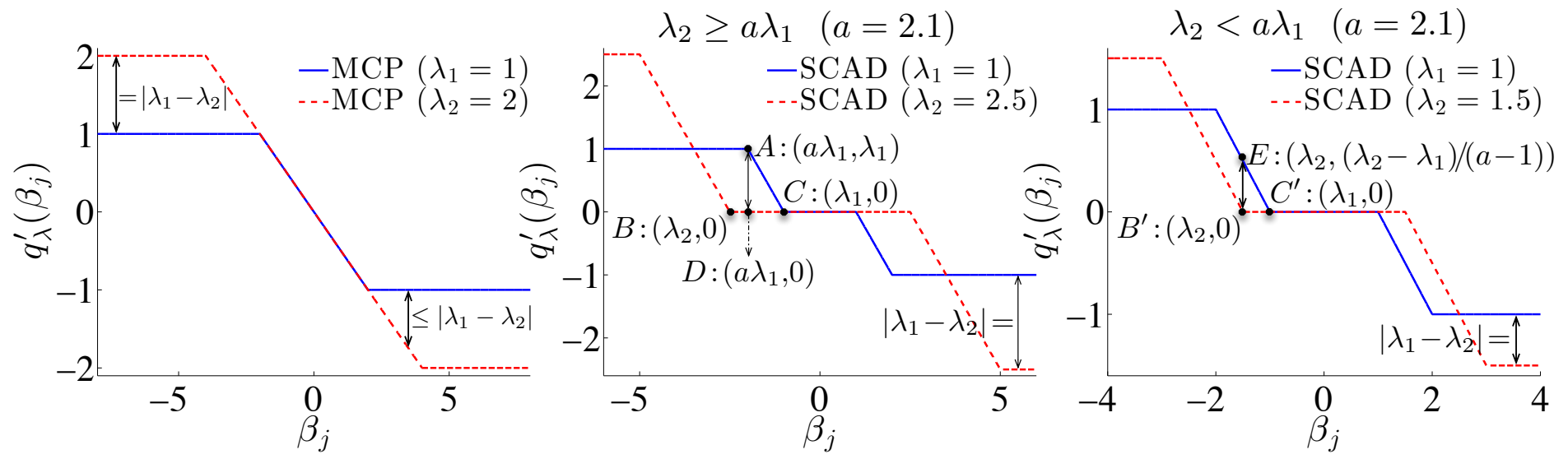

(a) (b) (c)

FIG 4. *An illustration of regularity condition* (e) *in Wang et al. (2014a) for MCP and SCAD: (a) Plots of $q'_{\lambda_1}(\beta_j)$ and $q'_{\lambda_2}(\beta_j)$ for MCP with $\lambda_1 = 1$, $\lambda_2 = 2$ and $b = 2$; (b) Plots of $q'_{\lambda_1}(\beta_j)$ and $q'_{\lambda_2}(\beta_j)$ for SCAD with $\lambda_1 = 1$, $\lambda_2 = 2.5$ and $a = 2.1$; (c) Plots of $q'_{\lambda_1}(\beta_j)$ and $q'_{\lambda_2}(\beta_j)$ for SCAD with $\lambda_1 = 1$, $\lambda_2 = 1.5$ and $a = 2.1$. Subfigure (a) shows that regularity condition* (e) *holds for MCP. For SCAD, we consider two cases: $\lambda_2 \geq a\lambda_1$, as illustrated in (b); $\lambda_2 < a\lambda_1$ as illustrated in (c). In the first case, $|AD| = \lambda_1 \leq (a-1)\lambda_1 \leq |\lambda_1 - \lambda_2|$ since $a > 2$ and $\lambda_2 \geq a\lambda_1$. In the second case, $|B'E| = (\lambda_2 - \lambda_1)/(a-1) \leq |\lambda_1 - \lambda_2|$, because the slope of $EC'$ is $(-1/(a-1))$ with $a > 2$.*

**A.3. Elliptical Distribution.** Before we present the definition of elliptical distribution, we first introduce some notation: If random vectors $\boldsymbol{Z}_1$ and $\boldsymbol{Z}_2$ have the same distribution, we denote by $\boldsymbol{Z}_1 \stackrel{d}{=} \boldsymbol{Z}_2$; The $d$-dimensional $\ell_2$ unit sphere $\{\boldsymbol{v} : \|\boldsymbol{v}\|_2 = 1, \boldsymbol{v} \in \mathbb{R}^d\}$ is denoted by $\mathbb{S}^{d-1}$; For a matrix $\mathbf{M} \in \mathbb{R}^{d\times d}$, we define $\mathrm{diag}(\mathbf{M})$ to be a diagonal matrix with diagonal entries $[\mathrm{diag}(\mathbf{M})]_{jj} = \mathbf{M}_{jj}$ $(j = 1, \ldots, d)$.

**Definition A.1** (Elliptical distribution)**.** For $\boldsymbol{\mu} = (\mu_1, \ldots, \mu_d)^T \in \mathbb{R}^d$ and $\boldsymbol{\Sigma} \in \mathbb{R}^{d\times d}$ with $\mathrm{rank}(\boldsymbol{\Sigma}) = r \leq d$, a random vector $\boldsymbol{W} = (W_1, \ldots, W_d)^T$ follows an elliptical distribution, denoted by $\mathrm{EC}_d(\boldsymbol{\mu}, \boldsymbol{\Sigma}, \Xi)$, if and only if

$$\boldsymbol{W} \stackrel{d}{=} \boldsymbol{\mu} + \Xi \mathbf{A}\boldsymbol{U}.$$

Here $\boldsymbol{U}$ is a random vector uniformly distributed on the unit sphere $\mathbb{S}^{r-1}$; $\Xi \geq 0$ is a scalar random variable independent of $\boldsymbol{U}$; $\mathbf{A} \in \mathbb{R}^{d\times r}$ is a deterministic matrix such that $\mathbf{A}\mathbf{A}^T = \boldsymbol{\Sigma}$. We call $\boldsymbol{\Sigma}$ the scatter matrix. The generalized correlation matrix is defined as $\boldsymbol{\Sigma}^0 = \mathrm{diag}(\boldsymbol{\Sigma})^{-1/2} \cdot \boldsymbol{\Sigma} \cdot \mathrm{diag}(\boldsymbol{\Sigma})^{-1/2}$. When $\mathbb{E}(\Xi^2)$ exists, $\boldsymbol{\Sigma}^0$ is the correlation matrix of $\boldsymbol{W}$.

**Remark A.2.** Note that simultaneously scaling $\Xi$ and $\boldsymbol{U}$ (e.g., $\Xi \to \Xi/C$ and $\boldsymbol{U} \to \boldsymbol{U}/C$, where $C$ is a constant) leads to the same elliptical distribution. To make this model identifiable, we assume $\mu_j = \mathbb{E}(W_j)$ and $\boldsymbol{\Sigma}_{jj} = \mathrm{Var}(W_j)$.

**Remark A.3.** The elliptical distribution family includes a variety of possibly heavy-tail distributions: multivariate Gaussian, multivariate Cauchy, Student's t, logistic, Kotz, symmetric Pearson type-II and type-VII distributions.

**A.4. Rank-based Covariance Matrix Estimation for Semiparametric Elliptical Design Regression.** We provide details of the two-step procedure for estimating the covariance matrix of the elliptically distributed random vector $\boldsymbol{Z} = (Y, \boldsymbol{X}^T)^T \in \mathbb{R}^{d+1}$ discussed in §2.2 of Wang et al. (2014a):

**Elliptical Covariance Matrix Estimation**

**S1.** First, we define a rank-based estimator $\widehat{\mathbf{R}}_{\boldsymbol{Z}}$ of the generalized correlation matrix $\boldsymbol{\Sigma}^0_{\boldsymbol{Z}}$ using the Kendall's tau statistic. Let $\mathbf{z}_1, \ldots, \mathbf{z}_n \in \mathbb{R}^{d+1}$ with $\mathbf{z}_i = (z_{i1}, \ldots, z_{i(d+1)})^T$ be $n$ independent observations of $\boldsymbol{Z}$. The Kendall's tau correlation coefficient is defined as

$$\widehat{\tau}_{jk}(\mathbf{z}_1, \ldots, \mathbf{z}_n) = \begin{cases} \displaystyle\sum_{1 \le i < i' < n} \frac{2\,\mathrm{sign}\left(z_{ij} - z_{i'j}\right) \mathrm{sign}\left(z_{ik} - z_{i'k}\right)}{n(n-1)}, & \text{for } j \neq k, \\ 1, & \text{for } j = k. \end{cases}$$

We define the Kendall's tau correlation matrix estimator as

$$\widehat{\mathbf{R}}_{\boldsymbol{Z}} = \left[\left(\widehat{\mathbf{R}}_{\boldsymbol{Z}}\right)_{jk}\right] = \left[\sin\left(\frac{\pi}{2}\widehat{\tau}_{jk}\left(\mathbf{z}_1, \ldots, \mathbf{z}_n\right)\right)\right]. \tag{A.1}$$

Liu et al. (2012); Han and Liu (2012, 2013) showed that $\widehat{\mathbf{R}}_{\boldsymbol{Z}}$ is a robust estimator of the population generalized correlation matrix $\boldsymbol{\Sigma}^0_{\boldsymbol{Z}}$, and is invariant to different distributions of the generating variable $\Xi$ within the whole elliptical family.

**S2.** Second, we construct a covariance matrix estimator

$$\widehat{\mathbf{K}}_{\boldsymbol{Z}} = \left[\left(\widehat{\mathbf{K}}_{\boldsymbol{Z}}\right)_{jk}\right] = \left[\left(\widehat{\mathbf{R}}_{\boldsymbol{Z}}\right)_{jk} \cdot \widehat{\sigma}_j \widehat{\sigma}_k\right], \tag{A.2}$$

where $\widehat{\sigma}_1, \ldots, \widehat{\sigma}_{d+1}$ are the estimators of the standard deviations of $Z_1, \ldots, Z_{d+1}$. We calculate $\widehat{\sigma}_1, \ldots, \widehat{\sigma}_{d+1}$ using the Catoni's $M$-estimator (Catoni, 2012) described in §E. The main advantage of the Cantoni's estimator is that, for a fixed level of confidence, it achieves the same deviation behavior as a Gaussian random variable under a weak moment condition.

## APPENDIX B: PROXIMAL-GRADIENT METHOD FOR NONCONVEX PROBLEMS

We provide details of the proximal-gradient method tailored to nonconvex problems. In particular, in §B.1 we provide the optimization update schemes for the specific nonconvex problems discussed in §2 of Wang et al. (2014a). In §B.2 we provide the detailed derivation of the closed-form expression of update scheme (3.9) in Wang et al. (2014a).

**B.1. Optimization Update Schemes for Specific Nonconvex Problems.** To obtain the specific optimization update schemes of the proximal-gradient method for the nonconvex problems discussed in §2 of Wang et al. (2014a), we only need to plug the following specific definitions of $\nabla\mathcal{L}(\boldsymbol{\beta})$ and $\nabla\mathcal{Q}_{\lambda_t}(\boldsymbol{\beta})$ into (3.11) of Wang et al. (2014a):

- For the (nonconvex) loss functions discussed in §2 of Wang et al. (2014a),

$$\nabla\mathcal{L}(\boldsymbol{\beta}) = \begin{cases} \dfrac{1}{n}\mathbf{X}^T\big(\mathbf{X}\boldsymbol{\beta}-\mathbf{y}\big), & \text{least squares loss,}\\ \dfrac{1}{n}\displaystyle\sum_{i=1}^{n}\mathbf{x}_i\left(\dfrac{\exp\big(\mathbf{x}_i^T\boldsymbol{\beta}\big)}{1+\exp\big(\mathbf{x}_i^T\boldsymbol{\beta}\big)}-y_i\right), & \text{logistic loss,}\\ \widehat{\mathbf{K}}_{\boldsymbol{X}}\boldsymbol{\beta}-\widehat{\mathbf{K}}_{\boldsymbol{X},Y}, & \text{semiparametric elliptical design loss,}\end{cases}$$

  where $\widehat{\mathbf{K}}_{\boldsymbol{X}}\in\mathbb{R}^{d\times d}$ and $\widehat{\mathbf{K}}_{\boldsymbol{X},Y}\in\mathbb{R}^{d\times 1}$ are the submatrices of $\widehat{\mathbf{K}}_{\boldsymbol{Z}}$, which is the semiparametric elliptical covariance matrix estimator defined in (A.2). More specifically,

$$\widehat{\mathbf{K}}_{\boldsymbol{Z}} = \begin{pmatrix} \widehat{\mathbf{K}}_{Y} & \widehat{\mathbf{K}}_{\boldsymbol{X},Y}^T\\ \widehat{\mathbf{K}}_{\boldsymbol{X},Y} & \widehat{\mathbf{K}}_{\boldsymbol{X}}\end{pmatrix}. \tag{B.1}$$

- For the nonconvex penalty functions discussed in §2 of Wang et al. (2014a),

$$\big(\nabla\mathcal{Q}_{\lambda_t}(\boldsymbol{\beta})\big)_j = \begin{cases} \dfrac{\lambda_t\,\mathrm{sign}(\beta_j)-\beta_j}{a-1}\cdot\mathbb{1}\big(\lambda_t<|\beta_j|\le a\lambda_t\big)\\ \qquad\qquad -\lambda_t\,\mathrm{sign}(\beta_j)\cdot\mathbb{1}\big(|\beta_j|>a\lambda_t\big), & \text{SCAD,}\\ -\dfrac{\beta_j}{b}\lambda_t\,\mathrm{sign}(\beta_j)\cdot\mathbb{1}\big(|\beta_j|\le b\lambda_t\big)\\ \qquad\qquad -\lambda_t\,\mathrm{sign}(\beta_j)\cdot\mathbb{1}\big(|\beta_j|>b\lambda_t\big), & \text{MCP,}\end{cases}$$

  where $a>2$, $b>0$.

**B.2. Derivation of Optimization Update Schemes.** For notational simplicity, we denote $L_t^k$ by $L$, $\boldsymbol{\beta}_t^{k-1}$ by $\boldsymbol{\beta}'$, and $\lambda_t$ by $\lambda$ in the rest of this section.

**Derivation of (3.10) of Wang et al. (2014a):** If $\Omega = \mathbb{R}^d$, then we have

$$\begin{aligned}\mathcal{T}_{L,\lambda}(\boldsymbol{\beta}'; +\infty) &= \operatorname*{argmin}_{\boldsymbol{\beta}\in\mathbb{R}^d} \left\{\psi_{L,\lambda}(\boldsymbol{\beta};\boldsymbol{\beta}')\right\} \\ &= \operatorname*{argmin}_{\boldsymbol{\beta}\in\mathbb{R}^d} \left\{\widetilde{\mathcal{L}}_\lambda(\boldsymbol{\beta}') + \nabla\widetilde{\mathcal{L}}_\lambda(\boldsymbol{\beta}')^T(\boldsymbol{\beta}-\boldsymbol{\beta}') + \frac{L}{2}\|\boldsymbol{\beta}-\boldsymbol{\beta}'\|_2^2 + \lambda\|\boldsymbol{\beta}\|_1\right\} \\ &= \operatorname*{argmin}_{\boldsymbol{\beta}\in\mathbb{R}^d} \left\{\frac{1}{2}\Bigg\|\boldsymbol{\beta} - \underbrace{\left(\boldsymbol{\beta}' - \frac{1}{L}\nabla\widetilde{\mathcal{L}}_\lambda(\boldsymbol{\beta}')\right)}_{\bar{\boldsymbol{\beta}}}\Bigg\|_2^2 + \frac{\lambda}{L}\|\boldsymbol{\beta}\|_1\right\}. \qquad \text{(B.2)}\end{aligned}$$

It is known that the minimizer of (B.2) can be obtained by soft-thresholding $\bar{\boldsymbol{\beta}}$ with the threshold of value $\lambda/L$, i.e.,

$$\big(\mathcal{T}_{L,\lambda}(\boldsymbol{\beta}'; +\infty)\big)_j = \begin{cases} 0 & \text{if } |\bar{\beta}_j| \le \lambda/L, \\ \operatorname{sign}(\bar{\beta}_j)(|\bar{\beta}_j| - \lambda/L) & \text{if } |\bar{\beta}_j| > \lambda/L. \end{cases} \qquad \text{(B.3)}$$

Therefore we obtain the first update scheme (3.10) in Wang et al. (2014a) for $\Omega = \mathbb{R}^d$.

**Derivation of (3.12) of Wang et al. (2014a):** If $\Omega = B_2(R) = \big\{\boldsymbol{\beta} : \|\boldsymbol{\beta}\|_2^2 \le R^2\big\}$, by Lagrangian duality we can transform the original optimization problem with constraint into an unconstraint optimization problem. Hence, there exists a Lagrangian multiplier $\tau \ge 0$ such that

$$\mathcal{T}_{L,\lambda}(\boldsymbol{\beta}'; R) = \operatorname*{argmin}_{\boldsymbol{\beta}\in B_2(R)} \left\{\psi_{L,\lambda}(\boldsymbol{\beta};\boldsymbol{\beta}')\right\} = \operatorname*{argmin}_{\boldsymbol{\beta}\in\mathbb{R}^d} \left\{\psi_{L,\lambda}(\boldsymbol{\beta};\boldsymbol{\beta}') + \frac{\tau}{2}\|\boldsymbol{\beta}\|_2^2\right\}.$$

Consequently, based on (B.2) we have

$$\begin{aligned}\mathcal{T}_{L,\lambda}(\boldsymbol{\beta}'; R) &= \operatorname*{argmin}_{\boldsymbol{\beta}\in\mathbb{R}^d} \Bigg\{\widetilde{\mathcal{L}}_\lambda(\boldsymbol{\beta}') + \nabla\widetilde{\mathcal{L}}_\lambda(\boldsymbol{\beta}')^T(\boldsymbol{\beta}-\boldsymbol{\beta}') \\ &\qquad\qquad + \frac{L}{2}\|\boldsymbol{\beta}-\boldsymbol{\beta}'\|_2^2 + \lambda\|\boldsymbol{\beta}\|_1 + \frac{\tau}{2}\|\boldsymbol{\beta}\|_2^2\Bigg\} \\ &= \operatorname*{argmin}_{\boldsymbol{\beta}\in\mathbb{R}^d} \left\{\frac{L+\tau}{2}\|\boldsymbol{\beta}\|_2^2 - \Big(L\cdot\boldsymbol{\beta}' - \nabla\widetilde{\mathcal{L}}_\lambda(\boldsymbol{\beta}')\Big)^T\boldsymbol{\beta} + \lambda\|\boldsymbol{\beta}\|_1\right\} \\ &= \operatorname*{argmin}_{\boldsymbol{\beta}\in\mathbb{R}^d} \left\{\frac{1}{2}\Bigg\|\boldsymbol{\beta} - \underbrace{\left(\frac{L}{L+\tau}\boldsymbol{\beta}' - \frac{1}{L+\tau}\nabla\widetilde{\mathcal{L}}_\lambda(\boldsymbol{\beta}')\right)}_{\frac{L}{L+\tau}\bar{\boldsymbol{\beta}}}\Bigg\|_2^2 + \frac{\lambda}{L+\tau}\|\boldsymbol{\beta}\|_1\right\}, \qquad \text{(B.4)}\end{aligned}$$

where $\bar{\boldsymbol{\beta}} = \boldsymbol{\beta}' - \nabla\widetilde{\mathcal{L}}_\lambda(\boldsymbol{\beta}')/L$. The minimizer of (B.4) can also be obtained by soft-thresholding, i.e.,

$$\left(\mathcal{T}_{L,\lambda}(\boldsymbol{\beta}';R)\right)_j = \begin{cases} 0 & \text{if } \frac{L}{L+\tau}|\bar{\beta}_j| \le \frac{\lambda}{L+\tau}, \\ \operatorname{sign}\left(\frac{L}{L+\tau}\bar{\beta}_j\right)\left(\frac{L}{L+\tau}|\bar{\beta}_j| - \frac{\lambda}{L+\tau}\right) & \text{if } \frac{L}{L+\tau}|\bar{\beta}_j| > \frac{\lambda}{L+\tau}. \end{cases} \tag{B.5}$$

Comparing (B.5) with (B.3), we have

$$\mathcal{T}_{L,\lambda}(\boldsymbol{\beta}';R) = \frac{L}{L+\tau}\mathcal{T}_{L,\lambda}(\boldsymbol{\beta}';+\infty). \tag{B.6}$$

Thus, we can obtain the constraint solution $\mathcal{T}_{L,\lambda}(\boldsymbol{\beta}';R)$ by first calculating the unconstraint solution $\mathcal{T}_{L,\lambda}(\boldsymbol{\beta}';+\infty)$, and then rescaling it by a factor of $L/(L+\tau)$. Note that here the Lagrangian multiplier $\tau$ is unknown. We discuss the following two cases:

- If the constraint $\boldsymbol{\beta} \in B_2(R)$ is inactive, then we have $\tau = 0$ by complementary slackness, which implies $\mathcal{T}_{L,\lambda}(\boldsymbol{\beta}';R) = \mathcal{T}_{L,\lambda}(\boldsymbol{\beta}';+\infty)$. Since the constraint is inactive, we have $\|\mathcal{T}_{L,\lambda}(\boldsymbol{\beta}';R)\|_2 = \|\mathcal{T}_{L,\lambda}(\boldsymbol{\beta}';+\infty)\|_2 < R$.
- If the constraint $\boldsymbol{\beta} \in B_2(R)$ is active, then we have $\tau \ge 0$ by complementary slackness. In this case, the minimizer $\mathcal{T}_{L,\lambda}(\boldsymbol{\beta}';R)$ lies on the boundary of $B_2(R)$. By (B.6) we have
$$\|\mathcal{T}_{L,\lambda}(\boldsymbol{\beta}';+\infty)\|_2 = \frac{L+\tau}{L}\|\mathcal{T}_{L,\lambda}(\boldsymbol{\beta}';R)\|_2 = \frac{L+\tau}{L}R \ge R.$$
To obtain $\mathcal{T}_{L,\lambda}(\boldsymbol{\beta}';R)$, we project $\mathcal{T}_{L,\lambda}(\boldsymbol{\beta}';+\infty)$ onto $B_2(R)$, which can be achieved by setting $\mathcal{T}_{L,\lambda}(\boldsymbol{\beta}';R) = R\cdot\mathcal{T}_{L,\lambda}(\boldsymbol{\beta}';+\infty)/\|\mathcal{T}_{L,\lambda}(\boldsymbol{\beta}';+\infty)\|_2$.

Therefore we obtain the second update scheme (3.12) in Wang et al. (2014a) for $\Omega = B_2(R)$.

## APPENDIX C: JUSTIFICATION OF ASSUMPTIONS

In §C.1 and §C.2, we prove that Assumption 4.1 and Assumption 4.4 in Wang et al. (2014a) hold with high probability respectively.

### C.1. Justification of Assumption 4.1 in Wang et al. (2014a).

Recall that Assumption 4.1 states that $\|\nabla\mathcal{L}(\boldsymbol{\beta}^*)\|_\infty$ can be upper bounded by $\lambda_{\text{tgt}}/8$. First we provide two lemmas on the upper bound of $\|\nabla\mathcal{L}(\boldsymbol{\beta}^*)\|_\infty$.

**Lemma C.1.** For least squares regression with sub-Gaussian noise and logistic regression, we assume that the columns of $\mathbf{X}$ are normalized in such a way that $\max_{j\in\{1,\ldots,d\}}\{\|\mathbf{X}_j\|_2\} \le \sqrt{n}$. Then we have

$$\|\nabla\mathcal{L}(\boldsymbol{\beta}^*)\|_\infty \le C\sqrt{\frac{\log d}{n}} \tag{C.1}$$

with probability at least $1 - d^{-1}$, where $C$ is a constant.

Proof. See Candés and Tao (2007); Zhang and Huang (2008); Zhang (2009); Bickel et al. (2009); Koltchinskii (2009a); Negahban et al. (2012); Wainwright (2009) for a detailed proof. □

**Lemma C.2.** For semiparametric elliptical design regression, we have

$$\|\nabla\mathcal{L}(\boldsymbol{\beta}^*)\|_\infty \le C\|\boldsymbol{\beta}^*\|_1\sqrt{\frac{\log d}{n}} \tag{C.2}$$

with probability at least $1 - (d+1)^{-5/2} - 2(d+1)^{-3}$, where $C$ is a constant.

Proof. See §E.3 for a detailed proof. □

Recall in Assumption 4.1 of Wang et al. (2014a) we set $\lambda_{\rm tgt} = C\sqrt{\log d/n}$ for least squares and logistic loss, and $\lambda_{\rm tgt} = C'\|\boldsymbol{\beta}^*\|_1\sqrt{\log d/n}$ for semiparametric elliptical design loss. Thus, according to Lemma C.1 and Lemma C.2, $\|\nabla\mathcal{L}(\boldsymbol{\beta}^*)\|_\infty \le \lambda_{\rm tgt}/8$ holds with high probability, i.e., Assumption 4.1 holds with high probability.

**C.2. Justification of Assumption 4.4 in Wang et al. (2014a).** In this section we show that, for semiparametric elliptical design loss and logistic loss, Assumption 4.4 holds with high probability.

**Semiparametric Elliptical Design Loss:** First we provide the following lemma on the largest and smallest sparse eigenvalues of the Hessian matrix $\nabla^2\mathcal{L}(\boldsymbol{\beta})$ of semiparametric elliptical design loss.

Let $n$ be the sample size, $d$ be the dimension of $\boldsymbol{\beta}$, and $\boldsymbol{Z} \in \mathbb{R}^{d+1}$ be the elliptically distributed random vector in §2.2 of Wang et al. (2014a). The corresponding covariance matrix estimator $\widehat{\mathbf{K}}_{\boldsymbol{Z}} \in \mathbb{R}^{(d+1)\times(d+1)}$ is defined in (A.2), while its submatrix $\widehat{\mathbf{K}}_{\boldsymbol{X}} \in \mathbb{R}^{d\times d}$ is defined in (B.1). Hence, the Hessian matrix of semiparametric elliptical design loss is $\nabla^2\mathcal{L}(\boldsymbol{\beta}) = \widehat{\mathbf{K}}_{\boldsymbol{X}}$. Let $s$ the sparsity level.

**Lemma C.3.** Under suitable conditions (see Han and Liu (2013) for details), for a sufficiently large $n$, there exists an $s$ such that $\rho_-\big(\nabla^2\mathcal{L}, s\big) > 0$ and $\rho_+\big(\nabla^2\mathcal{L}, s\big) < +\infty$ with probability at least $1-4d^{-1}-6d^{-2}$. Here $\rho_+\big(\nabla^2\mathcal{L}, s\big)$ and $\rho_-\big(\nabla^2\mathcal{L}, s\big)$ are defined in Definition 4.2 of Wang et al. (2014a).

PROOF. See §E.2 for a detailed proof. □

Equipped with Lemma C.3, we can justify Assumption 4.4. Recall $s^* = \|\boldsymbol{\beta}^*\|_0$, where $\boldsymbol{\beta}^*$ is the true parameter vector. Suppose that Lemma C.3 holds with $s = Cs^*$, $\rho_+\big(\nabla^2\mathcal{L}, s\big) = C'$ and $\rho_-\big(\nabla^2\mathcal{L}, s\big) = C''$, where $C$ satisfies

$$C \geq 2\left(144 \cdot \left(\frac{2C'}{C''}\right)^2 + 250 \cdot \left(\frac{2C'}{C''}\right)\right) + 1. \tag{C.3}$$

Meanwhile, let the concavity parameters of the nonconvex penalty be $\zeta_+ = 0$ and $\zeta_- = C''/2$. In the sequel we verify that, there exists an integer $\widetilde{s} = (C-1)/2 \cdot s^*$, where $C$ satisfies (C.3), that satisfies Assumption 4.4. Note that the condition number $\kappa$ defined in (4.5) is

$$\begin{aligned}\kappa = \frac{\rho_+\big(\nabla^2\mathcal{L}, s^* + 2\widetilde{s}\big) - \zeta_+}{\rho_-\big(\nabla^2\mathcal{L}, s^* + 2\widetilde{s}\big) - \zeta_-} = \frac{\rho_+\big(\nabla^2\mathcal{L}, Cs^*\big) - \zeta_+}{\rho_-\big(\nabla^2\mathcal{L}, Cs^*\big) - \zeta_-} &= \frac{\rho_+\big(\nabla^2\mathcal{L}, s\big) - \zeta_+}{\rho_-\big(\nabla^2\mathcal{L}, s\big) - \zeta_-} \\ &= \frac{C'}{C'' - C''/2} = \frac{2C'}{C''}.\end{aligned}$$

Since $\widetilde{s} = (C-1)/2 \cdot s^*$ where $C$ satisfies (C.3), we have

$$\widetilde{s} \geq \left(144 \cdot \left(\frac{2C'}{C''}\right)^2 + 250 \cdot \left(\frac{2C'}{C''}\right)\right) \cdot s^* = \big(144\kappa^2 + 250\kappa\big) \cdot s^*.$$

Thus we find an $\widetilde{s}$ that satisfies the requirements in Assumption 4.4. Therefore, Assumption 4.4 in Wang et al. (2014a) holds with probability at least $1 - 4d^{-1} - 6d^{-2}$.

**Logistic Loss:** Remind that, the Hessian matrix of logistic loss is defined in (4.2) of Wang et al. (2014a), while its sparse eigenvalues $\rho_-\big(\nabla^2\mathcal{L}, s, R\big)$ and $\rho_+\big(\nabla^2\mathcal{L}, s, R\big)$ are defined in Definition 4.3 of Wang et al. (2014a), where $R \in (0, +\infty)$ is an absolute constant.

In the sequel, we show that Assumption 4.4 is a weaker assumption than the assumption of restricted strong convexity and smoothness imposed by Loh and Wainwright (2013). Since they proved that their assumption holds with high probability, Assumption 4.4 also holds with high probability.

In fact, it suffices to show if the assumption of restricted strong convexity and smoothness holds, then for a sufficiently large $n$, there exists an $s$ such that $\rho_-\big(\nabla^2\mathcal{L}, s, R\big) > 0$ and $\rho_+\big(\nabla^2\mathcal{L}, s, R\big) < +\infty$. With this claim, we can justify Assumption 4.4 following the same argument as after Lemma C.3. Now we prove the previous claim.

Loh and Wainwright (2013) imposed the following assumption: For $\boldsymbol{\beta}, \boldsymbol{\beta}' \in$

$\mathbb{R}^d$ such that $\|\boldsymbol{\beta}\|_2 \le R$ and $\|\boldsymbol{\beta}'\|_2 \le R$, $\mathcal{L}(\boldsymbol{\beta})$ satisfies

$$\mathcal{L}(\boldsymbol{\beta}')-\mathcal{L}(\boldsymbol{\beta})-\nabla\mathcal{L}(\boldsymbol{\beta})^T(\boldsymbol{\beta}'-\boldsymbol{\beta}) \le C\|\boldsymbol{\beta}-\boldsymbol{\beta}'\|_2^2 + C'\cdot\frac{\log d}{n}\|\boldsymbol{\beta}-\boldsymbol{\beta}'\|_1^2, \tag{C.4}$$

$$\mathcal{L}(\boldsymbol{\beta}')-\mathcal{L}(\boldsymbol{\beta})-\nabla\mathcal{L}(\boldsymbol{\beta})^T(\boldsymbol{\beta}'-\boldsymbol{\beta}) \ge C''\|\boldsymbol{\beta}-\boldsymbol{\beta}'\|_2^2 - C'''\cdot\frac{\log d}{n}\|\boldsymbol{\beta}-\boldsymbol{\beta}'\|_1^2. \tag{C.5}$$

Here all the constants are positive. See equations (28) and (29) of their paper for details.

By Taylor's theorem and the mean value theorem, we have

$$\mathcal{L}(\boldsymbol{\beta}') = \mathcal{L}(\boldsymbol{\beta}) + \nabla\mathcal{L}(\boldsymbol{\beta})^T(\boldsymbol{\beta}'-\boldsymbol{\beta}) + \frac{1}{2}(\boldsymbol{\beta}'-\boldsymbol{\beta})^T\nabla^2\mathcal{L}\big(\gamma\boldsymbol{\beta}' + (1-\gamma)\boldsymbol{\beta}\big)(\boldsymbol{\beta}'-\boldsymbol{\beta})$$

for some $\gamma \in [0,1]$. Plugging this into the left-hand sides of (C.4) and (C.5), we have

$$\frac{1}{2}(\boldsymbol{\beta}'-\boldsymbol{\beta})^T\nabla^2\mathcal{L}\big(\gamma\boldsymbol{\beta}'+(1-\gamma)\boldsymbol{\beta}\big)(\boldsymbol{\beta}'-\boldsymbol{\beta}) \le C\|\boldsymbol{\beta}'-\boldsymbol{\beta}\|_2^2 + C'\cdot\frac{\log d}{n}\|\boldsymbol{\beta}'-\boldsymbol{\beta}\|_1^2, \tag{C.6}$$

$$\frac{1}{2}(\boldsymbol{\beta}'-\boldsymbol{\beta})^T\nabla^2\mathcal{L}\big(\gamma\boldsymbol{\beta}'+(1-\gamma)\boldsymbol{\beta}\big)(\boldsymbol{\beta}'-\boldsymbol{\beta}) \ge C''\|\boldsymbol{\beta}'-\boldsymbol{\beta}\|_2^2 - C'''\cdot\frac{\log d}{n}\|\boldsymbol{\beta}'-\boldsymbol{\beta}\|_1^2. \tag{C.7}$$

Suppose $\boldsymbol{\beta}$ and $\boldsymbol{\beta}'$ satisfy $\|\boldsymbol{\beta}'-\boldsymbol{\beta}\|_0 \le s$, which implies $\|\boldsymbol{\beta}'-\boldsymbol{\beta}\|_1 \le \sqrt{s}\cdot\|\boldsymbol{\beta}'-\boldsymbol{\beta}\|_2$. Plugging this upper bound of $\|\boldsymbol{\beta}'-\boldsymbol{\beta}\|_1$ into the right-hand sides of (C.6) and (C.7), we have

$$\frac{1}{2}(\boldsymbol{\beta}'-\boldsymbol{\beta})^T\nabla^2\mathcal{L}\big(\gamma\boldsymbol{\beta}'+(1-\gamma)\boldsymbol{\beta}\big)(\boldsymbol{\beta}'-\boldsymbol{\beta}) \le \left(C + C'\cdot\frac{s\log d}{n}\right)\cdot\|\boldsymbol{\beta}'-\boldsymbol{\beta}\|_2^2, \tag{C.8}$$

$$\frac{1}{2}(\boldsymbol{\beta}'-\boldsymbol{\beta})^T\nabla^2\mathcal{L}\big(\gamma\boldsymbol{\beta}'+(1-\gamma)\boldsymbol{\beta}\big)(\boldsymbol{\beta}'-\boldsymbol{\beta}) \ge \left(C'' - C'''\cdot\frac{s\log d}{n}\right)\cdot\|\boldsymbol{\beta}'-\boldsymbol{\beta}\|_2^2. \tag{C.9}$$

In (C.8) and (C.9), taking $n \ge \max\{2C'''/C'', 2C'/C\}\cdot s\log d/n$, and dividing $\|\boldsymbol{\beta}'-\boldsymbol{\beta}\|_2^2$ on both sides, we obtain

$$\frac{C''}{2} \le \frac{1}{2}\cdot\frac{(\boldsymbol{\beta}'-\boldsymbol{\beta})^T}{\|\boldsymbol{\beta}'-\boldsymbol{\beta}\|_2}\cdot\nabla^2\mathcal{L}\big(\gamma\boldsymbol{\beta}'+(1-\gamma)\boldsymbol{\beta}\big)\cdot\frac{(\boldsymbol{\beta}'-\boldsymbol{\beta})}{\|\boldsymbol{\beta}'-\boldsymbol{\beta}\|_2} \le \frac{3C}{2}. \tag{C.10}$$

Let $\boldsymbol{v} = (\boldsymbol{\beta}'-\boldsymbol{\beta})/\|\boldsymbol{\beta}'-\boldsymbol{\beta}\|_2$. Obviously, $\boldsymbol{v}$ is an arbitrary vector that satisfies $\|\boldsymbol{v}\|_2 = 1$ and $\|\boldsymbol{v}\|_0 \le s$. Taking $\boldsymbol{\beta}' \to \boldsymbol{\beta}$, we have $C'' \le \boldsymbol{v}^T\nabla^2\mathcal{L}(\boldsymbol{\beta})\boldsymbol{v} \le 3C$ for any $\boldsymbol{\beta} \le R$ and any $\boldsymbol{v}$ such that $\|\boldsymbol{v}\|_2 = 1$ and $\|\boldsymbol{v}\|_0 \le s$. By Definition 4.3, we have $\rho_-\big(\nabla^2\mathcal{L}, s, R\big) \ge C' > 0$ and $\rho_+\big(\nabla^2\mathcal{L}, s, R\big) \le 3C < +\infty$.

Therefore, if their assumption holds, then for a sufficiently large $n$, there exists some $s$ such that $\rho_-\big(\nabla^2\mathcal{L}, s, R\big) > 0$ and $\rho_+\big(\nabla^2\mathcal{L}, s, R\big) < +\infty$. Following the same argument as after Lemma C.3, we can show that Assumption 4.4 also holds true. That is to say, Assumption 4.4 of Wang et al. (2014a) is a weaker assumption.

Because their assumption holds with high probability for generalized linear models (not including Poisson model; see Proposition 1 of their paper) with sub-Gaussian design, our Assumption 4.4 holds with high probability in the same setting, including logistic loss as a special case in our paper.

## APPENDIX D: PROOF OF THEORETICAL RESULTS

To analyze the computational properties of our approximate regularization path following method, we first provide several useful lemmas on the proximal-gradient method that is used within each stage of the path following method.

**D.1. Preliminary Results about Proximal-Gradient Method.** Recall that the objective function can be formulated as $\phi_{\lambda_t}(\boldsymbol{\beta}) = \widetilde{\mathcal{L}}_{\lambda_t}(\boldsymbol{\beta}) + \lambda_t \|\boldsymbol{\beta}\|_1$ where $\widetilde{\mathcal{L}}_{\lambda_t}(\boldsymbol{\beta}) = \mathcal{L}(\boldsymbol{\beta}) + \mathcal{Q}_{\lambda_t}(\boldsymbol{\beta})$, while $\psi_{L_t^k,\lambda_t}\big(\boldsymbol{\beta};\boldsymbol{\beta}_t^{k-1}\big)$ is the local quadratic approximation of $\phi_{\lambda_t}(\boldsymbol{\beta})$ at $\boldsymbol{\beta}_t^{k-1}$, as defined in (3.7) of Wang et al. (2014a). The following lemma, which is adapted from Nesterov (2013), characterizes the decrement of the objective function.

**Lemma D.1.** Under Assumption 4.4, we assume $\big\|\big(\boldsymbol{\beta}_t^{k-1}\big)_{\overline{S^*}}\big\|_0 \leq \widetilde{s}$, where $\widetilde{s}$ is the positive integer specified in Assumption 4.4. For any $L_t^k > 0$ and fixed $\lambda_t \in [\lambda_{\text{tgt}}, \lambda_0]$, we have

$$\phi_{\lambda_t}\big(\boldsymbol{\beta}_t^k\big) \leq \phi_{\lambda_t}\big(\boldsymbol{\beta}_t^{k-1}\big) - \frac{L_t^k}{2}\big\|\boldsymbol{\beta}_t^k - \boldsymbol{\beta}_t^{k-1}\big\|_2^2.$$

Recall that, as defined in (3.15) of Wang et al. (2014a), $\omega_\lambda(\boldsymbol{\beta})$ characterizes the suboptimality of approximate solutions. The next lemma, which is also adapted from Nesterov (2013), provides an upper bound of $\omega_{\lambda_t}\big(\boldsymbol{\beta}_t^k\big)$ using $\big\|\boldsymbol{\beta}_t^k - \boldsymbol{\beta}_t^{k-1}\big\|_2$.

**Lemma D.2.** Under the assumptions of Lemma D.1, then we have

$$\omega_{\lambda_t}\big(\boldsymbol{\beta}_t^k\big) \leq \big(L_t^k + \rho_+ - \zeta_-\big)\big\|\boldsymbol{\beta}_t^k - \boldsymbol{\beta}_t^{k-1}\big\|_2,$$

where $\rho_+ = \rho_+\big(\nabla^2\mathcal{L}, s^* + 2\widetilde{s}\big)$ is the sparse eigenvalue specified in Assumption 4.4; As defined in regularity condition (a), $\zeta_+ > 0$ is the concavity parameter of the nonconvex penalty, which satisfies (4.6) in Wang et al. (2014a).

**D.2. Proof of Lemma 5.1 in Wang et al. (2014a).**

PROOF. Recall that $\mathcal{Q}_\lambda(\boldsymbol{\beta})$ is the concave component of the nonconvex penalty $\mathcal{P}_\lambda(\boldsymbol{\beta})$, which implies $-\mathcal{Q}_\lambda(\boldsymbol{\beta})$ is convex. Meanwhile, recall that $\mathcal{Q}_\lambda(\boldsymbol{\beta}) = \sum_{j=1}^d q_\lambda(\beta_j)$, where $q_\lambda(\beta_j)$ satisfies regularity condition (a) in Wang et al. (2014a). Hence we have

$$-\zeta_-(\beta_j' - \beta_j)^2 \le \left(q_\lambda'(\beta_j') - q_\lambda'(\beta_j)\right)(\beta_j' - \beta_j) \le -\zeta_+(\beta_j' - \beta_j)^2,$$

which implies the convex function $-\mathcal{Q}_\lambda(\boldsymbol{\beta})$ satisfies

$$\text{(D.1)} \qquad \Big(\nabla\big(-\mathcal{Q}_\lambda(\boldsymbol{\beta}')\big) - \nabla\big(-\mathcal{Q}_\lambda(\boldsymbol{\beta})\big)\Big)^T(\boldsymbol{\beta}' - \boldsymbol{\beta}) \le \zeta_-\|\boldsymbol{\beta}' - \boldsymbol{\beta}\|_2^2,$$

$$\text{(D.2)} \qquad \Big(\nabla\big(-\mathcal{Q}_\lambda(\boldsymbol{\beta}')\big) - \nabla\big(-\mathcal{Q}_\lambda(\boldsymbol{\beta})\big)\Big)^T(\boldsymbol{\beta}' - \boldsymbol{\beta}) \ge \zeta_+\|\boldsymbol{\beta}' - \boldsymbol{\beta}\|_2^2.$$

According to Nesterov (2004, Theorem 2.1.5 & Theorem 2.1.9), (D.1) and (D.2) are equivalent definitions of strong smoothness and strong convexity respectively. In other words, $-\mathcal{Q}_\lambda(\boldsymbol{\beta})$ satisfies

$$\text{(D.3)} \qquad -\mathcal{Q}_\lambda(\boldsymbol{\beta}') \le -\mathcal{Q}_\lambda(\boldsymbol{\beta}) - \nabla\mathcal{Q}(\boldsymbol{\beta})^T(\boldsymbol{\beta}' - \boldsymbol{\beta}) + \frac{\zeta_-}{2}\|\boldsymbol{\beta}' - \boldsymbol{\beta}\|_2^2,$$

$$\text{(D.4)} \qquad -\mathcal{Q}_\lambda(\boldsymbol{\beta}') \ge -\mathcal{Q}_\lambda(\boldsymbol{\beta}) - \nabla\mathcal{Q}(\boldsymbol{\beta})^T(\boldsymbol{\beta}' - \boldsymbol{\beta}) + \frac{\zeta_+}{2}\|\boldsymbol{\beta}' - \boldsymbol{\beta}\|_2^2.$$

For loss function $\mathcal{L}(\boldsymbol{\beta})$, by Taylor's theorem and the mean value theorem, we have

$$\text{(D.5)} \qquad \begin{aligned}\mathcal{L}(\boldsymbol{\beta}') =& \mathcal{L}(\boldsymbol{\beta}) + \nabla\mathcal{L}(\boldsymbol{\beta})^T(\boldsymbol{\beta}' - \boldsymbol{\beta}) \\ &+ \frac{1}{2}(\boldsymbol{\beta}' - \boldsymbol{\beta})^T\nabla^2\mathcal{L}\big(\gamma\boldsymbol{\beta} + (1-\gamma)\boldsymbol{\beta}'\big)(\boldsymbol{\beta}' - \boldsymbol{\beta}),\end{aligned}$$

where $\gamma \in [0,1]$. Note that we assume $\|(\boldsymbol{\beta}' - \boldsymbol{\beta})_{\overline{S^*}}\|_0 \le 2\widetilde{s}$, which implies $\|\boldsymbol{\beta}' - \boldsymbol{\beta}\|_0 \le s^* + 2\widetilde{s}$. For logistic loss, we assume $\|\boldsymbol{\beta}\|_2 \le R$ and $\|\boldsymbol{\beta}'\|_2 \le R$, which implies $\|\gamma\boldsymbol{\beta} + (1-\gamma)\boldsymbol{\beta}'\|_2 \le R$ by the convexity of $\ell_2$ norm. Hence, by Definition 4.2 and Definition 4.3 of Wang et al. (2014a), we have

$$\frac{(\boldsymbol{\beta}'-\boldsymbol{\beta})^T}{\|\boldsymbol{\beta}'-\boldsymbol{\beta}\|_2}\nabla^2\mathcal{L}\big(\gamma\boldsymbol{\beta}+(1-\gamma)\boldsymbol{\beta}'\big)\frac{(\boldsymbol{\beta}'-\boldsymbol{\beta})}{\|\boldsymbol{\beta}'-\boldsymbol{\beta}\|_2} \in \big[\rho_+\big(\nabla^2\mathcal{L}, s^*+2\widetilde{s}\big), \rho_-\big(\nabla^2\mathcal{L}, s^*+2\widetilde{s}\big)\big].$$

Plugging this into the right-hand side of (D.5), we have

$$\text{(D.6)} \qquad \mathcal{L}(\boldsymbol{\beta}') \ge \mathcal{L}(\boldsymbol{\beta}) + \nabla\mathcal{L}(\boldsymbol{\beta})^T(\boldsymbol{\beta}' - \boldsymbol{\beta}) + \frac{\rho_-\big(\nabla^2\mathcal{L}, s^* + 2\widetilde{s}\big)}{2}\|\boldsymbol{\beta}' - \boldsymbol{\beta}\|_2^2,$$

$$\text{(D.7)} \qquad \mathcal{L}(\boldsymbol{\beta}') \le \mathcal{L}(\boldsymbol{\beta}) + \nabla\mathcal{L}(\boldsymbol{\beta})^T(\boldsymbol{\beta}' - \boldsymbol{\beta}) + \frac{\rho_+\big(\nabla^2\mathcal{L}, s^* + 2\widetilde{s}\big)}{2}\|\boldsymbol{\beta}' - \boldsymbol{\beta}\|_2^2.$$

Recall that $\widetilde{\mathcal{L}}_\lambda(\boldsymbol{\beta}) = \mathcal{L}(\boldsymbol{\beta}) + \mathcal{Q}_\lambda(\boldsymbol{\beta})$. Subtracting (D.3) from (D.6), and (D.4)

from (D.7), we obtain

$$\widetilde{\mathcal{L}}_\lambda(\boldsymbol{\beta}') \geq \widetilde{\mathcal{L}}_\lambda(\boldsymbol{\beta}) + \nabla\widetilde{\mathcal{L}}_\lambda(\boldsymbol{\beta})^T(\boldsymbol{\beta}' - \boldsymbol{\beta}) + \frac{\rho_-\big(\nabla^2\mathcal{L}, s^* + 2\widetilde{s}\big) - \zeta_-}{2}\|\boldsymbol{\beta}' - \boldsymbol{\beta}\|_2^2$$

$$\widetilde{\mathcal{L}}_\lambda(\boldsymbol{\beta}') \leq \widetilde{\mathcal{L}}_\lambda(\boldsymbol{\beta}) + \nabla\widetilde{\mathcal{L}}_\lambda(\boldsymbol{\beta})^T(\boldsymbol{\beta}' - \boldsymbol{\beta}) + \frac{\rho_+\big(\nabla^2\mathcal{L}, s^* + 2\widetilde{s}\big) - \zeta_+}{2}\|\boldsymbol{\beta}' - \boldsymbol{\beta}\|_2^2.$$

Then we conclude the proof. □

### D.3. Proof of Lemma 5.2 in Wang et al. (2014a).

PROOF. **Statistical Recovery:** Since $\|\boldsymbol{\beta}_{\overline{S^*}}\|_0 \leq \widetilde{s}$ and $\|\boldsymbol{\beta}^*_{\overline{S^*}}\|_0 = 0$, we have $\|(\boldsymbol{\beta} - \boldsymbol{\beta}^*)_{\overline{S^*}}\| \leq \widetilde{s}$. For logistic loss, we further have $\|\boldsymbol{\beta}\|_2 \leq R$ and $\|\boldsymbol{\beta}^*\|_2 \leq R$. Thus Lemma 5.1 of Wang et al. (2014a) gives

$$\widetilde{\mathcal{L}}_\lambda(\boldsymbol{\beta}^*) \geq \widetilde{\mathcal{L}}_\lambda(\boldsymbol{\beta}) + (\boldsymbol{\beta}^* - \boldsymbol{\beta})^T\nabla\widetilde{\mathcal{L}}_\lambda(\boldsymbol{\beta}) + \frac{\rho_- - \zeta_-}{2}\|\boldsymbol{\beta}^* - \boldsymbol{\beta}\|_2^2, \tag{D.8}$$

$$\widetilde{\mathcal{L}}_\lambda(\boldsymbol{\beta}) \geq \widetilde{\mathcal{L}}_\lambda(\boldsymbol{\beta}^*) + (\boldsymbol{\beta} - \boldsymbol{\beta}^*)^T\nabla\widetilde{\mathcal{L}}_\lambda(\boldsymbol{\beta}^*) + \frac{\rho_- - \zeta_-}{2}\|\boldsymbol{\beta}^* - \boldsymbol{\beta}\|_2^2. \tag{D.9}$$

Adding (D.8) and (D.9) and moving $(\boldsymbol{\beta}^* - \boldsymbol{\beta})^T\nabla\widetilde{\mathcal{L}}_\lambda(\boldsymbol{\beta})$ to the left-hand side, we obtain

$$(\boldsymbol{\beta} - \boldsymbol{\beta}^*)^T\nabla\widetilde{\mathcal{L}}_\lambda(\boldsymbol{\beta}) \geq (\boldsymbol{\beta} - \boldsymbol{\beta}^*)^T\nabla\widetilde{\mathcal{L}}_\lambda(\boldsymbol{\beta}^*) + (\rho_- - \zeta_-)\|\boldsymbol{\beta}^* - \boldsymbol{\beta}\|_2^2. \tag{D.10}$$

Let $\boldsymbol{\xi} \in \partial\|\boldsymbol{\beta}\|_1$ be the subgradient that attains the minimum in

$$\omega_\lambda(\boldsymbol{\beta}) = \min_{\boldsymbol{\xi}' \in \partial\|\boldsymbol{\beta}\|_1} \max_{\boldsymbol{\beta}' \in \Omega} \left\{ \frac{(\boldsymbol{\beta} - \boldsymbol{\beta}')^T}{\|\boldsymbol{\beta} - \boldsymbol{\beta}'\|_1} \big(\nabla\widetilde{\mathcal{L}}_\lambda(\boldsymbol{\beta}) + \lambda\boldsymbol{\xi}'\big) \right\}.$$

Then we have

$$\omega_\lambda(\boldsymbol{\beta}) = \max_{\boldsymbol{\beta}' \in \Omega} \left\{ \frac{(\boldsymbol{\beta} - \boldsymbol{\beta}')^T}{\|\boldsymbol{\beta} - \boldsymbol{\beta}'\|_1} \big(\nabla\widetilde{\mathcal{L}}_\lambda(\boldsymbol{\beta}) + \lambda\boldsymbol{\xi}\big) \right\}. \tag{D.11}$$

Adding $\lambda(\boldsymbol{\beta} - \boldsymbol{\beta}^*)^T\boldsymbol{\xi}$ to the both sides of (D.10), we obtain

$$\begin{aligned} &(\boldsymbol{\beta} - \boldsymbol{\beta}^*)^T\big(\nabla\widetilde{\mathcal{L}}_\lambda(\boldsymbol{\beta}) + \lambda\boldsymbol{\xi}\big) \\ &\quad \geq (\boldsymbol{\beta} - \boldsymbol{\beta}^*)^T\nabla\widetilde{\mathcal{L}}_\lambda(\boldsymbol{\beta}^*) + (\rho_- - \zeta_-)\|\boldsymbol{\beta}^* - \boldsymbol{\beta}\|_2^2 + \lambda(\boldsymbol{\beta} - \boldsymbol{\beta}^*)^T\boldsymbol{\xi}. \end{aligned}$$

Since $\boldsymbol{\beta}^* \in \Omega$, by (D.11) we have

$$\begin{aligned} \frac{(\boldsymbol{\beta} - \boldsymbol{\beta}^*)^T}{\|\boldsymbol{\beta} - \boldsymbol{\beta}^*\|_1}\big(\nabla\widetilde{\mathcal{L}}_\lambda(\boldsymbol{\beta}) + \lambda\boldsymbol{\xi}\big) &\leq \max_{\boldsymbol{\beta}' \in \Omega} \left\{ \frac{(\boldsymbol{\beta} - \boldsymbol{\beta}')^T}{\|\boldsymbol{\beta} - \boldsymbol{\beta}'\|_1} \big(\nabla\widetilde{\mathcal{L}}_\lambda(\boldsymbol{\beta}) + \lambda\boldsymbol{\xi}\big) \right\} \\ &= \omega_\lambda(\boldsymbol{\beta}). \end{aligned} \tag{D.12}$$

Recall that we assume $\omega_\lambda(\boldsymbol{\beta}) \leq \lambda/2$, we obtain

$$(\boldsymbol{\beta} - \boldsymbol{\beta}^*)^T\big(\nabla\widetilde{\mathcal{L}}_\lambda(\boldsymbol{\beta}) + \lambda\boldsymbol{\xi}\big) \leq \lambda/2 \cdot \|\boldsymbol{\beta} - \boldsymbol{\beta}^*\|_1. \tag{D.13}$$

Plugging (D.13) into the left-hand side of (D.10), we obtain

$$\begin{aligned} &\lambda/2 \cdot \|\boldsymbol{\beta} - \boldsymbol{\beta}^*\|_1 \\ &\geq \underbrace{(\boldsymbol{\beta} - \boldsymbol{\beta}^*)^T \nabla \widetilde{\mathcal{L}}_\lambda(\boldsymbol{\beta}^*)}_{\text{(i)}} + (\rho_- - \zeta_-)\|\boldsymbol{\beta}^* - \boldsymbol{\beta}\|_2^2 + \underbrace{\lambda(\boldsymbol{\beta} - \boldsymbol{\beta}^*)^T \boldsymbol{\xi}}_{\text{(ii)}}. \end{aligned} \tag{D.14}$$

Now we provide lower bounds of terms (i) and (ii) in (D.14) respectively.

- **Bounding Term (i) in (D.14):** Recall that $\widetilde{\mathcal{L}}_\lambda(\boldsymbol{\beta}) = \mathcal{L}(\boldsymbol{\beta}) + \mathcal{Q}_\lambda(\boldsymbol{\beta})$. We have

$$(\boldsymbol{\beta} - \boldsymbol{\beta}^*)^T \nabla \widetilde{\mathcal{L}}_\lambda(\boldsymbol{\beta}^*) = \underbrace{(\boldsymbol{\beta} - \boldsymbol{\beta}^*)^T \nabla \mathcal{L}(\boldsymbol{\beta}^*)}_{\text{(i).a}} + \underbrace{(\boldsymbol{\beta} - \boldsymbol{\beta}^*)^T \nabla \mathcal{Q}_\lambda(\boldsymbol{\beta}^*)}_{\text{(i).b}}. \tag{D.15}$$

Separating the support of $\boldsymbol{\beta} - \boldsymbol{\beta}^*$ into $S^*$ and $\overline{S^*}$, we obtain

$$\|\boldsymbol{\beta} - \boldsymbol{\beta}^*\|_1 = \|(\boldsymbol{\beta} - \boldsymbol{\beta}^*)_{\overline{S^*}}\|_1 + \|(\boldsymbol{\beta} - \boldsymbol{\beta}^*)_{S^*}\|_1.$$

Then for term (i).a in (D.15), we have

$$\begin{aligned} &(\boldsymbol{\beta} - \boldsymbol{\beta}^*)^T \nabla \mathcal{L}(\boldsymbol{\beta}^*) \\ &\geq -\|\boldsymbol{\beta} - \boldsymbol{\beta}^*\|_1 \|\nabla \mathcal{L}(\boldsymbol{\beta}^*)\|_\infty \\ &= -\|(\boldsymbol{\beta} - \boldsymbol{\beta}^*)_{\overline{S^*}}\|_1 \|\nabla \mathcal{L}(\boldsymbol{\beta}^*)\|_\infty - \|(\boldsymbol{\beta} - \boldsymbol{\beta}^*)_{S^*}\|_1 \|\nabla \mathcal{L}(\boldsymbol{\beta}^*)\|_\infty. \end{aligned} \tag{D.16}$$

For term (i).b in (D.15), we have

$$\begin{aligned} &(\boldsymbol{\beta} - \boldsymbol{\beta}^*)^T \nabla \mathcal{Q}_\lambda(\boldsymbol{\beta}^*) \\ &= (\boldsymbol{\beta} - \boldsymbol{\beta}^*)_{S^*}^T \big(\nabla \mathcal{Q}_\lambda(\boldsymbol{\beta}^*)\big)_{S^*} + (\boldsymbol{\beta} - \boldsymbol{\beta}^*)_{\overline{S^*}}^T \big(\nabla \mathcal{Q}_\lambda(\boldsymbol{\beta}^*)\big)_{\overline{S^*}}. \end{aligned} \tag{D.17}$$

Note that $\mathcal{Q}_\lambda(\boldsymbol{\beta}^*)$ is separable. We have

$$\begin{aligned} (\boldsymbol{\beta} - \boldsymbol{\beta}^*)_{S^*}^T \big(\nabla \mathcal{Q}_\lambda(\boldsymbol{\beta}^*)\big)_{S^*} &= \sum_{j \in S^*} (\beta_j - \beta_j^*) \cdot q_\lambda'(\beta_j^*) \\ &= (\boldsymbol{\beta} - \boldsymbol{\beta}^*)_{S^*}^T \nabla \mathcal{Q}_\lambda(\boldsymbol{\beta}^*), \end{aligned} \tag{D.18}$$

$$\begin{aligned} (\boldsymbol{\beta} - \boldsymbol{\beta}^*)_{\overline{S^*}}^T \big(\nabla \mathcal{Q}_\lambda(\boldsymbol{\beta}^*)\big)_{\overline{S^*}} &= \sum_{j \in \overline{S^*}} (\beta_j - \beta_j^*) \cdot q_\lambda'(\beta_j^*) \\ &= \sum_{j \in \overline{S^*}} (\beta_j - \beta_j^*) \cdot q_\lambda'(0) = 0, \end{aligned} \tag{D.19}$$

where the second equation in (D.19) is because $\beta_j^* = 0$ for $j \in \overline{S^*}$, and the third is by regularity condition (c) that $q_\lambda'(0) = 0$. Plugging (D.18) and (D.19) into the right-hand side of (D.17), for term (i).b in (D.15)

we obtain

$$(\boldsymbol{\beta} - \boldsymbol{\beta}^*)^T \nabla \mathcal{Q}_\lambda(\boldsymbol{\beta}^*) = (\boldsymbol{\beta} - \boldsymbol{\beta}^*)_{S^*}^T \nabla \mathcal{Q}_\lambda(\boldsymbol{\beta}^*) \geq -\|(\boldsymbol{\beta} - \boldsymbol{\beta}^*)_{S^*}\|_1 \|\nabla \mathcal{Q}_\lambda(\boldsymbol{\beta}^*)\|_\infty. \tag{D.20}$$

Plugging (D.16) and (D.20) into the right-hand side of (D.15), then for term (i) in (D.14) we obtain

$$(\boldsymbol{\beta} - \boldsymbol{\beta}^*)^T \nabla \widetilde{\mathcal{L}}_\lambda(\boldsymbol{\beta}^*) \geq -\|(\boldsymbol{\beta} - \boldsymbol{\beta}^*)_{\overline{S^*}}\|_1 \|\nabla \mathcal{L}(\boldsymbol{\beta}^*)\|_\infty - \|(\boldsymbol{\beta} - \boldsymbol{\beta}^*)_{S^*}\|_1 \|\nabla \mathcal{L}(\boldsymbol{\beta}^*)\|_\infty - \|(\boldsymbol{\beta} - \boldsymbol{\beta}^*)_{S^*}\|_1 \|\nabla \mathcal{Q}_\lambda(\boldsymbol{\beta}^*)\|_\infty. \tag{D.21}$$

- **Bounding Term (ii) in (D.14):** For term (ii) in (D.14), by separating the support of $\boldsymbol{\beta} - \boldsymbol{\beta}^*$ into $S^*$ and $\overline{S^*}$ we have

$$\lambda (\boldsymbol{\beta} - \boldsymbol{\beta}^*)^T \boldsymbol{\xi} = \lambda \underbrace{(\boldsymbol{\beta} - \boldsymbol{\beta}^*)_{S^*}^T \boldsymbol{\xi}_{S^*}}_{\text{(ii).a}} + \lambda \underbrace{(\boldsymbol{\beta} - \boldsymbol{\beta}^*)_{\overline{S^*}}^T \boldsymbol{\xi}_{\overline{S^*}}}_{\text{(ii).b}}. \tag{D.22}$$

For term (ii).a in (D.22), since $\boldsymbol{\xi} \in \partial \|\boldsymbol{\beta}\|_1$, we have $\|\boldsymbol{\xi}_{S^*}\|_\infty \leq \|\boldsymbol{\xi}\|_\infty \leq 1$, which implies

$$(\boldsymbol{\beta} - \boldsymbol{\beta}^*)_{S^*}^T \boldsymbol{\xi}_{S^*} \geq -\|\boldsymbol{\xi}_{S^*}\|_\infty \|(\boldsymbol{\beta} - \boldsymbol{\beta}^*)_{S^*}\|_1 \geq -\|(\boldsymbol{\beta} - \boldsymbol{\beta}^*)_{S^*}\|_1. \tag{D.23}$$

For term (ii).b in (D.22), note that $\boldsymbol{\beta}^*_{\overline{S^*}} = \mathbf{0}$. Hence, $(\boldsymbol{\beta} - \boldsymbol{\beta}^*)_{\overline{S^*}} = \boldsymbol{\beta}_{\overline{S^*}}$. Recall $\boldsymbol{\xi} \in \partial \|\boldsymbol{\beta}\|_1$. For $\beta_j \neq 0$, since $\xi_j = \operatorname{sign}(\beta_j)$, we have $\beta_j \xi_j = |\beta_j|$. For $\beta_j = 0$, we have $\beta_j \xi_j = |\beta_j| = 0$. Therefore, we obtain

$$(\boldsymbol{\beta} - \boldsymbol{\beta}^*)_{\overline{S^*}}^T \boldsymbol{\xi}_{\overline{S^*}} = \boldsymbol{\beta}_{\overline{S^*}}^T \boldsymbol{\xi}_{\overline{S^*}} = \sum_{j \in \overline{S^*}} \beta_j \xi_j = \sum_{j \in \overline{S^*}} |\beta_j| = \|\boldsymbol{\beta}_{\overline{S^*}}\|_1 = \|(\boldsymbol{\beta} - \boldsymbol{\beta}^*)_{\overline{S^*}}\|_1. \tag{D.24}$$

Plugging (D.23) and (D.24) into the right-hand side of (D.22), we obtain

$$\lambda (\boldsymbol{\beta} - \boldsymbol{\beta}^*)^T \boldsymbol{\xi} \geq -\lambda \|(\boldsymbol{\beta} - \boldsymbol{\beta}^*)_{S^*}\|_1 + \lambda \|(\boldsymbol{\beta} - \boldsymbol{\beta}^*)_{\overline{S^*}}\|_1. \tag{D.25}$$

Plugging (D.21) and (D.25) into the right-hand side of (D.14), we obtain

$$\begin{aligned} &\lambda/2 \cdot \|\boldsymbol{\beta} - \boldsymbol{\beta}^*\|_1 \\ &\geq \underbrace{\begin{matrix} -\|(\boldsymbol{\beta} - \boldsymbol{\beta}^*)_{\overline{S^*}}\|_1 \|\nabla \mathcal{L}(\boldsymbol{\beta}^*)\|_\infty - \|(\boldsymbol{\beta} - \boldsymbol{\beta}^*)_{S^*}\|_1 \|\nabla \mathcal{L}(\boldsymbol{\beta}^*)\|_\infty \\ - \|(\boldsymbol{\beta} - \boldsymbol{\beta}^*)_{S^*}\|_1 \|\nabla \mathcal{Q}_\lambda(\boldsymbol{\beta}^*)\|_\infty \end{matrix}}_{\text{(i) in (D.14)}} \\ &\quad + (\rho_- - \zeta_-) \|\boldsymbol{\beta}^* - \boldsymbol{\beta}\|_2^2 \underbrace{-\lambda \|(\boldsymbol{\beta} - \boldsymbol{\beta}^*)_{S^*}\|_1 + \lambda \|(\boldsymbol{\beta} - \boldsymbol{\beta}^*)_{\overline{S^*}}\|_1}_{\text{(ii) in (D.14)}}. \end{aligned} \tag{D.26}$$

Again, we separate the left-hand side of (D.26) as $\lambda/2 \cdot \|\boldsymbol{\beta} - \boldsymbol{\beta}^*\|_1 = \lambda/2 \cdot$

$\|(\boldsymbol{\beta}-\boldsymbol{\beta}^*)_{\overline{S^*}}\|_1 + \lambda/2 \cdot \|(\boldsymbol{\beta}-\boldsymbol{\beta}^*)_{S^*}\|_1$. Rearranging the terms, we obtain

$$(\rho_- - \zeta_-)\|\boldsymbol{\beta}-\boldsymbol{\beta}^*\|_2^2 + \underbrace{\big(\lambda/2 - \|\nabla\mathcal{L}(\boldsymbol{\beta}^*)\|_\infty\big)\|(\boldsymbol{\beta}-\boldsymbol{\beta}^*)_{\overline{S^*}}\|_1}_{\text{(i)}} \le \big(3\lambda/2 + \underbrace{\|\nabla\mathcal{L}(\boldsymbol{\beta}^*)\|_\infty}_{\text{(ii)}} + \underbrace{\|\nabla\mathcal{Q}_\lambda(\boldsymbol{\beta}^*)\|_\infty}_{\text{(iii)}}\big)\|(\boldsymbol{\beta}-\boldsymbol{\beta}^*)_{S^*}\|_1. \tag{D.27}$$

For term (ii) in (D.27), by (4.1) in Assumption 4.1 of Wang et al. (2014a) and $\lambda \ge \lambda_{\mathrm{tgt}}$ we have

$$\|\nabla\mathcal{L}(\boldsymbol{\beta}^*)\|_\infty \le \lambda_{\mathrm{tgt}}/8 \le \lambda/8. \tag{D.28}$$

Meanwhile, (D.28) also implies that term (i) in (D.27) is positive. Recall that $\mathcal{Q}_\lambda(\boldsymbol{\beta}) = \sum_{j=1}^d q_\lambda(\beta_j)$, where $q_\lambda(\beta_j)$ satisfies regularity condition (d) in Wang et al. (2014a). Hence for term (iii) in (D.27) we have

$$\|\nabla\mathcal{Q}_\lambda(\boldsymbol{\beta}^*)\|_\infty = \max_{1\le j\le d} |q_\lambda'(\beta_j^*)| \le \lambda. \tag{D.29}$$

In summary, from (D.27) we obtain

$$\begin{aligned}(\rho_- - \zeta_-)\|\boldsymbol{\beta}-\boldsymbol{\beta}^*\|_2^2 &\le \big(3\lambda/2 + \|\nabla\mathcal{L}(\boldsymbol{\beta}^*)\|_\infty + \|\nabla\mathcal{Q}_\lambda(\boldsymbol{\beta}^*)\|_\infty\big)\|(\boldsymbol{\beta}-\boldsymbol{\beta}^*)_{S^*}\|_1\\ &\le (3\lambda/2 + \lambda/8 + \lambda)\|(\boldsymbol{\beta}-\boldsymbol{\beta}^*)_{S^*}\|_1\\ &\le 21\lambda/8 \cdot \sqrt{s^*}\|(\boldsymbol{\beta}-\boldsymbol{\beta}^*)_{S^*}\|_2\\ &\le 21\lambda/8 \cdot \sqrt{s^*}\|\boldsymbol{\beta}-\boldsymbol{\beta}^*\|_2. \end{aligned}\tag{D.30}$$

According to (4.3) of Wang et al. (2014a), we have $\rho_- - \zeta_- > 0$. Therefore, (D.30) gives

$$\|\boldsymbol{\beta}-\boldsymbol{\beta}^*\|_2 \le \frac{21/8}{\rho_- - \zeta_-}\lambda\sqrt{s^*}, \tag{D.31}$$

which implies the first conclusion.

**Results for the Objective Function Value:** Note that on the right-hand side of (D.9), we have $\rho_- - \zeta_- > 0$, which gives

$$\widetilde{\mathcal{L}}_\lambda(\boldsymbol{\beta}^*) \ge \widetilde{\mathcal{L}}_\lambda(\boldsymbol{\beta}) + (\boldsymbol{\beta}^*-\boldsymbol{\beta})^T\nabla\widetilde{\mathcal{L}}_\lambda(\boldsymbol{\beta}). \tag{D.32}$$

Meanwhile, since $\boldsymbol{\xi} \in \partial\|\boldsymbol{\beta}\|_1$, by the convexity of $\ell_1$ norm we have

$$\lambda\|\boldsymbol{\beta}^*\|_1 \ge \lambda\|\boldsymbol{\beta}\|_1 + \lambda(\boldsymbol{\beta}^*-\boldsymbol{\beta})^T\boldsymbol{\xi}. \tag{D.33}$$

Recall that $\phi_\lambda(\boldsymbol{\beta}) = \widetilde{\mathcal{L}}_\lambda(\boldsymbol{\beta}) + \lambda\|\boldsymbol{\beta}\|_1$. Adding (D.32) and (D.33), we obtain

$$\phi_\lambda(\boldsymbol{\beta}^*) \ge \phi_\lambda(\boldsymbol{\beta}) + (\boldsymbol{\beta}^*-\boldsymbol{\beta})^T\big(\nabla\widetilde{\mathcal{L}}_\lambda(\boldsymbol{\beta}) + \lambda\boldsymbol{\xi}\big), \tag{D.34}$$

which implies

$$\phi_\lambda(\boldsymbol{\beta}) - \phi_\lambda(\boldsymbol{\beta}^*) \le (\boldsymbol{\beta}-\boldsymbol{\beta}^*)^T\big(\nabla\widetilde{\mathcal{L}}_\lambda(\boldsymbol{\beta}) + \lambda\boldsymbol{\xi}\big) \le \lambda/2 \cdot \|\boldsymbol{\beta}-\boldsymbol{\beta}^*\|_1.$$

Here the second inequality follows from (D.13), which is a direct consequence of the assumption that $\omega_\lambda(\boldsymbol{\beta}) \le \lambda/2$. Separating the support of $\boldsymbol{\beta} - \boldsymbol{\beta}^*$ into $S^*$ and $\overline{S^*}$, we obtain

$$\begin{aligned} \phi_\lambda(\boldsymbol{\beta}) - \phi_\lambda(\boldsymbol{\beta}^*) &\le \lambda/2 \cdot \|\boldsymbol{\beta} - \boldsymbol{\beta}^*\|_1 \\ &\le \lambda/2 \cdot \|(\boldsymbol{\beta} - \boldsymbol{\beta}^*)_{S^*}\|_1 + \lambda/2 \cdot \|(\boldsymbol{\beta} - \boldsymbol{\beta}^*)_{\overline{S^*}}\|_1. \end{aligned} \tag{D.35}$$

Now we derive an upper bound of $\|(\boldsymbol{\beta} - \boldsymbol{\beta}^*)_{\overline{S^*}}\|_1$ on the right-hand side of (D.35). On the left-hand side of (D.27), we have $\rho_- - \zeta_- > 0$, which gives

$$\begin{aligned} &\big(\lambda/2 - \|\nabla \mathcal{L}(\boldsymbol{\beta}^*)\|_\infty\big)\|(\boldsymbol{\beta} - \boldsymbol{\beta}^*)_{\overline{S^*}}\|_1 \\ &\le \big(3\lambda/2 + \|\nabla \mathcal{L}(\boldsymbol{\beta}^*)\|_\infty + \|\nabla \mathcal{Q}_\lambda(\boldsymbol{\beta}^*)\|_\infty\big)\|(\boldsymbol{\beta} - \boldsymbol{\beta}^*)_{S^*}\|_1. \end{aligned} \tag{D.36}$$

Note that in (D.36) we have $\|\nabla \mathcal{L}(\boldsymbol{\beta}^*)\|_\infty \le \lambda/8$ by (D.28), and $\|\nabla \mathcal{Q}_\lambda(\boldsymbol{\beta}^*)\|_\infty \le \lambda$ by (D.29). Hence we have

$$(\lambda/2 - \lambda/8)\|(\boldsymbol{\beta} - \boldsymbol{\beta}^*)_{\overline{S^*}}\|_1 \le (3\lambda/2 + \lambda/8 + \lambda)\|(\boldsymbol{\beta} - \boldsymbol{\beta}^*)_{S^*}\|_1, \tag{D.37}$$

which implies $\|(\boldsymbol{\beta} - \boldsymbol{\beta}^*)_{\overline{S^*}}\|_1 \le 7\|(\boldsymbol{\beta} - \boldsymbol{\beta}^*)_{S^*}\|_1$. Plugging this into the right-hand side of (D.35), we obtain

$$\begin{aligned} \phi_\lambda(\boldsymbol{\beta}) - \phi_\lambda(\boldsymbol{\beta}^*) &\le (\lambda/2 + 7\lambda/2)\|(\boldsymbol{\beta} - \boldsymbol{\beta}^*)_{S^*}\|_1 \\ &\le 4\lambda\sqrt{s^*}\|(\boldsymbol{\beta} - \boldsymbol{\beta}^*)_{S^*}\|_2 \le 4\lambda\sqrt{s^*}\|\boldsymbol{\beta} - \boldsymbol{\beta}^*\|_2. \end{aligned} \tag{D.38}$$

Plugging the upper bound of $\|\boldsymbol{\beta} - \boldsymbol{\beta}^*\|_2$ in (D.31) into the right-hand side of (D.38), we obtain

$$\phi_\lambda(\boldsymbol{\beta}) - \phi_\lambda(\boldsymbol{\beta}^*) \le \frac{21/2}{\rho_- - \zeta_-}\lambda^2 s^*.$$

Hence we reach the second conclusion. □

### D.4. Proof of Lemma 5.3 in Wang et al. (2014a).

PROOF. Since $\|\boldsymbol{\beta}_{\overline{S^*}}\|_0 \le \widetilde{s}$ and $\|\boldsymbol{\beta}^*_{\overline{S^*}}\|_0 = 0$, we have $\|(\boldsymbol{\beta} - \boldsymbol{\beta}^*)_{\overline{S^*}}\|_0 \le \widetilde{s}$. For logistic loss, we further have $\|\boldsymbol{\beta}\|_2 \le R$ and $\|\boldsymbol{\beta}^*\|_2 \le R$, where $R$ is specified in Definition 4.3 of Wang et al. (2014a). Therefore, Lemma 5.1 of Wang et al. (2014a) gives

$$\widetilde{\mathcal{L}}_\lambda(\boldsymbol{\beta}^*) + (\boldsymbol{\beta} - \boldsymbol{\beta}^*)^T \nabla \widetilde{\mathcal{L}}_\lambda(\boldsymbol{\beta}^*) + \frac{\rho_- - \zeta_-}{2}\|\boldsymbol{\beta}^* - \boldsymbol{\beta}\|_2^2 \le \widetilde{\mathcal{L}}_\lambda(\boldsymbol{\beta}). \tag{D.39}$$

Recall that $\phi_\lambda(\boldsymbol{\beta}) = \widetilde{\mathcal{L}}_\lambda(\boldsymbol{\beta}) + \lambda\|\boldsymbol{\beta}\|_1$. Hence, from our assumption that

$$\phi_\lambda(\boldsymbol{\beta}) - \phi_\lambda(\boldsymbol{\beta}^*) \le \frac{21/2}{\rho_- - \zeta_-}\lambda^2 s^*$$

we obtain

$$\widetilde{\mathcal{L}}_\lambda(\boldsymbol{\beta}) - \widetilde{\mathcal{L}}_\lambda(\boldsymbol{\beta}^*) + \lambda(\|\boldsymbol{\beta}\|_1 - \|\boldsymbol{\beta}^*\|_1) \le \frac{21/2}{\rho_- - \zeta_-}\lambda^2 s^*. \tag{D.40}$$

Plugging (D.39) into the left-hand side of (D.40), we have

$$(\boldsymbol{\beta}-\boldsymbol{\beta}^*)^T\nabla\widetilde{\mathcal{L}}_\lambda(\boldsymbol{\beta}^*) + \frac{\rho_- - \zeta_-}{2}\|\boldsymbol{\beta}^* - \boldsymbol{\beta}\|_2^2 + \lambda(\|\boldsymbol{\beta}\|_1 - \|\boldsymbol{\beta}^*\|_1) \le \frac{21/2}{\rho_- - \zeta_-}\lambda^2 s^*.$$

Moving $(\boldsymbol{\beta} - \boldsymbol{\beta}^*)^T\nabla\widetilde{\mathcal{L}}_\lambda(\boldsymbol{\beta}^*) + \lambda(\|\boldsymbol{\beta}\|_1 - \|\boldsymbol{\beta}^*\|_1)$ to its right-hand side yields

$$\begin{aligned}&\frac{\rho_- - \zeta_-}{2}\|\boldsymbol{\beta}^* - \boldsymbol{\beta}\|_2^2 \\ &\le \frac{21/2}{\rho_- - \zeta_-}\lambda^2 s^* \underbrace{-(\boldsymbol{\beta} - \boldsymbol{\beta}^*)^T\nabla\widetilde{\mathcal{L}}_\lambda(\boldsymbol{\beta}^*)}_{\text{(i)}} + \lambda\underbrace{\big(\|\boldsymbol{\beta}^*\|_1 - \|\boldsymbol{\beta}\|_1\big)}_{\text{(ii)}}.\end{aligned} \tag{D.41}$$

For term (i) in (D.41), following the same way we obtain the lower bound of term (i) in (D.14) (in the proof of Lemma 5.2), we can obtain the same result as in (D.21), which implies

$$\begin{aligned}-(\boldsymbol{\beta}-\boldsymbol{\beta}^*)^T\nabla\widetilde{\mathcal{L}}_\lambda(\boldsymbol{\beta}^*) \le \|(\boldsymbol{\beta}-\boldsymbol{\beta}^*)_{\overline{S^*}}\|_1\|\nabla\mathcal{L}(\boldsymbol{\beta}^*)\|_\infty + \|(\boldsymbol{\beta}-\boldsymbol{\beta}^*)_{S^*}\|_1\|\nabla\mathcal{L}(\boldsymbol{\beta}^*)\|_\infty \\ + \|(\boldsymbol{\beta} - \boldsymbol{\beta}^*)_{S^*}\|_1\|\nabla\mathcal{Q}_\lambda(\boldsymbol{\beta}^*)\|_\infty.\end{aligned} \tag{D.42}$$

For term (ii) in (D.41), separating the support of $\boldsymbol{\beta}$ and $\boldsymbol{\beta}^*$ into $S^*$ and $\overline{S^*}$ respectively, we obtain

$$\|\boldsymbol{\beta}^*\|_1 - \|\boldsymbol{\beta}\|_1 = \|\boldsymbol{\beta}^*_{S^*}\|_1 + \|\boldsymbol{\beta}^*_{\overline{S^*}}\|_1 - \big(\|\boldsymbol{\beta}_{S^*}\|_1 + \|\boldsymbol{\beta}_{\overline{S^*}}\|_1\big). \tag{D.43}$$

Note that $\boldsymbol{\beta}^*_{\overline{S^*}} = \mathbf{0}$, which gives $\boldsymbol{\beta}_{\overline{S^*}} = \boldsymbol{\beta}_{\overline{S^*}} - \boldsymbol{\beta}^*_{\overline{S^*}} = (\boldsymbol{\beta} - \boldsymbol{\beta}^*)_{\overline{S^*}}$. Hence, from (D.43) we have

$$\begin{aligned}\|\boldsymbol{\beta}^*\|_1 - \|\boldsymbol{\beta}\|_1 &= \|\boldsymbol{\beta}^*_{S^*}\|_1 - \|\boldsymbol{\beta}_{S^*}\|_1 - \|(\boldsymbol{\beta} - \boldsymbol{\beta}^*)_{\overline{S^*}}\|_1 \\ &\le \|(\boldsymbol{\beta} - \boldsymbol{\beta}^*)_{S^*}\|_1 - \|(\boldsymbol{\beta} - \boldsymbol{\beta}^*)_{\overline{S^*}}\|_1,\end{aligned} \tag{D.44}$$

where the inequality follows from the triangle inequality. Plugging (D.42) and (D.44) into the right-hand side of (D.41), we obtain

$$\begin{aligned}&\frac{\rho_- - \zeta_-}{2}\|\boldsymbol{\beta}^* - \boldsymbol{\beta}\|_2^2 \\ &\le \underbrace{\begin{aligned}\|(\boldsymbol{\beta} - \boldsymbol{\beta}^*)_{\overline{S^*}}\|_1\|\nabla\mathcal{L}(\boldsymbol{\beta}^*)\|_\infty + \|(\boldsymbol{\beta} - \boldsymbol{\beta}^*)_{S^*}\|_1\|\nabla\mathcal{L}(\boldsymbol{\beta}^*)\|_\infty \\ + \|(\boldsymbol{\beta} - \boldsymbol{\beta}^*)_{S^*}\|_1\|\nabla\mathcal{Q}_\lambda(\boldsymbol{\beta}^*)\|_\infty\end{aligned}}_{\text{(i) in (D.41)}} \\ &\quad + \lambda\underbrace{\big(\|(\boldsymbol{\beta} - \boldsymbol{\beta}^*)_{S^*}\|_1 - \|(\boldsymbol{\beta} - \boldsymbol{\beta}^*)_{\overline{S^*}}\|_1\big)}_{\text{(ii) in (D.41)}} + \frac{21/2}{\rho_- - \zeta_-}\lambda^2 s^*.\end{aligned} \tag{D.45}$$

Rearranging the terms in (D.45), we obtain

$$\frac{\rho_- - \zeta_-}{2}\|\boldsymbol{\beta}-\boldsymbol{\beta}^*\|_2^2 + \underbrace{\big(\lambda - \|\nabla\mathcal{L}(\boldsymbol{\beta}^*)\|_\infty\big)\|(\boldsymbol{\beta}-\boldsymbol{\beta}^*)_{\overline{S^*}}\|_1}_{\text{(i)}} \\ \le \big(\lambda + \underbrace{\|\nabla\mathcal{L}(\boldsymbol{\beta}^*)\|_\infty}_{\text{(ii)}} + \underbrace{\|\nabla\mathcal{Q}_\lambda(\boldsymbol{\beta}^*)\|_\infty}_{\text{(iii)}}\big)\|(\boldsymbol{\beta}-\boldsymbol{\beta}^*)_{S^*}\|_1 + \frac{21/2}{\rho_- - \zeta_-}\lambda^2 s^*. \tag{D.46}$$

By (4.1) in Assumption 4.1 of Wang et al. (2014a) and $\lambda \ge \lambda_{\text{tgt}}$, for term (ii) in (D.46), we have

$$\|\nabla\mathcal{L}(\boldsymbol{\beta}^*)\|_\infty \le \lambda_{\text{tgt}}/8 \le \lambda/8. \tag{D.47}$$

Moreover, (D.47) implies that term (i) in (D.46) is positive. For term (iii) in (D.46), since $\mathcal{Q}_\lambda(\boldsymbol{\beta}) = \sum_{j=1}^d q_\lambda(\beta_j)$, where $q_\lambda(\beta_j)$ satisfies regularity condition (d), we have

$$\|\nabla\mathcal{Q}_\lambda(\boldsymbol{\beta}^*)\|_\infty \le \max_{1\le j\le d}|q'_\lambda(\beta_j^*)| \le \lambda. \tag{D.48}$$

Therefore, from (D.48) we obtain

$$\begin{aligned}
&\frac{\rho_- - \zeta_-}{2}\|\boldsymbol{\beta}-\boldsymbol{\beta}^*\|_2^2 \\
&\le \big(\lambda + \|\nabla\mathcal{L}(\boldsymbol{\beta}^*)\|_\infty + \|\nabla\mathcal{Q}_\lambda(\boldsymbol{\beta}^*)\|_\infty\big)\|(\boldsymbol{\beta}-\boldsymbol{\beta}^*)_{S^*}\|_1 + \frac{21/2}{\rho_- - \zeta_-}\lambda^2 s^* \\
&\le (\lambda + \lambda/8 + \lambda)\|(\boldsymbol{\beta}-\boldsymbol{\beta}^*)_{S^*}\|_1 + \frac{21/2}{\rho_- - \zeta_-}\lambda^2 s^* \\
&\le 17/8\cdot\lambda\|(\boldsymbol{\beta}-\boldsymbol{\beta}^*)_{S^*}\|_1 + \frac{21/2}{\rho_- - \zeta_-}\lambda^2 s^*. \qquad \text{(D.49)}
\end{aligned}$$

To further obtain an upper bound of the right-hand side of (D.49), we discuss two cases regarding the relationship between $\|(\boldsymbol{\beta}-\boldsymbol{\beta}^*)_{S^*}\|_1$ and $\lambda s^*$.

- If $7/(\rho_- - \zeta_-)\cdot\lambda s^* < \|(\boldsymbol{\beta}-\boldsymbol{\beta}^*)_{S^*}\|_1$, then we have

  $$\frac{21/2}{\rho_- - \zeta_-}\lambda^2 s^* < 3/2\cdot\lambda\|(\boldsymbol{\beta}-\boldsymbol{\beta}^*)_{S^*}\|_1.$$

  Plugging this into the right-hand side of (D.49), we obtain

  $$\begin{aligned}
  \frac{\rho_- - \zeta_-}{2}\|\boldsymbol{\beta}-\boldsymbol{\beta}^*\|_2^2 &\le (17/8\cdot\lambda + 3/2\cdot\lambda)\|(\boldsymbol{\beta}-\boldsymbol{\beta}^*)_{S^*}\|_1 \\
  &\le 29/8\cdot\lambda\sqrt{s^*}\|(\boldsymbol{\beta}-\boldsymbol{\beta}^*)_{S^*}\|_2 \\
  &\le 29/8\cdot\lambda\sqrt{s^*}\|\boldsymbol{\beta}-\boldsymbol{\beta}^*\|_2.
  \end{aligned}$$

Dividing $\|\boldsymbol{\beta}^* - \boldsymbol{\beta}\|_2$ on both sides, we have

$$\|\boldsymbol{\beta} - \boldsymbol{\beta}^*\|_2 \leq \frac{29/4}{\rho_- - \zeta_-}\lambda\sqrt{s^*}. \tag{D.50}$$

- If $\|(\boldsymbol{\beta} - \boldsymbol{\beta}^*)_{S^*}\|_1 \leq 7/(\rho_- - \zeta_-) \cdot \lambda s^*$, then we have

$$17/8 \cdot \lambda\|(\boldsymbol{\beta} - \boldsymbol{\beta}^*)_{S^*}\|_1 < \frac{119/8}{\rho_- - \zeta_-}\lambda^2 s^*.$$

  Plugging this into the right-hand side of (D.49), we obtain

$$\begin{aligned}\frac{\rho_- - \zeta_-}{2}\|\boldsymbol{\beta} - \boldsymbol{\beta}^*\|_2^2 &\leq \frac{119/8}{\rho_- - \zeta_-}\lambda^2 s^* + \frac{21/2}{\rho_- - \zeta_-}\lambda^2 s^* \\ &= \frac{203/8}{\rho_- - \zeta_-}\lambda^2 s^*, \end{aligned} \tag{D.51}$$

  which implies

$$\|\boldsymbol{\beta} - \boldsymbol{\beta}^*\|_2 \leq \frac{\sqrt{203}/2}{\rho_- - \zeta_-}\lambda\sqrt{s^*}. \tag{D.52}$$

Combining (D.50) and (D.52), since $\max\{29/4, \sqrt{203}/2\} \leq 15/2$, we obtain

$$\|\boldsymbol{\beta} - \boldsymbol{\beta}^*\|_2 < \frac{15/2}{\rho_- - \zeta_-}\lambda\sqrt{s^*}.$$

Hence we conclude the proof. □

### D.5. Proof of Lemma 5.4 in Wang et al. (2014a).

Proof. Recall the proximal-gradient update step defined in (3.8) of Wang et al. (2014a) with $\Omega = \mathbb{R}^d$, i.e., $R = +\infty$, takes the form

$$\big(\mathcal{T}_{L,\lambda}(\boldsymbol{\beta}; +\infty)\big)_j = \begin{cases} 0 & \text{if } |\bar{\beta}_j| \leq \lambda/L, \\ \operatorname{sign}(\bar{\beta}_j)\big(|\bar{\beta}_j| - \lambda/L\big) & \text{if } |\bar{\beta}_j| > \lambda/L, \end{cases} \tag{D.53}$$

for $j = 1, \ldots, d$, where

$$\bar{\boldsymbol{\beta}} = \boldsymbol{\beta} - \frac{1}{L}\nabla\widetilde{\mathcal{L}}_\lambda(\boldsymbol{\beta}), \tag{D.54}$$

and $\bar{\beta}_j$ is the $j$-th dimension of $\bar{\boldsymbol{\beta}}$. Furthermore, if $\Omega = B_2(R)$ of radius $R \in (0, \infty)$, $\mathcal{T}_{L,\lambda}(\boldsymbol{\beta}; R)$ can be obtained by projecting $\mathcal{T}_{L,\lambda}(\boldsymbol{\beta}; +\infty)$ shown in (D.53) onto $B_2(R)$, i.e.,

$$\mathcal{T}_{L,\lambda}(\boldsymbol{\beta}; R) = \begin{cases} \mathcal{T}_{L,\lambda}(\boldsymbol{\beta}; +\infty) & \text{if } \|\mathcal{T}_{L,\lambda}(\boldsymbol{\beta}; +\infty)\|_2 < R, \\ \dfrac{R \cdot \mathcal{T}_{L,\lambda}(\boldsymbol{\beta}; +\infty)}{\|\mathcal{T}_{L,\lambda}(\boldsymbol{\beta}; +\infty)\|_2} & \text{if } \|\mathcal{T}_{L,\lambda}(\boldsymbol{\beta}; +\infty)\|_2 \geq R. \end{cases} \tag{D.55}$$

Note that $\mathcal{T}_{L,\lambda}(\boldsymbol{\beta}; +\infty)$ and $\mathcal{T}_{L,\lambda}(\boldsymbol{\beta}; R)$ have exactly the same sparsity pattern.

Hence we focus on analyzing the sparsity pattern of $\mathcal{T}_{L,\lambda}(\boldsymbol{\beta};+\infty)$ in the following.

In fact, update scheme (D.53) defines a soft-thresholding operation on $\bar{\boldsymbol{\beta}}$ defined in (D.54), with the threshold value $\lambda/L$. To show $\big\|\big(\mathcal{T}_{L,\lambda}(\boldsymbol{\beta};+\infty)\big)_{\overline{S^*}}\big\|_0 \leq \widetilde{s}$, we need to prove that, for $j \in \overline{S^*}$, the number of $j$'s such that $\big|\bar{\beta}_j\big| > \lambda/L$ is no more than $\widetilde{s}$. To achieve this goal, we first reformulate $\bar{\boldsymbol{\beta}}$ as

$$\bar{\boldsymbol{\beta}} = \boldsymbol{\beta} - \frac{1}{L}\nabla\widetilde{\mathcal{L}}_\lambda(\boldsymbol{\beta}) = \boldsymbol{\beta} - \frac{1}{L}\nabla\widetilde{\mathcal{L}}_\lambda(\boldsymbol{\beta}^*) + \frac{1}{L}\big(\nabla\widetilde{\mathcal{L}}_\lambda(\boldsymbol{\beta}^*) - \nabla\widetilde{\mathcal{L}}_\lambda(\boldsymbol{\beta})\big). \tag{D.56}$$

Then it suffices to prove there exist integers $\widetilde{s}_1$, $\widetilde{s}_2$ and $\widetilde{s}_3$, which satisfy $\widetilde{s}_1+\widetilde{s}_2+\widetilde{s}_3 \leq \widetilde{s}$, such that

$$\big|\big\{j \in \overline{S^*} : |\beta_j| \geq 1/4\cdot\lambda/L\big\}\big| \leq \widetilde{s}_1, \tag{D.57}$$

$$\Big|\Big\{j \in \overline{S^*} : \big|\big(\nabla\widetilde{\mathcal{L}}_\lambda(\boldsymbol{\beta}^*)/L\big)_j\big| > 1/8\cdot\lambda/L\Big\}\Big| \leq \widetilde{s}_2, \tag{D.58}$$

$$\Big|\Big\{j \in \overline{S^*} : \big|\big(\nabla\widetilde{\mathcal{L}}_\lambda(\boldsymbol{\beta})/L - \nabla\widetilde{\mathcal{L}}_\lambda(\boldsymbol{\beta}^*)/L\big)_j\big| \geq 5/8\cdot\lambda/L\Big\}\Big| \leq \widetilde{s}_3. \tag{D.59}$$

This is because, if (D.57)-(D.59) hold, then there are at most $\widetilde{s}_1+\widetilde{s}_2+\widetilde{s}_3 \leq \widetilde{s}$ coordinates $j \in \overline{S^*}$ such that

$$|\beta_j| + \big|\big(\nabla\widetilde{\mathcal{L}}_\lambda(\boldsymbol{\beta}^*)/L\big)_j\big| + \big|\big(\nabla\widetilde{\mathcal{L}}_\lambda(\boldsymbol{\beta})/L - \nabla\widetilde{\mathcal{L}}_\lambda(\boldsymbol{\beta}^*)/L\big)_j\big| > \lambda/L.$$

Since by the triangular inequality (D.56) implies

$$\big|\bar{\beta}_j\big| \leq |\beta_j| + \big|\big(\nabla\widetilde{\mathcal{L}}_\lambda(\boldsymbol{\beta}^*)/L\big)_j\big| + \big|\big(\nabla\widetilde{\mathcal{L}}_\lambda(\boldsymbol{\beta})/L - \nabla\widetilde{\mathcal{L}}_\lambda(\boldsymbol{\beta}^*)/L\big)_j\big|,$$

the number of coordinates $j \in \overline{S^*}$ such that $\big|\bar{\beta}_j\big| > \lambda/L$ is also upper bounded by $\widetilde{s}_1+\widetilde{s}_2+\widetilde{s}_3 \leq \widetilde{s}$. In the following, we will prove (D.58)-(D.59) and specify the corresponding $\widetilde{s}_1$, $\widetilde{s}_2$ and $\widetilde{s}_3$.

**Proof of (D.57):** Note that for $j \in \overline{S^*}$, we have $\beta^*_j = 0$. Hence we have

$$\big|\big\{j \in \overline{S^*} : |\beta_j| \geq 1/4\cdot\lambda/L\big\}\big| = \big|\big\{j \in \overline{S^*} : |\beta_j - \beta^*_j| \geq 1/4\cdot\lambda/L\big\}\big|. \tag{D.60}$$

Meanwhile, note that

$$\begin{aligned}
&\frac{\lambda}{4L}\big|\big\{j \in \overline{S^*} : |\beta_j - \beta^*_j| \geq 1/4\cdot\lambda/L\big\}\big| \\
&\quad\leq \sum_{j\in\overline{S^*}} |\beta_j - \beta^*_j| \cdot \mathbb{1}\big(|\beta_j - \beta^*_j| \geq 1/4\cdot\lambda/L\big) \\
&\quad\leq \sum_{j\in\overline{S^*}} |\beta_j - \beta^*_j| \\
&\quad= \|(\boldsymbol{\beta}-\boldsymbol{\beta}^*)_{\overline{S^*}}\|_1.
\end{aligned} \tag{D.61}$$

Plugging (D.61) into the right-hand side of (D.60), we obtain

$$\big|\big\{j \in \overline{S^*} : |\beta_j| \geq 1/4 \cdot \lambda/L\big\}\big| \leq \frac{4L}{\lambda}\|(\boldsymbol{\beta} - \boldsymbol{\beta}^*)_{\overline{S^*}}\|_1. \tag{D.62}$$

Now we provide an upper bound of $\|(\boldsymbol{\beta} - \boldsymbol{\beta}^*)_{\overline{S^*}}\|_1$. Following the same way we derive (D.46) in the proof of Lemma 5.3, we can obtain

$$\begin{aligned}&\frac{\rho_- - \zeta_-}{2}\|\boldsymbol{\beta} - \boldsymbol{\beta}^*\|_2^2 + \big(\lambda - \|\nabla\mathcal{L}(\boldsymbol{\beta}^*)\|_\infty\big)\|(\boldsymbol{\beta} - \boldsymbol{\beta}^*)_{\overline{S^*}}\|_1 \\ &\leq \big(\lambda + \|\nabla\mathcal{L}(\boldsymbol{\beta}^*)\|_\infty + \|\nabla\mathcal{Q}_\lambda(\boldsymbol{\beta}^*)\|_\infty\big)\|(\boldsymbol{\beta} - \boldsymbol{\beta}^*)_{S^*}\|_1 + \frac{21/2}{\rho_- - \zeta_-}\lambda^2 s^*. \end{aligned}\tag{D.63}$$

According to (4.3) in Wang et al. (2014a), we have $\rho_- - \zeta_- > 0$. Hence (D.63) implies

$$\begin{aligned}&\big(\lambda - \|\nabla\mathcal{L}(\boldsymbol{\beta}^*)\|_\infty\big)\|(\boldsymbol{\beta} - \boldsymbol{\beta}^*)_{\overline{S^*}}\|_1 \\ &\leq \big(\lambda + \|\nabla\mathcal{L}(\boldsymbol{\beta}^*)\|_\infty + \|\nabla\mathcal{Q}_\lambda(\boldsymbol{\beta}^*)\|_\infty\big)\|(\boldsymbol{\beta} - \boldsymbol{\beta}^*)_{S^*}\|_1 + \frac{21/2}{\rho_- - \zeta_-}\lambda^2 s^*. \end{aligned}\tag{D.64}$$

By (4.1) in Assumption 4.1 of Wang et al. (2014a) and $\lambda \geq \lambda_{\rm tgt}$, we have

$$\|\nabla\mathcal{L}(\boldsymbol{\beta}^*)\|_\infty \leq \lambda_{\rm tgt}/8 \leq \lambda/8. \tag{D.65}$$

Meanwhile, since $\mathcal{Q}_\lambda(\boldsymbol{\beta}) = \sum_{j=1}^d q_\lambda(\beta_j)$ and $q_\lambda(\beta_j)$ satisfies regularity condition (d) in Wang et al. (2014a), we have

$$\|\nabla\mathcal{Q}_\lambda(\boldsymbol{\beta}^*)\|_\infty = \max_{1\leq j\leq d}|q'_\lambda(\beta_j^*)| \leq \lambda. \tag{D.66}$$

Plugging (D.65) and (D.66) into (D.64) and dividing $\lambda$ on both sides, we obtain

$$7/8 \cdot \|(\boldsymbol{\beta} - \boldsymbol{\beta}^*)_{\overline{S^*}}\|_1 \leq 17/8 \cdot \|(\boldsymbol{\beta} - \boldsymbol{\beta}^*)_{S^*}\|_1 + \frac{21/2}{\rho_- - \zeta_-}\lambda s^*. \tag{D.67}$$

Now we discuss two cases regarding the relationship between $\|(\boldsymbol{\beta} - \boldsymbol{\beta}^*)_{S^*}\|_1$ and $\lambda s^*$.

- If $7/(\rho_- - \zeta_-) \cdot \lambda s^* < \|(\boldsymbol{\beta} - \boldsymbol{\beta}^*)_{S^*}\|_1$, then we have

  $$\frac{21/2}{\rho_- - \zeta_-}\lambda s^* \leq 3/2 \cdot \|(\boldsymbol{\beta} - \boldsymbol{\beta}^*)_{S^*}\|_1.$$

  Plugging this into the right-hand side of (D.67), we obtain

  $$\|(\boldsymbol{\beta} - \boldsymbol{\beta}^*)_{\overline{S^*}}\|_1 \leq 29/7 \cdot \|(\boldsymbol{\beta} - \boldsymbol{\beta}^*)_{S^*}\|_1,$$

  which implies

  $$\begin{aligned}\|(\boldsymbol{\beta} - \boldsymbol{\beta}^*)_{\overline{S^*}}\|_1 \leq 29/7 \cdot \|(\boldsymbol{\beta} - \boldsymbol{\beta}^*)_{S^*}\|_1 &\leq 29/7 \cdot \sqrt{s^*}\|(\boldsymbol{\beta} - \boldsymbol{\beta}^*)_{S^*}\|_2 \\ &\leq 29/7 \cdot \sqrt{s^*}\|\boldsymbol{\beta} - \boldsymbol{\beta}^*\|_2. \end{aligned}\tag{D.68}$$

Plugging the upper bound of $\|\boldsymbol{\beta}-\boldsymbol{\beta}^*\|_2$ in Lemma 5.3 of Wang et al. (2014a) into the right-hand side of (D.68), we obtain

$$\|(\boldsymbol{\beta}-\boldsymbol{\beta}^*)_{\overline{S^*}}\|_1 \le 29/7 \cdot \sqrt{s^*} \cdot \frac{15/2}{\rho_- - \zeta_-}\lambda\sqrt{s^*} = \frac{435/14}{\rho_- - \zeta_-}\lambda s^*. \tag{D.69}$$

- If $\|(\boldsymbol{\beta}-\boldsymbol{\beta}^*)_{S^*}\|_1 \le 7/(\rho_- - \zeta_-)\cdot\lambda s^*$, then plugging this into the right-hand side of (D.67), we obtain

$$\|(\boldsymbol{\beta}-\boldsymbol{\beta}^*)_{\overline{S^*}}\|_1 \le 8/7 \cdot \frac{17/8\cdot 7 + 21/2}{\rho_- - \zeta_-}\lambda s^* = \frac{29}{\rho_- - \zeta_-}\lambda s^*. \tag{D.70}$$

Combining (D.69) and (D.70), we obtain

$$\|(\boldsymbol{\beta}-\boldsymbol{\beta}^*)_{\overline{S^*}}\|_1 \le \frac{\max\{435/14, 29\}}{\rho_- - \zeta_-}\lambda s^* \le \frac{435/14}{\rho_- - \zeta_-}\lambda s^*.$$

Plugging this into the right-hand side of (D.62), we obtain

$$\big|\{j \in \overline{S^*} : |\beta_j| \ge 1/4\cdot\lambda/L\}\big| \le \frac{4L}{\lambda}\cdot\frac{435/14}{\rho_- - \zeta_-}\lambda s^* < \frac{125L}{\rho_- - \zeta_-}s^*.$$

Meanwhile, since we assume $L < 2(\rho_+ - \zeta_+)$, we have

$$\big|\{j \in \overline{S^*} : |\beta_j| \ge 1/4\cdot\lambda/L\}\big| < 250\cdot\frac{\rho_+ - \zeta_+}{\rho_- - \zeta_-}\cdot s^* = 250\kappa s^*,$$

where the last equality follows from the definition of the condition number $\kappa$ in (4.5). Therefore we obtain (D.57) by setting $\widetilde{s}_1 = 250\kappa s^*$.

**Proof of (D.58):** Recall that $\nabla\widetilde{\mathcal{L}}_\lambda(\boldsymbol{\beta}) = \mathcal{L}(\boldsymbol{\beta}) + \mathcal{Q}_\lambda(\boldsymbol{\beta})$. Hence we have

$$\big\|\big(\nabla\widetilde{\mathcal{L}}_\lambda(\boldsymbol{\beta}^*)\big)_{\overline{S^*}}\big\|_\infty \le \big\|\big(\nabla\mathcal{L}(\boldsymbol{\beta}^*)\big)_{\overline{S^*}}\big\|_\infty + \big\|\big(\nabla\mathcal{Q}_\lambda(\boldsymbol{\beta}^*)\big)_{\overline{S^*}}\big\|_\infty. \tag{D.71}$$

By (4.1) in Assumption 4.1 of Wang et al. (2014a), we have

$$\big\|\big(\nabla\mathcal{L}(\boldsymbol{\beta}^*)\big)_{\overline{S^*}}\big\|_\infty \le \|\nabla\mathcal{L}(\boldsymbol{\beta}^*)\|_\infty \le \lambda/8. \tag{D.72}$$

Recall $\mathcal{Q}_\lambda(\boldsymbol{\beta}) = \sum_{j=1}^d q_\lambda(\beta_j)$, where $q_\lambda(\beta_j)$ satisfies regularity condition (c) that $q'_\lambda(0) = 0$. Hence we have

$$\big\|\big(\nabla\mathcal{Q}_\lambda(\boldsymbol{\beta}^*)\big)_{\overline{S^*}}\big\|_\infty = \max_{j\in\overline{S^*}}\big|q'_\lambda(\beta_j^*)\big| = \max_{j\in\overline{S^*}}\big|q'_\lambda(0)\big| = 0, \tag{D.73}$$

where the second equation follows from the fact that $\beta_j^* = 0$ for $j \in \overline{S^*}$. Plugging (D.73) and (D.72) into the right-hand side of (D.71), we obtain $\big\|\big(\nabla\widetilde{\mathcal{L}}_\lambda(\boldsymbol{\beta}^*)\big)_{\overline{S^*}}\big\|_\infty = \max_{j\in\overline{S^*}}\big|\big(\nabla\widetilde{\mathcal{L}}_\lambda(\boldsymbol{\beta}^*)/L\big)_j\big| \le \lambda/8$. Hence we have

$$\Big|\Big\{j \in \overline{S^*} : \big|\big(\nabla\widetilde{\mathcal{L}}_\lambda(\boldsymbol{\beta}^*)/L\big)_j\big| > 1/8\cdot\lambda/L\Big\}\Big| = 0.$$

Therefore, by setting $\widetilde{s}_2 = 0$, we obtain (D.58).

**Proof of** (D.59)**:** Consider an arbitrary subset $S'$ such that

$$S' \subseteq \Big\{ j : \big| \big( \nabla \widetilde{\mathcal{L}}_\lambda(\boldsymbol{\beta}) - \nabla \widetilde{\mathcal{L}}_\lambda(\boldsymbol{\beta}^*) \big)_j \big| \geq 5/8 \cdot \lambda \Big\}. \tag{D.74}$$

Let $s' = |S'|$. In the sequel we provide an upper bound of $s'$. Suppose $\boldsymbol{v} \in \mathbb{R}^d$ is chosen such that $v_j = \mathrm{sign}\big\{ \big( \nabla \widetilde{\mathcal{L}}_\lambda(\boldsymbol{\beta}) - \nabla \widetilde{\mathcal{L}}_\lambda(\boldsymbol{\beta}^*) \big)_j \big\}$ for $j \in S'$, and $v_j = 0$ for $j \notin S'$. Hence we have

$$\begin{aligned} \boldsymbol{v}^T \big( \nabla \widetilde{\mathcal{L}}_\lambda(\boldsymbol{\beta}) - \nabla \widetilde{\mathcal{L}}_\lambda(\boldsymbol{\beta}^*) \big) &= \sum_{j \in S'} v_j \big( \nabla \widetilde{\mathcal{L}}_\lambda(\boldsymbol{\beta}) - \nabla \widetilde{\mathcal{L}}_\lambda(\boldsymbol{\beta}^*) \big)_j \\ &= \sum_{j \in S'} \big| \big( \nabla \widetilde{\mathcal{L}}_\lambda(\boldsymbol{\beta}) - \nabla \widetilde{\mathcal{L}}_\lambda(\boldsymbol{\beta}^*) \big)_j \big| \geq 5/8 \cdot \lambda s'. \end{aligned} \tag{D.75}$$

Meanwhile, by Cauchy Schwarz inequality we have

$$\begin{aligned} \boldsymbol{v}^T \big( \nabla \widetilde{\mathcal{L}}_\lambda(\boldsymbol{\beta}) - \nabla \widetilde{\mathcal{L}}_\lambda(\boldsymbol{\beta}^*) \big) &\leq \|\boldsymbol{v}\|_2 \big\| \nabla \widetilde{\mathcal{L}}_\lambda(\boldsymbol{\beta}) - \nabla \widetilde{\mathcal{L}}_\lambda(\boldsymbol{\beta}^*) \big\|_2 \\ &\leq \sqrt{s'} \big\| \nabla \widetilde{\mathcal{L}}_\lambda(\boldsymbol{\beta}) - \nabla \widetilde{\mathcal{L}}_\lambda(\boldsymbol{\beta}^*) \big\|_2, \end{aligned} \tag{D.76}$$

where the last inequality follows from the fact that $\|\boldsymbol{v}\|_2 \leq \sqrt{s'} \|\boldsymbol{v}\|_\infty = \sqrt{s'}$, because $\boldsymbol{v}$ is chosen such that $\|\boldsymbol{v}\|_0 = s'$. Combining (D.75) and (D.76) gives

$$5/8 \cdot \lambda s' \leq \boldsymbol{v}^T \big( \nabla \widetilde{\mathcal{L}}_\lambda(\boldsymbol{\beta}) - \nabla \widetilde{\mathcal{L}}_\lambda(\boldsymbol{\beta}^*) \big) \leq \sqrt{s'} \big\| \nabla \widetilde{\mathcal{L}}_\lambda(\boldsymbol{\beta}) - \nabla \widetilde{\mathcal{L}}_\lambda(\boldsymbol{\beta}^*) \big\|_2. \tag{D.77}$$

Since $\|\boldsymbol{\beta}_{\overline{S^*}}\|_0 \leq \widetilde{s}$ and $\|\boldsymbol{\beta}^*_{\overline{S^*}}\|_0 = 0$, we have $\|(\boldsymbol{\beta} - \boldsymbol{\beta}^*)_{\overline{S^*}}\| \leq \widetilde{s}$. In the setting of logistic loss, we further have $\|\boldsymbol{\beta}\|_2 \leq R$ and $\|\boldsymbol{\beta}^*\|_2 \leq R$, where $R$ is specified in Definition 4.3 of Wang et al. (2014a). Therefore, Lemma 5.1 in Wang et al. (2014a) implies that $\widetilde{\mathcal{L}}_\lambda(\boldsymbol{\beta})$ is restricted strongly smooth. Hence we have

$$\widetilde{\mathcal{L}}_\lambda(\boldsymbol{\beta}) \leq \widetilde{\mathcal{L}}_\lambda(\boldsymbol{\beta}^*) + (\boldsymbol{\beta} - \boldsymbol{\beta}^*)^T \nabla \widetilde{\mathcal{L}}_\lambda(\boldsymbol{\beta}^*) + \frac{\rho_+ - \zeta_+}{2} \|\boldsymbol{\beta}^* - \boldsymbol{\beta}\|_2^2. \tag{D.78}$$

According to Nesterov (2004, Theorem 2.1.9), the strong smoothness of $\widetilde{\mathcal{L}}_\lambda(\boldsymbol{\beta})$ is equivalent to the Lipschitz continuity of its gradient, i.e.,

$$\big\| \nabla \widetilde{\mathcal{L}}_\lambda(\boldsymbol{\beta}) - \nabla \widetilde{\mathcal{L}}_\lambda(\boldsymbol{\beta}^*) \big\|_2 \leq (\rho_+ - \zeta_+) \|\boldsymbol{\beta} - \boldsymbol{\beta}^*\|_2. \tag{D.79}$$

Plugging (D.79) into the right-hand side of (D.77), we obtain

$$5/8 \cdot \lambda s' \leq (\rho_+ - \zeta_+) \cdot \sqrt{s'} \|\boldsymbol{\beta} - \boldsymbol{\beta}^*\|_2. \tag{D.80}$$

Plugging the upper bound of $\|\boldsymbol{\beta} - \boldsymbol{\beta}^*\|_2$ in Lemma 5.3 of Wang et al. (2014a) into the right-hand side of (D.80), we obtain

$$\begin{aligned} \sqrt{s'} &\leq \frac{8}{5\lambda} \cdot (\rho_+ - \zeta_+) \|\boldsymbol{\beta} - \boldsymbol{\beta}^*\|_2 \\ &\leq \frac{8}{5\lambda} \cdot (\rho_+ - \zeta_+) \cdot \frac{15/2}{\rho_- - \zeta_-} \lambda \sqrt{s^*} = 12\kappa \sqrt{s^*}, \end{aligned} \tag{D.81}$$

where the last equality follows from the definition of the condition number $\kappa$

in (4.5) of Wang et al. (2014a). Hence we obtain $s' \le 144\kappa^2 s^*$. Note that $S'$ is defined as an arbitrary subset of $\{j : |(\nabla\widetilde{\mathcal{L}}_\lambda(\boldsymbol{\beta}) - \nabla\widetilde{\mathcal{L}}_\lambda(\boldsymbol{\beta}^*))_j| \ge 5/8\cdot\lambda\}$ and

$$\Big\{j \in \overline{S^*} : \big|\big(\nabla\widetilde{\mathcal{L}}_\lambda(\boldsymbol{\beta}) - \nabla\widetilde{\mathcal{L}}_\lambda(\boldsymbol{\beta}^*)\big)_j\big| \ge 5/8\cdot\lambda\Big\} \subseteq \Big\{j : \big|\big(\nabla\widetilde{\mathcal{L}}_\lambda(\boldsymbol{\beta}) - \nabla\widetilde{\mathcal{L}}_\lambda(\boldsymbol{\beta}^*)\big)_j\big| \ge 5/8\cdot\lambda\Big\}.$$

Hence we have

$$\Big|\Big\{j \in \overline{S^*} : \big|\big(\nabla\widetilde{\mathcal{L}}_\lambda(\boldsymbol{\beta})/L - \nabla\widetilde{\mathcal{L}}_\lambda(\boldsymbol{\beta}^*)/L\big)_j\big| \ge 5/8\cdot\lambda/L\Big\}\Big| \le 144\kappa^2 s^*.$$

Therefore, by setting $\widetilde{s}_3 = 144\kappa^2 s^*$, we obtain (D.59).

In summary, we prove that (D.58)-(D.59) hold with $\widetilde{s}_1 = 250\kappa s^*$, $\widetilde{s}_2 = 0$ and $\widetilde{s}_2 = 144\kappa^2 s^*$. In Assumption 4.4, we assume $\widetilde{s} \ge 144\kappa^2 + 250\kappa$, which implies $\widetilde{s}_1 + \widetilde{s}_2 + \widetilde{s}_3 \le \widetilde{s}$. Therefore we have $\big\|\big(\mathcal{T}_{L,\lambda}(\boldsymbol{\beta}; +\infty)\big)_{\overline{S^*}}\big\|_0 < \widetilde{s}$. Since $\mathcal{T}_{L,\lambda}(\boldsymbol{\beta}; R)$ has the same sparsity pattern as $\mathcal{T}_{L,\lambda}(\boldsymbol{\beta}; +\infty)$, we also have that $\big\|\big(\mathcal{T}_{L,\lambda}(\boldsymbol{\beta}; R)\big)_{\overline{S^*}}\big\|_0 < \widetilde{s}$ for $R \in (0, +\infty)$. Hence we conclude the proof. □

**D.6. Proof of Theorem 5.5 in Wang et al. (2014a).** We first provide a useful lemma. It states that if $\boldsymbol{\beta}$ is $\epsilon$-suboptimal with respect to the regularization parameter $\lambda$ and sufficiently sparse, then for $\lambda' \le \lambda$ the objective function value $\phi_{\lambda'}(\boldsymbol{\beta})$ is close to $\phi_{\lambda'}(\widehat{\boldsymbol{\beta}}_{\lambda'})$. Here $\widehat{\boldsymbol{\beta}}_{\lambda'}$ is the exact local solution corresponding to $\lambda'$.

**Lemma D.3.** Let $\lambda \ge \lambda_{\rm tgt}$ and $\lambda' \in [\lambda_{\rm tgt}, \lambda]$. Suppose $\|\boldsymbol{\beta}_{\overline{S^*}}\|_0 \le \widetilde{s}$ and $\omega_\lambda(\boldsymbol{\beta}) \le \epsilon$. Let $\widehat{\boldsymbol{\beta}}_{\lambda'}$ be the exact local solution corresponding to $\lambda'$, which satisfies the exact optimality condition in (3.13) of Wang et al. (2014a) and $\big\|(\widehat{\boldsymbol{\beta}}_{\lambda'})_{\overline{S^*}}\big\|_0 \le \widetilde{s}$. For logistic loss, we further assume $\max\{\|\boldsymbol{\beta}\|_2, \|\widehat{\boldsymbol{\beta}}_{\lambda'}\|_2\} \le R$, where $R$ is specified in Definition 4.3 of Wang et al. (2014a). Under Assumption 4.1 and Assumption 4.4 of Wang et al. (2014a), we have

$$\phi_{\lambda'}(\boldsymbol{\beta}) - \phi_{\lambda'}(\widehat{\boldsymbol{\beta}}_{\lambda'}) \le C\big(\epsilon + 2(\lambda - \lambda')\big)\cdot(\lambda' + \lambda)s^*, \quad \text{where} \quad C = \frac{21}{\rho_- - \zeta_-}.$$

Proof. Since $\|\boldsymbol{\beta}_{\overline{S^*}}\|_0 \le \widetilde{s}$ and $\big\|(\widehat{\boldsymbol{\beta}}_{\lambda'})_{\overline{S^*}}\big\|_0 \le \widetilde{s}$, we have $\big\|(\boldsymbol{\beta} - \widehat{\boldsymbol{\beta}}_{\lambda'})_{\overline{S^*}}\big\| \le 2\widetilde{s}$. In the setting of logistic loss, we further have $\|\boldsymbol{\beta}\|_2 \le R$ and $\|\widehat{\boldsymbol{\beta}}_{\lambda'}\|_2 \le R$. Therefore, Lemma 5.1 of Wang et al. (2014a) gives

$$\begin{aligned}\widetilde{\mathcal{L}}_{\lambda'}(\widehat{\boldsymbol{\beta}}_{\lambda'}) &\ge \widetilde{\mathcal{L}}_{\lambda'}(\boldsymbol{\beta}) + (\widehat{\boldsymbol{\beta}}_{\lambda'} - \boldsymbol{\beta})^T\nabla\widetilde{\mathcal{L}}_{\lambda'}(\boldsymbol{\beta}) + \frac{\rho_- - \zeta_-}{2}\big\|\widehat{\boldsymbol{\beta}}_{\lambda'} - \boldsymbol{\beta}\big\|_2^2 \\ &\ge \widetilde{\mathcal{L}}_{\lambda'}(\boldsymbol{\beta}) + (\widehat{\boldsymbol{\beta}}_{\lambda'} - \boldsymbol{\beta})^T\nabla\widetilde{\mathcal{L}}_{\lambda'}(\boldsymbol{\beta}),\end{aligned} \tag{D.82}$$

where the second inequality is because $\rho_- - \zeta_- > 0$, which follows from (4.3) in Wang et al. (2014a).

Let $\boldsymbol{\xi} \in \partial\|\boldsymbol{\beta}\|_1$ be the subgradient that attains the minimum in

$$\omega_\lambda(\boldsymbol{\beta}) = \min_{\boldsymbol{\xi}' \in \partial\|\boldsymbol{\beta}\|_1} \max_{\boldsymbol{\beta}' \in \Omega} \left\{ \frac{(\boldsymbol{\beta} - \boldsymbol{\beta}')^T}{\|\boldsymbol{\beta} - \boldsymbol{\beta}'\|_1} \big(\nabla \widetilde{\mathcal{L}}_\lambda(\boldsymbol{\beta}) + \lambda \boldsymbol{\xi}'\big) \right\}, \tag{D.83}$$

where $\Omega = B_2(R)$ in the setting of logistic loss and $\Omega = \mathbb{R}^d$ in other settings. Since $\boldsymbol{\xi}$ is a minimizer, we have

$$\omega_\lambda(\boldsymbol{\beta}) = \max_{\boldsymbol{\beta}' \in \Omega} \left\{ \frac{(\boldsymbol{\beta} - \boldsymbol{\beta}')^T}{\|\boldsymbol{\beta} - \boldsymbol{\beta}'\|_1} \big(\nabla \widetilde{\mathcal{L}}_\lambda(\boldsymbol{\beta}) + \lambda \boldsymbol{\xi}\big) \right\}. \tag{D.84}$$

By the convexity of $\ell_1$ norm, we also have

$$\lambda' \big\|\widehat{\boldsymbol{\beta}}_{\lambda'}\big\|_1 \geq \lambda' \|\boldsymbol{\beta}\|_1 + \lambda' \boldsymbol{\xi}^T \big(\widehat{\boldsymbol{\beta}}_{\lambda'} - \boldsymbol{\beta}\big). \tag{D.85}$$

Recall that the objective function $\phi_\lambda(\boldsymbol{\beta})$ is defined as $\phi_\lambda(\boldsymbol{\beta}) = \widetilde{\mathcal{L}}_\lambda(\boldsymbol{\beta}) + \lambda\|\boldsymbol{\beta}\|_1$. Adding (D.82) and (D.85), we obtain

$$\phi_{\lambda'}\big(\widehat{\boldsymbol{\beta}}_{\lambda'}\big) \geq \phi_{\lambda'}(\boldsymbol{\beta}) + \big(\nabla \widetilde{\mathcal{L}}_{\lambda'}(\boldsymbol{\beta}) + \lambda' \boldsymbol{\xi}\big)^T \big(\widehat{\boldsymbol{\beta}}_{\lambda'} - \boldsymbol{\beta}\big).$$

Hence we have

$$\begin{aligned}
&\phi_{\lambda'}(\boldsymbol{\beta}) - \phi_{\lambda'}\big(\widehat{\boldsymbol{\beta}}_{\lambda'}\big) \\
&\quad \leq \big(\nabla \widetilde{\mathcal{L}}_{\lambda'}(\boldsymbol{\beta}) + \lambda' \boldsymbol{\xi}\big)^T \big(\boldsymbol{\beta} - \widehat{\boldsymbol{\beta}}_{\lambda'}\big) \\
&\quad = \Big(\big(\overbrace{\nabla \mathcal{L}(\boldsymbol{\beta}) + \nabla \mathcal{Q}_\lambda(\boldsymbol{\beta})}^{\nabla \widetilde{\mathcal{L}}_\lambda(\boldsymbol{\beta})} + \lambda \boldsymbol{\xi}\big) + \big(\nabla \mathcal{Q}_{\lambda'}(\boldsymbol{\beta}) - \nabla \mathcal{Q}_\lambda(\boldsymbol{\beta})\big) \\
&\qquad\qquad + \big(\lambda' \boldsymbol{\xi} - \lambda \boldsymbol{\xi}\big)\Big)^T \big(\boldsymbol{\beta} - \widehat{\boldsymbol{\beta}}_{\lambda'}\big) \\
&\quad \leq \underbrace{\big(\nabla \widetilde{\mathcal{L}}_\lambda(\boldsymbol{\beta}) + \lambda \boldsymbol{\xi}\big)^T \big(\boldsymbol{\beta} - \widehat{\boldsymbol{\beta}}_{\lambda'}\big)}_{\text{(i)}} + \underbrace{\big\|\nabla \mathcal{Q}_{\lambda'}(\boldsymbol{\beta}) - \nabla \mathcal{Q}_\lambda(\boldsymbol{\beta})\big\|_\infty}_{\text{(ii)}} \underbrace{\big\|\boldsymbol{\beta} - \widehat{\boldsymbol{\beta}}_{\lambda'}\big\|_1}_{\text{(iv)}} \\
&\qquad\qquad + \underbrace{\|\lambda' \boldsymbol{\xi} - \lambda \boldsymbol{\xi}\|_\infty}_{\text{(iii)}} \underbrace{\big\|\boldsymbol{\beta} - \widehat{\boldsymbol{\beta}}_{\lambda'}\big\|_1}_{\text{(iv)}}.
\end{aligned} \tag{D.86}$$

Now we provide upper bounds of terms (i)-(iv) correspondingly.

**Bounding Term (i) in (D.86):** According to (D.84), we have

$$\frac{\big(\boldsymbol{\beta} - \widehat{\boldsymbol{\beta}}_{\lambda'}\big)^T}{\big\|\boldsymbol{\beta} - \widehat{\boldsymbol{\beta}}_{\lambda'}\big\|_1} \big(\nabla \widetilde{\mathcal{L}}_\lambda(\boldsymbol{\beta}) + \lambda \boldsymbol{\xi}\big) \leq \max_{\boldsymbol{\beta}' \in \Omega} \left\{ \frac{(\boldsymbol{\beta} - \boldsymbol{\beta}')^T}{\|\boldsymbol{\beta} - \boldsymbol{\beta}'\|_1} \big(\nabla \widetilde{\mathcal{L}}_\lambda(\boldsymbol{\beta}) + \lambda \boldsymbol{\xi}\big) \right\} = \omega_\lambda(\boldsymbol{\beta}) \leq \epsilon,$$

where the last inequality is our assumption. Therefore we obtain

$$\big(\nabla \widetilde{\mathcal{L}}_\lambda(\boldsymbol{\beta}) + \lambda \boldsymbol{\xi}\big)^T \big(\boldsymbol{\beta} - \widehat{\boldsymbol{\beta}}_{\lambda'}\big) \leq \epsilon \cdot \big\|\boldsymbol{\beta} - \widehat{\boldsymbol{\beta}}_{\lambda'}\big\|_1. \tag{D.87}$$

We will provide an upper bound of $\big\|\boldsymbol{\beta} - \widehat{\boldsymbol{\beta}}_{\lambda'}\big\|_1$ when we handle term (iv).

**Bounding Term (ii) in (D.86):** Recall $\mathcal{Q}_\lambda(\boldsymbol{\beta}) = \sum_{i=1}^d q_\lambda(\beta_j)$. We have

$$\begin{aligned} \big\|\nabla \mathcal{Q}_{\lambda'}(\boldsymbol{\beta}) - \nabla \mathcal{Q}_{\lambda}(\boldsymbol{\beta})\big\|_\infty &= \max_{1\le j\le d}\big|q_{\lambda'}(\beta_j) - q_\lambda(\beta_j)\big| \\ &\le \max_{1\le j\le d}|\lambda' - \lambda| = \lambda - \lambda', \end{aligned} \tag{D.88}$$

where the inequality follows from regularity condition (e), the last equality is because $\lambda \ge \lambda'$.

**Bounding Term (iii) in (D.86):** Since $\boldsymbol{\xi} \in \partial\|\boldsymbol{\beta}\|_1$, we have $\|\boldsymbol{\xi}\|_\infty \le 1$. Then we obtain

$$\|\lambda'\boldsymbol{\xi} - \lambda\boldsymbol{\xi}\|_\infty = |\lambda' - \lambda|\|\boldsymbol{\xi}\|_\infty \le |\lambda - \lambda'| = \lambda - \lambda'. \tag{D.89}$$

**Bounding Term (iv) in (D.86):** Note that

$$\big\|\boldsymbol{\beta} - \widehat{\boldsymbol{\beta}}_{\lambda'}\big\|_1 \le \underbrace{\|\boldsymbol{\beta} - \boldsymbol{\beta}^*\|_1}_{\text{(iv).a}} + \underbrace{\big\|\widehat{\boldsymbol{\beta}}_{\lambda'} - \boldsymbol{\beta}^*\big\|_1}_{\text{(iv).b}}. \tag{D.90}$$

For term (iv).a, since $\boldsymbol{\beta}$ satisfies $\|\boldsymbol{\beta}_{\overline{S^*}}\|_0 \le \widetilde{s}$, $\omega_\lambda(\boldsymbol{\beta}) \le \lambda/2$, and $\|\boldsymbol{\beta}\|_2 \le R$ for logistic loss, we have that $\boldsymbol{\beta}$ satisfies the assumptions of Lemma 5.2 in Wang et al. (2014a). Following the same way we obtain (D.37) in the proof of Lemma 5.2 in Wang et al. (2014a), we can get

$$(\lambda/2 - \lambda/8)\|(\boldsymbol{\beta} - \boldsymbol{\beta}^*)_{\overline{S^*}}\|_1 \le (3\lambda/2 + \lambda/8 + \lambda)\|(\boldsymbol{\beta} - \boldsymbol{\beta}^*)_{S^*}\|_1,$$

which implies $\|(\boldsymbol{\beta} - \boldsymbol{\beta}^*)_{\overline{S^*}}\|_1 \le 7\|(\boldsymbol{\beta} - \boldsymbol{\beta}^*)_{S^*}\|_1$. Hence we obtain

$$\begin{aligned} \|\boldsymbol{\beta} - \boldsymbol{\beta}^*\|_1 \le \|(\boldsymbol{\beta} - \boldsymbol{\beta}^*)_{\overline{S^*}}\|_1 + \|(\boldsymbol{\beta} - \boldsymbol{\beta}^*)_{S^*}\|_1 &\le 8\|(\boldsymbol{\beta} - \boldsymbol{\beta}^*)_{S^*}\|_1 \\ &\le 8\sqrt{s^*}\|(\boldsymbol{\beta} - \boldsymbol{\beta}^*)_{S^*}\|_2 \\ &\le 8\sqrt{s^*}\|\boldsymbol{\beta} - \boldsymbol{\beta}^*\|_2. \end{aligned}$$

With the upper bound of $\|\boldsymbol{\beta} - \boldsymbol{\beta}^*\|_2$ in Lemma 5.2 of Wang et al. (2014a), we obtain

$$\|\boldsymbol{\beta} - \boldsymbol{\beta}^*\|_1 \le \frac{21}{\rho_- - \zeta_-}\lambda s^*. \tag{D.91}$$

Meanwhile, for term (iv).b, note that we assume $\widehat{\boldsymbol{\beta}}_{\lambda'}$ satisfies $\big\|\big(\widehat{\boldsymbol{\beta}}_{\lambda'}\big)_{\overline{S^*}}\big\|_0 \le \widetilde{s}$ and $\big\|\widehat{\boldsymbol{\beta}}_{\lambda'}\big\|_2 \le R$ for logistic loss. Since $\widehat{\boldsymbol{\beta}}_{\lambda'}$ is an exact local solution, it satisfies the exact optimality condition $\omega\big(\widehat{\boldsymbol{\beta}}_{\lambda'}\big) \le 0$, which gives $\omega\big(\widehat{\boldsymbol{\beta}}_{\lambda'}\big) < \lambda'/2$. Hence $\widehat{\boldsymbol{\beta}}_{\lambda'}$ also satisfies the conditions of Lemma 5.2 of Wang et al. (2014a). Similar to (D.91), we have

$$\big\|\widehat{\boldsymbol{\beta}}_{\lambda'} - \boldsymbol{\beta}^*\big\|_1 \le \frac{21}{\rho_- - \zeta_-}\lambda' s^*. \tag{D.92}$$

Plugging (D.92) and (D.91) into (D.90), for term (iv) in (D.86), we obtain

$$\|\boldsymbol{\beta} - \widehat{\boldsymbol{\beta}}_{\lambda'}\|_1 \le \frac{21}{\rho_- - \zeta_-}(\lambda' + \lambda)s^*. \tag{D.93}$$

Plugging (D.87)-(D.89) and (D.93) into the right-hand side of (D.86), we obtain

$$\begin{aligned}
&\phi_{\lambda'}(\boldsymbol{\beta}) - \phi_{\lambda'}(\widehat{\boldsymbol{\beta}}_{\lambda'}) \\
&\le \underbrace{\epsilon \cdot \frac{21}{\rho_- - \zeta_-}(\lambda' + \lambda)s^*}_{\text{(i) in (D.86)}} + \Big(\underbrace{(\lambda - \lambda')}_{\text{(ii) in (D.86)}} + \underbrace{(\lambda - \lambda')}_{\text{(iii) in (D.86)}}\Big) \cdot \underbrace{\frac{21}{\rho_- - \zeta_-}(\lambda' + \lambda)s^*}_{\text{(iv) in (D.86)}} \\
&\le \frac{21}{\rho_- - \zeta_-}\big(\epsilon + 2(\lambda - \lambda')\big) \cdot (\lambda' + \lambda)s^*,
\end{aligned}$$

where the upper bound of term (i) in (D.86) is obtained by plugging (D.93) into the right-hand side of (D.87). Hence we conclude the proof. □

Now we are ready to prove Theorem 5.5 of Wang et al. (2014a).

PROOF. **Sparsity of $\{\boldsymbol{\beta}_t^k\}_{k=0}^{\infty}$ within the $t$-th Stage:** In the following, we provide results concerning the sparsity of the sequence $\{\boldsymbol{\beta}_t^k\}_{k=0}^{\infty}$ within the $t$-th path following stage. In the following we prove this by induction. Note that the initialization satisfies

$$\big\|(\boldsymbol{\beta}_t^0)_{\overline{S^*}}\big\|_0 \le \widetilde{s}, \quad \omega_{\lambda_t}(\boldsymbol{\beta}_t^0) \le \lambda_t/2, \quad \text{and} \quad L_t^0 \le 2(\rho_+ - \zeta_+). \tag{D.94}$$

By Lemma 5.2 of Wang et al. (2014a) we have

$$\phi_{\lambda_t}(\boldsymbol{\beta}_t^0) - \phi_{\lambda_t}(\boldsymbol{\beta}^*) \le \frac{21/2}{\rho_- - \zeta_-}\lambda_t^2 s^*, \tag{D.95}$$

Suppose that, at the $(k-1)$-th iteration of the proximal-gradient method (Lines 5-9 of Algorithm 3 in Wang et al. (2014a)), we have

(D.96)

$$\big\|(\boldsymbol{\beta}_t^{k-1})_{\overline{S^*}}\big\|_0 \le \widetilde{s}, \ L_t^{k-1} \le 2(\rho_+ - \zeta_+), \ \phi_{\lambda_t}(\boldsymbol{\beta}_t^{k-1}) - \phi_{\lambda_t}(\boldsymbol{\beta}^*) \le \frac{21/2}{\rho_- - \zeta_-}\lambda_t^2 s^*,$$

Then by Lemma 5.4 in Wang et al. (2014a), we have that $\boldsymbol{\beta}_t^k = \mathcal{T}_{L_t^k, \lambda_t}(\boldsymbol{\beta}_t^{k-1}; R)$ satisfies

$$\big\|(\boldsymbol{\beta}_t^k)_{\overline{S^*}}\big\|_0 \le \widetilde{s}. \tag{D.97}$$

Note that, in the setting of logistic loss, we always have $\|\boldsymbol{\beta}_t^k\|_2 \le R$ for $k = 0, 1, \ldots$ because of the $\ell_2$ constraint $\Omega = B_2(R)$. Since $\|(\boldsymbol{\beta}_t^{k-1})_{\overline{S^*}}\|_0 \le \widetilde{s}$ and $\|(\boldsymbol{\beta}_t^k)_{\overline{S^*}}\|_0 \le \widetilde{s}$ imply $\|(\boldsymbol{\beta}_t^{k-1} - \boldsymbol{\beta}_t^k)_{\overline{S^*}}\| \le 2\widetilde{s}$, from Lemma 5.1 of Wang

et al. (2014a) we have

$$\widetilde{\mathcal{L}}_{\lambda_t}\big(\boldsymbol{\beta}_t^k\big) \geq \widetilde{\mathcal{L}}_{\lambda_t}\big(\boldsymbol{\beta}_t^{k-1}\big) + \nabla\widetilde{\mathcal{L}}_{\lambda_t}\big(\boldsymbol{\beta}_t^{k-1}\big)^T\big(\boldsymbol{\beta}_t^k - \boldsymbol{\beta}_t^{k-1}\big) + \frac{\rho_- - \zeta_-}{2}\big\|\boldsymbol{\beta}_t^k - \boldsymbol{\beta}_t^{k-1}\big\|_2^2, \tag{D.98}$$

$$\widetilde{\mathcal{L}}_{\lambda_t}\big(\boldsymbol{\beta}_t^k\big) \leq \widetilde{\mathcal{L}}_{\lambda_t}\big(\boldsymbol{\beta}_t^{k-1}\big) + \nabla\widetilde{\mathcal{L}}_{\lambda_t}\big(\boldsymbol{\beta}_t^{k-1}\big)^T\big(\boldsymbol{\beta}_t^k - \boldsymbol{\beta}_t^{k-1}\big) + \frac{\rho_+ - \zeta_+}{2}\big\|\boldsymbol{\beta}_t^k - \boldsymbol{\beta}_t^{k-1}\big\|_2^2. \tag{D.99}$$

Now we prove that (D.99) guarantees the line-search method in Algorithm 2 of Wang et al. (2014a) produces $L_t^k \leq 2(\rho_+ - \zeta_+)$. We prove by contradiction: We assume that, when the line-search method stops, it outputs $L_t^k > 2(\rho_+ - \zeta_+)$. Recall that we double $L_t^k$ at each line-search iteration (Line 6 of Algorithm 2 in Wang et al. (2014a)). Then at the line-search iteration right before the line-search method stops, we have $L_t^{k\prime} = L_t^k/2 > (\rho_+ - \zeta_+)$. Recall that the objective function $\phi_\lambda(\boldsymbol{\beta}) = \widetilde{\mathcal{L}}_\lambda(\boldsymbol{\beta}) + \lambda\|\boldsymbol{\beta}\|_1$. Adding $\lambda_t\big\|\boldsymbol{\beta}_t^k\big\|_1$ to the both sides of (D.99), we obtain

$$\begin{aligned}
\phi_{\lambda_t}\big(\boldsymbol{\beta}_t^k\big) &= \widetilde{\mathcal{L}}_{\lambda_t}\big(\boldsymbol{\beta}_t^k\big) + \lambda_t\big\|\boldsymbol{\beta}_t^k\big\|_1 \\
&\leq \widetilde{\mathcal{L}}_{\lambda_t}\big(\boldsymbol{\beta}_t^{k-1}\big) + \nabla\widetilde{\mathcal{L}}_{\lambda_t}\big(\boldsymbol{\beta}_t^{k-1}\big)^T\big(\boldsymbol{\beta}_t^k - \boldsymbol{\beta}_t^{k-1}\big) \\
&\qquad + \frac{\rho_+ - \zeta_+}{2}\big\|\boldsymbol{\beta}_t^k - \boldsymbol{\beta}_t^{k-1}\big\|_2^2 + \lambda_t\big\|\boldsymbol{\beta}_t^k\big\|_1 \\
&\leq \widetilde{\mathcal{L}}_{\lambda_t}\big(\boldsymbol{\beta}_t^{k-1}\big) + \nabla\widetilde{\mathcal{L}}_{\lambda_t}\big(\boldsymbol{\beta}_t^{k-1}\big)^T\big(\boldsymbol{\beta}_t^k - \boldsymbol{\beta}_t^{k-1}\big) \\
&\qquad + \frac{L_t^{k\prime}}{2}\big\|\boldsymbol{\beta}_t^k - \boldsymbol{\beta}_t^{k-1}\big\|_2^2 + \lambda_t\big\|\boldsymbol{\beta}_t^k\big\|_1 \\
&= \psi_{L_t^{k\prime},\lambda_t}\big(\boldsymbol{\beta}_t^k;\boldsymbol{\beta}_t^{k-1}\big),
\end{aligned}$$

where the last equality follows from (3.7) of Wang et al. (2014a). The stopping criterion of Algorithm 2 in Wang et al. (2014a) implies that the line-search method should have already stopped and give $\big(L_t^k\big)' = L_t^k/2$, which contradicts our assumption that the line-search method outputs $L_t^k$. Therefore we have

$$L_t^k \leq 2(\rho_+ - \zeta_+). \tag{D.100}$$

Moreover, according to (D.98) and (D.99), Lemma D.1 holds, i.e.,

$$\phi_{\lambda_t}\big(\boldsymbol{\beta}_t^k\big) \leq \phi_{\lambda_t}\big(\boldsymbol{\beta}_t^{k-1}\big) - \frac{L_t^k}{2}\big\|\boldsymbol{\beta}_t^k - \boldsymbol{\beta}_t^{k-1}\big\|_2^2, \tag{D.101}$$

which implies

$$\phi_{\lambda_t}(\boldsymbol{\beta}_t^k) - \phi_{\lambda_t}(\boldsymbol{\beta}^*) \le \phi_{\lambda_t}(\boldsymbol{\beta}_t^{k-1}) - \frac{L_t^k}{2}\|\boldsymbol{\beta}_t^k - \boldsymbol{\beta}_t^{k-1}\|_2^2 - \phi_{\lambda_t}(\boldsymbol{\beta}^*)$$

$$\le \frac{21/2}{\rho_- - \zeta_-}\lambda_t^2 s^*. \tag{D.102}$$

According to (D.97) and (D.100)-(D.102), now we have

(D.103)

$$\|(\boldsymbol{\beta}_t^k)_{\overline{S^*}}\|_0 \le \widetilde{s}, \quad L_t^k \le 2(\rho_+ - \zeta_+), \quad \phi_{\lambda_t}(\boldsymbol{\beta}_t^k) - \phi_{\lambda_t}(\boldsymbol{\beta}^*) \le \frac{21/2}{\rho_- - \zeta_-}\lambda_t^2 s^*.$$

Combining (D.94), (D.96) and (D.103), by induction we prove that (D.103) holds for all $k = 0, 1, \ldots$ within the $t$-th path following stage. Furthermore, by Lemma 5.3 of Wang et al. (2014a), all $\boldsymbol{\beta}_t^k$'s have nice statistical recovery properties, i.e.,

$$\|\boldsymbol{\beta}_t^k - \boldsymbol{\beta}^*\|_2 \le \frac{15/2}{\rho_- - \zeta_-}\lambda_t\sqrt{s^*}, \quad \text{for} \quad k = 0, 1, \ldots.$$

**Convergence to Unique Local Solution:** In the following, we prove that, within the $t$-th path following stage, the limit point of the sequence $\{\boldsymbol{\beta}_t^k\}_{k=0}^\infty$ generated by Algorithm 3 in Wang et al. (2014a) is unique and also an exact local solution. Since $\|(\boldsymbol{\beta}_t^0)_{\overline{S^*}}\| \le \widetilde{s}$, the restricted strong convexity of $\widetilde{\mathcal{L}}_\lambda(\boldsymbol{\beta})$ in Lemma 5.1 of Wang et al. (2014a) implies that the sub-level set

$$\{\boldsymbol{\beta} : \phi_{\lambda_t}(\boldsymbol{\beta}) \le \phi_{\lambda_t}(\boldsymbol{\beta}_t^0), \ \|(\boldsymbol{\beta}_t^0 - \boldsymbol{\beta})_{\overline{S^*}}\| \le 2\widetilde{s}\}$$

is bounded. From (D.101) and (D.103) we have

$$\phi_{\lambda_t}(\boldsymbol{\beta}_t^k) \le \phi_{\lambda_t}(\boldsymbol{\beta}_t^0) \quad \text{and} \quad \|(\boldsymbol{\beta}_t^k)_{\overline{S^*}}\|_0 \le \widetilde{s}, \quad \text{for} \quad k = 1, 2, \ldots.$$

Thus $\{\boldsymbol{\beta}_t^k\}_{k=0}^\infty$ is bounded, which implies that $\{\phi_{\lambda_t}(\boldsymbol{\beta}_t^k)\}_{k=0}^\infty$ is also bounded. Meanwhile, (D.101) implies that $\{\phi_{\lambda_t}(\boldsymbol{\beta}_t^k)\}_{k=0}^\infty$ decreases monotonically. By the Bolzano-Weierstrass theorem, the limit point of $\{\phi_{\lambda_t}(\boldsymbol{\beta}_t^k)\}_{k=0}^\infty$ is unique, which implies

$$\lim_{k\to\infty}\Big\{\phi_{\lambda_t}(\boldsymbol{\beta}_t^k) - \phi_{\lambda_t}(\boldsymbol{\beta}_t^{k-1})\Big\} = 0.$$

Consequently, by (D.101) we have that, for any limit point of $\{\boldsymbol{\beta}^k\}_{k=0}^\infty$,

$$\lim_{k\to\infty}\Big\{\|\boldsymbol{\beta}_t^k - \boldsymbol{\beta}_t^{k-1}\|_2\Big\} \le \frac{2}{L_t^k} \cdot \lim_{k\to\infty}\Big\{\phi_{\lambda_t}(\boldsymbol{\beta}_t^k) - \phi_{\lambda_t}(\boldsymbol{\beta}_t^{k-1})\Big\} = 0.$$

Moreover, Lemma D.2 implies

$$\lim_{k\to\infty}\Big\{\omega_{\lambda_t}(\boldsymbol{\beta}_t^k)\Big\} \le \Big(L_t^k + (\rho_+ - \zeta_+)\Big) \cdot \lim_{k\to\infty}\Big\{\|\boldsymbol{\beta}_t^k - \boldsymbol{\beta}_t^{k-1}\|_2\Big\} = 0.$$

In other words, the sequence $\{\boldsymbol{\beta}_t^k\}_{k=0}^{\infty}$ has a convergent subsequence, which satisfies $\lim_{k\to\infty}\{\omega_{\lambda_t}(\boldsymbol{\beta}_t^k)\} \le 0$. Furthermore, it implies that this convergent subsequence of $\{\boldsymbol{\beta}_t^k\}_{k=0}^{\infty}$ converges towards an exact local solution $\widehat{\boldsymbol{\beta}}_{\lambda_t}$ that satisfies the optimal condition in (3.13) of Wang et al. (2014a). By (D.103) we have $\|(\boldsymbol{\beta}_t^k)_{\overline{S^*}}\|_0 \le \widetilde{s}$ $(k = 1, 2, \ldots)$, which implies $\|(\widehat{\boldsymbol{\beta}}_{\lambda_t})_{\overline{S^*}}\|_0 \le \widetilde{s}$.

Now we prove the uniqueness of this exact local solution by contradiction. Let $\boldsymbol{\xi} \in \partial\|\widehat{\boldsymbol{\beta}}_{\lambda_t}\|_1$ be the subgradient that attains the minimum in

$$\omega_{\lambda_t}(\widehat{\boldsymbol{\beta}}_{\lambda_t}) = \min_{\boldsymbol{\xi}'\in\partial\|\widehat{\boldsymbol{\beta}}_{\lambda_t}\|_1} \max_{\boldsymbol{\beta}'\in\Omega} \left\{ \frac{(\widehat{\boldsymbol{\beta}}_{\lambda_t} - \boldsymbol{\beta}')^T}{\|\widehat{\boldsymbol{\beta}}_{\lambda_t} - \boldsymbol{\beta}'\|_1} \left(\nabla\widetilde{\mathcal{L}}_{\lambda_t}(\widehat{\boldsymbol{\beta}}_{\lambda_t}) + \lambda_t\boldsymbol{\xi}'\right) \right\}. \tag{D.104}$$

Since $\omega_{\lambda_t}(\widehat{\boldsymbol{\beta}}_{\lambda_t}) \le 0$, we have

$$\max_{\boldsymbol{\beta}'\in\Omega} \left\{ \frac{(\widehat{\boldsymbol{\beta}}_{\lambda_t} - \boldsymbol{\beta}')^T}{\|\widehat{\boldsymbol{\beta}}_{\lambda_t} - \boldsymbol{\beta}'\|_1} \left(\nabla\widetilde{\mathcal{L}}_{\lambda_t}(\widehat{\boldsymbol{\beta}}_{\lambda_t}) + \lambda_t\boldsymbol{\xi}\right) \right\} \le 0. \tag{D.105}$$

We assume there exists another local solution $\widehat{\boldsymbol{\beta}}'_{\lambda_t}$, which is the limit point of another convergent subsequence of $\{\boldsymbol{\beta}_t^k\}_{k=0}^{\infty}$. Since $\|(\widehat{\boldsymbol{\beta}}'_{\lambda_t})_{\overline{S^*}}\|_0 \le \widetilde{s}$, we have $\|(\widehat{\boldsymbol{\beta}}'_{\lambda_t} - \widehat{\boldsymbol{\beta}}_{\lambda_t})_{\overline{S^*}}\| \le 2\widetilde{s}$. In the setting of logistic loss, we have $\|\widehat{\boldsymbol{\beta}}'_{\lambda_t}\|_2 \le R$ and $\|\widehat{\boldsymbol{\beta}}_{\lambda_t}\|_2 \le R$ by the $\ell_2$ constraint. Hence Lemma 5.1 of Wang et al. (2014a) implies

$$\begin{aligned}\widetilde{\mathcal{L}}_{\lambda_t}(\widehat{\boldsymbol{\beta}}'_{\lambda_t}) \ge \widetilde{\mathcal{L}}_{\lambda_t}(\widehat{\boldsymbol{\beta}}_{\lambda_t}) &+ (\widehat{\boldsymbol{\beta}}'_{\lambda_t} - \widehat{\boldsymbol{\beta}}_{\lambda_t})^T\nabla\widetilde{\mathcal{L}}_{\lambda_t}(\widehat{\boldsymbol{\beta}}_{\lambda_t}) \\ &+ \frac{\rho_- - \zeta_-}{2}\|\widehat{\boldsymbol{\beta}}'_{\lambda_t} - \widehat{\boldsymbol{\beta}}_{\lambda_t}\|_2^2.\end{aligned} \tag{D.106}$$

Meanwhile, the convexity of $\ell_1$ norm implies

$$\lambda_t\|\widehat{\boldsymbol{\beta}}'_{\lambda_t}\|_1 \ge \lambda_t\|\widehat{\boldsymbol{\beta}}_{\lambda_t}\|_1 + \lambda_t(\widehat{\boldsymbol{\beta}}'_{\lambda_t} - \widehat{\boldsymbol{\beta}}_{\lambda_t})^T\boldsymbol{\xi}. \tag{D.107}$$

Recall that the objective function $\phi_\lambda(\boldsymbol{\beta}) = \widetilde{\mathcal{L}}_\lambda(\boldsymbol{\beta}) + \lambda\|\boldsymbol{\beta}\|_1$. Adding (D.106) and (D.107), we obtain

$$\begin{aligned}&\phi_{\lambda_t}(\widehat{\boldsymbol{\beta}}'_{\lambda_t}) - \phi_{\lambda_t}(\widehat{\boldsymbol{\beta}}_{\lambda_t}) \\ &\ge \underbrace{\left(\nabla\widetilde{\mathcal{L}}_{\lambda_t}(\widehat{\boldsymbol{\beta}}_{\lambda_t}) + \lambda_t\boldsymbol{\xi}\right)^T(\widehat{\boldsymbol{\beta}}'_{\lambda_t} - \widehat{\boldsymbol{\beta}}_{\lambda_t})}_{\text{(i)}} + \frac{\rho_- - \zeta_-}{2}\|\widehat{\boldsymbol{\beta}}'_{\lambda_t} - \widehat{\boldsymbol{\beta}}_{\lambda_t}\|_2^2.\end{aligned} \tag{D.108}$$

Since (D.105) implies

$$\begin{aligned}\frac{(\widehat{\boldsymbol{\beta}}_{\lambda_t} - \widehat{\boldsymbol{\beta}}'_{\lambda_t})^T}{\|\widehat{\boldsymbol{\beta}}_{\lambda_t} - \widehat{\boldsymbol{\beta}}'_{\lambda_t}\|_1}\left(\nabla\widetilde{\mathcal{L}}_{\lambda_t}(\widehat{\boldsymbol{\beta}}_{\lambda_t}) + \lambda_t\boldsymbol{\xi}\right) &\le \max_{\boldsymbol{\beta}'\in\Omega}\left\{\frac{(\widehat{\boldsymbol{\beta}}_{\lambda_t} - \boldsymbol{\beta}')^T}{\|\widehat{\boldsymbol{\beta}}_{\lambda_t} - \boldsymbol{\beta}'\|_1}\left(\nabla\widetilde{\mathcal{L}}_{\lambda_t}(\widehat{\boldsymbol{\beta}}_{\lambda_t}) + \lambda_t\boldsymbol{\xi}\right)\right\} \\ &\le 0,\end{aligned}$$

term (i) in (D.108) is nonnegative. Hence we obtain

$$\phi_{\lambda_t}(\widehat{\boldsymbol{\beta}}'_{\lambda_t}) - \phi_{\lambda_t}(\widehat{\boldsymbol{\beta}}_{\lambda_t}) \geq \frac{\rho_- - \zeta_-}{2} \big\|\widehat{\boldsymbol{\beta}}'_{\lambda_t} - \widehat{\boldsymbol{\beta}}_{\lambda_t}\big\|_2^2. \tag{D.109}$$

Recall we already know that the limit point of $\{\phi_{\lambda_t}(\boldsymbol{\beta}_t^k)\}_{k=0}^{\infty}$ is unique, which implies $\phi_{\lambda_t}(\widehat{\boldsymbol{\beta}}'_{\lambda_t}) - \phi_{\lambda_t}(\widehat{\boldsymbol{\beta}}_{\lambda_t}) = 0$. Then we obtain $\big\|\widehat{\boldsymbol{\beta}}'_{\lambda_t} - \widehat{\boldsymbol{\beta}}_{\lambda_t}\big\|_2^2 = 0$, which contradicts our assumption that $\widehat{\boldsymbol{\beta}}'_{\lambda_t} \neq \widehat{\boldsymbol{\beta}}_{\lambda_t}$. In other words, we prove that the sequence $\{\boldsymbol{\beta}_t^k\}_{k=0}^{\infty}$ converges to a unique local solution $\widehat{\boldsymbol{\beta}}_{\lambda_t}$.

**Geometric Convergence Rate of Algorithm 3 in Wang et al. (2014a):** Now we establish the geometric rate of convergence of Algorithm 3. According to the stopping criterion of Algorithm 2 in Wang et al. (2014a), we have

$$\begin{aligned}
\phi_{\lambda_t}(\boldsymbol{\beta}_t^k) &\leq \psi_{L_t^k,\lambda_t}(\boldsymbol{\beta}_t^k; \boldsymbol{\beta}_t^{k-1}) \\
&= \min_{\boldsymbol{\beta}} \Big\{ \widetilde{\mathcal{L}}_{\lambda_t}(\boldsymbol{\beta}_t^{k-1}) + \nabla \widetilde{\mathcal{L}}_{\lambda_t}(\boldsymbol{\beta}_t^{k-1})^T (\boldsymbol{\beta} - \boldsymbol{\beta}_t^{k-1}) \\
&\qquad + \frac{L_t^k}{2} \|\boldsymbol{\beta} - \boldsymbol{\beta}_t^{k-1}\|_2^2 + \lambda_t \|\boldsymbol{\beta}\|_1 \Big\} \\
&\leq \min_{\substack{\boldsymbol{\beta} = \alpha\widehat{\boldsymbol{\beta}}_{\lambda_t} + (1-\alpha)\boldsymbol{\beta}_t^{k-1} \\ \alpha \in [0,1]}} \Big\{ \overbrace{\widetilde{\mathcal{L}}_{\lambda_t}(\boldsymbol{\beta}_t^{k-1}) + \nabla \widetilde{\mathcal{L}}_{\lambda_t}(\boldsymbol{\beta}_t^{k-1})^T (\boldsymbol{\beta} - \boldsymbol{\beta}_t^{k-1})}^{\text{(i)}} \\
&\qquad + \frac{L_t^k}{2} \|\boldsymbol{\beta} - \boldsymbol{\beta}_t^{k-1}\|_2^2 + \lambda_t \|\boldsymbol{\beta}\|_1 \Big\}. 
\end{aligned} \tag{D.110}$$

For term (i), since $\big\|(\boldsymbol{\beta}_t^{k-1})_{\overline{S^*}}\big\|_0 \leq \widetilde{s}$, $\big\|(\widehat{\boldsymbol{\beta}}_{\lambda_t})_{\overline{S^*}}\big\|_0 \leq \widetilde{s}$ and $\boldsymbol{\beta} = \alpha\widehat{\boldsymbol{\beta}}_{\lambda_t} + (1-\alpha)\boldsymbol{\beta}_t^{k-1}$ with $\alpha \in [0,1]$, we obtain $\big\|(\boldsymbol{\beta} - \boldsymbol{\beta}_t^{k-1})_{\overline{S^*}}\big\|_0 \leq 2\widetilde{s}$. For logistic loss, since $\big\|\boldsymbol{\beta}_t^{k-1}\big\|_2 \leq R$ and $\big\|\widehat{\boldsymbol{\beta}}_{\lambda_t}\big\|_2 \leq R$, we have $\|\boldsymbol{\beta}\|_2 \leq R$, since the $\ell_2$ ball $B_2(R)$ is a convex set. Applying Lemma 5.1 in Wang et al. (2014a), we have

$$\begin{aligned}
\widetilde{\mathcal{L}}_{\lambda_t}(\boldsymbol{\beta}) &\geq \widetilde{\mathcal{L}}_{\lambda_t}(\boldsymbol{\beta}_t^{k-1}) + \nabla \widetilde{\mathcal{L}}_{\lambda_t}(\boldsymbol{\beta}_t^{k-1})^T (\boldsymbol{\beta} - \boldsymbol{\beta}_t^{k-1}) \\
&\qquad + \frac{\rho_- - \zeta_-}{2} \|\boldsymbol{\beta} - \boldsymbol{\beta}_t^{k-1}\|_2^2 \\
&\geq \widetilde{\mathcal{L}}_{\lambda_t}(\boldsymbol{\beta}_t^{k-1}) + \nabla \widetilde{\mathcal{L}}_{\lambda_t}(\boldsymbol{\beta}_t^{k-1})^T (\boldsymbol{\beta} - \boldsymbol{\beta}_t^{k-1}),
\end{aligned} \tag{D.111}$$

where the second inequality follows from (4.3) in Wang et al. (2014a). Plugging (D.111) into (D.110), we obtain

$$\phi_{\lambda_t}(\boldsymbol{\beta}_t^k) \leq \min_{\substack{\boldsymbol{\beta} = \alpha\widehat{\boldsymbol{\beta}}_{\lambda_t} + (1-\alpha)\boldsymbol{\beta}_t^{k-1} \\ \alpha \in [0,1]}} \Big\{ \widetilde{\mathcal{L}}_{\lambda_t}(\boldsymbol{\beta}) + \frac{L_t^k}{2} \|\boldsymbol{\beta} - \boldsymbol{\beta}_t^{k-1}\|_2^2 + \lambda_t \|\boldsymbol{\beta}\|_1 \Big\}. \tag{D.112}$$

Since $\|(\beta_t^{k-1})_{\overline{S^*}}\|_0 \le \widetilde{s}$ and $\|(\widehat{\beta}_{\lambda_t})_{\overline{S^*}}\|_0 \le \widetilde{s}$ imply $\|(\widehat{\beta}_{\lambda_t} - \beta_t^{k-1})_{\overline{S^*}}\|_0 \le 2\widetilde{s}$, Lemma 5.1 in Wang et al. (2014a) implies that the strong convexity of $\widetilde{\mathcal{L}}_{\lambda_t}(\beta)$ holds for $\widehat{\beta}_{\lambda_t}$ and $\beta_t^{k-1}$. Hence we have

$$\begin{aligned}\widetilde{\mathcal{L}}_{\lambda_t}(\beta) &= \widetilde{\mathcal{L}}_{\lambda_t}\big(\alpha\widehat{\beta}_{\lambda_t} + (1-\alpha)\beta^{k-1}\big)\\ &\le \alpha\widetilde{\mathcal{L}}_{\lambda_t}\big(\widehat{\beta}_{\lambda_t}\big) + (1-\alpha)\widetilde{\mathcal{L}}_{\lambda_t}\big(\beta^{k-1}\big).\end{aligned} \tag{D.113}$$

Meanwhile, by the convexity of $\ell_1$ norm we have

$$\begin{aligned}\lambda_t\|\beta\|_1 &= \lambda_t\big\|\alpha\widehat{\beta}_{\lambda_t} + (1-\alpha)\beta^{k-1}\big\|_1\\ &\le \alpha\lambda_t\big\|\widehat{\beta}_{\lambda_t}\big\|_1 + (1-\alpha)\big\|\beta^{k-1}\big\|_1.\end{aligned} \tag{D.114}$$

Plugging (D.113) and (D.114) into the right-hand side of (D.112), we obtain

$$\begin{aligned}\phi_{\lambda_t}\big(\beta_t^k\big) &\le \min_{\alpha\in[0,1]}\Big\{\alpha\Big(\widetilde{\mathcal{L}}_{\lambda_t}\big(\widehat{\beta}_{\lambda_t}\big) + \lambda_t\big\|\widehat{\beta}_{\lambda_t}\big\|_1\Big)\\ &\qquad + (1-\alpha)\Big(\widetilde{\mathcal{L}}_{\lambda_t}\big(\beta_t^{k-1}\big) + \lambda_t\big\|\beta_t^{k-1}\big\|_1\Big)\\ &\qquad + \frac{L_t^k}{2}\big\|\alpha\widehat{\beta}_{\lambda_t} + (1-\alpha)\beta_t^{k-1} - \beta_t^{k-1}\big\|_2^2\Big\}\\ &= \min_{\alpha\in[0,1]}\Big\{\alpha\phi_{\lambda_t}\big(\widehat{\beta}_{\lambda_t}\big) + (1-\alpha)\phi_{\lambda_t}\big(\beta_t^{k-1}\big)\\ &\qquad + \frac{L_t^k}{2}\big\|\alpha\widehat{\beta}_{\lambda_t} + (1-\alpha)\beta_t^{k-1} - \beta_t^{k-1}\big\|_2^2\Big\}\\ &\le \min_{\alpha\in[0,1]}\Big\{\phi_{\lambda_t}\big(\beta_t^{k-1}\big) - \alpha\Big(\phi_{\lambda_t}\big(\beta_t^{k-1}\big) - \phi_{\lambda_t}\big(\widehat{\beta}_{\lambda_t}\big)\Big)\\ &\qquad + \frac{\alpha^2 L_t^k}{2}\underbrace{\big\|\beta_t^{k-1} - \widehat{\beta}_{\lambda_t}\big\|_2^2}_{\text{(i)}}\Big\}.\end{aligned} \tag{D.115}$$

For term (i), similar to (D.109), applying the exact optimality condition of $\widehat{\beta}_{\lambda_t}$ and the restricted strong convexity of $\widetilde{\mathcal{L}}_{\lambda_t}(\beta)$, we obtain

$$\phi_{\lambda_t}\big(\beta_t^{k-1}\big) - \phi_{\lambda_t}\big(\widehat{\beta}_{\lambda_t}\big) \ge \frac{\rho_- - \zeta_-}{2}\big\|\beta_t^{k-1} - \widehat{\beta}_{\lambda_t}\big\|_2^2.$$

Plugging this into the right-hand side of (D.115), we obtain

$$\begin{aligned}\phi_{\lambda_t}\big(\beta_t^k\big) &\le \min_{\alpha\in[0,1]}\Big\{\phi_{\lambda_t}\big(\beta_t^{k-1}\big) - \alpha\Big(\phi_{\lambda_t}\big(\beta_t^{k-1}\big) - \phi_{\lambda_t}\big(\widehat{\beta}_{\lambda_t}\big)\Big)\\ &\qquad + \frac{\alpha^2 L_t^k}{2}\cdot\frac{2}{\rho_- - \zeta_-}\Big(\phi_{\lambda_t}\big(\beta_t^{k-1}\big) - \phi_{\lambda_t}\big(\widehat{\beta}_{\lambda_t}\big)\Big)\Big\}.\end{aligned} \tag{D.116}$$

The right-hand side of (D.116) attains its minimum if $\alpha = (\rho_- - \zeta_-)/\big(2L_t^k\big)$.

Plugging this value of $\alpha$ into (D.116), we obtain

$$\phi_{\lambda_t}\big(\boldsymbol{\beta}_t^k\big) \le \phi_{\lambda_t}\big(\boldsymbol{\beta}_t^{k-1}\big) - \frac{\rho_- - \zeta_-}{4L_t^k}\Big(\phi_{\lambda_t}\big(\boldsymbol{\beta}_t^{k-1}\big) - \phi_{\lambda_t}\big(\widehat{\boldsymbol{\beta}}_{\lambda_t}\big)\Big),$$

which implies

$$\begin{aligned} &\phi_{\lambda_t}\big(\boldsymbol{\beta}_t^k\big) - \phi_{\lambda_t}\big(\widehat{\boldsymbol{\beta}}_{\lambda_t}\big) \\ &\quad\le \Big(\phi_{\lambda_t}\big(\boldsymbol{\beta}_t^{k-1}\big) - \phi_{\lambda_t}\big(\widehat{\boldsymbol{\beta}}_{\lambda_t}\big)\Big) - \frac{\rho_- - \zeta_-}{4L_t^k}\Big(\phi_{\lambda_t}\big(\boldsymbol{\beta}_t^{k-1}\big) - \phi_{\lambda_t}\big(\widehat{\boldsymbol{\beta}}_{\lambda_t}\big)\Big) \\ &\quad= \bigg(1 - \frac{\rho_- - \zeta_-}{4L_t^k}\bigg)\Big(\phi_{\lambda_t}\big(\boldsymbol{\beta}_t^{k-1}\big) - \phi_{\lambda_t}\big(\widehat{\boldsymbol{\beta}}_{\lambda_t}\big)\Big). \end{aligned} \tag{D.117}$$

Recall that in (D.103) we have $L_t^k \le 2(\rho_+ - \zeta_+)$ $(k = 0, 1, \ldots)$. Plugging in this into the right-hand side of (D.117), we obtain

$$\begin{aligned} \phi_{\lambda_t}\big(\boldsymbol{\beta}_t^k\big) - \phi_{\lambda_t}\big(\widehat{\boldsymbol{\beta}}_{\lambda_t}\big) &\le \bigg(1 - \frac{1}{8}\cdot\underbrace{\frac{\rho_- - \zeta_-}{\rho_+ - \zeta_+}}_{1/\kappa}\bigg)\Big(\phi_{\lambda_t}\big(\boldsymbol{\beta}_t^{k-1}\big) - \phi_{\lambda_t}\big(\widehat{\boldsymbol{\beta}}_{\lambda_t}\big)\Big) \\ &= \Big(1 - \frac{1}{8\kappa}\Big)^2\Big(\phi_{\lambda_t}\big(\boldsymbol{\beta}_t^{k-2}\big) - \phi_{\lambda_t}\big(\widehat{\boldsymbol{\beta}}_{\lambda_t}\big)\Big) \\ &\;\;\vdots \\ &= \Big(1 - \frac{1}{8\kappa}\Big)^k\Big(\phi_{\lambda_t}\big(\boldsymbol{\beta}_t^{0}\big) - \phi_{\lambda_t}\big(\widehat{\boldsymbol{\beta}}_{\lambda_t}\big)\Big), \end{aligned} \tag{D.118}$$

where $\kappa$ is the condition number defined in (4.5) of Wang et al. (2014a). Now we can characterize the total number of proximal-gradient steps required to obtain an approximate solution $\widetilde{\boldsymbol{\beta}}_t = \boldsymbol{\beta}_t^{k+1}$ that satisfies

$$\omega_{\lambda_t}\big(\widetilde{\boldsymbol{\beta}}_t\big) \le \lambda_t/4 \quad (t = 1, \ldots, N-1), \quad \text{ or } \quad \omega_{\lambda_t}\big(\widetilde{\boldsymbol{\beta}}\big) \le \epsilon_{\text{opt}} \quad (t = N). \tag{D.119}$$

From Lemma D.2, we have

$$\begin{aligned} \omega_{\lambda_t}\big(\boldsymbol{\beta}_t^{k+1}\big) &\le \Big(L_t^{k+1} + (\rho_+ - \zeta_+)\Big)\big\|\boldsymbol{\beta}_t^{k+1} - \boldsymbol{\beta}_t^k\big\|_2 \\ &= L_t^{k+1}\left(1 + \frac{\rho_+ - \zeta_+}{L_t^{k+1}}\right)\big\|\boldsymbol{\beta}_t^{k+1} - \boldsymbol{\beta}_t^k\big\|_2. \end{aligned} \tag{D.120}$$

Note that the stopping criterion of the line-search method (Line 7 of Algorithm 2 in in Wang et al. (2014a)) implies $L_t^{k+1} \ge \rho_- - \zeta_-$. Otherwise, we assume that $L_t^{k+1} < \rho_- - \zeta_-$. Since $\big\|\big(\boldsymbol{\beta}_t^{k+1}\big)_{\overline{S^*}}\big\|_0 \le \widetilde{s}$ and $\big\|\big(\boldsymbol{\beta}_t^{k}\big)_{\overline{S^*}}\big\|_0 \le \widetilde{s}$

imply $\|(\boldsymbol{\beta}_t^k - \boldsymbol{\beta}_t^{k+1})_{\overline{S^*}}\|_0 \le 2\widetilde{s}$, by Lemma 5.1 in Wang et al. (2014a) we have

$$\begin{aligned}
&\psi_{L_t^{k+1},\lambda_t}(\boldsymbol{\beta}_t^{k+1};\boldsymbol{\beta}_t^k)\\
&\quad= \widetilde{\mathcal{L}}_{\lambda_t}(\boldsymbol{\beta}_t^k) + \nabla\widetilde{\mathcal{L}}_{\lambda_t}(\boldsymbol{\beta}_t^k)^T(\boldsymbol{\beta}_t^{k+1}-\boldsymbol{\beta}_t^k) + \frac{L^{k+1}}{2}\|\boldsymbol{\beta}_t^{k+1}-\boldsymbol{\beta}_t^k\|_2^2\\
&\qquad\qquad + \lambda_t\|\boldsymbol{\beta}_t^{k+1}\|_1\\
&\quad< \widetilde{\mathcal{L}}_{\lambda_t}(\boldsymbol{\beta}_t^k) + \nabla\widetilde{\mathcal{L}}_{\lambda_t}(\boldsymbol{\beta}_t^k)^T(\boldsymbol{\beta}_t^{k+1}-\boldsymbol{\beta}_t^k) + \frac{\rho_- - \zeta_-}{2}\|\boldsymbol{\beta}_t^{k+1}-\boldsymbol{\beta}_t^k\|_2^2\\
&\qquad\qquad + \lambda_t\|\boldsymbol{\beta}_t^{k+1}\|_1\\
&\quad\le \widetilde{\mathcal{L}}_{\lambda_t}(\boldsymbol{\beta}_t^{k+1}) + \lambda_t\|\boldsymbol{\beta}_t^{k+1}\|_1\\
&\quad= \phi_{\lambda_t}(\boldsymbol{\beta}_t^{k+1}).
\end{aligned}$$

Here the first equality is from the definition in (3.7) of Wang et al. (2014a), the first inequality is from our assumption that $L_t^{k+1} < \rho_- - \zeta_-$, the second inequality follows from the restricted strong convexity by Lemma 5.1 in Wang et al. (2014a). However, this contradicts the stopping criterion $\phi_{\lambda_t}(\boldsymbol{\beta}_t^{k+1}) \le \psi_{L_t^{k+1},\lambda_t}(\boldsymbol{\beta}_t^{k+1};\boldsymbol{\beta}_t^k)$. Therefore we have proved $L_t^{k+1} \ge \rho_- - \zeta_-$. From (D.120) we have

$$\begin{aligned}
\omega_{\lambda_t}(\boldsymbol{\beta}_t^{k+1}) &\le L_t^{k+1}\left(1+\frac{\rho_+ - \zeta_+}{\rho_- - \zeta_-}\right)\|\boldsymbol{\beta}_t^{k+1}-\boldsymbol{\beta}_t^k\|_2\\
&= L_t^{k+1}(1+\kappa)\|\boldsymbol{\beta}_t^{k+1}-\boldsymbol{\beta}_t^k\|_2.
\end{aligned} \tag{D.121}$$

Moreover, by Lemma D.1 we have

$$\frac{L_t^{k+1}}{2}\|\boldsymbol{\beta}_t^{k+1}-\boldsymbol{\beta}_t^k\|_2^2 \le \phi_\lambda(\boldsymbol{\beta}_t^k) - \phi_\lambda(\boldsymbol{\beta}_t^{k+1}).$$

Plugging this into the right-hand side of (D.121), we obtain

$$\begin{aligned}
\omega_{\lambda_t}(\boldsymbol{\beta}_t^{k+1}) &\le (1+\kappa)L_t^{k+1}\|\boldsymbol{\beta}_t^{k+1}-\boldsymbol{\beta}_t^k\|_2\\
&\le (1+\kappa)\sqrt{2L_t^{k+1}\Big(\phi_{\lambda_t}(\boldsymbol{\beta}_t^k) - \phi_{\lambda_t}(\boldsymbol{\beta}_t^{k+1})\Big)}.
\end{aligned}$$

According to (D.101), the sequence $\{\phi_{\lambda_t}(\boldsymbol{\beta}_t^k)\}_{k=0}^\infty$ decreases monotonically. Therefore, we have $\phi_{\lambda_t}(\boldsymbol{\beta}_t^{k+1}) \ge \phi_{\lambda_t}(\widehat{\boldsymbol{\beta}}_{\lambda_t})$, which implies

$$\omega_{\lambda_t}(\boldsymbol{\beta}_t^{k+1}) \le (1+\kappa)\sqrt{2L_t^{k+1}\Big(\phi_{\lambda_t}(\boldsymbol{\beta}_t^k) - \phi_{\lambda_t}(\widehat{\boldsymbol{\beta}}_{\lambda_t})\Big)}. \tag{D.122}$$

Now we provide an upper bound of the right-hand side of (D.122). Recall that in (D.103) we have $L_t^k \le 2(\rho_+ - \zeta_+)$ $(k = 0, 1, \ldots)$, and in (D.118) we have $\phi_{\lambda_t}(\boldsymbol{\beta}_t^k) - \phi_{\lambda_t}(\widehat{\boldsymbol{\beta}}_{\lambda_t}) \le (1 - 1/(8\kappa))^k\Big(\phi_{\lambda_t}(\boldsymbol{\beta}_t^0) - \phi_{\lambda_t}(\widehat{\boldsymbol{\beta}}_{\lambda_t})\Big)$. Note that we assume $\|(\boldsymbol{\beta}_t^0)_{\overline{S^*}}\|_0 \le \widetilde{s}$ and $\omega_{\lambda_t}(\boldsymbol{\beta}_t^0) \le \lambda_t/2$. In Lemma D.3, we set

$\lambda' = \lambda = \lambda_t$ and $\epsilon = \lambda_t/2$, then we have

$$\phi_{\lambda_t}(\boldsymbol{\beta}_t^0) - \phi_{\lambda_t}(\widehat{\boldsymbol{\beta}}_{\lambda_t}) \le \frac{21}{\rho_- - \zeta_-}{\lambda_t}^2 s^*.$$

Plugging these into the right-hand side of (D.122), we obtain

$$\begin{aligned}\omega_{\lambda_t}(\boldsymbol{\beta}_t^{k+1}) &\le (1+\kappa)\sqrt{4(\rho_+ - \zeta_+)\cdot\Big(1-\frac{1}{8\kappa}\Big)^k\frac{21}{\rho_- - \zeta_-}{\lambda_t}^2 s^*}\\ &= (1+\kappa)\sqrt{84\kappa\Big(1-\frac{1}{8\kappa}\Big)^k}\cdot\lambda_t\sqrt{s^*}.\end{aligned}$$

Therefore, for $t = 1,\ldots,N-1$, to ensure that $\boldsymbol{\beta}_t^{k+1}$ satisfies $\omega_{\lambda_t}(\boldsymbol{\beta}_t^{k+1}) \le \lambda_t/4$, it suffices to make $k$ satisfy

$$(1+\kappa)\sqrt{84\kappa\Big(1-\frac{1}{8\kappa}\Big)^k}\cdot\lambda_t\sqrt{s^*} \le \lambda_t/4,$$

which implies

$$k \ge 2\log\big(8\sqrt{21}\cdot\sqrt{\kappa}(1+\kappa)\cdot\sqrt{s^*}\big)\Big/\log\Big(1-\frac{1}{8\kappa}\Big).$$

Similarly, for $t = N$, to ensure that $\boldsymbol{\beta}_t^{k+1}$ satisfies $\omega_{\lambda_t}(\boldsymbol{\beta}^{k+1}) \le \epsilon_{\text{opt}}$, $k$ should satisfy

$$k \ge 2\log\big(2\sqrt{21}\cdot\sqrt{\kappa}(1+\kappa)\cdot\sqrt{s^*}\lambda_t/\epsilon_{\text{opt}}\big)\Big/\log\Big(1-\frac{1}{8\kappa}\Big).$$

Therefore we conclude the proof of Theorem 5.5. □

**D.7. Proof of Theorem 4.5 in Wang et al. (2014a).** First we present a useful lemma. It ensures that the approximate solution $\widetilde{\boldsymbol{\beta}}_{t-1}$, which is obtained from the $(t-1)$-th path following stage, is $(\lambda_t/2)$-suboptimal with respect to regularization parameter $\lambda_t$, i.e., $\omega_{\lambda_t}(\widetilde{\boldsymbol{\beta}}_{t-1}) \le \lambda_t/2$.

**Lemma D.4.** Let $\widetilde{\boldsymbol{\beta}}_{t-1}$ $(t = 1,\ldots,N)$ be the approximate solution obtained from the $(t-1)$-th path following stage (Line 8 of Algorithm 1 in Wang et al. (2014a)). If $\omega_{\lambda_{t-1}}(\widetilde{\boldsymbol{\beta}}_{t-1}) \le \lambda_{t-1}/4$. Under Assumption 4.1 and Assumption 4.4 in Wang et al. (2014a), we have

$$\omega_{\lambda_t}(\widetilde{\boldsymbol{\beta}}_{t-1}) \le \lambda_t/2,$$

where $\lambda_t = \eta\lambda_{t-1}$ with $\eta \in [0.9, 1)$.

Proof. Consider the regularization parameter $\lambda_{t-1}$. Let $\boldsymbol{\xi} \in \partial\big\|\widetilde{\boldsymbol{\beta}}_{t-1}\big\|_1$

be the subgradient that attains the minimum in

(D.123)
$$\omega_{\lambda_{t-1}}(\widetilde{\boldsymbol{\beta}}_{t-1}) = \min_{\boldsymbol{\xi}' \in \partial \|\widetilde{\boldsymbol{\beta}}_{t-1}\|_1} \max_{\boldsymbol{\beta}' \in \Omega} \left\{ \frac{(\widetilde{\boldsymbol{\beta}}_{t-1} - \boldsymbol{\beta}')^T}{\|\widetilde{\boldsymbol{\beta}}_{t-1} - \boldsymbol{\beta}'\|_1} \Big( \nabla \widetilde{\mathcal{L}}_{\lambda_{t-1}}(\widetilde{\boldsymbol{\beta}}_{t-1}) + \lambda_{t-1} \boldsymbol{\xi}' \Big) \right\},$$

which implies

$$\omega_{\lambda_{t-1}}(\widetilde{\boldsymbol{\beta}}_{t-1}) = \max_{\boldsymbol{\beta}' \in \Omega} \left\{ \frac{(\widetilde{\boldsymbol{\beta}}_{t-1} - \boldsymbol{\beta}')^T}{\|\widetilde{\boldsymbol{\beta}}_{t-1} - \boldsymbol{\beta}'\|_1} \Big( \nabla \widetilde{\mathcal{L}}_{\lambda_{t-1}}(\widetilde{\boldsymbol{\beta}}_{t-1}) + \lambda_{t-1} \boldsymbol{\xi} \Big) \right\}. \tag{D.124}$$

Now we consider regularization parameter $\lambda_t$. We have

$$\begin{aligned} \omega_{\lambda_t}(\widetilde{\boldsymbol{\beta}}_{t-1}) &= \min_{\boldsymbol{\xi}' \in \partial \|\widetilde{\boldsymbol{\beta}}_{t-1}\|_1} \max_{\boldsymbol{\beta}' \in \Omega} \left\{ \frac{(\widetilde{\boldsymbol{\beta}}_{t-1} - \boldsymbol{\beta}')^T}{\|\widetilde{\boldsymbol{\beta}}_{t-1} - \boldsymbol{\beta}'\|_1} \Big( \nabla \widetilde{\mathcal{L}}_{\lambda_t}(\widetilde{\boldsymbol{\beta}}_{t-1}) + \lambda_t \boldsymbol{\xi}' \Big) \right\} \\ &\le \max_{\boldsymbol{\beta}' \in \Omega} \left\{ \frac{(\widetilde{\boldsymbol{\beta}}_{t-1} - \boldsymbol{\beta}')^T}{\|\widetilde{\boldsymbol{\beta}}_{t-1} - \boldsymbol{\beta}'\|_1} \Big( \nabla \widetilde{\mathcal{L}}_{\lambda_t}(\widetilde{\boldsymbol{\beta}}_{t-1}) + \lambda_t \boldsymbol{\xi} \Big) \right\}, \end{aligned} \tag{D.125}$$

where $\boldsymbol{\xi}$ is defined as the minimizer of (D.123). Recall that $\nabla \widetilde{\mathcal{L}}_{\lambda_t}(\widetilde{\boldsymbol{\beta}}_{t-1}) = \nabla \mathcal{L}(\widetilde{\boldsymbol{\beta}}_{t-1}) + \nabla \mathcal{Q}_{\lambda_t}(\widetilde{\boldsymbol{\beta}}_{t-1})$. We have

$$\begin{aligned} \nabla \widetilde{\mathcal{L}}_{\lambda_t}(\widetilde{\boldsymbol{\beta}}_{t-1}) + \lambda_t \boldsymbol{\xi} = {} & \Big( \nabla \mathcal{L}(\widetilde{\boldsymbol{\beta}}_{t-1}) + \nabla \mathcal{Q}_{\lambda_{t-1}}(\widetilde{\boldsymbol{\beta}}_{t-1}) + \lambda_t \boldsymbol{\xi} \Big) + (\lambda_{t-1} \boldsymbol{\xi} - \lambda_t \boldsymbol{\xi}) \\ & + \Big( \nabla \mathcal{Q}_{\lambda_t}(\widetilde{\boldsymbol{\beta}}_{t-1}) - \nabla \mathcal{Q}_{\lambda_{t-1}}(\widetilde{\boldsymbol{\beta}}_{t-1}) \Big). \end{aligned}$$

Plugging this into the right-hand side of (D.125), we obtain

(D.126)
$$\begin{aligned} \omega_{\lambda_t}(\widetilde{\boldsymbol{\beta}}_{t-1}) \le {} & \underbrace{\max_{\boldsymbol{\beta}' \in \Omega} \left\{ \frac{(\widetilde{\boldsymbol{\beta}}_{t-1} - \boldsymbol{\beta}')^T}{\|\widetilde{\boldsymbol{\beta}}_{t-1} - \boldsymbol{\beta}'\|_1} \Big( \nabla \widetilde{\mathcal{L}}_{\lambda_{t-1}}(\widetilde{\boldsymbol{\beta}}_{t-1}) + \lambda_{t-1} \boldsymbol{\xi} \Big) \right\}}_{\text{(i)}} \\ & + \underbrace{\max_{\boldsymbol{\beta}' \in \Omega} \left\{ \frac{(\widetilde{\boldsymbol{\beta}}_{t-1} - \boldsymbol{\beta}')^T}{\|\widetilde{\boldsymbol{\beta}}_{t-1} - \boldsymbol{\beta}'\|_1} (\lambda_{t-1} \boldsymbol{\xi} - \lambda_t \boldsymbol{\xi}) \right\}}_{\text{(ii)}} \\ & + \underbrace{\max_{\boldsymbol{\beta}' \in \Omega} \left\{ \frac{(\widetilde{\boldsymbol{\beta}}_{t-1} - \boldsymbol{\beta}')^T}{\|\widetilde{\boldsymbol{\beta}}_{t-1} - \boldsymbol{\beta}'\|_1} \Big( \nabla \mathcal{Q}_{\lambda_t}(\widetilde{\boldsymbol{\beta}}_{t-1}) - \nabla \mathcal{Q}_{\lambda_{t-1}}(\widetilde{\boldsymbol{\beta}}_{t-1}) \Big) \right\}}_{\text{(iii)}}. \end{aligned}$$

According to (D.124), term (i) in (D.126) is equal to $\omega_{\lambda_{t-1}}(\widetilde{\boldsymbol{\beta}}_{t-1})$, which is upper bounded by $\lambda_{t-1}/4$ by our assumption. For term (ii) in (D.126), we

have

$$\max_{\boldsymbol{\beta}'\in\Omega}\left\{\frac{\big(\widetilde{\boldsymbol{\beta}}_{t-1}-\boldsymbol{\beta}'\big)^T}{\big\|\widetilde{\boldsymbol{\beta}}_{t-1}-\boldsymbol{\beta}'\big\|_1}\big(\lambda_{t-1}\boldsymbol{\xi}-\lambda_t\boldsymbol{\xi}\big)\right\}\le\max_{\boldsymbol{\beta}'\in\mathbb{R}^d}\left\{\frac{\big(\widetilde{\boldsymbol{\beta}}_{t-1}-\boldsymbol{\beta}'\big)^T}{\big\|\widetilde{\boldsymbol{\beta}}_{t-1}-\boldsymbol{\beta}'\big\|_1}\big(\lambda_{t-1}\boldsymbol{\xi}-\lambda_t\boldsymbol{\xi}\big)\right\}$$
$$=\big\|\lambda_{t-1}\boldsymbol{\xi}-\lambda_t\boldsymbol{\xi}\big\|_\infty$$
$$\le\lambda_{t-1}-\lambda_t,$$

where first inequality is due to the duality between $\ell_1$ and $\ell_\infty$ norm, while the second inequality is due to the fact that $\lambda_{t-1}>\lambda_t$ and $\|\boldsymbol{\xi}\|_\infty\le 1$, which follows from $\boldsymbol{\xi}\in\partial\big\|\widetilde{\boldsymbol{\beta}}_{t-1}\big\|_1$. Similarly, for term (iii) we have

$$\max_{\boldsymbol{\beta}'\in\Omega}\left\{\frac{\big(\widetilde{\boldsymbol{\beta}}_{t-1}-\boldsymbol{\beta}'\big)^T}{\big\|\widetilde{\boldsymbol{\beta}}_{t-1}-\boldsymbol{\beta}'\big\|_1}\Big(\nabla\mathcal{Q}_{\lambda_t}\big(\widetilde{\boldsymbol{\beta}}_{t-1}\big)-\nabla\mathcal{Q}_{\lambda_{t-1}}\big(\widetilde{\boldsymbol{\beta}}_{t-1}\big)\Big)\right\}$$
$$\le\big\|\nabla\mathcal{Q}_{\lambda_t}\big(\widetilde{\boldsymbol{\beta}}_{t-1}\big)-\nabla\mathcal{Q}_{\lambda_{t-1}}\big(\widetilde{\boldsymbol{\beta}}_{t-1}\big)\big\|_\infty$$
$$=\max_{1\le j\le d}\Big|q'_{\lambda_t}\big((\widetilde{\beta}_{t-1})_j\big)-q'_{\lambda_{t-1}}\big((\widetilde{\beta}_{t-1})_j\big)\Big|$$
$$\le\lambda_{t-1}-\lambda_t,$$

where the second inequality follows from regularity condition (e) in Wang et al. (2014a). Hence, from (D.126) we obtain

$$\omega_{\lambda_t}\big(\widetilde{\boldsymbol{\beta}}_{t-1}\big)\le\overbrace{\lambda_{t-1}/4}^{\text{(i) in (D.126)}}+\overbrace{\lambda_{t-1}-\lambda_t}^{\text{(ii) in (D.126)}}+\overbrace{\lambda_{t-1}-\lambda_t}^{\text{(iii) in(D.126)}}$$
$$\le\big(1/(4\eta)+1/\eta-1+1/\eta-1\big)\lambda_t\le\lambda_t/2,$$

where the last inequality is obtained by plugging in $\eta\in[0.9,1)$. Hence we conclude the proof. □

Now we are ready to prove Theorem 4.5 in Wang et al. (2014a).

PROOF. **Geometric Rate of Convergence within Each Stage:** The stopping criterion of Algorithm 3 (Line 9) in Wang et al. (2014a) implies

$$\omega_{\lambda_{t-1}}\big(\widetilde{\boldsymbol{\beta}}_{t-1}\big)\le\lambda_{t-1}/4,\quad\text{for}\ \ t=1,\ldots,N.$$

By Lemma D.4 we have

$$\omega_{\lambda_t}\big(\widetilde{\boldsymbol{\beta}}_{t-1}\big)\le\lambda_t/2,\quad\text{for}\ \ t=1,\ldots,N. \tag{D.127}$$

Recall we initialize the $t$-th stage with $\widetilde{\boldsymbol{\beta}}_{t-1}=\boldsymbol{\beta}_t^0$ and $L_{t-1}=L_t^0$ (Line 8 of Algorithm 1). By Theorem 5.5 in Wang et al. (2014a), as long as $\big\|\big(\widetilde{\boldsymbol{\beta}}_{t-1}\big)_{\overline{S^*}}\big\|_0\le\widetilde{s}$ and $L_{(t-1)}\le 2(\rho_+-\zeta_+)$, we have

$$\big\|\big(\boldsymbol{\beta}_t^k\big)_{\overline{S^*}}\big\|_0\le\widetilde{s},\quad L_t^k\le 2(\rho_+-\zeta_+),\quad\text{for}\ \ k=1,2,\ldots,$$

which implies $\big\|(\widetilde{\boldsymbol{\beta}}_t)_{\overline{S^*}}\big\|_0 \le \widetilde{s}$ and $L_t \le 2(\rho_+ - \zeta_+)$. Remind that we initialize the entire path following procedure with $\widetilde{\boldsymbol{\beta}}_0 = \mathbf{0}$ and $L_0 = L_{\min} \le 2(\rho_+ - \zeta_+)$ (Line 4 of Algorithm 1 in Wang et al. (2014a)). By induction we obtain

$$\big\|(\widetilde{\boldsymbol{\beta}}_t)_{\overline{S^*}}\big\|_0 \le \widetilde{s}, \quad L_t \le 2(\rho_+ - \zeta_+), \quad \text{for} \;\; t = 1, \ldots, N.$$

By setting $\lambda = \lambda_t$ and $\widetilde{\boldsymbol{\beta}} = \widetilde{\boldsymbol{\beta}}_t$ $(t = 1, \ldots, N)$ in Theorem 5.5 of Wang et al. (2014a), we obtain that, within the $t$-th stage $(t = 1, \ldots, N-1)$, the total number of proximal-gradient iterations is no more than

$$2\log\big(8\sqrt{21}\cdot\sqrt{\kappa}(1+\kappa)\cdot\sqrt{s^*}\big)\Big/\log\left(\frac{1}{1-1/(8\kappa)}\right),$$

while within the $N$-th stage, the total number of proximal-gradient steps is no more that

$$2\log\big(2\sqrt{21}\cdot\sqrt{\kappa}(1+\kappa)\cdot\sqrt{s^*}\lambda_{\mathrm{tgt}}/\epsilon_{\mathrm{opt}}\big)\Big/\log\left(\frac{1}{1-1/(8\kappa)}\right).$$

Hence we obtain the first conclusion.

**Geometric Rate of Convergence over the Full Path:** Now we prove the second statement about the total number of proximal-gradient steps along the entire solution path. The total number of path following stages is

$$N = \log(\lambda_{\mathrm{tgt}}/\lambda_0)/\log\eta.$$

Together with the first result, we have that the total number of proximal-gradient steps is no more than

$$(N-1)C'\log\big(4C\sqrt{s^*}\big) + C'\log\big(C\sqrt{s^*}\lambda_{\mathrm{tgt}}/\epsilon_{\mathrm{opt}}\big).$$

where

$$C = 2\sqrt{21}\cdot\sqrt{\kappa}(1+\kappa), \quad C' = 2\Big/\log\left(\frac{1}{1-1/(8\kappa)}\right).$$

**Geometric Rate of Convergence of the Objective Function Values:** Now we prove the third statement concerning the objective function value. For $t = 1, \ldots, N-1$, by (D.127) we have $\omega_{\lambda_{t+1}}\big(\widetilde{\boldsymbol{\beta}}_t\big) \le \lambda_{t+1}/2$. Setting $\lambda' = \lambda_{\mathrm{tgt}}$, $\lambda = \lambda_{t+1}$, $\boldsymbol{\beta} = \widetilde{\boldsymbol{\beta}}_t$ and $\epsilon = \lambda_{t+1}/2$ in Lemma D.3, we obtain

$$\phi_{\lambda_{\mathrm{tgt}}}\big(\widetilde{\boldsymbol{\beta}}_t\big) - \phi_{\lambda_{\mathrm{tgt}}}\big(\widehat{\boldsymbol{\beta}}_{\lambda_{\mathrm{tgt}}}\big) \le \frac{21}{\rho_- - \zeta_-}\big(\lambda_{t+1}/2 + 2(\lambda_{t+1} - \lambda_{\mathrm{tgt}})\big)\cdot(\lambda_{\mathrm{tgt}} + \lambda_{t+1})s^*.$$

Since $\lambda_{\mathrm{tgt}} \le \lambda_{t+1}$, we have

$$\phi_{\lambda_{\mathrm{tgt}}}\big(\widetilde{\boldsymbol{\beta}}_t\big) - \phi_{\lambda_{\mathrm{tgt}}}\big(\widehat{\boldsymbol{\beta}}_{\lambda_{\mathrm{tgt}}}\big) \le \frac{21}{\rho_- - \zeta_-}\big(\lambda_{t+1}/2 + 2\lambda_{t+1}\big)\cdot 2\lambda_{t+1}s^* = \frac{105\cdot\lambda_{t+1}^2 s^*}{\rho_- - \zeta_-}.$$

Since $\lambda_{t+1} = \eta^{t+1}\lambda_0$, we obtain

$$\phi_{\lambda_{\rm tgt}}\big(\widetilde{\boldsymbol{\beta}}_t\big) - \phi_{\lambda_{\rm tgt}}\big(\widehat{\boldsymbol{\beta}}_{\lambda_{\rm tgt}}\big) \le \eta^{2(t+1)}\frac{105\cdot\lambda_0^2 s^*}{\rho_- - \zeta_-}, \quad \text{for } t = 1,\ldots,N-1.$$

Similarly, for $t = N$, we have $\omega_{\lambda_{\rm tgt}}\big(\widetilde{\boldsymbol{\beta}}_N\big) \le \epsilon_{\rm opt}$. By setting $\lambda = \lambda' = \lambda_{\rm tgt}$ and $\epsilon = \epsilon_{\rm opt}$ in Lemma D.3, we have

$$\phi_{\lambda_{\rm tgt}}\big(\widetilde{\boldsymbol{\beta}}_t\big) - \phi_{\lambda_{\rm tgt}}\big(\widehat{\boldsymbol{\beta}}_{\lambda_{\rm tgt}}\big) \le \frac{21\cdot\lambda_{\rm tgt}s^*}{\rho_- - \zeta_-}\epsilon_{\rm opt}.$$

Therefore we conclude the proof of Theorem 4.5 in Wang et al. (2014a). □

### D.8. Proof of Theorem 4.7 in Wang et al. (2014a).

Proof. Recall $\widetilde{\boldsymbol{\beta}}_t$ is the approximate local solution obtained from the $t$-th path following stage (Lines 8 and 12 of Algorithm 1 in Wang et al. (2014a)). Therefore, it satisfies the stopping criterion of the proximal-gradient method (Line 9 of Algorithm 3 in Wang et al. (2014a)), i.e., for $t = 1,\ldots,N-1$ we have $\omega_{\lambda_t}\big(\widetilde{\boldsymbol{\beta}}_t\big) \le \lambda_t/4 < \lambda_t/2$, while for $t = N$ we have $\omega_{\lambda_t}\big(\widetilde{\boldsymbol{\beta}}_t\big) \le \epsilon_{\rm opt} \ll \lambda_{\rm tgt}/4 < \lambda_t/2$. Meanwhile, by (5.2) in Theorem 5.5 of Wang et al. (2014a), $\widetilde{\boldsymbol{\beta}}_t$ satisfies $\big\|\big(\widetilde{\boldsymbol{\beta}}_t\big)_{\overline{S^*}}\big\|_0 \le \widetilde{s}$. For logistic loss, we further have $\|\widetilde{\boldsymbol{\beta}}_t\|_2 \le R$ due to the $\ell_2$ constraint. Therefore Lemma 5.2 in Wang et al. (2014a) gives

$$\big\|\widetilde{\boldsymbol{\beta}}_t - \boldsymbol{\beta}^*\big\|_2 \le \frac{21/8}{\rho_- - \zeta_-}\lambda_t\sqrt{s^*}, \quad \text{for } \ t = 1,\ldots,N,$$

which concludes the proof. □

### D.9. Proof of Theorem 4.8 in Wang et al. (2014a).

Proof. We denote the subgradients by $\boldsymbol{\xi}^* \in \partial\|\boldsymbol{\beta}^*\|_1$ and $\widehat{\boldsymbol{\xi}} \in \partial\big\|\widehat{\boldsymbol{\beta}}_{\lambda_t}\big\|_1$. In particular, we set $\widehat{\boldsymbol{\xi}}$ to be the subgradient that attains the minimum in

$$\omega_{\lambda_t}\big(\widehat{\boldsymbol{\beta}}_{\lambda_t}\big) = \min_{\boldsymbol{\xi}'\in\partial\|\widehat{\boldsymbol{\beta}}_{\lambda_t}\|_1}\ \max_{\boldsymbol{\beta}'\in\Omega}\left\{\frac{\big(\widehat{\boldsymbol{\beta}}_{\lambda_t} - \boldsymbol{\beta}'\big)^T}{\big\|\widehat{\boldsymbol{\beta}}_{\lambda_t} - \boldsymbol{\beta}'\big\|_1}\Big(\nabla\widetilde{\mathcal{L}}_{\lambda_t}\big(\widehat{\boldsymbol{\beta}}_{\lambda_t}\big) + \lambda_t\boldsymbol{\xi}'\Big)\right\}.$$

Recall that $\widehat{\boldsymbol{\beta}}_{\lambda_t}$ satisfies the exact optimality condition that $\omega_{\lambda_t}\big(\widehat{\boldsymbol{\beta}}_{\lambda_t}\big) \le 0$, hence we have

$$\max_{\boldsymbol{\beta}'\in\Omega}\left\{\big(\widehat{\boldsymbol{\beta}}_{\lambda_t} - \boldsymbol{\beta}'\big)^T\Big(\nabla\widetilde{\mathcal{L}}_{\lambda_t}\big(\widehat{\boldsymbol{\beta}}_{\lambda_t}\big) + \lambda_t\widehat{\boldsymbol{\xi}}\Big)\right\} \le 0. \tag{D.128}$$

Theorem 5.5 of Wang et al. (2014a) gives $\big\|\big(\widehat{\boldsymbol{\beta}}_{\lambda_t}\big)_{\overline{S^*}}\big\|_0 \le \widetilde{s}$. Since

$$\big\|\big(\widehat{\boldsymbol{\beta}}_{\lambda_t} - \boldsymbol{\beta}^*\big)_{\overline{S^*}}\big\|_0 \le \widetilde{s},$$

according to Lemma 5.1 of Wang et al. (2014a) the restricted convexity holds for $\widetilde{\mathcal{L}}_{\lambda_t}(\boldsymbol{\beta})$ at $\boldsymbol{\beta}_t$ and $\boldsymbol{\beta}^*$, i.e.,

$$\widetilde{\mathcal{L}}_{\lambda_t}(\widehat{\boldsymbol{\beta}}_{\lambda_t}) \geq \widetilde{\mathcal{L}}_{\lambda_t}(\boldsymbol{\beta}^*) + \nabla\widetilde{\mathcal{L}}_{\lambda_t}(\boldsymbol{\beta}^*)^T(\widehat{\boldsymbol{\beta}}_{\lambda_t} - \boldsymbol{\beta}^*) + \frac{\rho_- - \zeta_-}{2}\|\widehat{\boldsymbol{\beta}}_{\lambda_t} - \boldsymbol{\beta}^*\|_2^2, \tag{D.129}$$

$$\widetilde{\mathcal{L}}_{\lambda_t}(\boldsymbol{\beta}^*) \geq \widetilde{\mathcal{L}}_{\lambda_t}(\widehat{\boldsymbol{\beta}}_{\lambda_t}) + \nabla\widetilde{\mathcal{L}}_{\lambda_t}(\widehat{\boldsymbol{\beta}}_{\lambda_t})^T(\boldsymbol{\beta}^* - \widehat{\boldsymbol{\beta}}_{\lambda_t}) + \frac{\rho_- - \zeta_-}{2}\|\boldsymbol{\beta}^* - \widehat{\boldsymbol{\beta}}_{\lambda_t}\|_2^2. \tag{D.130}$$

Meanwhile, by the convexity of $\ell_1$ norm, we have

$$\lambda_t\|\widehat{\boldsymbol{\beta}}_{\lambda_t}\|_1 \geq \lambda_t\|\boldsymbol{\beta}^*\|_1 + \lambda_t(\widehat{\boldsymbol{\beta}}_{\lambda_t} - \boldsymbol{\beta}^*)^T\boldsymbol{\xi}^*, \tag{D.131}$$

$$\lambda_t\|\boldsymbol{\beta}^*\|_1 \geq \lambda_t\|\widehat{\boldsymbol{\beta}}_{\lambda_t}\|_1 + \lambda_t(\boldsymbol{\beta}^* - \widehat{\boldsymbol{\beta}}_{\lambda_t})^T\widehat{\boldsymbol{\xi}}. \tag{D.132}$$

Recall that $\widetilde{\mathcal{L}}_\lambda(\boldsymbol{\beta}) = \mathcal{L}(\boldsymbol{\beta}) + \mathcal{Q}_\lambda(\boldsymbol{\beta})$. Adding (D.129)-(D.132), we obtain

$$\begin{aligned} 0 \geq\ & \underbrace{\Big(\nabla\mathcal{L}(\boldsymbol{\beta}^*) + \nabla\mathcal{Q}_{\lambda_t}(\boldsymbol{\beta}^*) + \lambda_t\boldsymbol{\xi}^*\Big)^T(\widehat{\boldsymbol{\beta}}_{\lambda_t} - \boldsymbol{\beta}^*)}_{\text{(i)}} \\ & + \underbrace{\Big(\nabla\widetilde{\mathcal{L}}_{\lambda_t}(\widehat{\boldsymbol{\beta}}_{\lambda_t}) + \lambda_t\widehat{\boldsymbol{\xi}}\Big)^T(\boldsymbol{\beta}^* - \widehat{\boldsymbol{\beta}}_{\lambda_t})}_{\text{(ii)}} + (\rho_- - \zeta_-)\|\widehat{\boldsymbol{\beta}}_{\lambda_t} - \boldsymbol{\beta}^*\|_2^2. \end{aligned} \tag{D.133}$$

According to (D.128) we have

$$\Big(\nabla\widetilde{\mathcal{L}}_{\lambda_t}(\widehat{\boldsymbol{\beta}}_{\lambda_t}) + \lambda_t\widehat{\boldsymbol{\xi}}\Big)^T(\widehat{\boldsymbol{\beta}}_{\lambda_t} - \boldsymbol{\beta}^*) \leq \max_{\boldsymbol{\beta}' \in \Omega}\Big\{(\widehat{\boldsymbol{\beta}}_{\lambda_t} - \boldsymbol{\beta}')^T\Big(\nabla\widetilde{\mathcal{L}}_{\lambda_t}(\widehat{\boldsymbol{\beta}}_{\lambda_t}) + \lambda\widehat{\boldsymbol{\xi}}\Big)\Big\} \leq 0,$$

which implies that term (ii) in (D.133) is nonnegative. Moving term (i) in (D.133) to its left-hand side, we obtain

$$\begin{aligned} & (\rho_- - \zeta_-)\|\widehat{\boldsymbol{\beta}}_{\lambda_t} - \boldsymbol{\beta}^*\|_2^2 \\ & \quad \leq \big(\nabla\mathcal{L}(\boldsymbol{\beta}^*) + \nabla\mathcal{Q}_{\lambda_t}(\boldsymbol{\beta}^*) + \lambda_t\boldsymbol{\xi}^*\big)^T(\widehat{\boldsymbol{\beta}}_{\lambda_t} - \boldsymbol{\beta}^*) \\ & \quad \leq \min_{\boldsymbol{\xi}^* \in \partial\|\boldsymbol{\beta}^*\|_1}\Bigg\{\sum_{j=1}^d\Big(\Big|\big(\nabla\mathcal{L}(\boldsymbol{\beta}^*) + \nabla\mathcal{Q}_{\lambda_t}(\boldsymbol{\beta}^*) + \lambda_t\boldsymbol{\xi}^*\big)_j\Big| \cdot \Big|(\boldsymbol{\beta}^* - \widehat{\boldsymbol{\beta}}_{\lambda_t})_j\Big|\Big)\Bigg\}. \end{aligned} \tag{D.134}$$

In the sequel, we decompose the summation on the right-hand side of (D.134) into three parts: $j \in \overline{S^*}$, $j \in S_1^*$ and $j \in S_2^*$, where $S_1^* = \{j : |\beta_j| \geq \nu_t\}$ and $S_2^* = \{j : |\beta_j| < \nu_t\}$. Here $\nu_t > 0$ is defined in (4.16) of Wang et al. (2014a).

- For $j \in \overline{S^*}$, by regularity condition (c) in Wang et al. (2014a), we have

$$\big(\nabla\mathcal{Q}_{\lambda_t}(\boldsymbol{\beta}^*)\big)_j = q'_{\lambda_t}(\beta_j^*) = q'_{\lambda_t}(0) = 0, \quad \text{for} \ \ j \in \overline{S^*}.$$

By (4.1) in Assumption 4.1 of Wang et al. (2014a), we have

$$\max_{j\in\overline{S^*}}\Big|\big(\nabla\mathcal{L}(\boldsymbol{\beta}^*)\big)_j\Big| \le \max_{1\le j\le d}\Big|\big(\nabla\mathcal{L}(\boldsymbol{\beta}^*)\big)_j\Big| = \|\nabla\mathcal{L}(\boldsymbol{\beta}^*)\|_\infty$$
$$\le \lambda_{\rm tgt}/8 \le \lambda_t/8 < \lambda_t.$$

Hence we have

$$\max_{j\in\overline{S^*}}\Big|\big(\nabla\mathcal{L}(\boldsymbol{\beta}^*) + \mathcal{Q}_{\lambda_t}(\boldsymbol{\beta}^*)\big)_j\Big| \le \lambda_t.$$

Meanwhile, since $\boldsymbol{\xi}^* \in \partial\|\boldsymbol{\beta}^*\|_1$, we have $\lambda_t\boldsymbol{\xi}_j^* \in [-\lambda_t, \lambda_t]$. Therefore, for any $j \in \overline{S^*}$, we can always find a $\xi_j^*$ such that

$$\Big|\big(\nabla\mathcal{L}(\boldsymbol{\beta}^*) + \nabla\mathcal{Q}_{\lambda_t}(\boldsymbol{\beta}^*)\big)_j + \lambda_t\xi_j^*\Big| = 0,$$

which implies

$$\min_{\boldsymbol{\xi}^*\in\partial\|\boldsymbol{\beta}^*\|_1}\Big\{\Big|\big(\nabla\mathcal{L}(\boldsymbol{\beta}^*) + \nabla\mathcal{Q}_{\lambda_t}(\boldsymbol{\beta}^*) + \lambda_t\boldsymbol{\xi}^*\big)_j\Big|\Big\} = 0, \quad \text{for} \ \ j \in \overline{S^*}.$$

Thus we obtain

(D.135)

$$\min_{\boldsymbol{\xi}^*\in\partial\|\boldsymbol{\beta}^*\|_1}\Big\{\sum_{j\in\overline{S^*}}\Big|\big(\nabla\mathcal{L}(\boldsymbol{\beta}^*)+\nabla\mathcal{Q}_{\lambda_t}(\boldsymbol{\beta}^*)+\lambda_t\boldsymbol{\xi}^*\big)_j\Big|\cdot\Big|\big(\boldsymbol{\beta}^*-\widehat{\boldsymbol{\beta}}_{\lambda_t}\big)_j\Big|\Big\} = 0.$$

- For $j \in S_1^* \subseteq S^*$, we have $|\beta_j^*| \ge \nu_t$. Recall that $\mathcal{P}_\lambda(\boldsymbol{\beta}) = \mathcal{Q}_\lambda(\boldsymbol{\beta}) + \lambda\|\boldsymbol{\beta}\|_1$. By our assumption on $\mathcal{P}_{\lambda_t}(\boldsymbol{\beta})$ in (4.16) of Wang et al. (2014a), we have

$$\big(\nabla\mathcal{Q}_{\lambda_t}(\boldsymbol{\beta}^*) + \lambda_t\boldsymbol{\xi}^*\big)_j = p'_{\lambda_t}(\beta_j^*) = 0, \quad \text{for} \ \ j \in S_1^*,$$

  which implies

  (D.136)

$$\min_{\boldsymbol{\xi}^*\in\partial\|\boldsymbol{\beta}^*\|_1}\Big\{\sum_{j\in S_1^*}\Big(\Big|\big(\nabla\mathcal{L}(\boldsymbol{\beta}^*) + \nabla\mathcal{Q}_{\lambda_t}(\boldsymbol{\beta}^*) + \lambda_t\boldsymbol{\xi}^*\big)_j\Big|\cdot\Big|\big(\boldsymbol{\beta}^* - \widehat{\boldsymbol{\beta}}_{\lambda_t}\big)_j\Big|\Big)\Big\}$$
$$= \sum_{j\in S_1^*}\Big|\big(\nabla\mathcal{L}(\boldsymbol{\beta}^*)\big)_j\Big|\cdot\Big|\big(\boldsymbol{\beta}^* - \widehat{\boldsymbol{\beta}}_{\lambda_t}\big)_j\Big|$$
$$\le \big\|\big(\nabla\mathcal{L}(\boldsymbol{\beta}^*)\big)_{S_1^*}\big\|_2\cdot\big\|\boldsymbol{\beta}^* - \widehat{\boldsymbol{\beta}}_{\lambda_t}\big\|_2.$$

- For $j \in S_2^* \subseteq S^*$, we have $|\beta_j^*| < \nu_t$. According to (4.1) in Assumption 4.1 of Wang et al. (2014a), we have

$$\max_{j\in S_2^*}\Big|\big(\nabla\mathcal{L}(\boldsymbol{\beta}^*)\big)_j\Big| \le \max_{1\le j\le d}\Big|\big(\nabla\mathcal{L}(\boldsymbol{\beta}^*)\big)_j\Big| = \|\nabla\mathcal{L}(\boldsymbol{\beta}^*)\|_\infty \le \lambda_t/8 \le \lambda_t/8.$$

Meanwhile we have

$$\max_{j\in S_2^*}\left|\big(\nabla \mathcal{Q}_{\lambda_t}(\boldsymbol{\beta}^*)\big)_j\right| = \max_{j\in S_2^*}\left|q'_{\lambda_t}(\beta_j^*)\right| \le \max_{1\le j\le d}\left|q'_{\lambda_t}(\beta_j^*)\right| \le \lambda_t,$$

where the last inequality follows from regularity condition (d) in Wang et al. (2014a). Also, since $\boldsymbol{\xi}^* \in \partial\|\boldsymbol{\beta}^*\|_1$, we have $|\xi_j^*| \le 1$. Therefore we obtain that, for $j \in S_2^*$,

$$\begin{aligned}&\left|\big(\nabla\mathcal{L}(\boldsymbol{\beta}^*) + \nabla\mathcal{Q}_{\lambda_t}(\boldsymbol{\beta}^*) + \lambda_t\boldsymbol{\xi}^*\big)_j\right| \\ &\quad\le \max_{j\in S_2^*}\left|\big(\nabla\mathcal{L}(\boldsymbol{\beta}^*)\big)_j\right| + \max_{j\in S_2^*}\left|\big(\nabla\mathcal{Q}_{\lambda_t}(\boldsymbol{\beta}^*)\big)_j\right| + \lambda_t \le 3\lambda_t.\end{aligned}$$

which implies

(D.137)

$$\begin{aligned}&\min_{\boldsymbol{\xi}^*\in\partial\|\boldsymbol{\beta}^*\|_1}\left\{\sum_{j\in S_2^*}\left|\big(\nabla\mathcal{L}(\boldsymbol{\beta}^*) + \nabla\mathcal{Q}_{\lambda_t}(\boldsymbol{\beta}^*) + \lambda_t\boldsymbol{\xi}^*\big)_j\right|\cdot\left|\big(\boldsymbol{\beta}^* - \widehat{\boldsymbol{\beta}}_{\lambda_t}\big)_j\right|\right\} \\ &\le 3\lambda_t\sum_{j\in S_2^*}\left|\big(\boldsymbol{\beta}^* - \widehat{\boldsymbol{\beta}}_{\lambda_t}\big)_j\right| \\ &= 3\lambda_t\left\|\big(\boldsymbol{\beta}^* - \widehat{\boldsymbol{\beta}}_{\lambda_t}\big)_{\overline{S_2^*}}\right\|_1 \\ &\le 3\lambda_t\sqrt{s^*}\left\|\big(\boldsymbol{\beta}^* - \widehat{\boldsymbol{\beta}}_{\lambda_t}\big)_{\overline{S_2^*}}\right\|_2 \\ &\le 3\lambda_t\sqrt{s_2^*}\left\|\boldsymbol{\beta}^* - \widehat{\boldsymbol{\beta}}_{\lambda_t}\right\|_2.\end{aligned}$$

Plugging (D.135)-(D.137) into the right-hand side of (D.134), we obtain

$$\left\|\widehat{\boldsymbol{\beta}}_{\lambda_t} - \boldsymbol{\beta}^*\right\|_2 \le \frac{1}{\rho_- - \zeta_-}\left(\left\|\big(\nabla\mathcal{L}(\boldsymbol{\beta}^*)\big)_{S_1^*}\right\|_2 + 3\lambda_t\sqrt{s_2^*}\right),$$

which concludes the proof of Theorem 4.8 in Wang et al. (2014a). □

**D.10. Proof for Lemma 4.9 and Theorem 4.10.** First, we prove Lemma 4.9 in Wang et al. (2014a), which states that the oracle estimator $\widehat{\boldsymbol{\beta}}_{\mathrm{O}}$ is uniquely defined and has nice statistical recovery property.

PROOF. To prove that the global minimizer of (4.19) in Wang et al. (2014a) is unique even for nonconvex loss functions, in the following we show that $\mathcal{L}(\boldsymbol{\beta})$ is strongly convex on the sparse set $\{\boldsymbol{\beta} : \mathrm{supp}\,(\boldsymbol{\beta}) \subseteq S^*\}$. We assume that $\boldsymbol{\beta}$ and $\boldsymbol{\beta}'$ satisfy $\mathrm{supp}(\boldsymbol{\beta}) \subseteq S^*$ and $\mathrm{supp}(\boldsymbol{\beta}') \subseteq S^*$. By Taylor's theorem and the mean value theorem, we have

(D.138)

$$\mathcal{L}(\boldsymbol{\beta}') = \mathcal{L}(\boldsymbol{\beta}) + \nabla\mathcal{L}(\boldsymbol{\beta})^T(\boldsymbol{\beta}'-\boldsymbol{\beta}) + \frac{1}{2}(\boldsymbol{\beta}'-\boldsymbol{\beta})^T\nabla^2\mathcal{L}\big(\gamma\boldsymbol{\beta}' + (1-\gamma)\boldsymbol{\beta}\big)(\boldsymbol{\beta}'-\boldsymbol{\beta}),$$

where $\gamma \in [0,1]$. Note that we have $\|\boldsymbol{\beta}' - \boldsymbol{\beta}\|_0 = s^* < s^* + 2\widetilde{s}$. By Definition 4.2 and Definition 4.3 in Wang et al. (2014a), we have

$$\frac{(\boldsymbol{\beta}' - \boldsymbol{\beta})^T}{\|\boldsymbol{\beta}' - \boldsymbol{\beta}\|_2} \nabla^2 \mathcal{L}\big(\gamma\boldsymbol{\beta} + (1-\gamma)\boldsymbol{\beta}'\big) \frac{(\boldsymbol{\beta}' - \boldsymbol{\beta})}{\|\boldsymbol{\beta}' - \boldsymbol{\beta}\|_2} \geq \rho_-\big(\nabla^2\mathcal{L}, s^* + 2\widetilde{s}\big).$$

Plugging this into the right-hand side of (D.138), we obtain

$$\mathcal{L}(\boldsymbol{\beta}') \geq \mathcal{L}(\boldsymbol{\beta}) + \nabla\mathcal{L}(\boldsymbol{\beta})^T(\boldsymbol{\beta}' - \boldsymbol{\beta}) + \frac{\rho_-}{2}\|\boldsymbol{\beta}' - \boldsymbol{\beta}\|_2^2, \tag{D.139}$$

where $\rho_- = \rho_-\big(\nabla^2\mathcal{L}, s^* + 2\widetilde{s}\big)$ is a positive constant according to Assumption 4.4. Note that (D.139) holds for any $\boldsymbol{\beta}$ and $\boldsymbol{\beta}'$ such that $\mathrm{supp}(\boldsymbol{\beta}) \subseteq S^*$ and $\mathrm{supp}(\boldsymbol{\beta}') \subseteq S^*$. Therefore, $\mathcal{L}(\boldsymbol{\beta})$ is strongly convex on this sparse set, which implies the minimizer of (4.19) in Wang et al. (2014a) is unique.

Now we prove the statistical recovery property of the oracle estimator $\widehat{\boldsymbol{\beta}}_{\mathrm{O}}$ in the setting where $\mathcal{L}(\boldsymbol{\beta})$ is least squares loss. Let $\widehat{\boldsymbol{\beta}}'_{\mathrm{O}}, \boldsymbol{\beta}^{*\prime} \in \mathbb{R}^{s^*}$ be the restrictions of $\widehat{\boldsymbol{\beta}}_{\mathrm{O}}, \boldsymbol{\beta}^* \in \mathbb{R}^d$ to $S^*$ respectively, and $\mathbf{X}_{S^*} \in \mathbb{R}^{n \times s^*}$ be a new matrix containing the columns of $\mathbf{X}$, i.e., $\mathbf{X}_j$, that satisfy $j \in S^*$. Since $\widehat{\boldsymbol{\beta}}'_{\mathrm{O}}$ is the solution to the ordinary least squares problem

$$\widehat{\boldsymbol{\beta}}'_{\mathrm{O}} = \mathop{\mathrm{argmin}}_{\boldsymbol{\beta}' \in \mathbb{R}^{s^*}} \frac{1}{2n}\|\mathbf{X}_{S^*}\boldsymbol{\beta}' - \mathbf{y}\|_2^2,$$

it has the closed-form expression of

$$\widehat{\boldsymbol{\beta}}'_{\mathrm{O}} = (\mathbf{X}_{S^*}^T\mathbf{X}_{S^*})^{-1}\mathbf{X}_{S^*}^T\mathbf{y}.$$

Here we still need to prove that $\mathbf{X}_{S^*}^T\mathbf{X}_{S^*} \in \mathbb{R}^{s^* \times s^*}$ is invertible. Note that the smallest eigenvalue of $\mathbf{X}_{S^*}^T\mathbf{X}_{S^*}$ is defined as

$$\Lambda_{\min}\big(\mathbf{X}_{S^*}^T\mathbf{X}_{S^*}\big) = \inf\left\{\boldsymbol{v}^T\mathbf{X}_{S^*}^T\mathbf{X}_{S^*}\boldsymbol{v} : \|\boldsymbol{v}\|_2 = 1,\ \boldsymbol{v} \in \mathbb{R}^{s^*}\right\},$$

which satisfies

$$\begin{aligned}
\Lambda_{\min}\big(\mathbf{X}_{S^*}^T\mathbf{X}_{S^*}\big) &= \inf\left\{\boldsymbol{v}^T\mathbf{X}^T\mathbf{X}\boldsymbol{v} : \|\boldsymbol{v}\|_2 = 1,\ \boldsymbol{v} \in \mathbb{R}^d,\ \ \mathrm{supp}(\boldsymbol{v}) = S^*\right\} \\
&\geq \inf\left\{\boldsymbol{v}^T\mathbf{X}^T\mathbf{X}\boldsymbol{v} : \|\boldsymbol{v}\|_2 = 1,\ \boldsymbol{v} \in \mathbb{R}^d,\ \ \|\boldsymbol{v}\|_0 \leq s^*\right\} \\
&\geq \inf\left\{\boldsymbol{v}^T\mathbf{X}^T\mathbf{X}\boldsymbol{v} : \|\boldsymbol{v}\|_2 = 1,\ \boldsymbol{v} \in \mathbb{R}^d,\ \ \|\boldsymbol{v}\|_0 \leq s^* + 2\widetilde{s}\right\} \\
&= n\rho_-\big(\nabla^2\mathcal{L}, s^* + 2\widetilde{s}\big) \qquad \text{(D.140)} \\
&> 0.
\end{aligned}$$

Here the first and second inequality are due to $\{\boldsymbol{v} : \mathrm{supp}(\boldsymbol{v}) = S^*\} \subseteq \{\boldsymbol{v} : \|\boldsymbol{v}\|_0 \leq s^*\} \subseteq \{\boldsymbol{v} : \|\boldsymbol{v}\|_0 \leq s^* + 2\widetilde{s}\}$, while the second equality follows from Definition 4.2 in Wang et al. (2014a), since for least squares loss $\nabla^2\mathcal{L}(\boldsymbol{\beta}) = \mathbf{X}^T\mathbf{X}/n$, and the last inequality follows from Assumption 4.4 in Wang et al. (2014a). Therefore the smallest eigenvalue of $\mathbf{X}_{S^*}^T\mathbf{X}_{S^*}$ is positive, which implies that $\mathbf{X}_{S^*}^T\mathbf{X}_{S^*}$ is invertible.

By our assumption on $(Y|\boldsymbol{X}=\mathbf{x}_i)$, we have $\mathbf{y}=\mathbf{X}\boldsymbol{\beta}^*+\boldsymbol{\epsilon}=\mathbf{X}_{S^*}\boldsymbol{\beta}^{*\prime}+\boldsymbol{\epsilon}$, where $\boldsymbol{\epsilon}\in\mathbb{R}^n$ is a zero mean sub-Gaussian random vector with independent entries and variance proxy $\sigma^2$. Therefore, we have

$$\widehat{\boldsymbol{\beta}}_{\mathrm{O}}' - \boldsymbol{\beta}^{*\prime} = (\mathbf{X}_{S^*}^T\mathbf{X}_{S^*})^{-1}\mathbf{X}_{S^*}^T\mathbf{y} - \boldsymbol{\beta}^{*\prime} = (\mathbf{X}_{S^*}^T\mathbf{X}_{S^*})^{-1}\mathbf{X}_{S^*}^T(\mathbf{X}\boldsymbol{\beta}^*+\boldsymbol{\epsilon}) - \boldsymbol{\beta}^{*\prime}$$
$$= (\mathbf{X}_{S^*}^T\mathbf{X}_{S^*})^{-1}\mathbf{X}_{S^*}^T\boldsymbol{\epsilon}.$$

Now we provide an upper bound of $\|\widehat{\boldsymbol{\beta}}_{\mathrm{O}}' - \boldsymbol{\beta}^{*\prime}\|_\infty$. Note that the $j$-th entry of $(\mathbf{X}_{S^*}^T\mathbf{X}_{S^*})^{-1}\mathbf{X}_{S^*}^T\boldsymbol{\epsilon}\in\mathbb{R}^{s^*}$ could be denoted as $\boldsymbol{e}_j(\mathbf{X}_{S^*}^T\mathbf{X}_{S^*})^{-1}\mathbf{X}_{S^*}^T\boldsymbol{\epsilon}$. Here $\boldsymbol{e}_j\in\mathbb{R}^{s^*}$ denotes a vector that is all-zero expect an "1" in its $j$-th coordinate. Hence, for any $j$, $\boldsymbol{e}_j(\mathbf{X}_{S^*}^T\mathbf{X}_{S^*})^{-1}\mathbf{X}_{S^*}^T\boldsymbol{\epsilon}$ is sub-Gaussian with variance proxy $\|\boldsymbol{e}_j(\mathbf{X}_{S^*}^T\mathbf{X}_{S^*})^{-1}\mathbf{X}_{S^*}^T\|_2^2\sigma^2$. Therefore we have

$$\mathbb{P}\big(\big|\boldsymbol{e}_j(\mathbf{X}_{S^*}^T\mathbf{X}_{S^*})^{-1}\mathbf{X}_{S^*}^T\boldsymbol{\epsilon}\big|>t\big) \le 2\exp\Big(-t^2/\big(\|\boldsymbol{e}_j(\mathbf{X}_{S^*}^T\mathbf{X}_{S^*})^{-1}\mathbf{X}_{S^*}^T\|_2^2\sigma^2\big)\Big),$$

which implies

$$\mathbb{P}\left(\max_{j\in\{1,\dots,s^*\}}\big|\boldsymbol{e}_j(\mathbf{X}_{S^*}^T\mathbf{X}_{S^*})^{-1}\mathbf{X}_{S^*}^T\boldsymbol{\epsilon}\big|>t\right)$$
$$\le 2s^*\exp\left(-t^2\Big/\Big(\max_{j\in\{1,\dots,s^*\}}\|\boldsymbol{e}_j(\mathbf{X}_{S^*}^T\mathbf{X}_{S^*})^{-1}\mathbf{X}_{S^*}^T\|_2^2\sigma^2\Big)\right).$$

Taking $t=C\max_{j\in\{1,\dots,s^*\}}\|\boldsymbol{e}_j(\mathbf{X}_{S^*}^T\mathbf{X}_{S^*})^{-1}\mathbf{X}_{S^*}^T\|_2\sigma\cdot\sqrt{2\log s^*}$ with $C>0$, we have that

$$\|\widehat{\boldsymbol{\beta}}_{\mathrm{O}}' - \boldsymbol{\beta}^{*\prime}\|_\infty = \|(\mathbf{X}_{S^*}^T\mathbf{X}_{S^*})^{-1}\mathbf{X}_{S^*}^T\boldsymbol{\epsilon}\|_\infty$$
$$= \max_{j\in\{1,\dots,s^*\}}\big|\boldsymbol{e}_j(\mathbf{X}_{S^*}^T\mathbf{X}_{S^*})^{-1}\mathbf{X}_{S^*}^T\boldsymbol{\epsilon}\big|$$
$$\le C\max_{j\in\{1,\dots,s^*\}}\|\boldsymbol{e}_j(\mathbf{X}_{S^*}^T\mathbf{X}_{S^*})^{-1}\mathbf{X}_{S^*}^T\|_2\sigma\cdot\sqrt{2\log s^*} \tag{D.141}$$

holds with probability at least $1-2\exp(-C^2)/s^*$. In other words, there exists a constant $C>0$ sufficiently large such that (D.141) holds with high probability. Note that, for any $j\in\{1,\dots,d\}$

$$\|\boldsymbol{e}_j(\mathbf{X}_{S^*}^T\mathbf{X}_{S^*})^{-1}\mathbf{X}_{S^*}^T\|_2^2 = \boldsymbol{e}_j(\mathbf{X}_{S^*}^T\mathbf{X}_{S^*})^{-1}\mathbf{X}_{S^*}^T\mathbf{X}_{S^*}(\mathbf{X}_{S^*}^T\mathbf{X}_{S^*})^{-1}\boldsymbol{e}_j^T$$
$$= \boldsymbol{e}_j(\mathbf{X}_{S^*}^T\mathbf{X}_{S^*})^{-1}\boldsymbol{e}_j^T$$
$$\le \Lambda_{\max}\big((\mathbf{X}_{S^*}^T\mathbf{X}_{S^*})^{-1}\big)$$
$$= 1/\Lambda_{\min}\big(\mathbf{X}_{S^*}^T\mathbf{X}_{S^*}\big)$$
$$\le 1/(n\rho_-),$$

where the last inequality follows from (D.140). Plugging this into (D.141),

we obtain

$$\big\|\widehat{\boldsymbol{\beta}}_{\rm O}' - \boldsymbol{\beta}^{*\prime}\big\|_\infty \le C\sigma\sqrt{2/\rho_-}\cdot\sqrt{\frac{\log s^*}{n}}.$$

We remind that $\widehat{\boldsymbol{\beta}}_{\rm O}'$ and $\boldsymbol{\beta}^{*\prime}$ are the restrictions of $\widehat{\boldsymbol{\beta}}_{\rm O}$ and $\boldsymbol{\beta}^*$ to $S^*$, and $\mathrm{supp}\big(\widehat{\boldsymbol{\beta}}_{\rm O}\big) \subseteq S^*$. Therefore we obtain

$$\big\|\widehat{\boldsymbol{\beta}}_{\rm O} - \boldsymbol{\beta}^{*}\big\|_\infty \le C\sigma\sqrt{2/\rho_-}\cdot\sqrt{\frac{\log s^*}{n}},$$

which concludes the proof. □

Now we prove Theorem 4.10 in Wang et al. (2014a).

PROOF. Let $\widehat{\boldsymbol{\xi}} \in \partial\big\|\widehat{\boldsymbol{\beta}}_{\lambda_t}\big\|_1$. We set $\widehat{\boldsymbol{\xi}}$ as the subgradient which attains the minimum in

$$\omega_{\lambda_t}\big(\widehat{\boldsymbol{\beta}}_{\lambda_t}\big) = \min_{\boldsymbol{\xi}'\in\partial\|\widehat{\boldsymbol{\beta}}_{\lambda_t}\|_1} \max_{\boldsymbol{\beta}'\in\Omega} \left\{ \frac{\big(\widehat{\boldsymbol{\beta}}_{\lambda_t} - \boldsymbol{\beta}'\big)^T}{\big\|\widehat{\boldsymbol{\beta}}_{\lambda_t} - \boldsymbol{\beta}'\big\|_1}\Big(\nabla\widetilde{\mathcal{L}}_{\lambda_t}\big(\widehat{\boldsymbol{\beta}}_{\lambda_t}\big) + \lambda_t\boldsymbol{\xi}'\Big)\right\}.$$

Since $\widehat{\boldsymbol{\beta}}_{\lambda_t}$ satisfies the exact optimality condition that $\omega_{\lambda_t}\big(\widehat{\boldsymbol{\beta}}_{\lambda_t}\big) \le 0$, we have

$$\max_{\boldsymbol{\beta}'\in\Omega}\left\{\big(\widehat{\boldsymbol{\beta}}_{\lambda_t} - \boldsymbol{\beta}'\big)^T\Big(\nabla\widetilde{\mathcal{L}}_{\lambda_t}\big(\widehat{\boldsymbol{\beta}}_{\lambda_t}\big) + \lambda_t\widehat{\boldsymbol{\xi}}\Big)\right\} \le 0. \tag{D.142}$$

Now we prove that there exists some $\boldsymbol{\xi}_{\rm O} \in \partial\big\|\widehat{\boldsymbol{\beta}}_{\rm O}\big\|_1$, such that $\widehat{\boldsymbol{\beta}}_{\rm O}$ satisfies the same exact optimality condition

$$\max_{\boldsymbol{\beta}'\in\Omega}\left\{\big(\widehat{\boldsymbol{\beta}}_{\rm O} - \boldsymbol{\beta}'\big)^T\Big(\nabla\widetilde{\mathcal{L}}_{\lambda_t}\big(\widehat{\boldsymbol{\beta}}_{\rm O}\big) + \lambda_t\boldsymbol{\xi}_{\rm O}\Big)\right\} \le 0. \tag{D.143}$$

Recall that $\widetilde{\mathcal{L}}_\lambda(\boldsymbol{\beta}) = \mathcal{L}(\boldsymbol{\beta}) + \mathcal{Q}_\lambda(\boldsymbol{\beta})$. In (D.143), we have

(D.144)

$$\begin{aligned}&\big(\widehat{\boldsymbol{\beta}}_{\rm O} - \boldsymbol{\beta}'\big)^T\Big(\nabla\widetilde{\mathcal{L}}_{\lambda_t}\big(\widehat{\boldsymbol{\beta}}_{\rm O}\big) + \lambda_t\boldsymbol{\xi}_{\rm O}\Big)\\ &= \underbrace{\sum_{j\in S^*}\big(\widehat{\boldsymbol{\beta}}_{\rm O} - \boldsymbol{\beta}'\big)_j\Big(\nabla\widetilde{\mathcal{L}}_{\lambda_t}\big(\widehat{\boldsymbol{\beta}}_{\rm O}\big) + \lambda_t\boldsymbol{\xi}_{\rm O}\Big)_j}_{\text{(i)}} + \underbrace{\sum_{j\in \overline{S^*}}\big(\widehat{\boldsymbol{\beta}}_{\rm O} - \boldsymbol{\beta}'\big)_j\Big(\nabla\widetilde{\mathcal{L}}_{\lambda_t}\big(\widehat{\boldsymbol{\beta}}_{\rm O}\big) + \lambda_t\boldsymbol{\xi}_{\rm O}\Big)_j}_{\text{(ii)}}.\end{aligned}$$

For term (i) in (D.144), according to Lemma 4.9 of Wang et al. (2014a) we have, for $n$ sufficiently large,

$$\big|\big(\widehat{\boldsymbol{\beta}}_{\rm O}\big)_j\big| \ge |\beta_j^*| - \big\|\widehat{\boldsymbol{\beta}}_{\rm O} - \boldsymbol{\beta}^*\big\|_\infty \ge 2\nu_t - \sigma\sqrt{2/\rho_-}\cdot\sqrt{\frac{\log s^*}{n}} \ge \nu_t. \tag{D.145}$$

Recall that $\mathcal{P}_\lambda(\boldsymbol{\beta}) = \mathcal{Q}_\lambda(\boldsymbol{\beta}) + \lambda\|\boldsymbol{\beta}\|_1$. Hence we have

$$\Big(\nabla\mathcal{Q}_{\lambda_t}(\widehat{\boldsymbol{\beta}}_{\mathrm{O}}) + \lambda_t\boldsymbol{\xi}_{\mathrm{O}}\Big)_j = \Big(\nabla\mathcal{P}_{\lambda_t}(\widehat{\boldsymbol{\beta}}_{\mathrm{O}})\Big)_j = p'_{\lambda_t}\Big((\widehat{\boldsymbol{\beta}}_{\mathrm{O}})_j\Big) = 0, \tag{D.146}$$

where the last equality is from (4.16) of Wang et al. (2014a). Then we have

$$\begin{aligned} &\sum_{j\in S^*}(\widehat{\boldsymbol{\beta}}_{\mathrm{O}}-\boldsymbol{\beta}')_j\Big(\nabla\widetilde{\mathcal{L}}_{\lambda_t}(\widehat{\boldsymbol{\beta}}_{\mathrm{O}})+\lambda_t\boldsymbol{\xi}_{\mathrm{O}}\Big)_j \\ &\quad= \sum_{j\in S^*}(\widehat{\boldsymbol{\beta}}_{\mathrm{O}}-\boldsymbol{\beta}')_j\Big(\nabla\mathcal{L}(\widehat{\boldsymbol{\beta}}_{\mathrm{O}})+\nabla\mathcal{Q}_{\lambda_t}(\widehat{\boldsymbol{\beta}}_{\mathrm{O}})+\lambda_t\boldsymbol{\xi}_{\mathrm{O}}\Big)_j \\ &\quad= \sum_{j\in S^*}(\widehat{\boldsymbol{\beta}}_{\mathrm{O}}-\boldsymbol{\beta}')_j\Big(\nabla\mathcal{L}(\widehat{\boldsymbol{\beta}}_{\mathrm{O}})\Big)_j, \end{aligned} \tag{D.147}$$

where the first equality follows from $\widetilde{\mathcal{L}}_\lambda(\boldsymbol{\beta}) = \mathcal{L}(\boldsymbol{\beta}) + \mathcal{Q}_\lambda(\boldsymbol{\beta})$, and the second follows from (D.146). We remind that $\widehat{\boldsymbol{\beta}}_{\mathrm{O}}$ is the global solution to the oracle minimization problem in (4.19) of Wang et al. (2014a). Hence $\widehat{\boldsymbol{\beta}}_{\mathrm{O}}$ satisfies the exact optimality condition of (4.19)

$$\max_{\boldsymbol{\beta}'\in\Omega}\bigg\{\sum_{j\in S^*}(\widehat{\boldsymbol{\beta}}_{\mathrm{O}}-\boldsymbol{\beta}')_j\Big(\nabla\mathcal{L}(\widehat{\boldsymbol{\beta}}_{\mathrm{O}})\Big)_j\bigg\} \le 0.$$

Thus, taking maximum over $\boldsymbol{\beta}' \in \Omega$ on both sides of (D.147), we have that, the maximum of term (i) over $\boldsymbol{\beta}' \in \Omega$ is upper bounded by zero.

For term (ii) in (D.144), remind that $\big(\widehat{\boldsymbol{\beta}}_{\mathrm{O}}\big)_j = 0$ for $j \in \overline{S^*}$. According to regularity condition (c) we have

$$\Big(\nabla\mathcal{Q}_{\lambda_t}(\widehat{\boldsymbol{\beta}}_{\mathrm{O}})\Big)_j = 0.$$

Meanwhile, for any $j \in \overline{S^*}$ it holds that

$$\Big|\Big(\nabla\mathcal{L}(\widehat{\boldsymbol{\beta}}_{\mathrm{O}})\Big)_j\Big| \le \|\nabla\mathcal{L}(\widehat{\boldsymbol{\beta}}_{\mathrm{O}})\|_\infty \le \Big\|\mathbf{X}^T(\mathbf{y}-\mathbf{X}\widehat{\boldsymbol{\beta}}_{\mathrm{O}})/n\Big\|_\infty. \tag{D.148}$$

On the right-hand side of (D.148), we have

$$\begin{aligned} \mathbf{y}-\mathbf{X}\widehat{\boldsymbol{\beta}}_{\mathrm{O}} &= \mathbf{y}-\mathbf{X}_{S^*}(\mathbf{X}_{S^*}^T\mathbf{X}_{S^*})^{-1}\mathbf{X}_{S^*}^T\mathbf{y} \\ &= \mathbf{X}_{S^*}\boldsymbol{\beta}^*_{S^*}+\boldsymbol{\epsilon}-\mathbf{X}_{S^*}(\mathbf{X}_{S^*}^T\mathbf{X}_{S^*})^{-1}\mathbf{X}_{S^*}^T(\mathbf{X}_{S^*}\boldsymbol{\beta}^*_{S^*}+\boldsymbol{\epsilon}) \\ &= \big(\mathbf{I}-\mathbf{X}_{S^*}(\mathbf{X}_{S^*}^T\mathbf{X}_{S^*})^{-1}\mathbf{X}_{S^*}^T\big)\boldsymbol{\epsilon}. \end{aligned} \tag{D.149}$$

Note that $\mathbf{X}_{S^*}(\mathbf{X}_{S^*}^T\mathbf{X}_{S^*})^{-1}\mathbf{X}_{S^*}^T$ is a projection matrix, which further implies $\mathbf{I}-\mathbf{X}_{S^*}(\mathbf{X}_{S^*}^T\mathbf{X}_{S^*})^{-1}\mathbf{X}_{S^*}^T$ is also a projection matrix. We define the right-hand side of (D.149) to be $\boldsymbol{\epsilon}'$. Hence we have that $\boldsymbol{\epsilon}'$ is sub-Gaussian with variance proxy no larger than that of $\boldsymbol{\epsilon}$, and

$$\Big|\Big(\nabla\mathcal{L}(\widehat{\boldsymbol{\beta}}_{\mathrm{O}})\Big)_j\Big| \le \|\nabla\mathcal{L}(\widehat{\boldsymbol{\beta}}_{\mathrm{O}})\|_\infty = \|\mathbf{X}^T\boldsymbol{\epsilon}'/n\|_\infty.$$

Note that

$$\nabla\mathcal{L}(\boldsymbol{\beta}^*) = \mathbf{X}^T(\mathbf{y} - \mathbf{X}\boldsymbol{\beta}^*)/n = \mathbf{X}^T\boldsymbol{\epsilon}/n.$$

Following the same proof for $\|\nabla\mathcal{L}(\boldsymbol{\beta}^*)\|_\infty \leq \lambda_{\text{tgt}}/8$, from (D.148) and (D.149) we obtain

$$\left|\Big(\nabla\mathcal{L}(\widehat{\boldsymbol{\beta}}_{\text{O}})\Big)_j\right| \leq \big\|\nabla\mathcal{L}(\widehat{\boldsymbol{\beta}}_{\text{O}})\big\|_\infty \leq \lambda_{\text{tgt}}/8.$$

Therefore, since $\boldsymbol{\xi}_{\text{O}} \in \partial\big\|\widehat{\boldsymbol{\beta}}_{\text{O}}\big\|_1$, for

$$\Big(\nabla\widetilde{\mathcal{L}}_{\lambda_t}(\widehat{\boldsymbol{\beta}}_{\text{O}}) + \lambda_t\boldsymbol{\xi}_{\text{O}}\Big)_j = \Big(\nabla\mathcal{L}(\widehat{\boldsymbol{\beta}}_{\text{O}}) + \nabla\mathcal{Q}_{\lambda_t}(\widehat{\boldsymbol{\beta}}_{\text{O}}) + \lambda_t\boldsymbol{\xi}_{\text{O}}\Big)_j = \Big(\nabla\mathcal{L}(\widehat{\boldsymbol{\beta}}_{\text{O}}) + \lambda_t\boldsymbol{\xi}_{\text{O}}\Big)_j,$$

we can set $(\boldsymbol{\xi}_{\text{O}})_j = -\Big(\nabla\mathcal{L}(\widehat{\boldsymbol{\beta}}_{\text{O}})/\lambda_t\Big)_j$. Then we obtain, for any $j \in \overline{S^*}$,

$$\Big(\nabla\widetilde{\mathcal{L}}_{\lambda_t}(\widehat{\boldsymbol{\beta}}_{\text{O}}) + \lambda_t\boldsymbol{\xi}_{\text{O}}\Big)_j = 0,$$

which further implies that term (ii) in (D.144) is zero. In summary, taking maximum over $\boldsymbol{\beta}' \in \Omega$ on both sides of (D.144), we obtain (D.143).

Now we are ready to prove that $\widehat{\boldsymbol{\beta}}_{\lambda_t} = \widehat{\boldsymbol{\beta}}_{\text{O}}$. Recall that the oracle estimator satisfies $\text{supp}(\widehat{\boldsymbol{\beta}}_{\text{O}}) \subseteq S^*$. Meanwhile, from Theorem 5.5 of Wang et al. (2014a) we have $\big\|(\widehat{\boldsymbol{\beta}}_{\lambda_t})_{\overline{S^*}}\big\|_0 \leq \widetilde{s}$. Hence, we have $\big\|(\widehat{\boldsymbol{\beta}}_{\lambda_t} - \widehat{\boldsymbol{\beta}}_{\text{O}})_{\overline{S^*}}\big\|_0 \leq \widetilde{s}$, and Lemma 5.1 of Wang et al. (2014a) yields

$$\widetilde{\mathcal{L}}_{\lambda_t}(\widehat{\boldsymbol{\beta}}_{\lambda_t}) \geq \widetilde{\mathcal{L}}_{\lambda_t}(\widehat{\boldsymbol{\beta}}_{\text{O}}) + \nabla\widetilde{\mathcal{L}}_{\lambda_t}(\widehat{\boldsymbol{\beta}}_{\text{O}})^T(\widehat{\boldsymbol{\beta}}_{\lambda_t} - \widehat{\boldsymbol{\beta}}_{\text{O}}) + \frac{\rho_- - \zeta_-}{2}\big\|\widehat{\boldsymbol{\beta}}_{\lambda_t} - \widehat{\boldsymbol{\beta}}_{\text{O}}\big\|_2^2, \tag{D.150}$$

$$\widetilde{\mathcal{L}}_{\lambda_t}(\widehat{\boldsymbol{\beta}}_{\text{O}}) \geq \widetilde{\mathcal{L}}_{\lambda_t}(\widehat{\boldsymbol{\beta}}_{\lambda_t}) + \nabla\widetilde{\mathcal{L}}_{\lambda_t}(\widehat{\boldsymbol{\beta}}_{\lambda_t})^T(\widehat{\boldsymbol{\beta}}_{\text{O}} - \widehat{\boldsymbol{\beta}}_{\lambda_t}) + \frac{\rho_- - \zeta_-}{2}\big\|\widehat{\boldsymbol{\beta}}_{\text{O}} - \widehat{\boldsymbol{\beta}}_{\lambda_t}\big\|_2^2. \tag{D.151}$$

Meanwhile, by the convexity of $\ell_1$ norm, we have

$$\lambda_t\big\|\widehat{\boldsymbol{\beta}}_{\lambda_t}\big\|_1 \geq \lambda_t\big\|\widehat{\boldsymbol{\beta}}_{\text{O}}\big\|_1 + \lambda_t(\widehat{\boldsymbol{\beta}}_{\lambda_t} - \widehat{\boldsymbol{\beta}}_{\text{O}})^T\boldsymbol{\xi}_{\text{O}}, \tag{D.152}$$

$$\lambda_t\big\|\widehat{\boldsymbol{\beta}}_{\text{O}}\big\|_1 \geq \lambda_t\big\|\widehat{\boldsymbol{\beta}}_{\lambda_t}\big\|_1 + \lambda_t(\widehat{\boldsymbol{\beta}}_{\text{O}} - \widehat{\boldsymbol{\beta}}_{\lambda_t})^T\widehat{\boldsymbol{\xi}}. \tag{D.153}$$

Adding (D.150)-(D.153), we obtain

$$0 \geq \underbrace{\Big(\nabla\widetilde{\mathcal{L}}_{\lambda_t}(\widehat{\boldsymbol{\beta}}_{\lambda_t}) + \lambda_t\widehat{\boldsymbol{\xi}}\Big)^T(\widehat{\boldsymbol{\beta}}_{\text{O}} - \widehat{\boldsymbol{\beta}}_{\lambda_t})}_{\text{(i)}} + \underbrace{\Big(\nabla\widetilde{\mathcal{L}}_{\lambda_t}(\widehat{\boldsymbol{\beta}}_{\text{O}}) + \lambda_t\boldsymbol{\xi}_{\text{O}}\Big)^T(\widehat{\boldsymbol{\beta}}_{\lambda_t} - \widehat{\boldsymbol{\beta}}_{\text{O}})}_{\text{(ii)}} + (\rho_- - \zeta_-)\big\|\widehat{\boldsymbol{\beta}}_{\lambda_t} - \widehat{\boldsymbol{\beta}}_{\text{O}}\big\|_2^2.$$

According to (D.142), we have

$$\big(\widehat{\boldsymbol{\beta}}_{\lambda_t} - \widehat{\boldsymbol{\beta}}_{\mathrm{O}}\big)^T \Big(\nabla \widetilde{\mathcal{L}}_{\lambda_t}\big(\widehat{\boldsymbol{\beta}}_{\lambda_t}\big) + \lambda_t \widehat{\boldsymbol{\xi}}\Big) \le \max_{\boldsymbol{\beta}' \in \Omega} \Big\{ \big(\widehat{\boldsymbol{\beta}}_{\lambda_t} - \boldsymbol{\beta}'\big)^T \Big(\nabla \widetilde{\mathcal{L}}_{\lambda_t}\big(\widehat{\boldsymbol{\beta}}_{\lambda_t}\big) + \lambda_t \widehat{\boldsymbol{\xi}}\Big) \Big\}$$
$$\le 0,$$

which implies term (i) is nonnegative. Similarly, according to (D.143), term (ii) is also nonnegative. Hence we have

$$(\rho_- - \zeta_-) \big\| \widehat{\boldsymbol{\beta}}_{\lambda_t} - \widehat{\boldsymbol{\beta}}_{\mathrm{O}} \big\|_2^2 \le 0.$$

By (4.3) of Wang et al. (2014a) we have $\rho_- - \zeta_- > 0$, which implies $\widehat{\boldsymbol{\beta}}_{\lambda_t} = \widehat{\boldsymbol{\beta}}_{\mathrm{O}}$. Thus, we conclude that $\widehat{\boldsymbol{\beta}}_{\lambda_t}$ is the oracle estimator $\widehat{\boldsymbol{\beta}}_{\mathrm{O}}$. Moreover, (D.145) implies that $\min_{j \in S^*} \big| \big(\widehat{\boldsymbol{\beta}}_{\mathrm{O}}\big)_j \big| > 0$. Together with $\mathrm{supp}\big(\widehat{\boldsymbol{\beta}}_{\mathrm{O}}\big) \subseteq S^*$, we have $\mathrm{supp}\big(\widehat{\boldsymbol{\beta}}_{\lambda_t}\big) = \mathrm{supp}\big(\widehat{\boldsymbol{\beta}}_{\mathrm{O}}\big) = \mathrm{supp}(\boldsymbol{\beta}^*)$. □

## APPENDIX E: THEORETICAL RESULTS ABOUT SEMIPARAMETRIC ELLIPTICAL DESIGN REGRESSION

In this section, we first introduce the Catoni's $M$-estimator of standard deviation, then we provide the detailed proofs of some necessary results on semiparametric elliptical design regression.[2]

**E.1. Catoni's $M$-Estimator of Standard Deviation.** Catoni (2012) proposed a novel estimator for the mean and standard deviation of heavy-tail distributions. Let $\boldsymbol{Z} = (Z_1, \dots, Z_{d+1})$ be the elliptically distributed random vector defined in §2.2 of Wang et al. (2014a). We consider the estimator of the marginal mean $\mathbb{E}(Z_j)$ $(j = 1, \dots, d+1)$. Let $h : \mathbb{R} \to \mathbb{R}$ be a continuous strictly increasing function satisfying

$$-\log(1 - x + x^2/2) \le h(x) \le \log(1 + x + x^2/2).$$

For instance, we choose $h(\cdot)$ to be

$$h(x) = \begin{cases} \log(1 + x + x^2/2), & \text{if } x \ge 0, \\ -\log(1 - x + x^2/2), & \text{otherwise.} \end{cases}$$

Let $\delta \in (0, 1)$ be such that $n \ge 2\log(1/\delta)$. We introduce

$$a_\delta = \sqrt{2\log(1/\delta) \Big/ \left( nv + \frac{2nv \log(1/\delta)}{n - 2\log(1/\delta)} \right)}, \tag{E.1}$$

[2]§E.1, Lemma E.1 and Corollary E.2 come from an unpublished internal technical report. We provide them here for completeness.

where $v$ is an upper bound of $\mathrm{Var}(Z_j)$ for all $j$. Catoni's estimator of $\mathbb{E}(Z_j)$ is defined as $\widehat{\mu}_j = \widehat{\mu}_j(n, \delta)$ such that

$$\sum_{i=1}^{n} h\big(\alpha_\delta(z_{i,j} - \widehat{\mu}_j)\big) = 0, \quad j = 1, \dots, d+1, \tag{E.2}$$

where $z_{i,j}$ is the $i$-th ($i = 1, \dots, n$) realizations of $Z_j$. As $h(\cdot)$ is differentiable everywhere, we can solve (E.2) with Newton's method efficiently. Similarly we can estimate $\mathbb{E}(Z_j^2)$ with $\widehat{m}_j$ defined in a similar way. Then we obtain an estimator of the marginal standard deviation $\sigma_j$

$$\widehat{\sigma}_j = \sqrt{\widehat{m}_j - \widehat{\mu}_j^2}, \quad j = 1, \dots, d+1. \tag{E.3}$$

**E.2. Proof of Lemma C.3.** To establish results concerning the smallest sparse eigenvalue for $\widehat{\mathbf{K}}_{\boldsymbol{X}}$, we need to prove several concentration results. The next lemma and proposition provide the concentration inequality for Catoni's estimator of marginal standard deviation, which is defined in (E.3). We first consider the estimator of variance in the following lemma.

**Lemma E.1.** Let $\boldsymbol{X} = (X_1, \dots, X_d)^T$ be a random vector and $\mathbf{x}_1, \dots, \mathbf{x}_n$ be $n$ independent realizations of $\boldsymbol{X}$ with $\mathrm{Var}(X_j) = v_j$ and $\mathbb{E}\big(X_j^4\big) \le M$, for $j = 1, \dots, d$. We assume that

$$\max_{1 \le j \le d} \{|\mathbb{E}(X_j)|\} \le \mu_{\max}, \quad v_{\max} = \max_{1 \le j \le d} \{v_j\}.$$

For the estimator $\widehat{v}_j = \widehat{m}_j - \widehat{\mu}_j^2$ with $\widehat{m}_j$ and $\widehat{\mu}_j$ defined in (E.2), if $n > 5 \log d$, we have, with probability at least $1 - 2d^{-3}$,

$$\max_{1 \le j \le d} \Big\{ \big| v_j - \widehat{v}_j \big| \Big\} \le C \sqrt{\frac{\log d}{n}},$$

where $C$ is a constant.

Proof. For $j \in \{1, \dots, d\}$, we use $\widehat{m}_j$ to estimate $\mathbb{E}\big(X_j^2\big)$. Catoni (2012) showed that

$$\mathbb{P}\Big( \big| \widehat{m}_j - \mathbb{E}\big(X_j^2\big) \big| > t \Big) \le \exp\Big( - \frac{nt^2}{M} \Big).$$

Taking a union bound, we have

$$\mathbb{P}\Big( \max_{1 \le j \le d} \Big\{ \big| \widehat{m}_j - \mathbb{E}\big(X_j^2\big) \big| \Big\} > t \Big) \le d \exp\Big( - \frac{nt^2}{M} \Big),$$

or equivalently, with probability at least $1 - d^{-3}$,

$$\max_{1 \le j \le d} \Big\{ \big| \widehat{m}_j - \mathbb{E}\big(X_j^2\big) \big| \Big\} \le 2\sqrt{M} \sqrt{\frac{\log d}{n}}. \tag{E.4}$$

Meanwhile, we use $\widehat{\mu}_j$ to estimate $\mathbb{E}(X_j)$. By similar arguments as above, we have

$$\max_{1\le j\le d}\Big\{\big|\widehat{\mu}_j-\mathbb{E}(X_j)\big|\Big\}\le 2\sqrt{v_{\max}}\sqrt{\frac{\log d}{n}} \tag{E.5}$$

with probability at least $1-d^{-3}$.

Note that

$$\max_{1\le j\le d}\Big\{\Big|\widehat{\mu}_j^2-\big(\mathbb{E}(X_j)\big)^2\Big|\Big\}\le\max_{1\le j\le d}\Big\{\big|\widehat{\mu}_j-\mathbb{E}(X_j)\big|\Big\}\cdot\max_{1\le j\le d}\Big\{\big|\widehat{\mu}_j+\mathbb{E}(X_j)\big|\Big\}.$$

Since we assume that $\max_{1\le j\le d}\{\mathbb{E}(X_j)\}\le\mu_{\max}$, we have

$$\max_{1\le j\le d}\Big\{\Big|\widehat{\mu}_j^2-\big(\mathbb{E}(X_j)\big)^2\Big|\Big\}\le\bigg(4\mu_{\max}+4\sqrt{v_{\max}}\sqrt{\frac{\log d}{n}}\bigg)\cdot\sqrt{v_{\max}}\sqrt{\frac{\log d}{n}} \tag{E.6}$$

with probability at least $1-d^{-3}$. Since $\log d/n<1$, from (E.6) we have,

$$\max_{1\le j\le d}\Big\{\Big|\widehat{\mu}_j^2-\big(\mathbb{E}(X_j)\big)^2\Big|\Big\}\le(4\mu_{\max}+4\sqrt{v_{\max}})\cdot\sqrt{v_{\max}}\sqrt{\frac{\log d}{n}}. \tag{E.7}$$

Combining (E.4) and (E.7), we have, with probability at least $1-2d^{-3}$,

$$\max_{1\le j\le d}\Big\{\big|\widehat{m}_j-\widehat{\mu}_j^2-\mathrm{Var}(X_j)\big|\Big\}\le C\sqrt{\frac{\log d}{n}},$$

where $C=2\sqrt{M}+\big(4\mu_{\max}+4\sqrt{v_{\max}}\big)\sqrt{v_{\max}}$. □

We use $\widehat{\sigma}_j=\sqrt{\widehat{v}_j}$ to estimate $\sigma_j=\sqrt{v_j}$. Using Lemma E.1, we derive a concentration inequality for $\widehat{\sigma}_j$ in the following corollary.

**Corollary E.2.** Let $\sigma_j=\sqrt{v_j}$ and $\widehat{\sigma}_j=\sqrt{\widehat{v}_j}$ for $j=1,\dots,d$. By assuming $\sigma_j\ge\sigma_{\min}>0$ for all $j=1,\dots,d$, we have, with probability at least $1-2d^{-3}$,

$$\max_{1\le j\le d}\Big\{|\sigma_j-\widehat{\sigma}_j|\Big\}\le C\sqrt{\frac{\log d}{n}},$$

where $C$ is a constant.

Proof. By Lemma E.1, we have, with probability at least $1-2d^{-3}$,

$$\max_{1\le j\le d}\Big\{|v_j-\widehat{v}_j|\Big\}\le C\sqrt{\frac{\log d}{n}}.$$

Since $|v_j-\widehat{v}_j|=|\sigma_j-\widehat{\sigma}_j|\cdot|\sigma_j+\widehat{\sigma}_j|$, it follows that

$$\max_{1\le j\le d}\Big\{|\sigma_j-\widehat{\sigma}_j|\Big\}\le\frac{C}{\min_{1\le j\le d}\Big\{|\sigma_j+\widehat{\sigma}_j|\Big\}}\sqrt{\frac{\log d}{n}}\le\frac{C}{\sigma_{\min}}\sqrt{\frac{\log d}{n}}.$$

As we assume that $\sigma_j > \sigma_{\min}$ for all $j$, we conclude the proof. $\square$

Before we establish the sparse eigenvalue condition for $\widehat{\mathbf{K}}_{\boldsymbol{X}}$, we provide a concentration result of $\widehat{\mathbf{R}}_{\boldsymbol{X}}$ in the following lemma.

**Lemma E.3** (Han and Liu (2013))**.** Let $\mathbf{x}_1, \ldots, \mathbf{x}_n$ be $n$ realizations of a random vector $\boldsymbol{X} \sim \mathrm{EC}_d(0, \boldsymbol{\Sigma}_{\boldsymbol{X}}, \Xi)$ as in Definition A.1. We assume that the smallest eigenvalue of the generalized correlation matrix $\boldsymbol{\Sigma}^0_{\boldsymbol{X}}$ is strictly positive. Under the sign sub-Gaussian condition (see Han and Liu (2013) for more details), the correlation matrix estimator $\widehat{\mathbf{R}}_{\boldsymbol{X}}$ defined in (A.1) satisfies that, with probability at least $1 - 2d^{-1} - d^{-2}$,

$$\sup_{\|\boldsymbol{v}\|_0 \le s} \left\{ \frac{\left| \boldsymbol{v}^T \big( \widehat{\mathbf{R}}_{\boldsymbol{X}} - \boldsymbol{\Sigma}^0_{\boldsymbol{X}} \big) \boldsymbol{v} \right|}{\|\boldsymbol{v}\|_2^2} \right\} \le C \sqrt{\frac{s \log d}{n}}$$

for $s \in \{1, \ldots, d\}$ and a sufficiently large $n$.

We now prove Lemma C.3.

Proof. Here we denote the diagonal matrix that has $x_1, \ldots, x_d$ on its diagonal by $\mathrm{diag}(x_1, \ldots, x_d)$. Let $\mathbf{D} = \mathrm{diag}(\sigma_1, \ldots, \sigma_d)$ and $\widehat{\mathbf{D}} = \mathrm{diag}(\widehat{\sigma}_1, \ldots, \widehat{\sigma}_d)$. First we consider the smallest sparse eigenvalue, which satisfies

$$\begin{aligned}
\rho_-\big(\nabla^2 \mathcal{L}, s\big) &= \inf_{\|\boldsymbol{v}\|_0 \le s} \left\{ \frac{\boldsymbol{v}^T \widehat{\mathbf{K}}_{\boldsymbol{X}} \boldsymbol{v}}{\|\boldsymbol{v}\|_2^2} \right\} \\
&= \inf_{\|\boldsymbol{v}\|_0 \le s} \left\{ \frac{\big(\widehat{\mathbf{D}}\boldsymbol{v}\big)^T \widehat{\mathbf{R}}_{\boldsymbol{X}} \big(\widehat{\mathbf{D}}\boldsymbol{v}\big)}{\big\|\widehat{\mathbf{D}}\boldsymbol{v}\big\|_2^2} \cdot \frac{\big\|\widehat{\mathbf{D}}\boldsymbol{v}\big\|_2^2}{\|\boldsymbol{v}\|_2^2} \right\} \\
&\ge \inf_{\|\boldsymbol{v}\|_0 \le s} \left\{ \frac{\boldsymbol{v}^T \widehat{\mathbf{R}}_{\boldsymbol{X}} \boldsymbol{v}}{\|\boldsymbol{v}\|_2^2} \right\} \cdot \min_{1 \le j \le d} \{\widehat{\sigma}_j\}. \qquad \text{(E.8)}
\end{aligned}$$

The first term on the right-hand side of (E.8) is the smallest sparse eigenvalue of $\widehat{\mathbf{R}}_{\boldsymbol{X}}$. Since we have from Lemma E.3 that, with probability at least $1 - 2d^{-1} - d^{-2}$,

$$\sup_{\|\boldsymbol{v}\|_0 \le s} \left\{ \frac{\left| \boldsymbol{v}^T \big( \widehat{\mathbf{R}}_{\boldsymbol{X}} - \boldsymbol{\Sigma}^0_{\boldsymbol{X}} \big) \boldsymbol{v} \right|}{\|\boldsymbol{v}\|_2^2} \right\} \le C \sqrt{\frac{s \log d}{n}}.$$

Then for a sufficiently large $n$, we have

$$\boldsymbol{v}^T \big( \boldsymbol{\Sigma}^0_{\boldsymbol{X}} - \widehat{\mathbf{R}}_{\boldsymbol{X}} \big) \boldsymbol{v} \le C \sqrt{\frac{s \log d}{n}} \le \frac{1}{2} \Lambda_{\min}\big(\boldsymbol{\Sigma}^0_{\boldsymbol{X}}\big), \quad \text{for} \quad \|\boldsymbol{v}\|_0 \le s.$$

Here $\Lambda_{\min}\big(\boldsymbol{\Sigma}^0_{\boldsymbol{X}}\big)$ denotes the smallest eigenvalue of $\boldsymbol{\Sigma}^0_{\boldsymbol{X}}$, which is strictly positive by assumption. Then we obtain

$$\frac{1}{2}\Lambda_{\min}\big(\boldsymbol{\Sigma}_{\boldsymbol{X}}^0\big) \le \boldsymbol{v}^T\boldsymbol{\Sigma}_{\boldsymbol{X}}^0\boldsymbol{v} - \frac{1}{2}\Lambda_{\min}\big(\boldsymbol{\Sigma}_{\boldsymbol{X}}^0\big) \le \boldsymbol{v}^T\widehat{\mathbf{R}}_{\boldsymbol{X}}\boldsymbol{v}, \quad \text{for} \;\; \|\boldsymbol{v}\|_0 \le s.$$

Taking infimum over both sides, we get

$$\inf_{\|\boldsymbol{v}\|_0\le s}\left\{\frac{\boldsymbol{v}^T\widehat{\mathbf{R}}_{\boldsymbol{X}}\boldsymbol{v}}{\|\boldsymbol{v}\|_2^2}\right\} \ge \frac{1}{2}\Lambda_{\min}\big(\boldsymbol{\Sigma}_{\boldsymbol{X}}^0\big) > 0. \tag{E.9}$$

We now consider $\min_{1\le j\le d}\{\widehat{\sigma}_j\}$ in (E.8). In Corollary E.2 we prove that, with probability at least $1-2d^{-3}$,

$$|\sigma_j - \widehat{\sigma}_j| \le C'\sqrt{\frac{\log d}{n}}, \quad \text{for} \;\; 1 \le j \le d,$$

where $C'$ is a constant. For a sufficiently large $n$, we have

$$\widehat{\sigma}_j \ge \frac{1}{2}\sigma_j > 0, \quad \text{for} \;\; 1 \le j \le d$$

with the same probability. Taking minimum over both sides, we get

$$\min_{1\le j\le d}\{\widehat{\sigma}_j\} \ge \frac{1}{2}\min_{1\le j\le d}\{\sigma_j\} > 0 \tag{E.10}$$

with probability at least $1-2d^{-2}$. Plugging (E.9) and (E.10) into the right-hand side of (E.8), we reach the conclusion that $\rho_-\big(\nabla^2\mathcal{L}, s\big) > 0$ holds with probability at least $1-2d^{-1}-3d^{-2}$.

Now we consider the largest sparse eigenvalue, which satisfies

$$\begin{aligned}
\rho_+\big(\nabla^2\mathcal{L}, s\big) &= \sup_{\|\boldsymbol{v}\|_0\le s}\left\{\frac{\boldsymbol{v}^T\widehat{\mathbf{K}}_{\boldsymbol{X}}\boldsymbol{v}}{\|\boldsymbol{v}\|_2^2}\right\} \\
&= \sup_{\|\boldsymbol{v}\|_0\le s}\left\{\frac{\big(\widehat{\mathbf{D}}\boldsymbol{v}\big)^T\widehat{\mathbf{R}}_{\boldsymbol{X}}\big(\widehat{\mathbf{D}}\boldsymbol{v}\big)}{\big\|\widehat{\mathbf{D}}\boldsymbol{v}\big\|_2^2}\cdot\frac{\big\|\widehat{\mathbf{D}}\boldsymbol{v}\big\|_2^2}{\|\boldsymbol{v}\|_2^2}\right\} \\
&\le \sup_{\|\boldsymbol{v}\|_0\le s}\left\{\frac{\boldsymbol{v}^T\widehat{\mathbf{R}}_{\boldsymbol{X}}\boldsymbol{v}}{\|\boldsymbol{v}\|_2^2}\right\}\cdot\max_{1\le j\le d}\{\widehat{\sigma}_j\}. \qquad \text{(E.11)}
\end{aligned}$$

The first term on the right-hand side of (E.11) is the largest sparse eigenvalue of $\widehat{\mathbf{R}}_{\boldsymbol{X}}$. Since we have from Lemma E.3 that, with probability at least $1-2d^{-1}-d^{-2}$,

$$\sup_{\|\boldsymbol{v}\|_0\le s}\left\{\frac{\big|\boldsymbol{v}^T\big(\widehat{\mathbf{R}}_{\boldsymbol{X}} - \boldsymbol{\Sigma}_{\boldsymbol{X}}^0\big)\boldsymbol{v}\big|}{\|\boldsymbol{v}\|_2^2}\right\} \le C\sqrt{\frac{s\log d}{n}}.$$

Then for a sufficiently large $n$, we have

$$\boldsymbol{v}^T\big(\widehat{\mathbf{R}}_{\boldsymbol{X}} - \boldsymbol{\Sigma}^0_{\boldsymbol{X}}\big)\boldsymbol{v} \le C\sqrt{\frac{s\log d}{n}} \le \frac{1}{2}\Lambda_{\max}\big(\boldsymbol{\Sigma}^0_{\boldsymbol{X}}\big), \quad \text{for} \ \ \|\boldsymbol{v}\|_0 \le s.$$

Here $\Lambda_{\max}\big(\boldsymbol{\Sigma}^0_{\boldsymbol{X}}\big)$ denotes the largest eigenvalue of $\boldsymbol{\Sigma}^0_{\boldsymbol{X}}$. Then we obtain

$$\boldsymbol{v}^T\widehat{\mathbf{R}}_{\boldsymbol{X}}\boldsymbol{v} \le \boldsymbol{v}^T\boldsymbol{\Sigma}^0_{\boldsymbol{X}}\boldsymbol{v} + \frac{1}{2}\Lambda_{\max}\big(\boldsymbol{\Sigma}^0_{\boldsymbol{X}}\big) \le \frac{3}{2}\Lambda_{\max}\big(\boldsymbol{\Sigma}^0_{\boldsymbol{X}}\big), \quad \text{for} \ \ \|\boldsymbol{v}\|_0 \le s.$$

Taking supremum over both sides, we get

$$\sup_{\|\boldsymbol{v}\|_0\le s}\left\{\frac{\boldsymbol{v}^T\widehat{\mathbf{R}}_{\boldsymbol{X}}\boldsymbol{v}}{\|\boldsymbol{v}\|_2^2}\right\} \le \frac{1}{2}\Lambda_{\max}\big(\boldsymbol{\Sigma}^0_{\boldsymbol{X}}\big) < +\infty. \tag{E.12}$$

We now consider $\max_{1\le j\le d}\{\widehat{\sigma}_j\}$ in (E.11). In Corollary E.2 we prove that, with probability at least $1-2d^{-3}$,

$$|\sigma_j - \widehat{\sigma}_j| \le C'\sqrt{\frac{\log d}{n}}, \quad \text{for} \ \ 1 \le j \le d,$$

where $C'$ is a constant. For a sufficiently large $n$, we have

$$\widehat{\sigma}_j \le \frac{3}{2}\sigma_j < +\infty, \quad \text{for} \ \ 1\le j\le d$$

with the same probability. Taking minimum over both sides, we get

$$\max_{1\le j\le d}\{\widehat{\sigma}_j\} \le \frac{3}{2}\max_{1\le j\le d}\{\sigma_j\} < +\infty \tag{E.13}$$

with probability at least $1-2d^{-2}$. Plugging (E.12) and (E.13) into the right-hand side of (E.11), we reach the conclusion that $\rho_+\big(\nabla^2\mathcal{L}, s\big) < +\infty$ holds with probability at least $1-2d^{-1}-3d^{-2}$. Thus we conclude the proof. □

### E.3. Proof of Lemma C.2.

Proof. For semiparametric elliptical design regression, we have

$$\nabla\mathcal{L}(\boldsymbol{\beta}^*) = \widehat{\mathbf{K}}_{\boldsymbol{X},Y} - \widehat{\mathbf{K}}_{\boldsymbol{X}}\boldsymbol{\beta}^* = \widehat{\mathbf{K}}_{\boldsymbol{X},Y} - \boldsymbol{\Sigma}_{\boldsymbol{X},Y} + \boldsymbol{\Sigma}_{\boldsymbol{X},Y} - \widehat{\mathbf{K}}_{\boldsymbol{X}}\boldsymbol{\beta}^*,$$

where $\widehat{\mathbf{K}}_{\boldsymbol{X}} \in \mathbb{R}^{d\times d}$ and $\widehat{\mathbf{K}}_{\boldsymbol{X},Y} \in \mathbb{R}^{d\times 1}$ are the submatrices of $\widehat{\mathbf{K}}_{\boldsymbol{Z}} \in \mathbb{R}^{(d+1)\times(d+1)}$ defined in (B.1). Since $\mathbb{E}(Y|\boldsymbol{X}=\mathbf{x}) = \mathbf{x}^T\boldsymbol{\beta}^*$, we have

$$\boldsymbol{\Sigma}_{\boldsymbol{X},Y} = \mathbb{E}(\boldsymbol{X}Y) = \mathbb{E}(\boldsymbol{X}\boldsymbol{X}^T\boldsymbol{\beta}^*) = \boldsymbol{\Sigma}_{\boldsymbol{X}}\boldsymbol{\beta}^*.$$

Hence we have

$$\begin{aligned}\|\nabla\mathcal{L}(\boldsymbol{\beta}^*)\|_\infty &= \big\|\widehat{\mathbf{K}}_{\boldsymbol{X},Y} - \boldsymbol{\Sigma}_{\boldsymbol{X},Y} + \boldsymbol{\Sigma}_{\boldsymbol{X}}\boldsymbol{\beta}^* - \widehat{\mathbf{K}}_{\boldsymbol{X}}\boldsymbol{\beta}^*\big\|_\infty \\ &\le \big\|\widehat{\mathbf{K}}_{\boldsymbol{X},Y} - \boldsymbol{\Sigma}_{\boldsymbol{X},Y}\big\|_\infty + \big\|\boldsymbol{\Sigma}_{\boldsymbol{X}}\boldsymbol{\beta}^* - \widehat{\mathbf{K}}_{\boldsymbol{X}}\boldsymbol{\beta}^*\big\|_\infty.\end{aligned}$$

Before we upper bound the two terms on the right-hand side, we establish a concentration inequality for $\widehat{\mathbf{K}}_{\boldsymbol{Z}}$. Let $\mathbf{D}_{\boldsymbol{Z}} = \mathrm{diag}(\sigma_1,\ldots,\sigma_{d+1})$ and $\widehat{\mathbf{D}}_{\boldsymbol{Z}} =$ $\operatorname{diag}(\widehat{\sigma}_1, \ldots, \widehat{\sigma}_{d+1})$, where $\sigma_1, \ldots, \sigma_{d+1}$ are the marginal standard deviations of $\boldsymbol{Z} \in \mathbb{R}^{(d+1)} = (Y, \boldsymbol{X})^T$ while $\widehat{\sigma}_1, \ldots, \widehat{\sigma}_{d+1}$ are the corresponding Catoni's estimators defined in (E.3). We have

$$\boldsymbol{\Sigma}_{\boldsymbol{Z}} = \mathbf{D}_{\boldsymbol{Z}} \boldsymbol{\Sigma}^0_{\boldsymbol{Z}} \mathbf{D}_{\boldsymbol{Z}}, \quad \widehat{\mathbf{K}}_{\boldsymbol{Z}} = \widehat{\mathbf{D}}_{\boldsymbol{Z}} \widehat{\mathbf{R}}_{\boldsymbol{Z}} \widehat{\mathbf{D}}_{\boldsymbol{Z}},$$

where $\widehat{\mathbf{R}}_{\boldsymbol{Z}}$ is the rank-based estimator of the generalized correlation matrix $\boldsymbol{\Sigma}^0_{\boldsymbol{Z}}$ defined in (A.1). Han and Liu (2012) proved that, with probability at least at least $1 - (d+1)^{-5/2}$,

$$\big\|\widehat{\mathbf{R}}_{\boldsymbol{Z}} - \boldsymbol{\Sigma}^0_{\boldsymbol{Z}}\big\|_{\max} \le C\sqrt{\frac{\log(d+1)}{n}},$$

where $\|\mathbf{M}\|_{\max} = \max_{1 \le i,j \le d}\big\{|M_{i,j}|\big\}$ for $\mathbf{M} \in \mathbb{R}^{d \times d}$. We have

$$\begin{aligned}
&\big\|\widehat{\mathbf{D}}_{\boldsymbol{Z}} \widehat{\mathbf{R}}_{\boldsymbol{Z}} \widehat{\mathbf{D}}_{\boldsymbol{Z}} - \mathbf{D}_{\boldsymbol{Z}} \boldsymbol{\Sigma}^0_{\boldsymbol{Z}} \mathbf{D}_{\boldsymbol{Z}}\big\|_{\max} \\
&\quad = \big\|\mathbf{D}_{\boldsymbol{Z}}(\widehat{\mathbf{R}}_{\boldsymbol{Z}} - \boldsymbol{\Sigma}^0_{\boldsymbol{Z}})\mathbf{D}_{\boldsymbol{Z}} + (\widehat{\mathbf{D}}_{\boldsymbol{Z}} - \mathbf{D}_{\boldsymbol{Z}})\widehat{\mathbf{R}}_{\boldsymbol{Z}} \mathbf{D}_{\boldsymbol{Z}} + \widehat{\mathbf{D}}_{\boldsymbol{Z}} \widehat{\mathbf{R}}_{\boldsymbol{Z}}(\widehat{\mathbf{D}}_{\boldsymbol{Z}} - \mathbf{D}_{\boldsymbol{Z}})\big\|_{\max} \\
&\quad \le \big\|\mathbf{D}_{\boldsymbol{Z}}(\widehat{\mathbf{R}}_{\boldsymbol{Z}} - \boldsymbol{\Sigma}^0_{\boldsymbol{Z}})\mathbf{D}_{\boldsymbol{Z}}\big\|_{\max} + \big\|(\widehat{\mathbf{D}}_{\boldsymbol{Z}} - \mathbf{D}_{\boldsymbol{Z}})\widehat{\mathbf{R}}_{\boldsymbol{Z}} \mathbf{D}_{\boldsymbol{Z}}\big\|_{\max} \\
&\qquad\qquad + \big\|\widehat{\mathbf{D}}_{\boldsymbol{Z}} \widehat{\mathbf{R}}_{\boldsymbol{Z}}(\widehat{\mathbf{D}}_{\boldsymbol{Z}} - \mathbf{D}_{\boldsymbol{Z}})\big\|_{\max} \\
&\quad \le \|\mathbf{D}_{\boldsymbol{Z}}\|^2_{\max} \big\|\widehat{\mathbf{R}}_{\boldsymbol{Z}} - \boldsymbol{\Sigma}^0_{\boldsymbol{Z}}\big\|^2_{\max} + \|\mathbf{D}_{\boldsymbol{Z}}\|_{\max} \big\|\widehat{\mathbf{D}}_{\boldsymbol{Z}} - \mathbf{D}_{\boldsymbol{Z}}\big\|_{\max} \\
&\qquad\qquad + \big\|\widehat{\mathbf{D}}_{\boldsymbol{Z}}\big\|_{\max} \big\|\widehat{\mathbf{D}}_{\boldsymbol{Z}} - \mathbf{D}_{\boldsymbol{Z}}\big\|_{\max}.
\end{aligned} \tag{E.14}$$

Following similar arguments in Corollary E.2, we have

$$\big\|\widehat{\mathbf{D}}_{\boldsymbol{Z}} - \mathbf{D}_{\boldsymbol{Z}}\big\|_{\max} \le C\sqrt{\frac{\log(d+1)}{n}}, \quad \big\|\widehat{\mathbf{D}}_{\boldsymbol{Z}}\big\|_{\max} \le \big\|\mathbf{D}_{\boldsymbol{Z}}\big\|_{\max} + C\sqrt{\frac{\log(d+1)}{n}}$$

with probability at least $1 - 2(d+1)^{-3}$. We assume that $\sigma_j$ $(1 \le j \le d+1)$ is upper bounded, from (E.14) we have, with probability at least $1 - (d+1)^{-5/2} - 2(d+1)^{-3}$,

$$\big\|\boldsymbol{\Sigma}_{\boldsymbol{Z}} - \widehat{\mathbf{K}}_{\boldsymbol{Z}}\big\|_{\max} \le C\sqrt{\frac{\log(d+1)}{n}},$$

which implies that with the same probability,

$$\begin{aligned}
&\big\|\widehat{\mathbf{K}}_{\boldsymbol{X},Y} - \boldsymbol{\Sigma}_{\boldsymbol{X},Y}\big\|_\infty \le C\sqrt{\frac{\log(d+1)}{n}}, \\
&\big\|\boldsymbol{\Sigma}_{\boldsymbol{X}} \boldsymbol{\beta}^* - \widehat{\mathbf{K}}_{\boldsymbol{X}} \boldsymbol{\beta}^*\big\|_\infty \le \|\boldsymbol{\beta}^*\|_1 \big\|\boldsymbol{\Sigma}_{\boldsymbol{X}} - \widehat{\mathbf{K}}_{\boldsymbol{X}}\big\|_{\max} \le C\|\boldsymbol{\beta}^*\|_1 \sqrt{\frac{\log(d+1)}{n}}.
\end{aligned}$$

Then we reach the conclusion. □

## APPENDIX F: DETAILED SETTINGS OF NUMERICAL EXPERIMENTS

The detailed settings of the first numerical experiment in §7 of Wang et al. (2014a) are as follows:

- The design matrix $\mathbf{X} \in \mathbb{R}^{n\times d}$ contains $n = 500$ independent realizations of a random vector $\boldsymbol{X} \in \mathbb{R}^d$ with $d = 2500$, which follows a $t$-distribution with 5 degrees of freedom, zero mean and correlation matrix $\mathbf{\Sigma}_{\boldsymbol{X}}^0$. We set the correlation matrix $\mathbf{\Sigma}_{\boldsymbol{X}}^0$ to be $(\mathbf{\Sigma}_{\boldsymbol{X}}^0)_{i,j} = 0.8^{|i-j|}$ $(1 \le i, j \le d)$. Meanwhile, in the $i$-th data sample the response $y_i$ follows a univariate $t$-distribution with 5 degrees of freedom, mean $\mathbf{x}_i^T\boldsymbol{\beta}^*$ and variance 0.01. Here $\mathbf{x}_i^T$ is the $i$-th row of the design matrix $\mathbf{X}$, and $\boldsymbol{\beta}^*$ is the true parameter vector specified as follows.
- For the true parameter vector $\boldsymbol{\beta}^* \in \mathbb{R}^d$, we set its first 100 coordinates to be independent realizations of a standard univariate Gaussian distribution (zero mean and unit variance), and other coordinates to be zero, i.e., we set $s^* = |\mathrm{supp}(\boldsymbol{\beta}^*)| = 100$.
- For the sequence of regularization parameters $\{\lambda_t\}_{t=0}^N$, we set $\lambda_{\mathrm{tgt}} = 0.05$ by cross-validation. Remind $\lambda_0 = \|\nabla\mathcal{L}(\mathbf{0})\|_\infty = \big\|\widehat{\mathbf{K}}_{\boldsymbol{X},Y}\big\|_\infty$, where $\widehat{\mathbf{K}}_{\boldsymbol{X},Y} \in \mathbb{R}^d$ is in (B.1). We fix the random seed in MATLAB to be 2. In this setting, we observe $\lambda_0 = 2.8516$. We set $\eta = 0.9015$, so that the total number of regularization parameters is $N = \log(\lambda_{\mathrm{tgt}}/\lambda_0)/\log\eta = 39$.
- For the MCP penalty defined in (2.2) of Wang et al. (2014a), we set $b = 1.1$. Meanwhile, we set the optimization precision within the $N$-th path following stage to be $\epsilon_{\mathrm{opt}} = 10^{-6}$, and $L_{\min} = 10^{-6}$.

The detailed settings of the second numerical experiment in §7 are as follows:

- The design matrix $\mathbf{X} \in \mathbb{R}^{n\times d}$ contains $n = 200$ independent realizations of a random vector $\boldsymbol{X} \in \mathbb{R}^d$ with $d = 2000$, which follows a zero mean Gaussian distribution with covariance matrix $(\mathbf{\Sigma}_{\boldsymbol{X}})_{i,j} = 0.9 \cdot \mathbb{1}\,(i \ne j) + \mathbb{1}\,(i = j)$. Meanwhile, we set $\mathbf{y} - \mathbf{X}\boldsymbol{\beta}^*$ to be an $n$-dimensional Gaussian random vector with zero mean and covariance matrix $\mathbf{I}$; $\boldsymbol{\beta}^*$ is set to be zero on its first 1990 dimensions, and takes $+2$ and $-2$ with equal probability on its last 10 dimensions.
- For the sequence of regularization parameters $\{\lambda_t\}_{t=0}^N$, we set $\lambda_{\mathrm{tgt}}$ by cross-validation, and $\lambda_0 = \|\nabla\mathcal{L}(\mathbf{0})\|_\infty = \|\mathbf{X}^T\mathbf{y}\|_\infty/n$. Other parameters are set to be the same as in the previous experiment.
- We compare with LLA (Zou and Li, 2008), the calibrated CCCP (Wang et al., 2013), SparseNet (Mazumder et al., 2011), and the multi-stage

convex relaxation method (Zhang, 2010b; Zhang et al., 2013). For SparseNet we use the same sequence of regularization parameters as in our setting. For the other procedures, we employ the `glmnet` package (Friedman et al., 2010) to compute the Lasso problem in (6.1) of Wang et al. (2014a) at each stage.

- For the multi-stage convex relaxation method, we set the maximum number of stages to be 20. For the calibrated CCCP, we set the tuning parameter $\tau = 1/\log(n)$ as suggested by Wang et al. (2013). To be fair, each method selects its most suitable regularization parameter using cross-validation. We repeat the experiment for 1000 times.

Department of Operations Research
and Financial Engineering
Princeton University
Princeton, New Jersey 08544
USA
E-mail: zhaoran@princeton.edu
hanliu@princeton.edu

Department of Statistics
Rutgers University
Piscataway, New Jersey 08854
USA
E-mail: tzhang@stat.rutgers.edu